\pdfoutput=1 
\documentclass[10pt,twocolumn,letterpaper]{article}

\usepackage{cvpr}              %

\usepackage{siunitx}  %
\usepackage{bbm}
\usepackage{comment}
\usepackage{xfrac}
\usepackage{tocloft}

\newcommand\blfootnote[1]{%
  \begingroup
  \renewcommand\thefootnote{}\footnote{#1}%
  \addtocounter{footnote}{-1}%
  \endgroup
}

\definecolor{ms_note}{RGB}{0, 181, 190}
\definecolor{dt_note}{RGB}{255, 50, 200}
\definecolor{bgcolor}{RGB}{190, 181, 190}
\definecolor{verify_color}{RGB}{255, 30, 80}

\PassOptionsToPackage{table}{xcolor}
\usepackage{xcolor} %
\usepackage{colortbl}
\usepackage{pifont}
\usepackage{tikz}

\newcommand{\absrel}{\textrm{rel}}
\newcommand{\threshI}{\tau}
\newcommand{\my}{\ding{51}}
\newcommand{\mn}{\ding{55}}
\newcommand{\bestresult}[1]{\textbf{#1}}
\newcommand{\kittishort}{KITTI}
\newcommand{\scannetshort}{ScanNet}
\newcommand{\ethdshort}{ETH3D}
\newcommand{\dtushort}{DTU}
\newcommand{\tanksandtemplesshort}{T\&T}
\definecolor{mylightgray}{RGB}{238,238,238} 
\colorlet{bgcolor}{mylightgray}
\newcommand{\vect}[1]{\mathbf{#1}}
\newcommand{\kitti}{KITTI}
\newcommand{\scannet}{ScanNet}
\newcommand{\dtu}{DTU}
\newcommand{\trainedsimto}[1]{(#1)}

\definecolor{firstcolor}{rgb}{1, 1, 1}
\definecolor{secondcolor}{rgb}{1, 1, 1}
\definecolor{thirdcolor}{rgb}{1,1, 1}

\renewcommand{\paragraph}[1]{\vspace{2pt}\noindent\textbf{#1}\hspace{6pt}}

\newcommand{\modelname}{MVSA\@\xspace}

\newcommand{\benchmarkname}{RMVDB}

\newcounter{ablationrow}
\renewcommand{\theablationrow}{\Alph{ablationrow}}
\newcommand{\row}[1]{\refstepcounter{ablationrow}\label{#1}\theablationrow}

\definecolor{cvprblue}{rgb}{0.21,0.49,0.74}
\usepackage[pagebackref,breaklinks,colorlinks,allcolors=cvprblue]{hyperref}

\def\paperID{13284} %
\def\confName{CVPR}
\def\confYear{2025}

\title{MVSAnywhere: Zero-Shot Multi-View Stereo}

\newcommand{\gap}{\hspace{5pt}}
\author{Sergio Izquierdo$^{3,*}$ \gap Mohamed Sayed$^{1}$ \gap Michael Firman$^{1}$  \gap Guillermo Garcia-Hernando$^{1}$  \\
Daniyar Turmukhambetov$^{1}$ 
Javier Civera$^{3}$ \gap Oisin Mac Aodha$^{2}$  \gap  Gabriel Brostow$^{1,4}$  \gap Jamie Watson$^{1,4}$ \\ 
$^{1}$Niantic \hspace{20pt} $^{2}$University of Edinburgh   \hspace{20pt}$^{3}$I3A, Universidad de Zaragoza \hspace{20pt}$^{4}$UCL  \\
 \url{https://nianticlabs.github.io/mvsanywhere/}}

\begin{document}
\maketitle

\begin{abstract}
Computing accurate depth from multiple views is a fundamental and longstanding challenge in computer vision.
However, most existing approaches do not generalize well across different domains and scene types (\eg indoor \vs outdoor). 
Training a general-purpose multi-view stereo model is challenging and raises several questions, \eg how to best make use of transformer-based architectures, how to incorporate additional metadata when there is a variable number of input views, and how to estimate the range of valid depths which can vary considerably across different scenes and is typically not known a priori? 
To address these issues, we introduce \textit{MVSA}, a novel and versatile \underline{M}ulti-\underline{V}iew \underline{S}tereo architecture that aims to work \underline{A}nywhere by generalizing across diverse domains and depth ranges.
\textit{MVSA} combines monocular and multi-view cues with an adaptive cost volume to deal with scale-related issues. 
We demonstrate state-of-the-art zero-shot depth estimation on the Robust Multi-View Depth Benchmark, surpassing  existing multi-view stereo and monocular baselines. 
\end{abstract}

\vspace{-5pt}
\section{Introduction}
\blfootnote{$^*$Work done during an internship at Niantic.}
\vspace{-2pt}
Estimating accurate depth from multiple RGB images is a core challenge in 3D vision, and a building block for downstream applications like 3D reconstruction and autonomous driving. 
Recent approaches in learning-based multi-view stereo (MVS) are capable of generating accurate depths~\cite{yao2018mvsnet,Yang2022,cao2022mvsformer}. 
However, existing methods typically struggle to generalize to scene and camera setups that differ significantly from those in their training data. 
As a result, there is a pressing need for general-purpose MVS methods that are more robust to differences between the training and test distributions.

\begin{figure}[t]
    \vspace{-5pt}
    \includegraphics[width=\columnwidth]{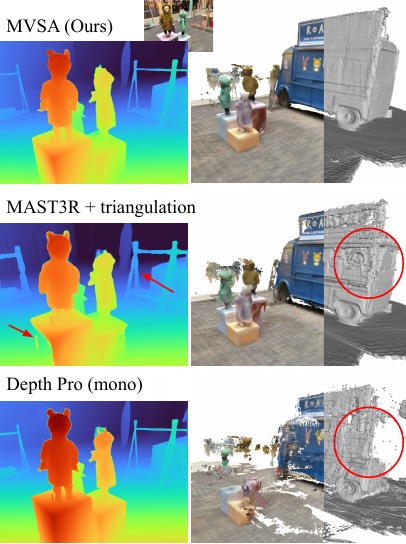}
    \vspace{-17pt}
    \caption{
        Our \textbf{\modelname} model results in high-quality reconstructions from posed images, and is superior to existing monocular and MVS methods. 
        Here we compare with Depth Pro~\cite{bochkovskii2024depthpro}, a recent monocular method which  produces sharp and good looking depth maps, but can have inconsistent scaling of depths, which are required for good meshes.
        We also include a variant of MAST3R~\cite{mast3r_arxiv24} that we have augmented with ground truth camera poses.
        Our model gives sharp depth maps which are also accurate and 3D consistent, producing high-quality meshes in zero-shot environments.
    }
    \label{fig:teaser}
    \vspace{-18pt}
\end{figure}

We take inspiration from the recent explosion in scene-agnostic \emph{single-view} depth models, which predict plausible  metric~\cite{yin2023metric,hu2024metric3d,bhat2023zoedepth,guizilini2024grin,piccinelli2024unidepth,wang2024moge} or up-to-scale~\cite{Ranftl2022,ke2023repurposing,depthanything,depth_anything_v2} depth using only a single image as input. 
These models are typically trained on large curated sets of synthetic and/or real RGB-D data, endowing them with impressive generalization performance on previously unseen data. %
Single-view models are, however, inherently limited by their input. For our specific depth prediction target, constraining the model's input to just one image forces it to use single-view geometry cues (\eg vanishing points) and learned patterns \cite{danier2024depthcues}, while losing the stronger multi-view signal.
While there are temporal extensions of these single view models~\cite{wang2019recurrent,patil2020don,watson2021temporal,shao2024learning,yang2024depthanyvideo}, their focus is on temporal perceptual consistency, and not necessarily multi-view consistency. 
In application contexts where multiple views are available at inference time, it stands to reason that these lead to significantly more accurate depth estimates~\cite{smolyanskiy2018importance,Luo2020VideoDepth,izquierdo2023sfm}. 

Developing a general-purpose MVS method, however, raises two significant challenges. 
Firstly, it should be able to deal with arbitrary depth ranges. %
Existing MVS methods typically require a known range of depths to `search' over along epipolar lines, corresponding to a discrete set of depth bins used to build a cost volume. 
These depths are typically either fixed (and chosen from the range of depths in the training data) \cite{sayed2022simplerecon} or are provided at test time for each image \cite{yao2018mvsnet,Wang2020,Yang2022}.
Secondly, the many emerging benefits of ViTs~\cite{dosovitskiy2020image} motivate us to find a way to `upgrade' parts of standard MVS architectures that are still CNNs.

To address these challenges, we introduce a new general-purpose  MVS method named \underline{M}ulti-\underline{V}iew \underline{S}tereo \underline{A}nywhere (\textbf{\modelname}). 
Similarly to recent performant monocular methods, it is trained on a large and diverse set of data, spanning diverse depth ranges.
Along with harmonizing these training signals, our main technical contributions are:

\vspace{-2pt}
\begin{itemize}
    \item A novel transformer-based architecture that processes the multi-view cost volume, while \emph{also} incorporating monocular features. %
    We propose a Cost Volume Patchifier that tokenizes the cost volume without loosing its details, while also incorporating features from a monocular ViT.

    \item We propose a view-count-agnostic and scale-agnostic mechanism to construct the cost volume using geometric metadata given any number of input source frames. This is in contrast to the established practice~\cite{sayed2022simplerecon} of concatenating geometric metadata from a fixed number of frames to build the cost volume.

\end{itemize}

\modelname predicts highly accurate and 3D-consistent \emph{depths}, obtaining state-of-the-art results on the Robust Multi-View Depth Benchmark~\cite{schroeppel2022robust}, which contains a variety of challenging held-out datasets. We also report scores for some new single- and multi-view methods for comparison.
Our better depths result in improved 3D mesh \emph{reconstruction} compared to alternative depth-based reconstruction methods (\cref{fig:teaser}). 
Code and pretrained models are available at \href{https://github.com/nianticlabs/mvsanywhere}{https://github.com/nianticlabs/mvsanywhere}.

\vspace{-5pt}
\section{Related Work}
\vspace{-5pt}
\paragraph{Multi-view stereo  (MVS).} MVS algorithms  estimate depth from posed multi-view images using epipolar geometry~\cite{wang2024learning}.
Given calibrated cameras, early methods estimated depth by matching image patches~\cite{furukawa2015multi,schoenberger2016mvs}.
Subsequently, deep learning approaches were introduced, first for stereo matching~\cite{zbontar2015computing}  and later improved via end-to-end learning, typically using plane-sweep cost volumes~\cite{Teed2020Deepv2d,kendall2017end,im2019dpsnet,wang2018mvdepthnet,huang2018deepmvs,yao2018mvsnet,gu2020cascade,Zhang2019GANet,cheng2019learning}.
Subsequent  methods introduced advances in architectures~\cite{ding2022transmvsnet,cao2022mvsformer,cao2024mvsformerplus}, increased robustness to occlusion and moving objects~\cite{Zhang2020,long2020occlusion,wimbauer2021monorec}, integrated temporal information~\cite{duzceker2021deepvideomvs}, improved model efficiency~\cite{sayed2022simplerecon,Yu2020},  jointly estimated camera pose~\cite{dust3r_cvpr24,mast3r_arxiv24} and ingested prior geometry estimates to improve depths~\cite{sayed2024doubletake}.

\begin{figure}
    \centering
    \includegraphics[width=.85\linewidth]{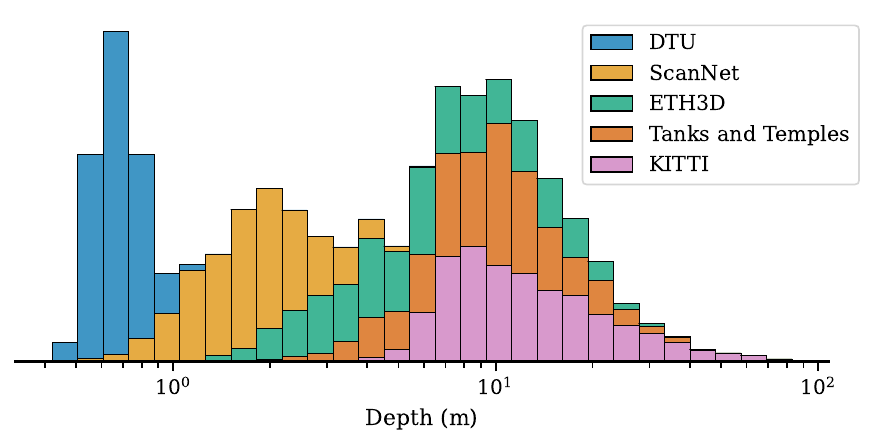}
    \vspace{-7pt}
    \caption{\textbf{MVS datasets cover a wide range of depth values.} 
    Here we show the distribution of \% depths in the DTU~\cite{jensen2014large}, ScanNet~\cite{dai2017scannet}, ETH3D~\cite{schoeps2017cvpr}, Tanks and Temples~\cite{Knapitsch2017}, and KITTI~\cite{Geiger2012CVPR} datasets, as a stacked bar chart. Note the log x-axis.
    This wide range of depth values can be challenging when it comes to constructing meaningful cost volumes and predicting the final depths.
    }
    \vspace{-13pt}
    \label{fig:range-of-depths}    
\end{figure}

With some exceptions~\cite{zhang2019domaininvariant}, earlier stereo and MVS methods were traditionally both trained and tested on the same dataset/domain, and were limited in their ability to generalize to out-of-distribution data. 
This domain generalization issue is a consequence of most performant learning-based MVS methods being data-hungry. 
Approaches such as a training on synthetic~\cite{mayer2016large,zhang2024robust} or pseudo-labeled depth~\cite{hu2024DepthCrafter} can be effective, but so far, struggle to span a diverse range of scene types and scales.  
Self-supervised approaches can be trained without depth supervision, but current methods produce inferior depths compared to fully supervised approaches~\cite{khot2019learning,dai2019mvs2,yang2021self}. 
Concurrent with our work, \cite{guo2024stereo,wen2025foundationstereo} trained large \emph{binocular} stereo models on large synthetic datasets.

\paragraph{Adaptive cost volumes.} 
One of the challenges in developing a general-purpose domain-agnostic MVS method is that different scenes can contain wildly different depth ranges, \eg indoor scenes are limited to a few meters, while outdoor ones can span much larger distances. 
This is a problem as conventional cost volumes require a known depth range, which is typically just estimated based on the minimum and maximum depth values in the training set.  
As a result, there is a need for cost volumes that are not restricted to a pre-defined range or bins, and instead are adaptive. 
In the context of self-supervised learning with unscaled poses, \cite{watson2021temporal} estimated bin ranges at training time via an exponential moving average of the depth predictions.  %
Another approach is to predict bin centers iteratively in a coarse-to-fine manner, where the outputs from the previous iteration are used to seed the range in the next~\cite{gu2020cascade,mi2022generalized,zhang2023arai}.
Alternatively, the bin offsets can be predicted by a learned network~\cite{Conti_2024_3DV} or from estimated depth uncertainty~\cite{cheng2020deep,li2023nr}. 
We estimate cost volume depth ranges to enable us adapt to any range of depths, while prior work has done this when the test time range is  known, but they wish to reduce computation or enhance detail.

\paragraph{Single-view depth.}
Monocular depth methods, trained using supervised learning,  do not have access to multi-view images at inference time so instead aim to estimate a depth map from a \emph{single} image~\cite{eigen2014depth,eigen2015predicting,fu2018deep,MegaDepthLi18}.
Building on highly advanced and effective image backbones~\cite{oquab2024dinov2}, more recent monocular methods have focused on making \emph{general-purpose} depth estimation models, which aim to work on arbitrary scenes~\cite{chen2016single,Ranftl2022}.
Further works have scaled up the size of models and datasets, training on combinations of real and/or synthetic data~\cite{depthanything,depth_anything_v2}, and have used  stronger image-level priors~\cite{ke2023repurposing,he2024lotus}.
One of the limitations of models trained from  stereo-image-derived supervision without known baselines~\cite{Ranftl2022}
 or human annotations~\cite{chen2016single} is that these only enable a relative, and not metric (\eg in meters), depth prediction.

Other monocular models predict \emph{metric} depth~\cite{yin2023metric,hu2024metric3d,bhat2023zoedepth,bochkovskii2024depthpro,wang2024moge}.
Not only does this rely on appropriate training data, but also requires an understanding of camera intrinsics, which are often a required additional input to the network.
Conventional monocular methods are inherently limited by only incorporating information from single views at inference time, even when multi-view information is available~\cite{watson2021temporal}. 
On the other hand, with recent advances, they can still provide a very valuable signal when only one image is available.  
As in~\cite{wang2022mvster,xu2024depthsplat,bartolomei2024stereo}, we combine features extracted from a monocular depth model with a multi-view cost volume to better leverage monocular and multi-view cues.

\section{General-purpose Multi-View Stereo}

Our model takes as input a $H \times W$ reference image $I_r$ together with neighboring source frames  $I_{i \in \{1...N\}}$, each with their relative poses and intrinsics. 
At test time we aim to predict a dense depth map ${\hat{D}_r}$ for $I_r$.
For ours to be a general-purpose MVS method, we seek to:
\begin{enumerate}
    \item \textbf{Generalize to any domain}. Most current MVS methods are typically trained on and tested on data from similar domains, \eg indoor only or driving only. 
    \item \textbf{Generalize to any range of depths}. 
    Predicted depth maps need to be accurate for nearby surfaces  (\eg for robotics) or for more distant ones (\eg for drones and autonomous driving). In some scenarios like SfM, the depths and camera poses are in a non-metric up-to-scale coordinate system. Hence, general-purpose MVS should be robust to the scale of the coordinate system.

    \item \textbf{Be robust to the number and selection of source frames}.
    Traditional MVS systems can struggle when there is little overlap between source and reference frames. We also want MVS methods to be agnostic to the number of source frames available at test time.
   
    \item \textbf{Predict 3D-consistent depths.}
    Depths from one viewpoint should be consistent with those predicted from different viewpoints. Fusion of consistent depth maps will produce a mesh with accurate estimates of 3D surfaces. 
    
\end{enumerate}

\noindent While prior works have tackled these problems in turn, we are the first model, to the best of our knowledge, to tackle all four problems in a single system.

\begin{figure}[t]
    \includegraphics[width=\columnwidth]{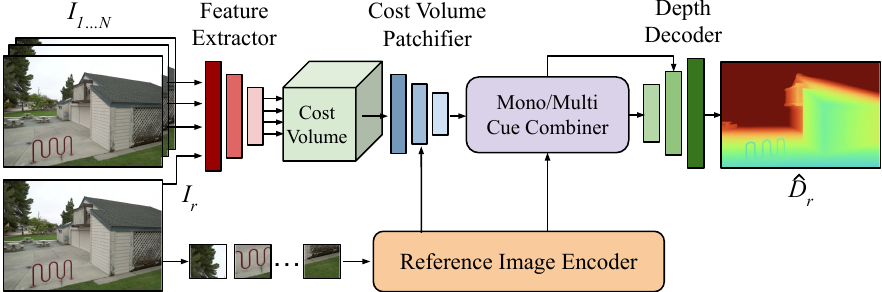}
    \vspace{-15pt}
    \caption{
        \textbf{Our general-purpose multi-view depth estimation model}. 
        We start with a cost-volume based architecture, which matches deep features between views at different hypothesized depths. 
        Key for performance are our Cost Volume Patchifier and Mono/Multi Cue Combiner. 
        These also fuse single-view information coming from the Reference Image Encoder and source views. 
    }
    \vspace{-10pt}
    \label{fig:method_overview}
\end{figure}

\subsection{MVSAnywhere}%
We introduce \textbf{MVSAnywhere} (MVSA), a novel general-purpose MVS system which is designed to embody each of the previous properties.
To help us learn from diverse datasets and hence \textbf{generalize to any domain}, we use a large transformer-based architecture, which takes as input: 
(1) multi-view information from the reference and source images,
and 
(2) single-view information, which is extracted directly from the reference image via a monocular \emph{reference image encoder}.
The overall architecture (\cref{fig:method_overview} and supplementary) is broadly inspired by recent MVS approaches, \eg \cite{sayed2022simplerecon}.
It comprises five key components: 
\vspace{-4pt}
\begin{description}[itemsep=1pt, parsep=0pt, leftmargin=8pt]
    \item[Feature extractor.]
        This encodes the source and reference images into deep feature maps $\mathcal{F}_r$ and $\mathcal{F}_{i \in \{1...N\}}$, that will be processed via a cost volume.
        We use the first two blocks of a ResNet18~\cite{he2016deep} for this encoder, producing feature maps at resolution $H/4 \times W/4$.
    \item[Cost volume.]
        Following \eg \cite{kendall2017end, wang2018mvdepthnet,huang2018deepmvs,cao2024mvsformerplus}, we warp feature maps $\mathcal{F}_{i}$ from each source view to the reference one using a set of hypothesized depth values (\ie bins) $\mathcal{D}$.
        We then concatenate these warped features and $\mathcal{F}_{r}$ with appropriate \emph{metadata}, following \cite{sayed2022simplerecon}.
        See \cref{sec:view-count-agnostic-metadata} for our specific novel contributions in this matter.
    \item[Reference image encoder.]
        This extracts powerful deep monocular features for $I_r$.
        We use the ViT Base~\cite{dosovitskiy2020image} encoder from Depth Anything V2~\cite{depth_anything_v2}, with their pretrained weights for relative monocular depth estimation, which help us to be \textbf{robust to limited overlaps} between source and reference frames. As ViT Base operates on $14 \times 14$ patches, the reference image is resized to $\frac{14H}{16} \times \frac{14W}{16}$ resolution before feeding to ViT Base, such that the extracted features are size $\frac{H}{16} \times \frac{W}{16}$.
    \item[Mono/Multi Cue Combiner.]
        This converts the ``patchified'' features of the cost volume and reference image into a sequence of features which go to our depth decoder.
        Monocular and multi-view cues are combined by a novel component described in \cref{sec:mono_multi_cue_combiner}.
        
    \item[Depth Decoder.]
        Based on the decoder from~\cite{Ranftl2021}, MVSA progressively upsamples and processes features from the Mono/Multi Cue Combiner module to produce the final depth map at the reference image resolution.
\end{description}

\begin{figure}
    \centering
    \includegraphics[width=1.0\columnwidth]{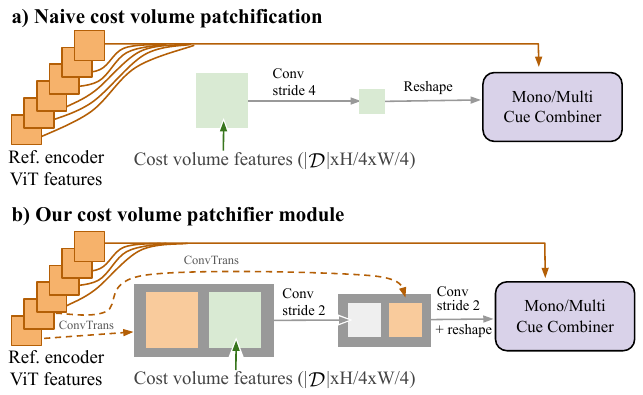}
    \vspace{-16pt}
    \caption{\textbf{Our cost volume patchifier} enables high-quality information to be extracted from a $|\mathcal{D}| \times \frac{H}{4} \times \frac{W}{4}$ cost volume, ready for input to the Mono/Multi Cue Combiner ViT. (a) Shows the naive approach to patchification.  (b) Our approach makes better use of the reference image features. 
    }
    \label{fig:cost_vol_patchifier}
    \vspace{-8pt}
\end{figure}

\subsection{Metadata Agnostic to View Count and Scale} 
\label{sec:view-count-agnostic-metadata}

SimpleRecon~\cite{sayed2022simplerecon} demonstrated that readily available metadata, \eg geometric and camera pose information, can be incorporated into the cost volume to improve depths.  
For each pixel location $(u_r, v_r)$ in $I_r$ and depth bin $k$ in $\mathcal{D}$, we backproject the pixel to a 3D point $P$ and then reproject it into every source view $I_i$.
The specific metadata for the bin with coordinates $(u_r, v_r, k)$ in the cost volume includes: the dot product between feature vector $\mathcal{F}_r(u_r, v_r)$ and corresponding pixels in $\mathcal{F}_i(u_i, v_i)$, ray directions from source and reference origin to $P$, depths in reference and source views, angle between rays from reference and source views, relative poses between the reference and source cameras, and depth validity masks (in case $P$ is outside a source frame frustum). 
See our supplementary for full details.  

SimpleRecon~\cite{sayed2022simplerecon}'s cost volume concatenates metadata from all eight source frames and runs an MLP to produce one single cost (matching score) per spatial location and depth hypothesis. 
While this gives good scores, its limitation is that it requires \emph{exactly} eight source frames for every training and test reference image, limiting the model's flexibility (note though that traditional MVS methods are typically already view-count agnostic). 
To address this limitation, we introduce a \emph{view-count-agnostic metadata} component which enables a single model to \textbf{generalize to any number of source views}.
For each source frame, we run an MLP that ingests the metadata from the reference frame and the source frame and predicts two values: a score and a weight. 
This results in $N$ scores and $N$ weights.
A weighted sum of the $N$ scores is computed after the $N$ weights go through softmax.
This weighted sum is used as the value in the cost volume at every pixel location $(u, v)$ and depth hypothesis $k$.
Our novel module enables aggregation of the matching score and confidence for each source frame, while allowing for a variable number of source frames for each $I_r$.

The source camera poses may be close to the reference, or far from it.
To be more invariant to this possible range of scales, we also make the metadata \textbf{scene scale-agnostic}. 
To this end, we normalize the relative pose measures of the metadata using a maximum across all the source frames for a given reference frame. 
We also normalize the depth hypothesis metadata using the maximum and minimum of $\mathcal{D}$.

As the scene scale information is not provided to the rest of our network, we rescale the depth predictions to match the scale of the input poses. Our depths are predicted with a sigmoid function $\sigma$ over the logit $x$. To align the prediction of the network with the cost volume, the sigmoid output is scaled by the depth range of the cost volume, so
\begin{align}
\hat{D}_r = \exp\left(\log(d_{\min}) + \log(d_{\max} / d_{\min}) \cdot \sigma(x)\right).
\end{align}

\subsection{Mono/Multi Cue Combiner} %
\label{sec:mono_multi_cue_combiner}

Given the cost volume of shape 
$|\mathcal{D}| \times \frac{H}{4} \times \frac{W}{4}$, and the reference image encoder features of shape $C \times \frac{H}{16} \times \frac{W}{16}$ (outputs of different blocks of the reference image encoder), we pose the question: how can we best combine these features to provide the strongest signal for the decoder?
Motivated by the recent success of transformer architectures in single-view depth prediction~\cite{depth_anything_v2, bochkovskii2024depthpro, Ranftl2021}, we use a ViT-Base network to process these features in a \emph{Mono/Multi Cue Combiner} network, which produces a sequence of tokens for the decoder to transform into a depth prediction.

To effectively achieve this we need to i) convert the cost volume into a token sequence without sacrificing information and ii) incorporate the monocular cues to help in decoding sharp depth.
For i), a naive approach would be to apply a strided convolution projecting to the ViT token dimensions, resembling how RGB images are patchified. However this is suboptimal, for it lacks contextual information on how to achieve this downsampling.
Instead, we propose a \textbf{cost volume patchifier} module.
This guides the downsampling process with information from the first two blocks of the reference image encoder. 
We convert the cost volume into tokens using two strided convolutions, but first, concatenate  each of them with the monocular features of the first two blocks, transposed and projected at 1/4 and 1/8 of the input resolution, respectively. The output of this module is a sequence of $\frac{H}{16}\times\frac{W}{16}$ tokens, matching the sequence length of the monocular features. These tokens are then fed into a ViT-B initialized with DINOv2 weights (see Fig.~\ref{fig:cost_vol_patchifier}).

For ii) we add the tokens from the cost volume with the ones from the reference image encoder after projecting the latter with a linear layer. We repeat this process at blocks 2, 5, 9, and 11 of the ViT to incorporate multiple levels of monocular cues. This simple mechanism allows the network to refine and regularize the cost volume with the help of the reference image structure.

\subsection{Generalizing to Any Range of Depths }
\label{sec:depth_generalization}
When building a cost volume, a set of depth hypotheses (\ie bins) $\mathcal{D}$ are used to warp feature maps $\mathcal{F}_i$ to $I_r$. 
This raises the question: How do we choose $\mathcal{D}$ to \textbf{generalize to any range of depths}? 
Depth ranges vary hugely across datasets (see \cref{fig:range-of-depths}), so using the same fixed range is suboptimal.

We address this with a cascaded cost volume approach, first introduced in 3D stereo matching~\cite{gu2020cascade,mi2022generalized,zhang2023arai}. 
While these works start from a known `ground truth' depth range, we use the known intrinsics and extrinsics to infer the minimum and maximum depths that could be matched between $I_r$ and each $I_i$.
We space our initial depth bins logarithmically within this range, then make an initial depth prediction. 
The min and max values of this initial estimate are then used to rebuild the cost volume for a final depth prediction. 
This iterative process occurs only at test time; during training, we use the known depth range. 
Full details are in the supplementary.
Importantly, previous methods that are provided with an exact depth range learn to predict depths that cover all the depth hypotheses.
Thus, when using a rough estimate of the range, these methods fail to align the prediction to the actual valid depths.
To further mitigate this issue, we augment the ground truth ranges via a random perturbation during training.

\begin{table}
    \centering
    \resizebox{1\columnwidth}{!}{%
    \begin{tabular}{llllllll}
    \toprule
\textbf{Name}	&			\textbf{Scenes}	&	 \textbf{\#total}	& 	\textbf{\# total}	& 	\textbf{\# training}	&	\textbf{Metric}	&	\textbf{Moving}	\\
	& 				& 	\textbf{scenes}	& 	\textbf{images}	& 	\textbf{tuples}	& 	\textbf{poses?}	& 	\textbf{objects?}	\\
\midrule											
Hypersim~\cite{roberts2021hypersim}	&			Indoor	&	461	&	~77K	&	~45K	&	Yes	&	No	\\
TartanAIR~\cite{wang2020tartanair}	&			Indoor, Outdoor	&	30	&	~1M	&	~92K	&	Yes	&	Yes	\\
BlendedMVG~\cite{yao2020blendedmvs}	&			Indoor, Outdoor, Aerial	&	389	&	~110K	&	~97K	&	No	&	No	\\
MatrixCity~\cite{li2023matrixcity}	&			Outdoor, Aerial	&	1	&	~519K	&	~40K	&	Yes	&	No	\\
VKITTI2~\cite{cabon2020vkitti2,gaidon2016virtual}	&			Outdoor	&	5	&	~21K	&	~40K	&	Yes	&	Yes	\\
Dynamic Replica~\cite{karaev2023dynamicstereo} 	&			Indoor	&	484	&	~145K	&	~70K	&	Yes	&	Yes	\\
MVSSynth~\cite{huang2018deepmvs}	&			Outdoor	&	117	&	~12K	&	~3K	&	No	&	Yes	\\
SAIL-VOS 3D~\cite{HuCVPR2019,HuCVPR2021}	&			Indoor, Outdoor 	&	6807	&	~237K	&	~21K	&	Yes	&	Yes	\\
\bottomrule
    \end{tabular}
    }
    \vspace{-7pt}
    \caption{We train on eight MVS datasets from a variety of domains.
    All these datasets are synthetically rendered, giving them perfect ground truth depths and camera calibration. However, BlendedMVG uses real textures on their assets.
    }
    \vspace{-12pt}
    \label{tab:training-datasets}
\end{table}

\subsection{Implementation Details}

\paragraph{Losses.}
We use the supervised losses from~\cite{sayed2022simplerecon}.
These comprise an L1 loss between the log of the ground truth and the log of the predicted depth values, and a gradient and normals loss. 
Training losses are applied to four output scales of the decoder. 
At inference, only the final largest-scale prediction is used. 
We take as input $640 \times 480$ images, and output depth maps at the same resolution. We use 64 depth bins in $\mathcal{D}$ sampled in log space.

\paragraph{Keyframes.}
For datasets with dense sequences, we choose reference and source frames with the strategy of~\cite{duzceker2021deepvideomvs,sayed2022simplerecon}. 
To be robust to sparser sets of frames, we also select tuples based on geometry overlap, obtaining tuples of not necessarily consecutive frames.
Full details on our architecture and training strategy are provided in the supplementary.

\begin{table*}[t!]
    
    \newcommand{\baselinename}{Robust MVD Baseline}
    \newcommand{\absrelname}{Absolute Relative Error}
    \newcommand{\otherview}{\other{} view}
    \newcommand{\otherviews}{\otherview{}s}
    \newcommand{\threshIname}{Inlier Ratio}%
    \newcommand{\other}{source}
    \centering
    \resizebox{\textwidth}{!}{
        \setlength{\tabcolsep}{0.8mm}

\begin{tabular}{|l|c|c|c
|c >{\columncolor{bgcolor}}c
|c >{\columncolor{bgcolor}}c
|c >{\columncolor{bgcolor}}c
|c >{\columncolor{bgcolor}}c
|c >{\columncolor{bgcolor}}c
|c >{\columncolor{bgcolor}}c c
|}

\hline
    \textbf{Approach}
    & \textbf{\scriptsize{GT}}
    & \textbf{\scriptsize{GT}} 
    & \textbf{Align}
    & \multicolumn{2}{c|}{\textbf{\kittishort{}}}
    & \multicolumn{2}{c|}{\textbf{\scannetshort{}}}
    & \multicolumn{2}{c|}{\textbf{\ethdshort{}}}
    & \multicolumn{2}{c|}{\textbf{\dtushort{}}}
    & \multicolumn{2}{c|}{\textbf{\tanksandtemplesshort{}}}
    & \multicolumn{3}{c|}{\textbf{Average}}
    \\

    & \textbf{\scriptsize{Poses}}
    & \textbf{\scriptsize{Range}}
    &
    & $\absrel\downarrow$ & $\threshI\uparrow$
    & $\absrel\downarrow$ & $\threshI\uparrow$
    & $\absrel\downarrow$ & $\threshI\uparrow$
    & $\absrel\downarrow$ & $\threshI\uparrow$
    & $\absrel\downarrow$ & $\threshI\uparrow$
    & $\absrel\downarrow$ & $\threshI\uparrow$ & time [s] $\downarrow$
    \\
    \hline
    \hline

    \multicolumn{17}{|l|}{\textbf{a) Depth from frames (w/o poses)}}
    \\

	DeMoN~\cite{Ummenhofer2016}
	& \mn
	& \mn
	& $\Vert \vect t \Vert$
	& 15.5
	& 15.2
	& \bestresult{12.0}
	& \bestresult{21.0}
	& 17.4
	& 15.4
	& 21.8
	& 16.6
	& 13.0
	& 23.2
	& 16.0
	& 18.3
	& {0.08}
	\\ %

	DeepV2D \scriptsize{\kitti{}}~\normalsize{\cite{Teed2020Deepv2d}}
	& \mn
	& \mn
	& med
	& (\bestresult{3.1})
	& (\bestresult{74.9})
	& {23.7}
	& {11.1}
	& {27.1}
	& {10.1}
	& {24.8}
	& {8.1}
	& {34.1}
	& {9.1}
	& {22.6}
	& {22.7}
	& {2.07}
	\\ %

	DeepV2D \scriptsize{\scannet{}}\normalsize{~\cite{Teed2020Deepv2d}}
	& \mn
	& \mn
	& med
	& {10.0}
	& {36.2}
	& ({4.4})
	& ({54.8})
	& {11.8}
	& {29.3}
	& {7.7}
	& {33.0}
	& \bestresult{8.9}
	& \bestresult{46.4}
	& {8.6}
	& {39.9}
	& 3.57
	\\ %

 	MAST3R~\cite{mast3r_arxiv24} (raw output)
	& \mn
	& \mn
	& med
	& \bestresult{3.3}
	& \bestresult{67.7}
	& (\bestresult{4.3})
	& (\bestresult{64.0})
	& \bestresult{2.7}
	& \bestresult{79.0}
	& \bestresult{3.5}
	& \bestresult{66.7}
	& (\bestresult{2.4})
	& (\bestresult{81.6})
	& \bestresult{3.3}
	& \bestresult{71.8}
	& \bestresult{0.07}
	\\ %

        MAST3R~\cite{mast3r_arxiv24}  (raw output)  
	& \mn
	& \mn
	& \mn
    &  61.4 %
    &  0.4
    &  (12.8) %
    &  (19.4)
    &  43.8 %
    &  3.1
    &  145.8 %
    &  0.5
    &  (66.9) %
    &  (0.0)
    &  66.1
    &  4.7
    & \bestresult{0.07}
    \\

\hline
\hline

    \multicolumn{17}{|l|}{\textbf{b) Depth from frames and poses (with per-image range provided)}}
    \\

	MVSNet\normalsize{~\cite{yao2018mvsnet}}	& \my
	& \my
	& \mn
	& 22.7
	& 36.1
	& 24.6
	& 20.4
	& 35.4
	& 31.4
	& (1.8)
	& (86.0)
	& 8.3
	& 73.0
	& 18.6
	& 49.4
	& \bestresult{0.07}
	\\ %

	MVSNet \scriptsize{Inv. Depth}\normalsize{~\cite{yao2018mvsnet}}
	& \my
	& \my
	& \mn
	& 18.6
	& 30.7
	& 22.7
	& 20.9
	& 21.6
	& 35.6
	& (1.8)
	& (86.7)
	& 6.5
	& 74.6
	& 14.2
	& 49.7
	& 0.32
	\\ %

	Vis-MVSNet~\cite{Zhang2020}
	& \my
	& \my
	& \mn
	& {9.5}
	& {55.4}
	& 8.9
	& 33.5
	& {10.8}
	& {43.3}
	& (1.8)
	& (87.4)
	& {4.1}
	& {87.2}
	& {7.0}
	& {61.4}
	& 0.70
	\\ %

	PatchmatchNet~\cite{Wang2020}
	& \my
	& \my
	& \mn
	& 10.8
	& 45.8
	& {8.5}
	& {35.3}
	& 19.1
	& 34.8
	& (2.1)
	& (82.8)
	& 4.8
	& 82.9
	& 9.1
	& 56.3
	& 0.28
	\\ %

        MVSFormer++ \scriptsize{\dtu{}+BlendedMVG}~\cite{cao2024mvsformerplus}
	& \my
	& \my
	& \mn
        &  {4.4}
        &  {65.7}
        &  {7.9}
        &  {39.4}
        &  {7.8}
        &  {50.4}
        &  (\bestresult{0.9})
        &  (\bestresult{95.3})
        &  {3.2}
        &  {88.1}
        &  {4.8}
        &  {67.8}
        &  0.78
        \\

        MVSFormer++ \scriptsize{CE our data}~\cite{cao2024mvsformerplus}
	& \my
	& \my
	& \mn
        &  {4.4}
        &  {63.9}
        &  {6.4}
        &  {43.3}
        &  {6.7}
        &  {56.4}
        &  ({1.2})
        &  ({90.7})
        &  {2.6}
        &  {88.3}
        &  {4.3}
        &  {68.5}
        &  0.78
        \\

        MVSFormer++ \scriptsize{regression our data}~\cite{cao2024mvsformerplus}
	& \my
	& \my
	& \mn
        &  \bestresult{3.7}
        &  \bestresult{67.2}
        &  \bestresult{5.7}
        &  \bestresult{45.1}
        &  \bestresult{5.3}
        &  \bestresult{57.5}
        &  ({1.2})
        &  ({90.5})
        &  \bestresult{2.2}
        &  \bestresult{88.6}
        &  \bestresult{3.6}
        &  \bestresult{69.8}
        &  0.78
        \\

        \hline
        \hline  
            \multicolumn{17}{|l|}{\textbf{c) Single-view depth}}
    \\

        Depth Pro~\cite{bochkovskii2024depthpro} $\dagger$
        & \mn
        & \mn
        & med
        &  6.1
        &  39.6
        &  (4.3)
        &  (58.4)
        &  6.1
        &  53.5
        &  5.6
        &  49.6
        &  5.6
        &  57.5
        &  5.6
        &  51.7
        &  5.16
        \\

        Depth Pro~\cite{bochkovskii2024depthpro} $\dagger$
        & \mn
        & \mn
        & \mn
        &  13.6
        &  14.3
        &  9.2
        &  19.7
        &  28.5
        &  8.7
        &  161.8
        &  3.5
        &  38.3
        &  4.4
        &  50.3
        &  10.1
        &  5.16
        \\

        Metric3D~\cite{hu2024metric3d} $\dagger$
        & \mn
        & \mn
        & med
        &  5.1
        &  44.1
        &  \bestresult{2.4}
        &  \bestresult{78.3}
        &  4.4
        &  54.5
        &  10.1
        &  39.5
        &  6.2
        &  48.0
        &  5.6
        &  52.9
        &  0.46
        \\

        Metric3D~\cite{hu2024metric3d} $\dagger$
        & \mn
        & \mn
        & \mn
        &  8.7
        &  13.2
        &  6.2
        &  19.3
        &  12.7
        &  13.0
        &  890.5
        &  1.4
        &  16.7
        &  13.7
        &  187.0
        &  12.1
        &  0.46
        \\
    
        UniDepthV2~\cite{piccinelli2024unidepth} $\dagger$ 
        & \mn
        & \mn
        & med        
        &  \bestresult{4.0}
        &  \bestresult{55.3}
        &  (2.1)
        &  (82.6)
        &  \bestresult{3.7}
        &  \bestresult{66.2}
        &  3.2
        &  72.3
        &  \bestresult{3.6}
        &  \bestresult{68.4}
        &  \bestresult{3.3}
        &  \bestresult{68.9}
        &  0.29
        \\
    
        UniDepthV2~\cite{piccinelli2024unidepth} $\dagger$ 
        & \mn
        & \mn
        & \mn
        &  13.7
        &  4.8
        &  (3.2)
        &  (61.3)
        &  15.4
        &  11.9
        &  964.8
        &  1.3
        &  16.7
        &  12.7
        &  202.7
        &  18.4
        &  0.29
        \\

        UniDepthV1~\cite{piccinelli2024unidepth} $\dagger$ 
        & \mn
        & \mn
        & med 
            &  4.4
    &  51.6
    &  (\bestresult{1.9})
    &  (\bestresult{84.3})
    &  5.4
    &  48.4
    &  9.3
    &  31.8
    &  9.6
    &  38.7
    &  6.1
    &  51.0
    &  0.21
    \\  
        UniDepthV1~\cite{piccinelli2024unidepth} $\dagger$ 
        & \mn
        & \mn
        & \mn
            &  5.2
    &  39.5
    &  (2.7)
    &  (69.4)
    &  48.2
    &  1.8
    &  583.3
    &  1.0
    &  30.7
    &  4.2
    &  134.0
    &  23.2
    &  0.20
        \\

	DepthAnything V2 (ViT-B)~\cite{depth_anything_v2}
	& \mn
	& \mn
	& lstsq $\dagger$
	& 6.6
	& 38.6
	& 4.0
	& 58.6
	& 4.7
	& 56.5
	& \bestresult{2.6}
	& \bestresult{74.7}
	& 4.5
	& 57.5
	& 4.8
	& 54.1
	& \bestresult{0.05}
	\\ %

    \hline
    \hline
    
    \multicolumn{17}{|l|}{\textbf{d) Depth from frames and poses (w/o per-image range)}}
    \\

	Fast-MVSNet~\cite{Yu2020}
	& \my
	& \mn
	& \mn
	& 12.1
	& 37.4
	& 287.1
	& 9.4
	& 131.2
	& 9.6
	& (540.4)
	& (1.9)
	& 33.9
	& 47.2
	& 200.9
	& 21.1
	& 0.35
	\\ %

	MVS2D \scriptsize{\scannet{}}\normalsize{~\cite{Yang2022}}  %
	& \my
	& \mn
	& \mn
	& 73.4
	& 0.0
	& (4.5)
	& (54.1)
	& 30.7
	& 14.4
	& 5.0
	& 57.9
	& 56.4
	& 11.1
	& 34.0
	& 27.5
	& \bestresult{0.05}
	\\ %

	MVS2D \scriptsize{\dtu{}}\normalsize{~\cite{Yang2022}}  %
	& \my
	& \mn
	& \mn
	& 93.3
	& 0.0
	& 51.5
	& 1.6
	& 78.0
	& 0.0
	& (\bestresult{1.6})
	& (92.3)
	& 87.5
	& 0.0
	& 62.4
	& 18.8
	& 0.06
	\\ %

	Robust MVD Baseline \cite{schroeppel2022robust}
	& \my
	& \mn
	& \mn
	& {7.1}
	& 41.9
	& {7.4}
	& {38.4}
	& {9.0}
	& {42.6}
	& {2.7}
	& {82.0}
	& {5.0}
	& {75.1}
	& {6.3}
	& {56.0}
	& 0.06
	\\ %

  	MAST3R (plus our triangulation)
	& \my
	& \mn
	& \mn
	& 3.4
	& 66.6
	& (4.5)
	& (\bestresult{63.0})
	& \bestresult{3.1}
	& \bestresult{72.9}
	& 3.4
	& 67.3
	& (2.4)
	& (83.3)
	& 3.4
	& 70.1
	& 0.72
	\\ %

        \textbf{\modelname} (Ours)%
	& \my
	& \mn
	& \mn
        &  \bestresult{3.2}
        &  \bestresult{68.8}
        &  \bestresult{3.7}
        &  \bestresult{62.9}
        &  {3.2}
        &  68.0
        &  \bestresult{1.3}
        &  \bestresult{95.0}
        &  \bestresult{2.1}
        &  \bestresult{90.5}
        &  \bestresult{2.7}
        &  \bestresult{77.0}
        &  0.12
        \\

    \hline
\end{tabular}

    }
    \vspace{-6pt}
    \caption{
        \textbf{We set a new SOTA in depth estimation on the RMVDB}.
        See \cref{sec:Experiments} for details of the metrics, baselines and groupings.
        A full version of this table, including older baselines, appears in the supplementary.
        Monocular methods with $\dagger$ are given ground truth intrinsics.
        The best result for each section appears in \bestresult{bold}, and (parentheses) indicate results where the evaluation dataset is in the training set.
    }
    \label{tab:robust-multi-view-depth}
    \vspace{-6pt}
\end{table*}

\paragraph{Training data.} 
For MVSA to \textbf{generalize across domains}, we train on a large and diverse set of synthetic datasets, as listed in \cref{tab:training-datasets}.
A subset of these training datasets contain moving objects. %
Our reference image encoder is initialized from Depth Anything V2 (DAV2)~\cite{depth_anything_v2}, which uses a teacher network trained on synthetic datasets similar to ours, and a student network distilled using various real images that do not overlap with our evaluation benchmarks. 
DAV2 was initialized from a pretrained DINOv2~\cite{oquab2024dinov2} network, in turn trained on internet images.

\section{Experiments}\label{sec:Experiments}
We evaluate \modelname on both depth estimation and 3D reconstruction tasks. 
We also implement and report scores for a set of new baselines, which have never before been evaluated on the benchmarks we use.

\subsection{Baselines} 
Where possible, we obtain results directly from prior works ~\cite{schroeppel2022robust,dust3r_cvpr24}.
We also  evaluate and implement other strong baselines that did not previously report performance on diverse MVS benchmarks. 
These include: 
(i) A strong monocular baseline in the form of DAV2 \cite{depth_anything_v2}. To account for the unknown affine transform, we align its predictions to the ground truth using least squares.
(ii) MAST3R~\cite{mast3r_arxiv24} (raw depth estimate) which involves passing the reference and one other source image as input and taking the $z$ component of the point cloud as the depth prediction. 
(iii) MAST3R (plus our triangulation) which is a novel extension of MAST3R so that it can use provided extrinsics and intrinsics, when available.  
For each of the available source images, we use MAST3R descriptors to match points with the reference image. 
We then triangulate points from such matches, rescale the raw depth predictions, and aggregate the point clouds from the different views using a sum weighted by the predicted confidences. 
Note, this method requires one forward pass and thousands of triangulations per source view, significantly reducing its speed.
MAST3R trains on ScanNet~\cite{dai2017scannet} and MegaDepth~\cite{MegaDepthLi18} (which contains a subset of the Tanks and Temples dataset~\cite{Knapitsch2017}).
Baseline implementation details  are in the supplementary.

\noindent{\bf Benchmark.} We evaluate `zero-shot' depth estimation performance on the five multi-view datasets from the RMVDB benchmark~\cite{schroeppel2022robust}, which are  not included in our training data.
It contains the KITTI~\cite{Geiger2012CVPR} ScanNet~\cite{dai2017scannet}, ETH3D~\cite{schoeps2017cvpr}, DTU~\cite{jensen2014large}, and Tanks and Temples~\cite{Knapitsch2017} datasets and represents a diverse set of evaluation scenarios, \eg driving sequences, room scans, building scans, and tabletop objects, among others.
We use the evaluation procedure and source view selection procedure from~\cite{schroeppel2022robust}, allowing  direct comparison to previous  approaches. 

Methods are grouped into four different types (a-d) depending on the information they are provided, \eg if they are given GT cameras, GT depth ranges, \etc.  
\modelname naturally fits into type (d), where all methods are given GT poses, so need to predict depth directly in metric scale and hence do not need any alignment or knowledge of the GT depth range. 
Note, some methods train on the training splits of one, or more, of the benchmark datasets, thus achieving very high scores in those cases. 
We denote these in \cref{tab:robust-multi-view-depth} with a parenthesis around them.

\noindent{\bf Metrics.} We report two commonly used metrics to compare the predicted $\hat{d}$ and GT depth $d$.
The absolute relative depth (rel) is computed per-pixel as $|\hat{d} - d| / d$, while the inlier percentage $\tau$, with threshold $1.03$, is computed per-pixel as $[\max(\hat{d}/d, d/\hat{d}) < 1.03]$, 
where $[ ]$ is the Iverson bracket.
Both metrics are averaged over all valid GT pixels in each test image, before averaging over all images.

\noindent{\bf Results.}
\cref{tab:robust-multi-view-depth} depicts the quantitative results, where we outputperform all baselines across most metrics.
Qualitative results in \cref{fig:qualitative_depths} demonstrate that our \modelname model produces depth maps with superior edge detail and consistent scaling across a variety of scenes, visually outperforming prior methods. 
\modelname also performs well on moving objects, \eg as found in KITTI; see also Fig.~\ref{fig:dynamic}.
Both \textbf{MAST3R triangulated} and the \textbf{Robust MVD Baseline} exhibit poor edge quality, limiting their suitability for applications such as single-image novel view synthesis~\cite{shih20203d}, which requires sharp depth boundaries. 
While Depth Pro produces sharp edges, it frequently displays incorrect depth scaling.
In contrast, our \modelname model combines competitive quantitative performance with sharper edges, making it ideal for tasks that demand both visual and depth accuracy. 
Finally, GT depth-based median and least squares scaling of monocular methods and depth from frames methods (w/o poses) is crucial for good scores, while \modelname consistently predicts high-quality and metric depths.

\begin{table}
    \centering
    {\fontsize{8pt}{10pt}\selectfont

        \setlength{\tabcolsep}{0.8mm}
        \begin{tabular}{|l
        |c >{\columncolor{bgcolor}}c
        |c >{\columncolor{bgcolor}}c
        |}
        
        \hline
            \textbf{Approach}
            & \multicolumn{2}{c|}{\textbf{\scannetshort{}}}
            & \multicolumn{2}{c|}{\textbf{\ethdshort{}}}
            \\

            & $\absrel\downarrow$ & $\threshI\uparrow$
            & $\absrel\downarrow$ & $\threshI\uparrow$
            \\
            \hline
            \hline
            Robust MVD Baseline \cite{schroeppel2022robust}
                & 6.02
                & 47.83
                & 5.75
                & 71.64
                \\
            MAST3R Triangulated
                &  (3.88)
                &  (68.68)
                &  2.37
                &  84.90
                \\

            \textbf{\modelname} (Ours) %
                & \textbf{3.22}
                & \textbf{69.45}
                & \textbf{1.27}
                & \textbf{93.24}
                \\
                \hline
        
        \end{tabular}
    }
    \vspace{-5pt}
    \caption{\textbf{Our variant of \benchmarkname}. We use better test-time tuples for ScanNet, and for ETH3D we use the undistorted test images.
    \label{tab:alternative-benchmark}
    }
    \vspace{-5pt}
\end{table}

\begin{figure}
    \centering
    \newcommand{\masterfigwidth}{0.32\columnwidth}
    \footnotesize
    
    \newcommand{\addArrowToImage}[5]{ %
        \begin{tikzpicture}
            \node[anchor=south west,inner sep=0] (image) at (0,0) {\includegraphics[width=\masterfigwidth]{#1}};
            \begin{scope}[x={(image.south east)},y={(image.north west)}]
                \draw[-latex, ultra thick, white] (#2,#3) -- (#4,#5);
            \end{scope}
        \end{tikzpicture}
    }
    
    \begin{tabular}{
        @{\hskip 0mm}c
        @{\hskip 1mm}c
        @{\hskip 1mm}c
        @{\hskip 1mm}
        }
        $I_r$ (RGB) & \modelname (Ours) & MVSFormer++ \\
        \includegraphics[width=\masterfigwidth]{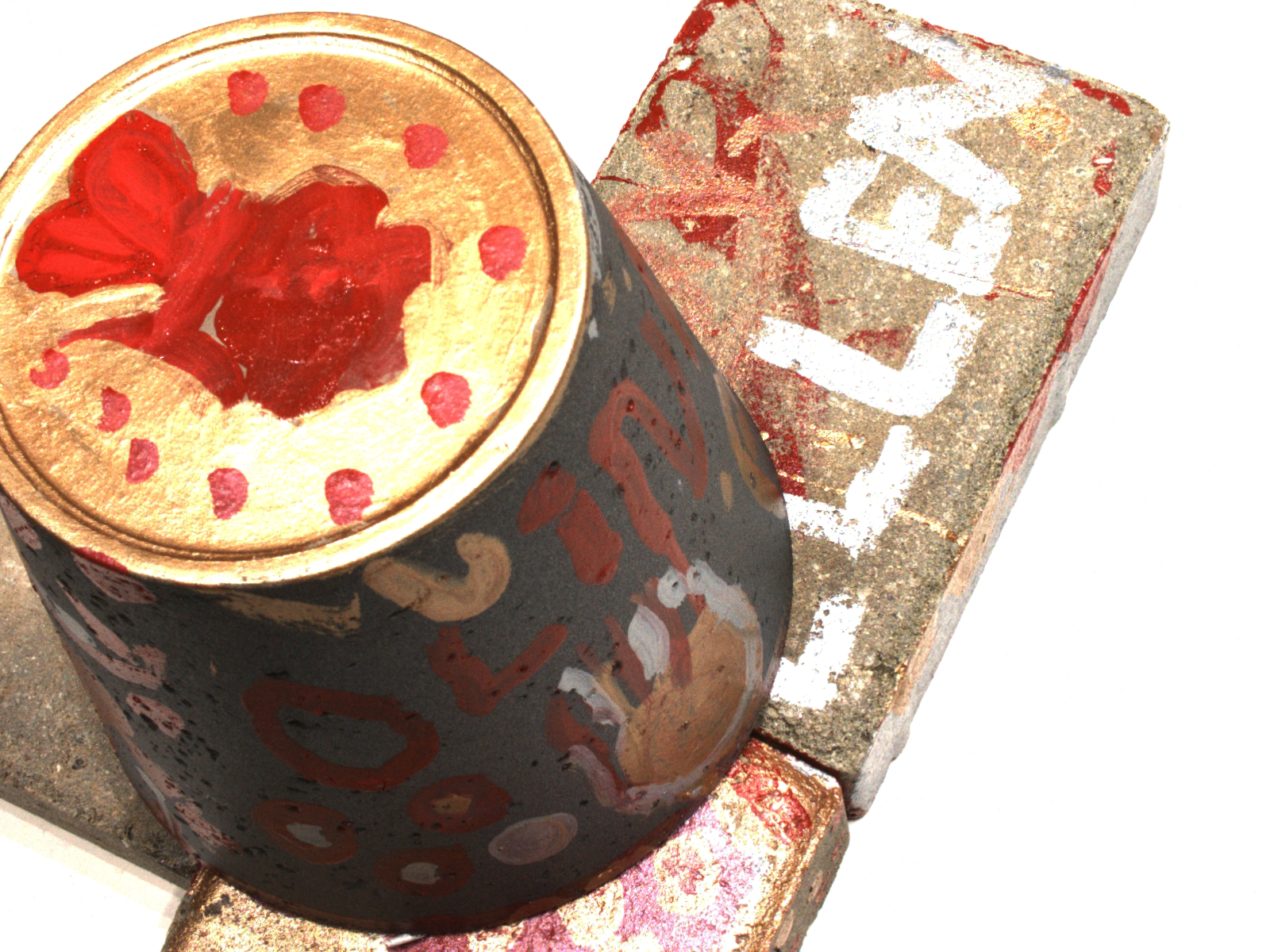} &
        \includegraphics[width=\masterfigwidth]{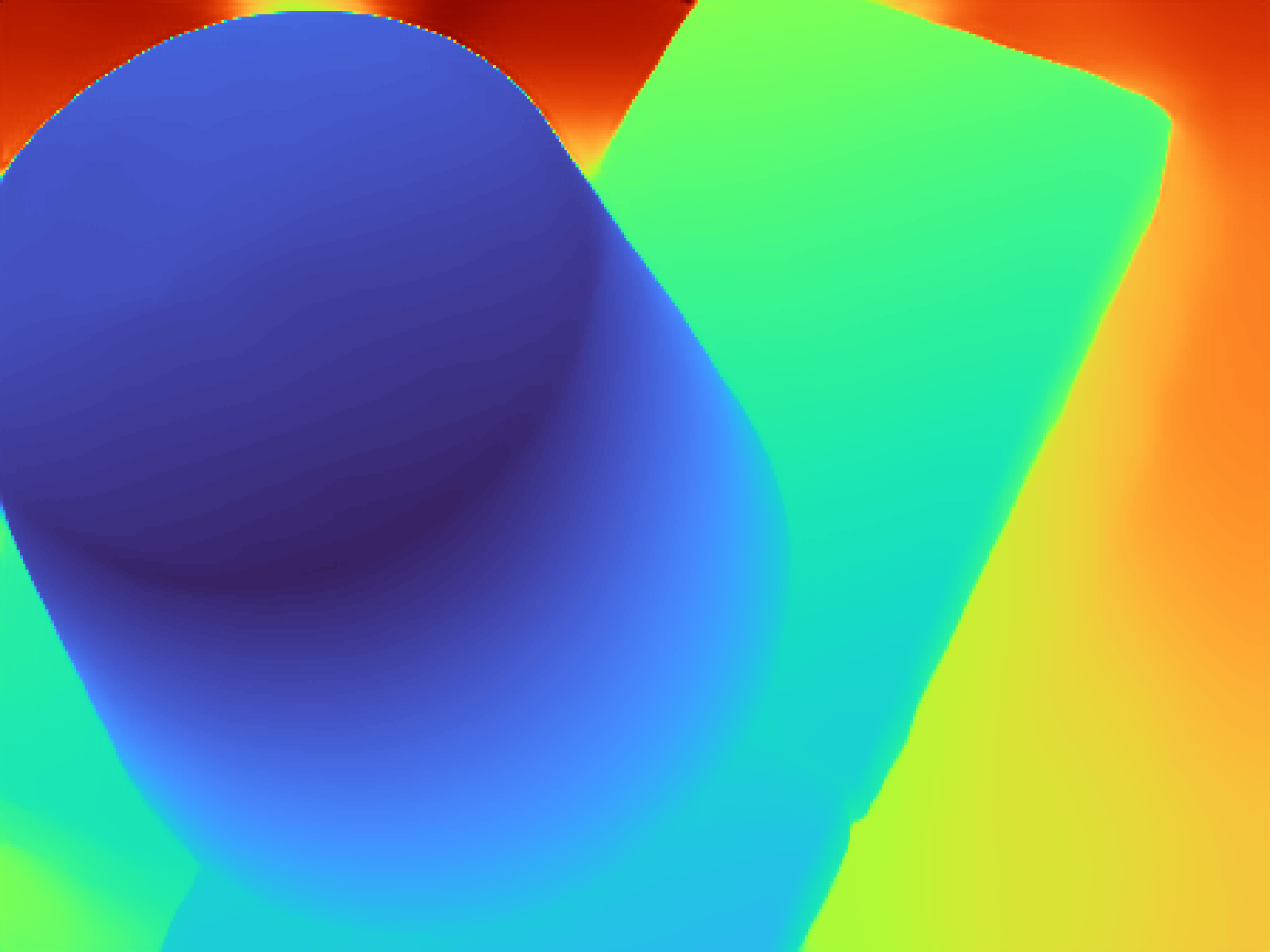} &
        \includegraphics[width=\masterfigwidth]{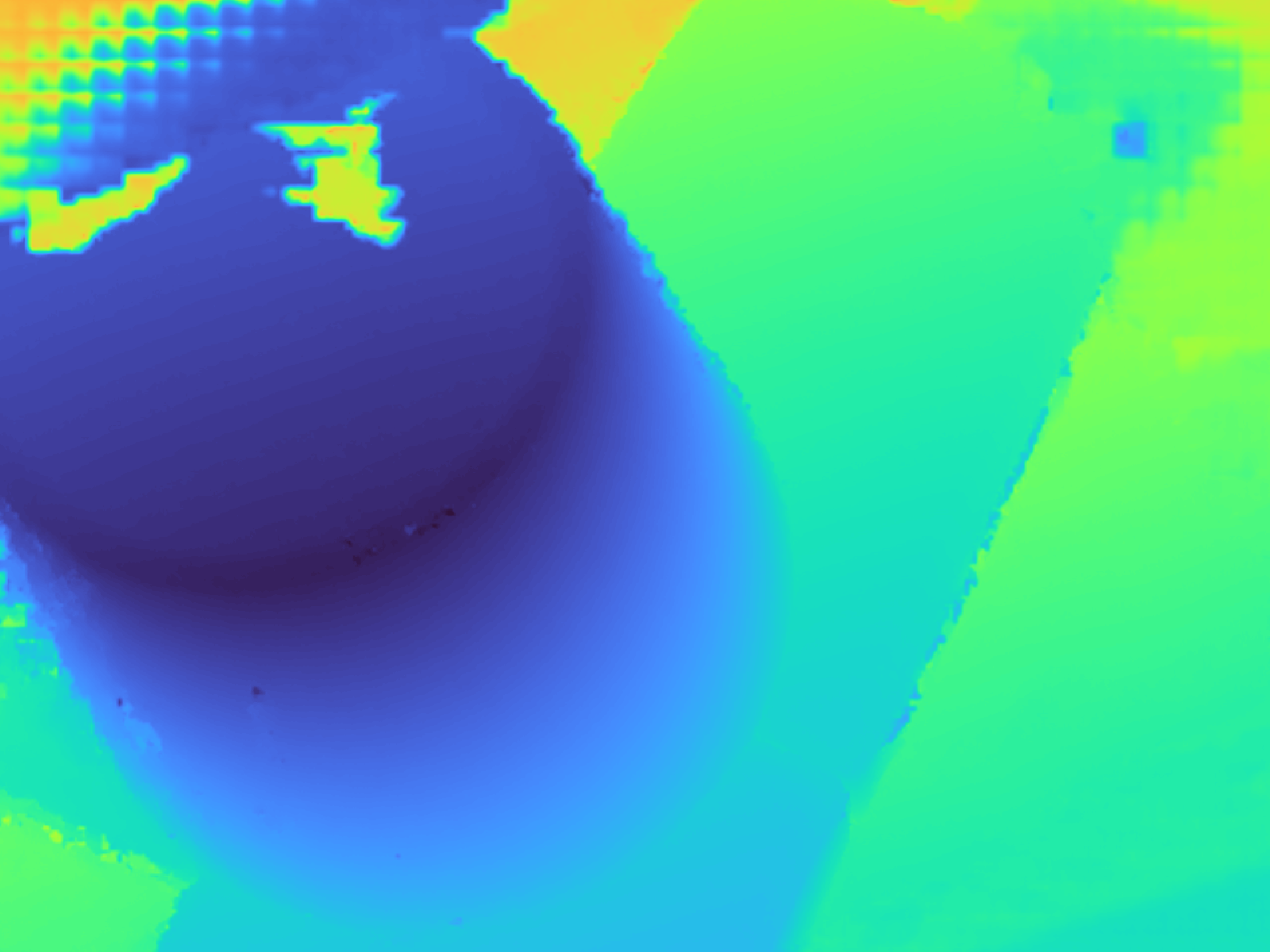} \\
        
    \end{tabular}
    \vspace{-10pt}
    \caption{
        \textbf{Many MVS models fail in areas of poor frame overlap}. Here we show  how MVSFormer++ (right) fails to recover geometry in areas of the image where there are no matching pixels between source and target views (see the top left corner). Our model (middle) handles this situation gracefully. 
    }
    \vspace{-12pt}
\end{figure}
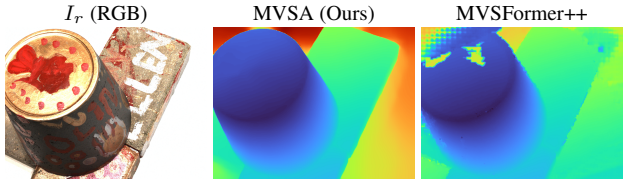

\noindent{\bf Alternative Variant of \benchmarkname.}
We further evaluate some of the leading models on a RMVDB variant, in which we change some conditions to better reflect real-world scenarios.
In this variant, for ScanNet we use keyframes using the strategy of \cite{duzceker2021deepvideomvs}, rather than the temporally sequential keyframes provided by the benchmark.
For ETH3D we undistort both the images and the ground truth using their provided Thin-Prism~\cite{weng1992camera} camera parameters.
Results are shown in \cref{tab:alternative-benchmark}.
On this revised benchmark, we more comprehensively outperform the baselines.

\newcommand{\rowref}[1]{Row \textbf{\ref{#1}}}

\subsection{Ablations Study}
\label{sec:ablations}
In \cref{tab:robust-multi-view-depth-ablation} we validate our design decisions by turning on and off sections of our system. 
We train all ablations at a smaller resolution ($512 \times 384$ input), and without using metadata, for efficiency.
At this resolution, \rowref{row:metadata_low_res} is `ours' and all other rows are ablations relative to this.
\rowref{row:vit_small} replaces our standard ViT-B with the smaller ViT Small, both for the cost volume ViT and the reference image encoder.
The reference image encoder is initialized from Depth Anything v2 (small).
\rowref{row:sr_arch} uses our training data and pipeline, but with the fully-convolutional architecture from SimpleRecon~\cite{sayed2022simplerecon} (without metadata).
\rowref{row:wo_noise} is our system but without adding noise to the ground truth range at training time (Section~\ref{sec:depth_generalization}).
Although this method can excel when the initial range is accurate, it can fail to generalize (see DTU).
\rowref{row:wo_dav2_weights}
is our full architecture but without the pretrained encoder weights from \cite{depth_anything_v2}. Instead we initialize with DINOv2 weights.
\rowref{row:fixed_bins}
is our system without a cascaded cost volume, and instead uses a fixed set of depth bins, losing the ability to refine depth bins and work with arbitrary scales or scene sizes. 
\rowref{row:wo_bin_refinement}
is our full model, but where we take the first depth prediction from the model as our final output, without re-building the cost volume. 
Even though these bins capture the full range of depths in the test datasets~(\cref{fig:range-of-depths}), we see that performance degrades.
\rowref{row:wo_mmcc_vit} uses CNN layers instead of ViT to combine mono/multi features.
\rowref{row:naive_patchify}
uses naive patchification to preprocess the cost volume for input to the mono/multi cue combiner ViT, as outlined in \cref{fig:cost_vol_patchifier}.
These results confirm our design decisions, and full implementation details are in the supplementary.

\paragraph{Robustness to pose rescaling.}
In the supplementary we include results where we scale ScanNet poses $\times 100$, and show our depths are robust to this scaling (\vs ours w/o normalization of the metadata, which performs badly).

\begin{figure}
    \noindent\includegraphics[width=1.0\columnwidth]{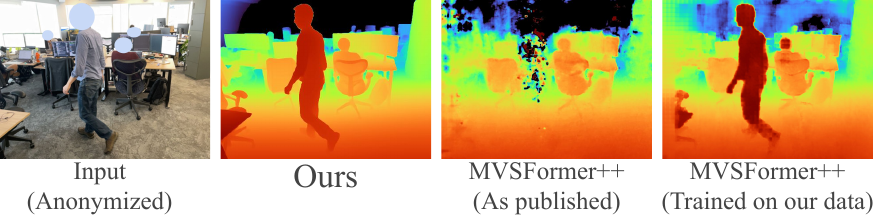}
    \vspace{-18pt}
    \caption{We handle \textbf{dynamic objects} significantly better than traditional MVS \eg MVSFormer++. See supplementary for details of the MVSFormer++ variants shown here.}
    \label{fig:dynamic}
\end{figure}

\begin{figure*}
    \centering
    \newcommand{\turnheightnew}{60pt}

    \newcommand{\addArrowToKITTIImage}[5]{ %
        \begin{tikzpicture}
            \node[anchor=south west,inner sep=0] (image) at (0,0) {\includegraphics[height=\turnheightnew, trim=6cm 0cm 5cm 0cm, clip]{#1}};
            \begin{scope}[x={(image.south east)},y={(image.north west)}]
                \draw[-latex, ultra thick, white] (#2,#3) -- (#4,#5);
            \end{scope}
        \end{tikzpicture}}

    \newcommand{\addArrowToImage}[5]{ %
        \begin{tikzpicture}
            \node[anchor=south west,inner sep=0] (image) at (0,0) {\includegraphics[height=\turnheightnew]{#1}};
            \begin{scope}[x={(image.south east)},y={(image.north west)}]
                \draw[-latex, ultra thick, white] (#2,#3) -- (#4,#5);
            \end{scope}
        \end{tikzpicture}}
    
    \resizebox{0.98\textwidth}{!}{
    \begin{tabular}{@{\hskip -2mm}c@{\hskip 1mm}c@{\hskip 1mm}c@{\hskip 1mm}c@{\hskip 1mm}c@{\hskip 1mm}c@{\hskip 1mm}c@{}}
        & KITTI & ScanNet & ETH3D & DTU & {T\&T} \\
    
        {\rotatebox{90}{\hspace{5mm}{\small RGB ($I_r$)}}} &
        \includegraphics[height=\turnheightnew, trim=6cm 0cm 5cm 0cm, clip]{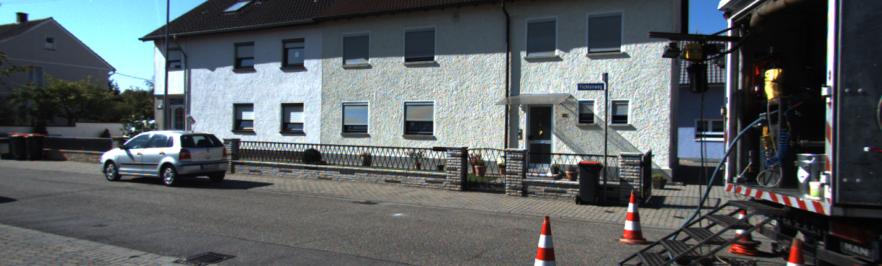} &
        \includegraphics[height=\turnheightnew]{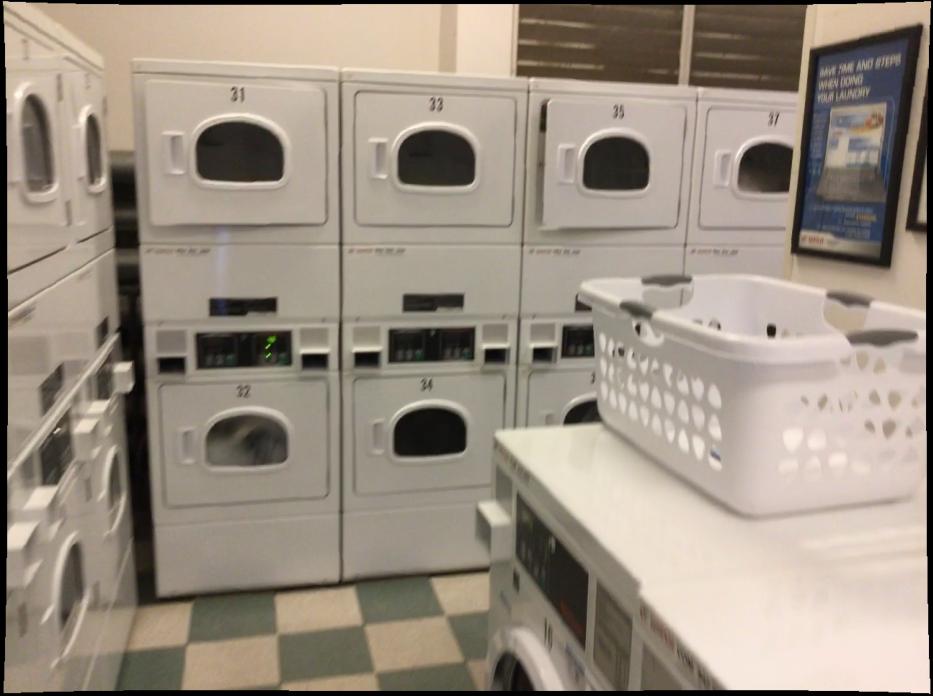} &
        \includegraphics[height=\turnheightnew]{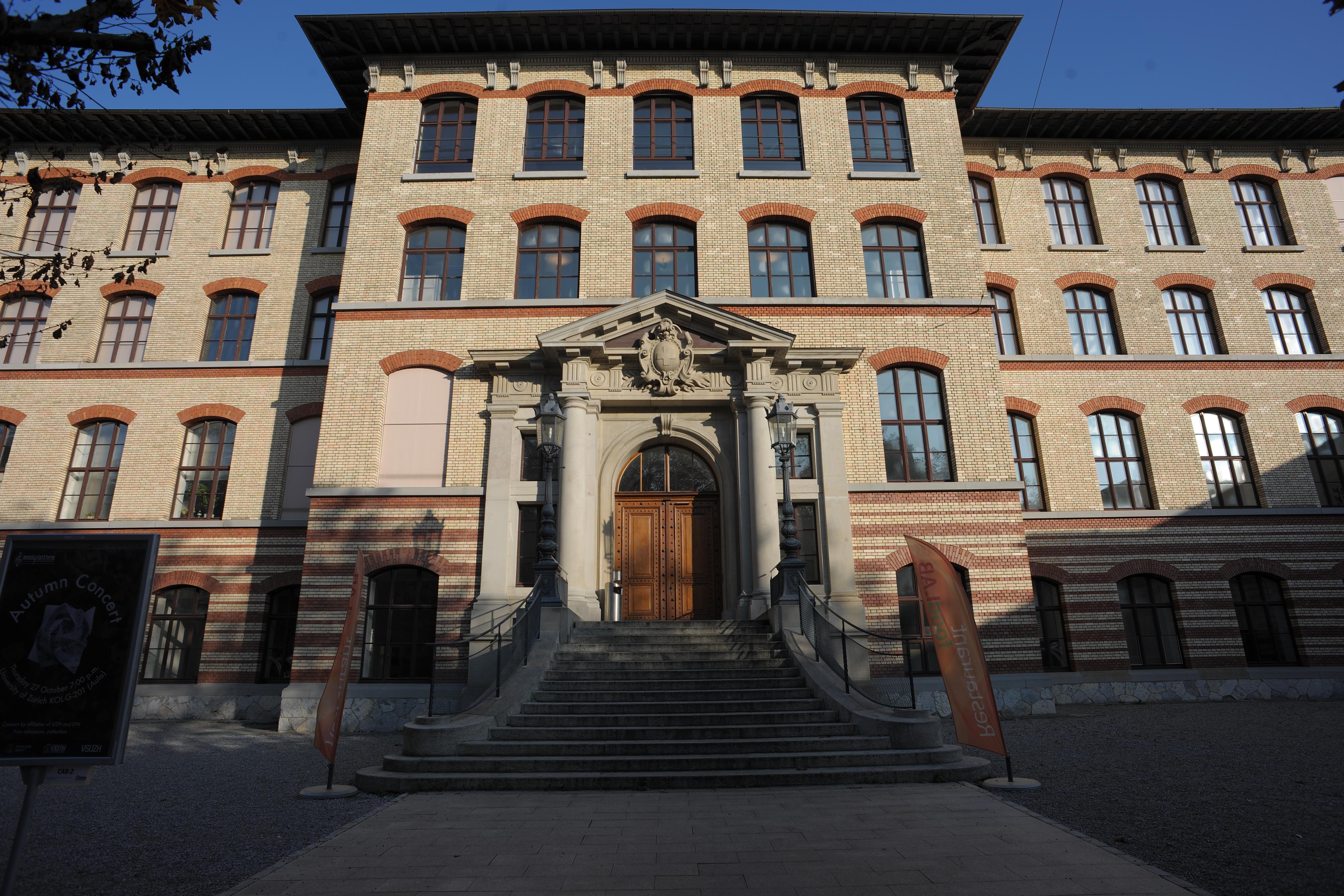} &
        \includegraphics[height=\turnheightnew]{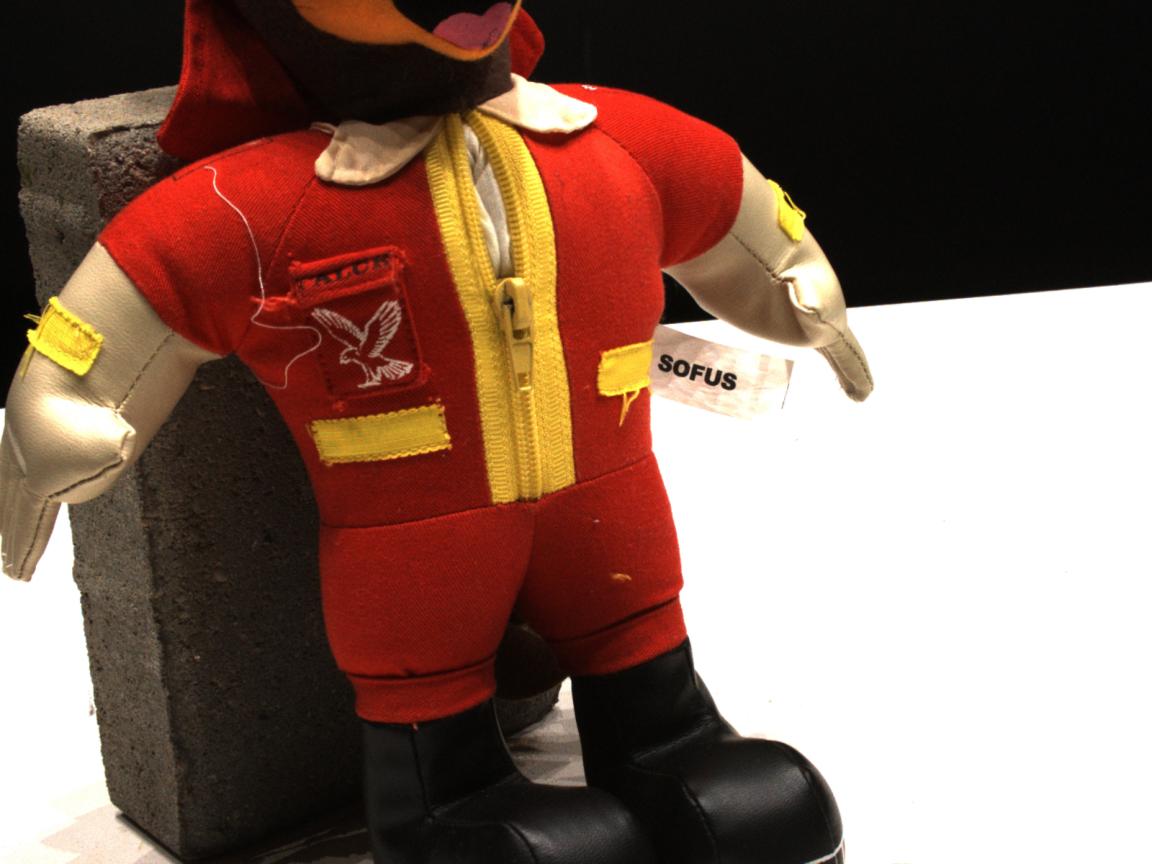} &
        \includegraphics[height=\turnheightnew]{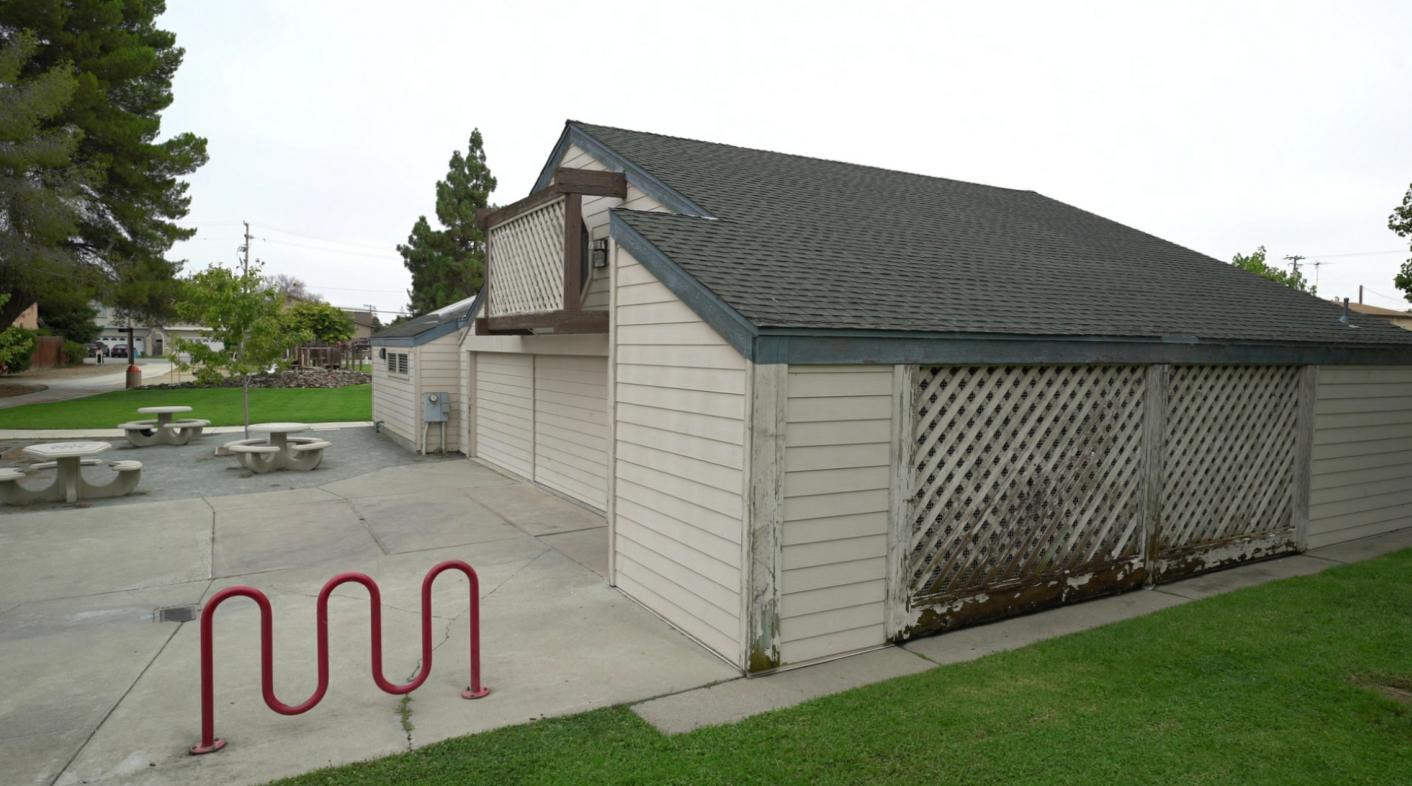}\\
        
        {\rotatebox{90}{\hspace{2mm}{\small Depth Pro~\protect\cite{bochkovskii2024depthpro}}}} &
        \includegraphics[height=\turnheightnew, trim=6cm 0cm 5cm 0cm, clip]{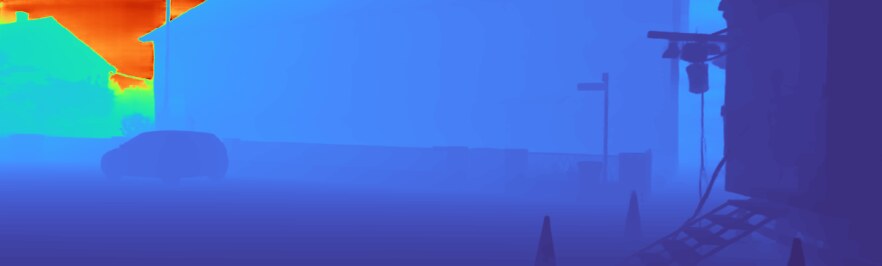} &
        \includegraphics[height=\turnheightnew]{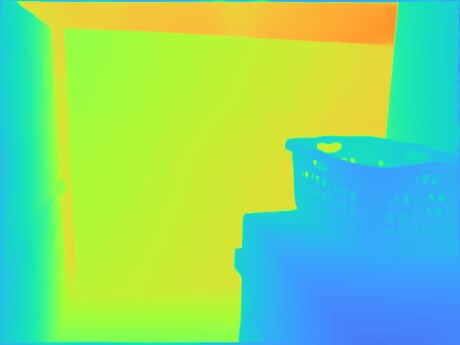} &
        \includegraphics[height=\turnheightnew]{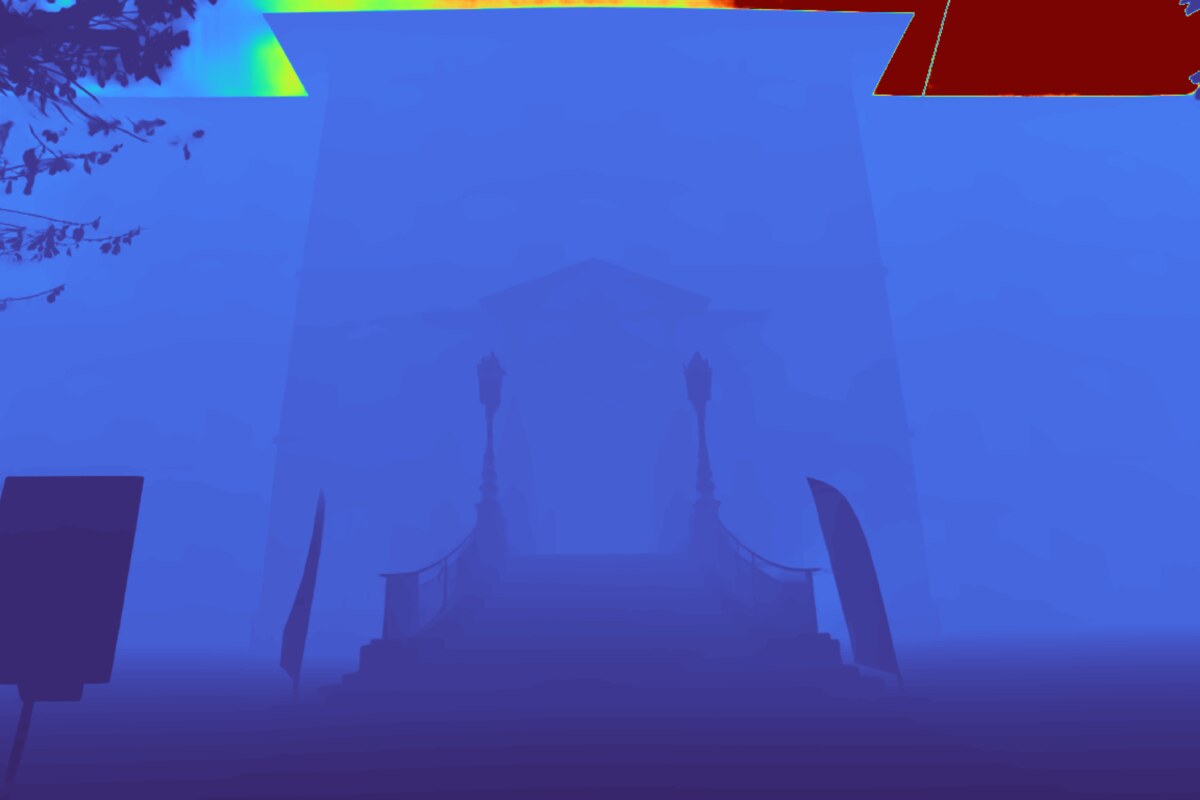} &
        \includegraphics[height=\turnheightnew]{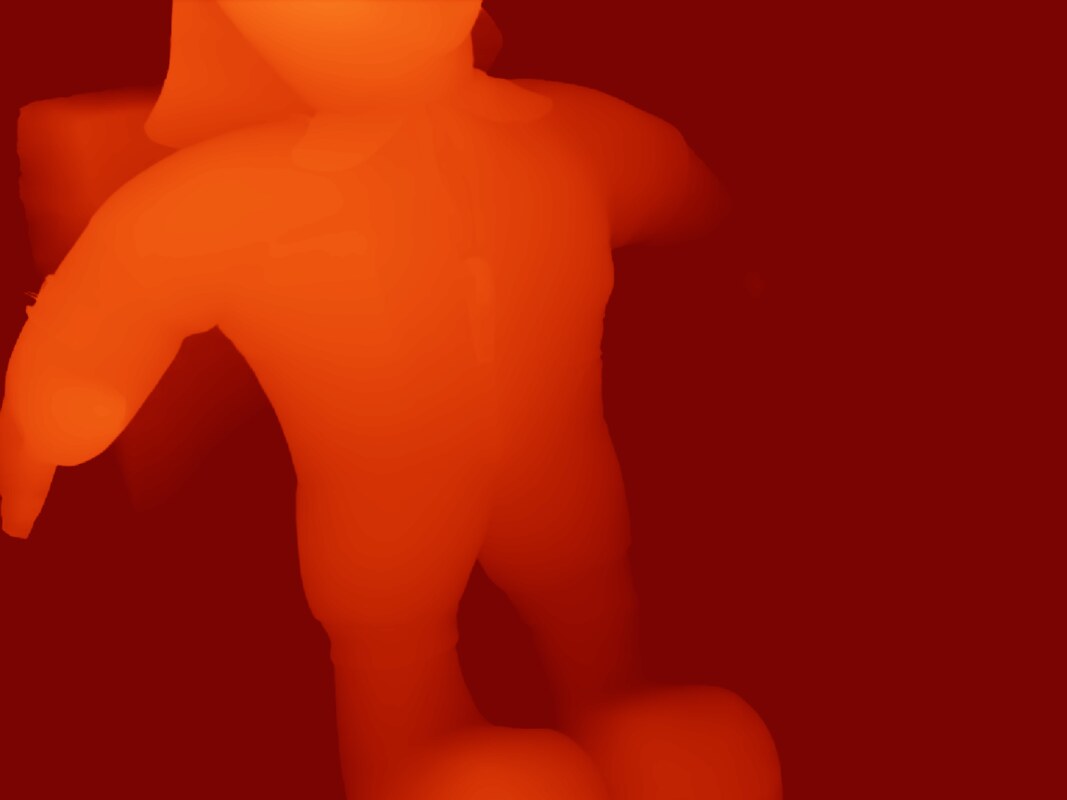} &
        \includegraphics[height=\turnheightnew]{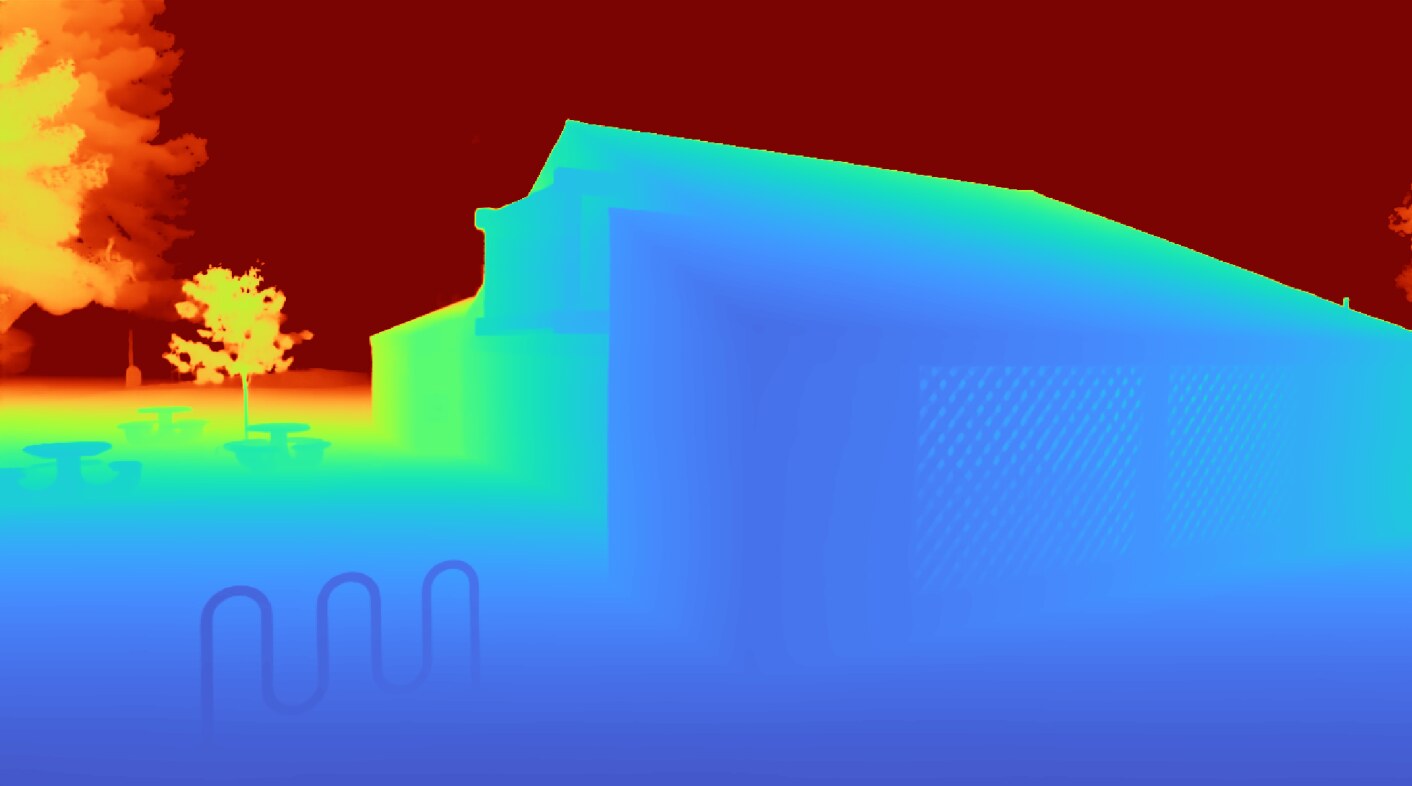} \\
        
        {\rotatebox{90}{\hspace{4mm}{\small rMVD~\protect\cite{schroeppel2022robust}}}} &
        \includegraphics[height=\turnheightnew, trim=6cm 0cm 5cm 0cm, clip]{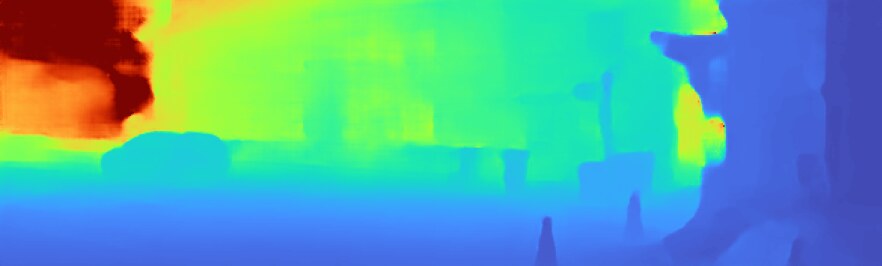} &
        \includegraphics[height=\turnheightnew]{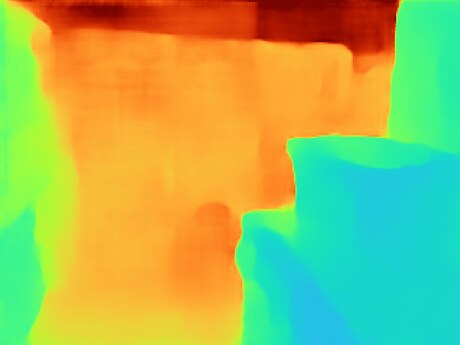} &
        \includegraphics[height=\turnheightnew]{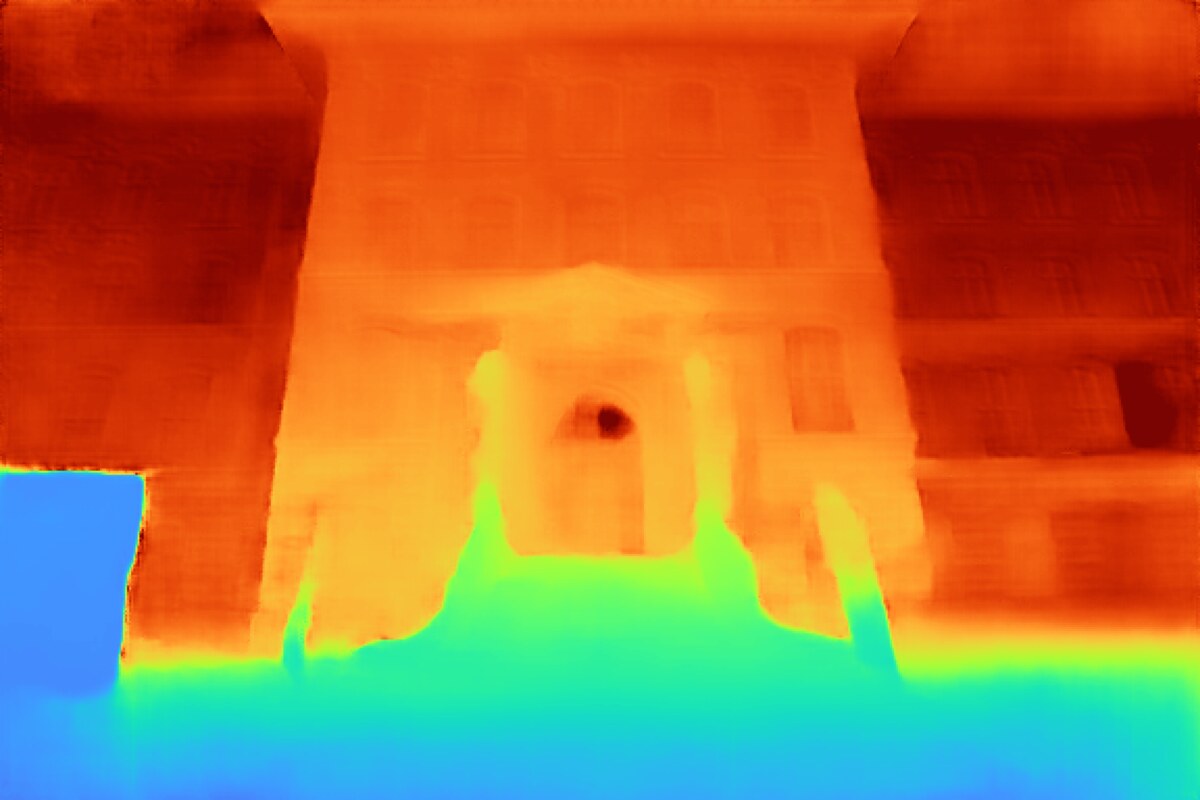} &
        \includegraphics[height=\turnheightnew]{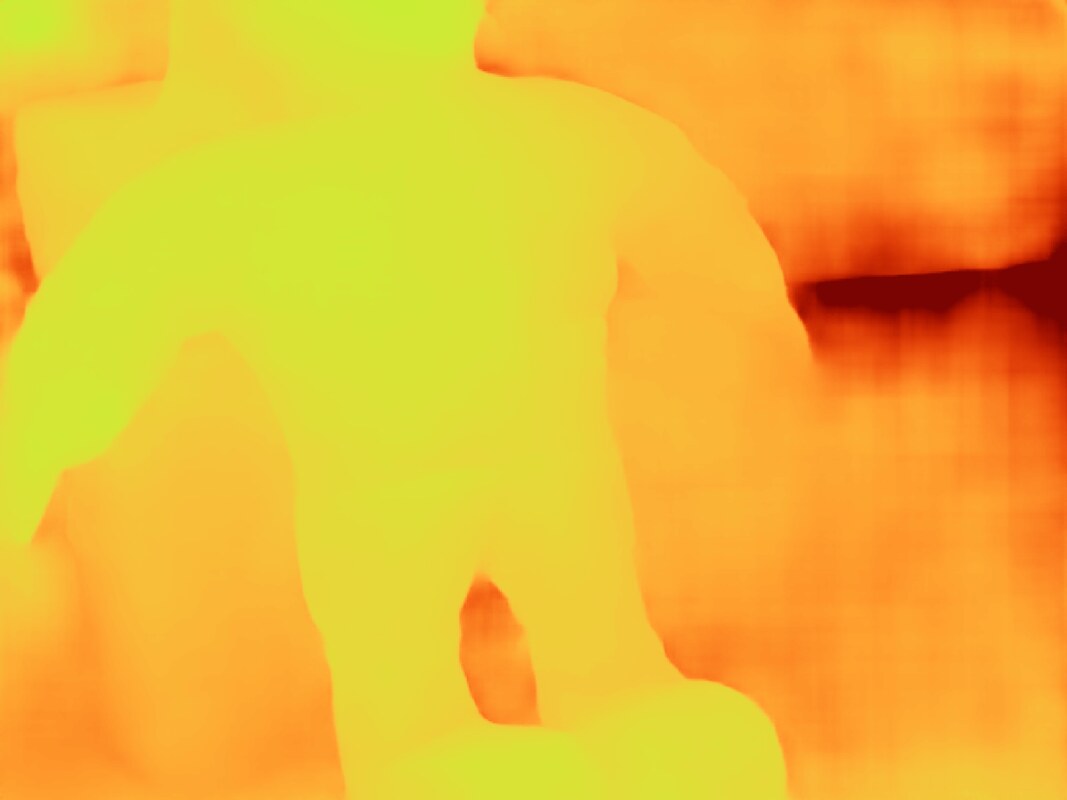} &
        \includegraphics[height=\turnheightnew]{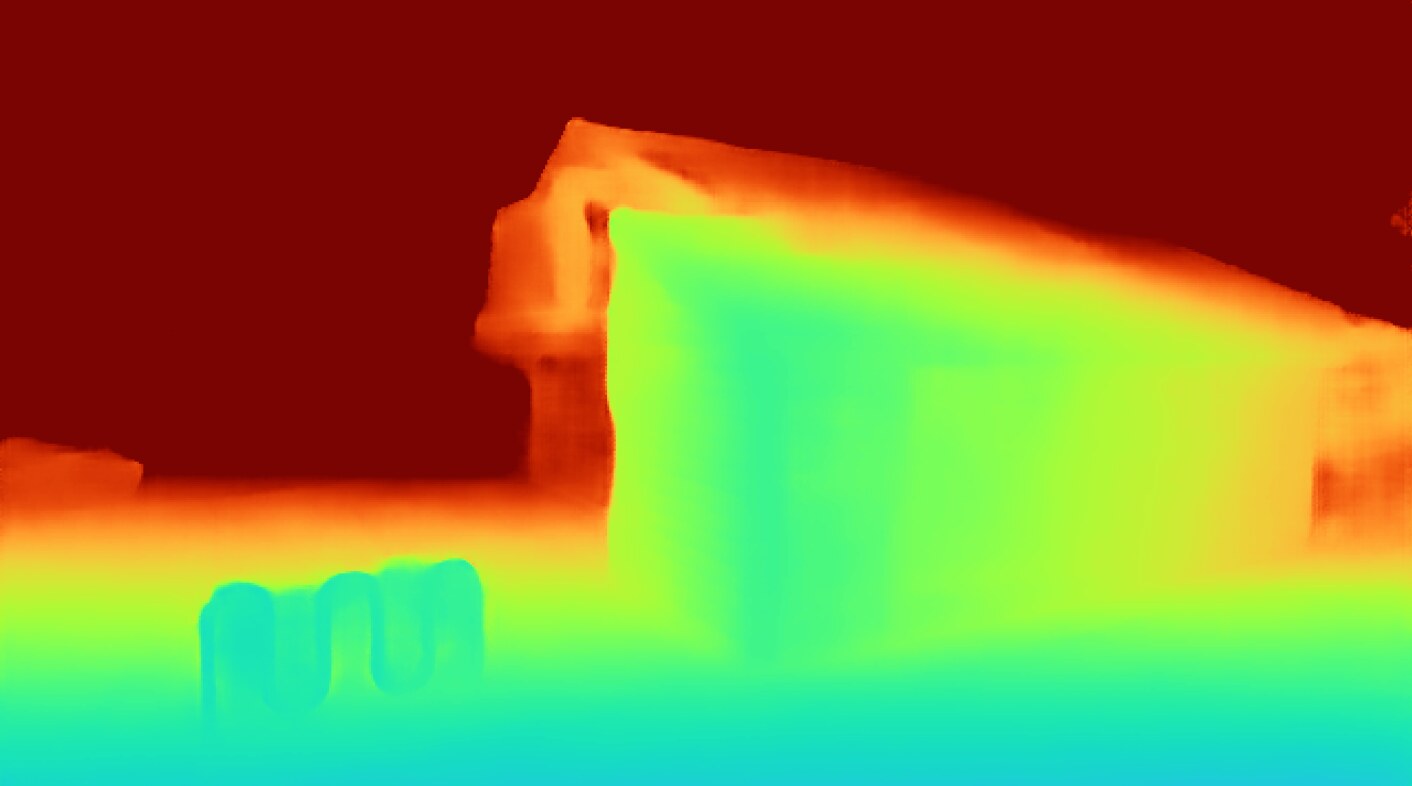} \\

        {\rotatebox{90}{\hspace{1mm}{\small MAST3R~\protect\cite{mast3r_arxiv24}}}} &
        \includegraphics[height=\turnheightnew, trim=6cm 0cm 5cm 0cm, clip]{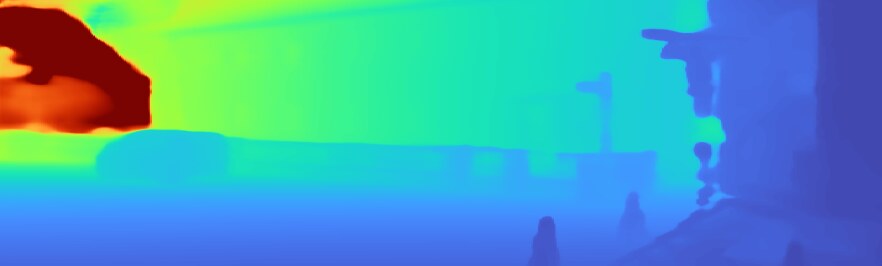} &
        \includegraphics[height=\turnheightnew]{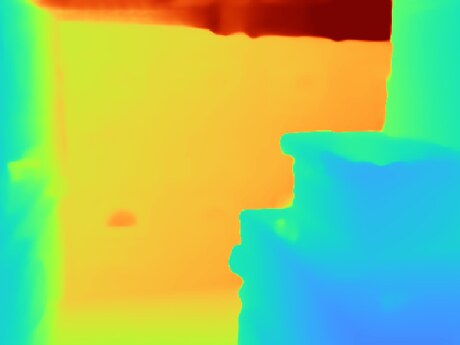} &
        \includegraphics[height=\turnheightnew]{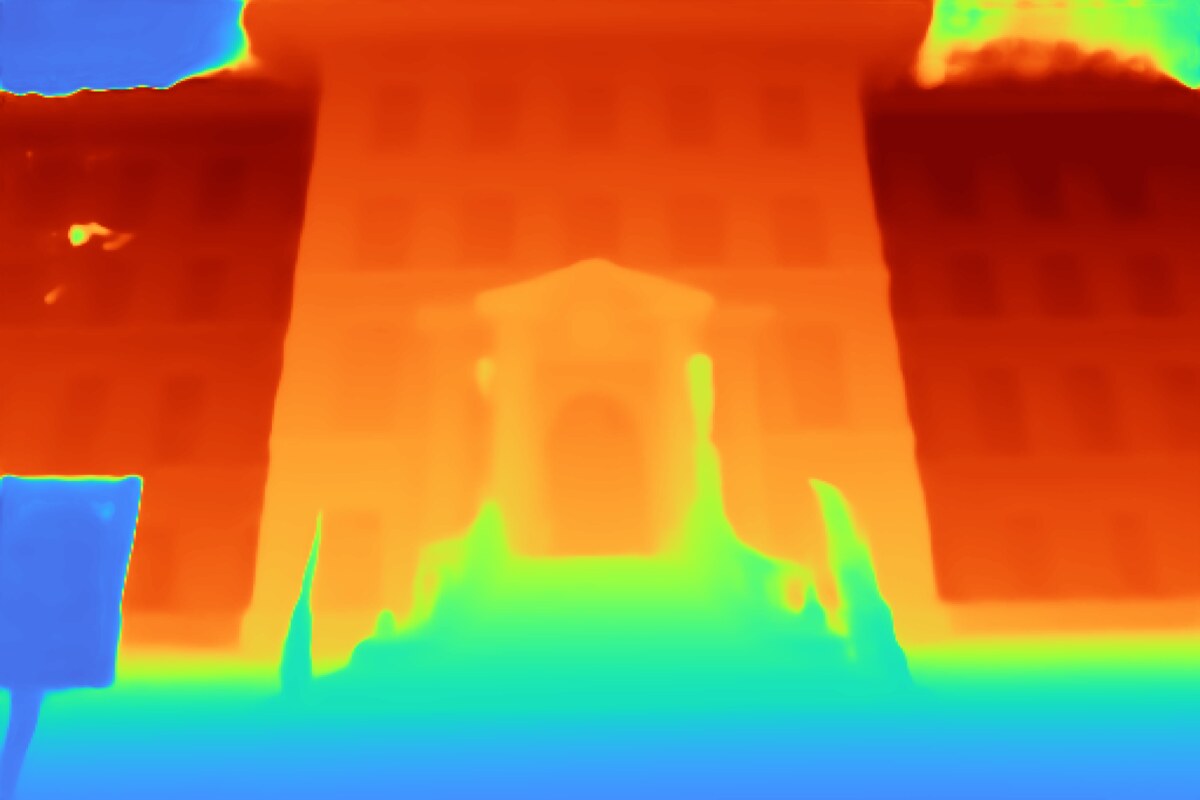} &
        \includegraphics[height=\turnheightnew]{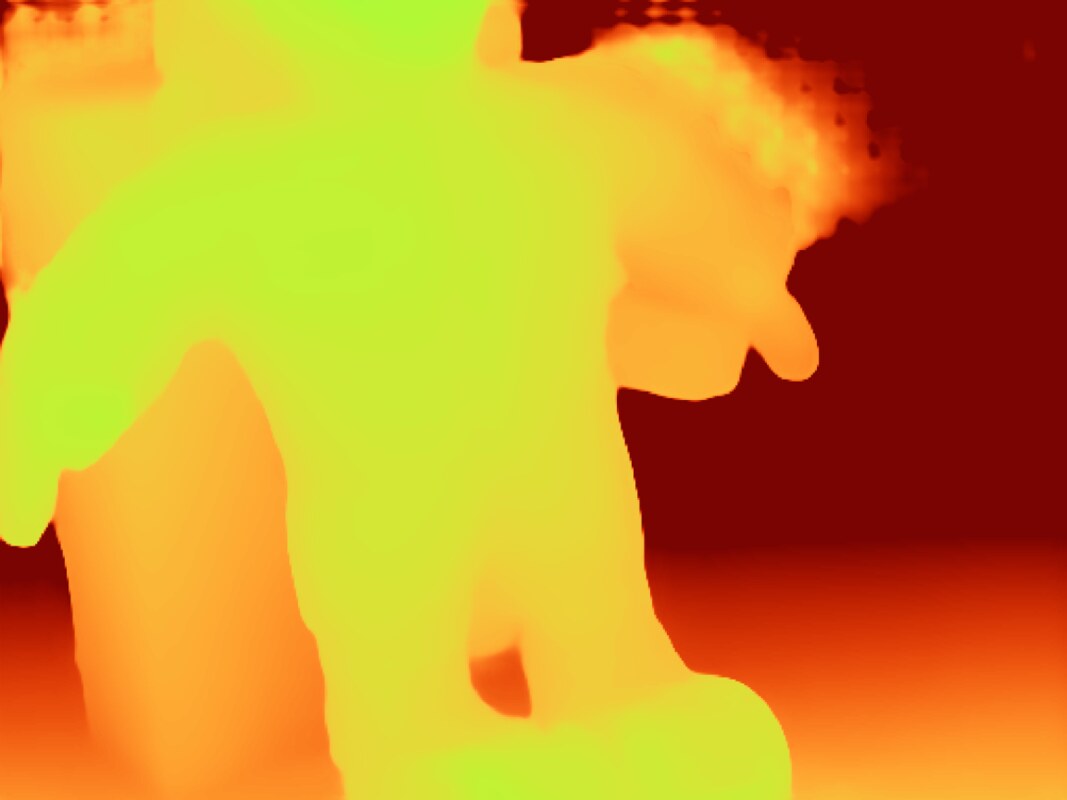} &
        \includegraphics[height=\turnheightnew]{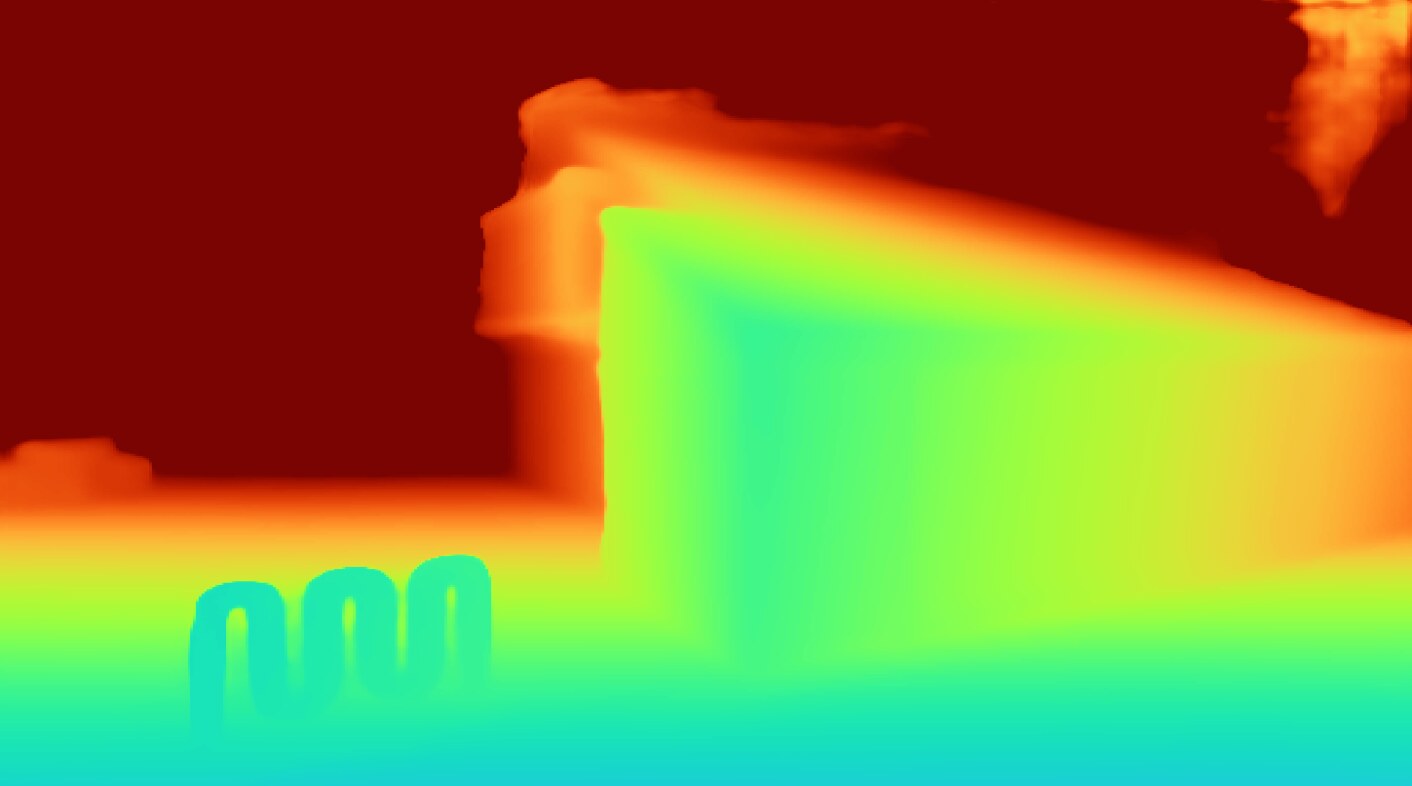} \\
        
        {\rotatebox{90}{\hspace{1.7mm}{\small MVSA (Ours)}}} &
        \addArrowToKITTIImage{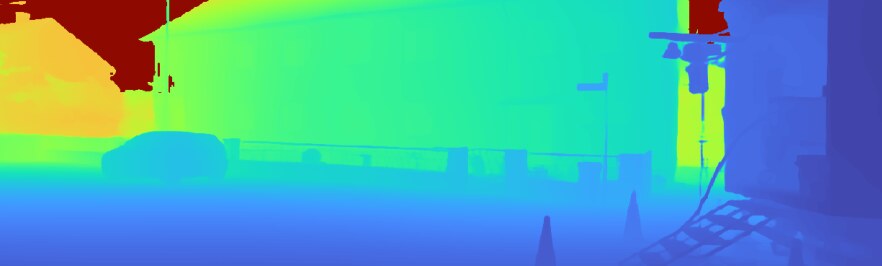}{0.7}{0.2}{0.85}{0.1} &
        
        \addArrowToImage{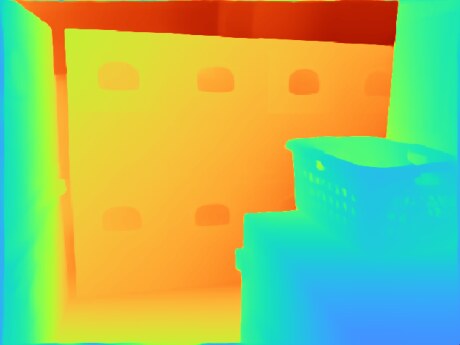}{0.4}{0.7}{0.62}{0.61} &
        
        \addArrowToImage{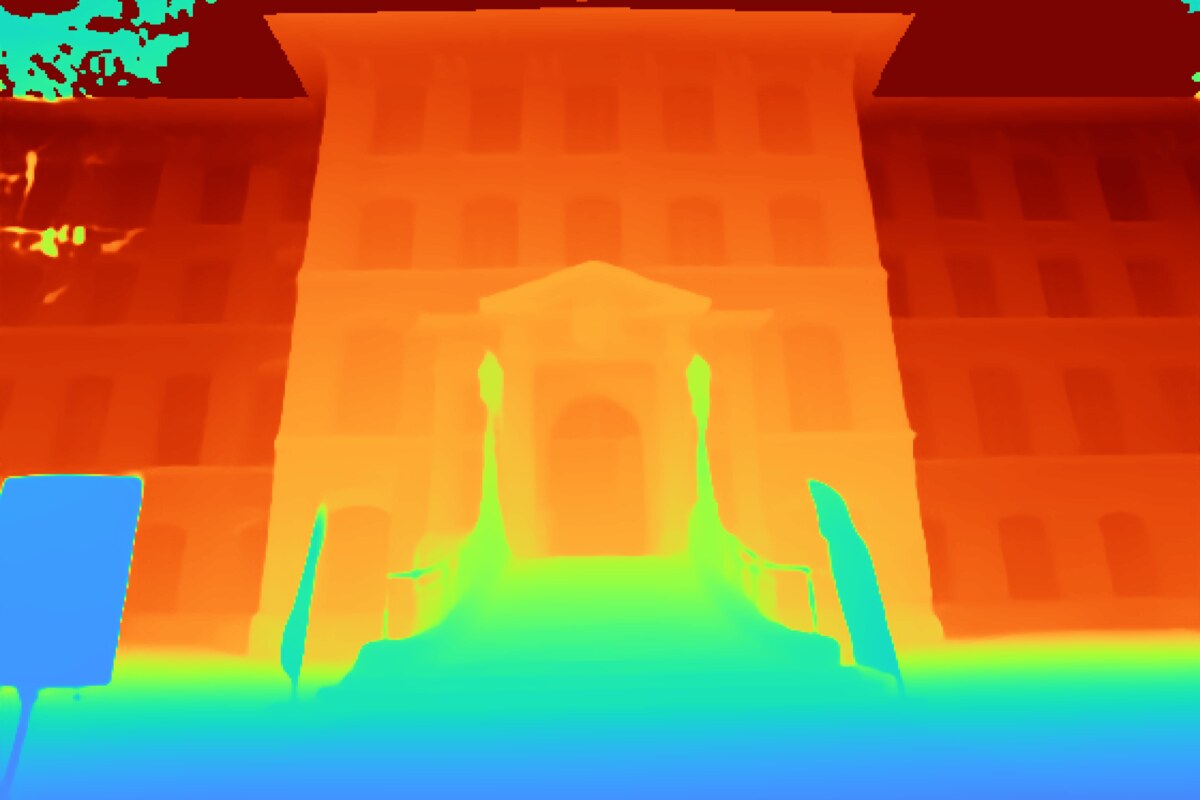}{0.77}{0.7}{0.6}{0.55} &
        
        \addArrowToImage{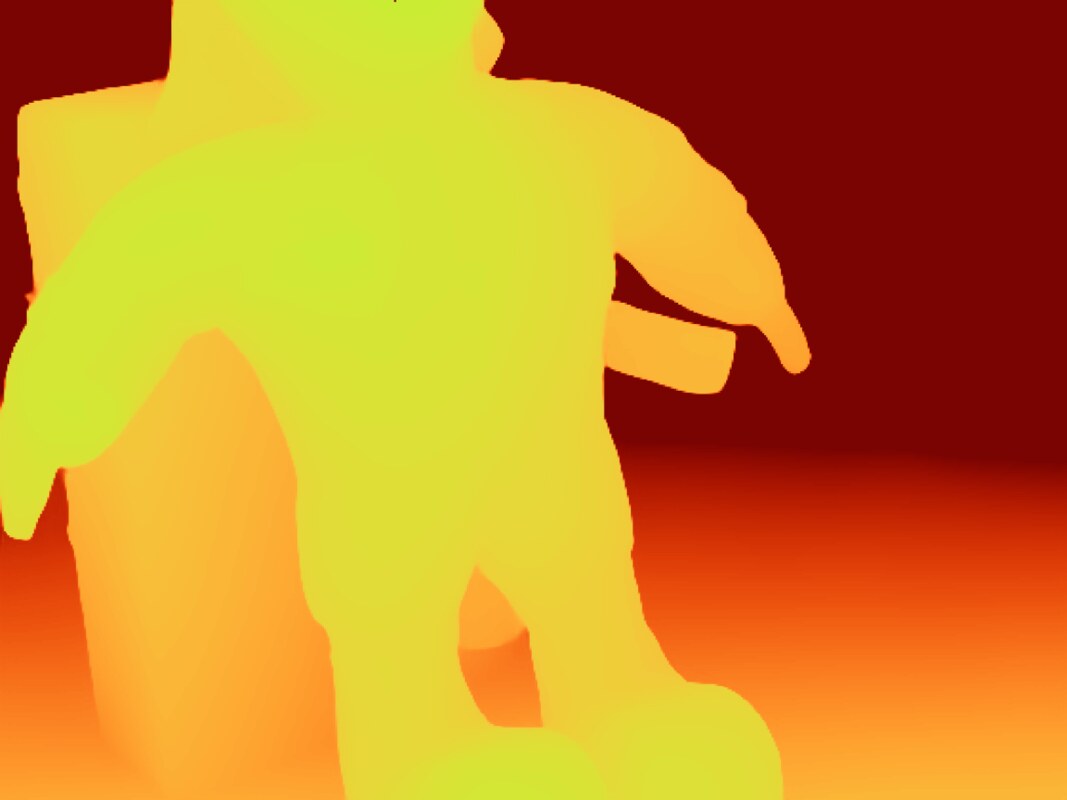}{0.9}{0.3}{0.7}{0.5} & 
        
        \addArrowToImage{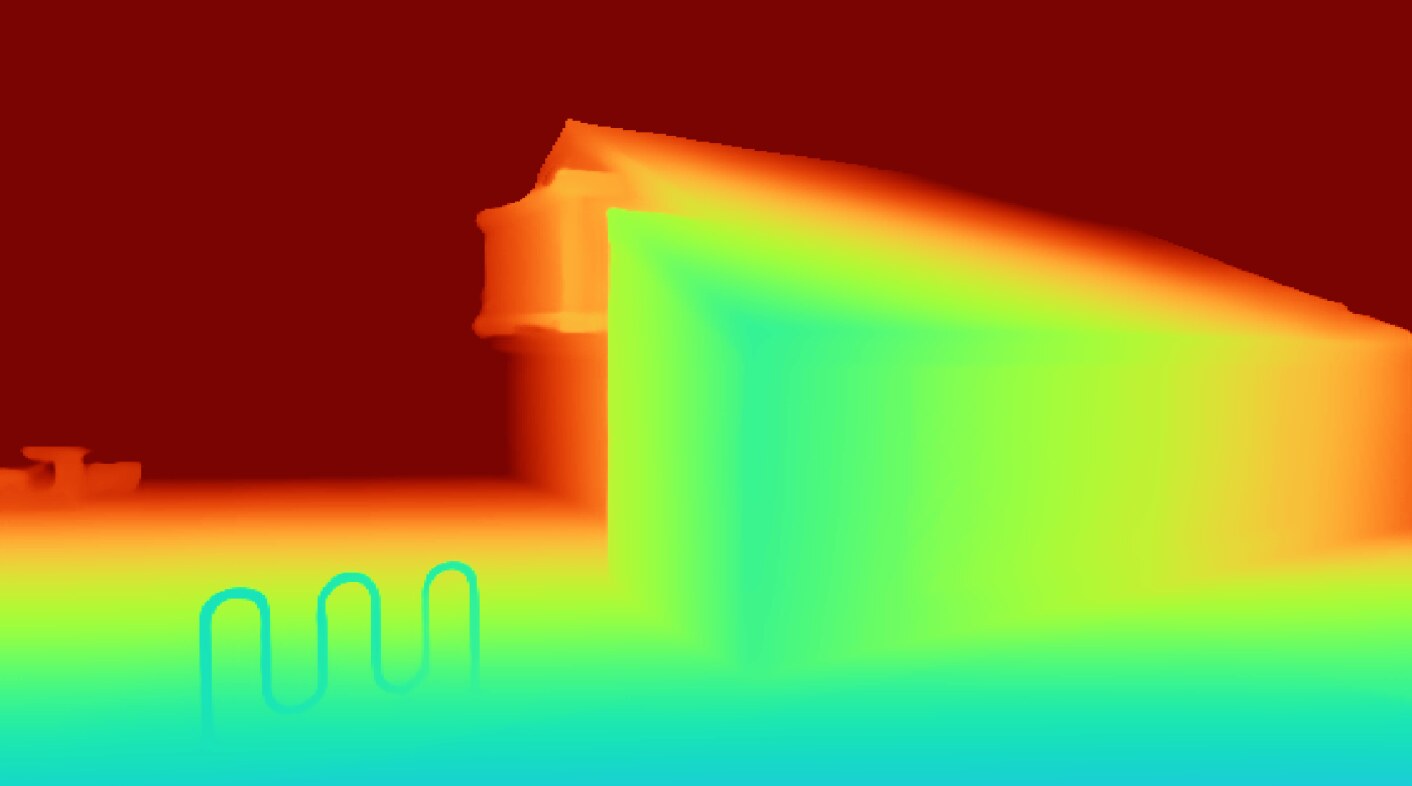}{0.2}{0.65}{0.25}{0.3} \\

        {\rotatebox{90}{\hspace{7mm}{\small GT}}} &
        \includegraphics[height=\turnheightnew, trim=6cm 0cm 5cm 0cm, clip]{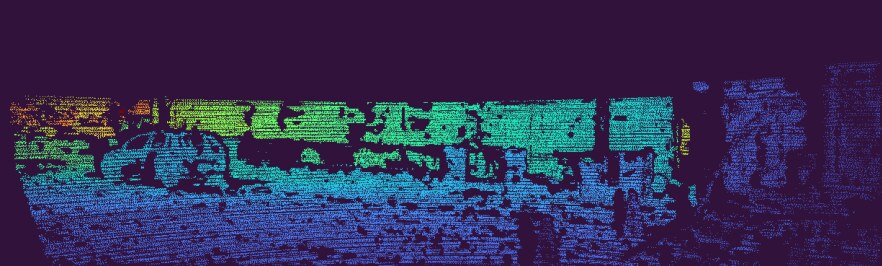} &
        \includegraphics[height=\turnheightnew]{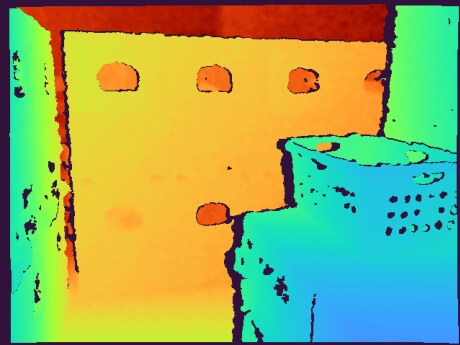} &
        \includegraphics[height=\turnheightnew]{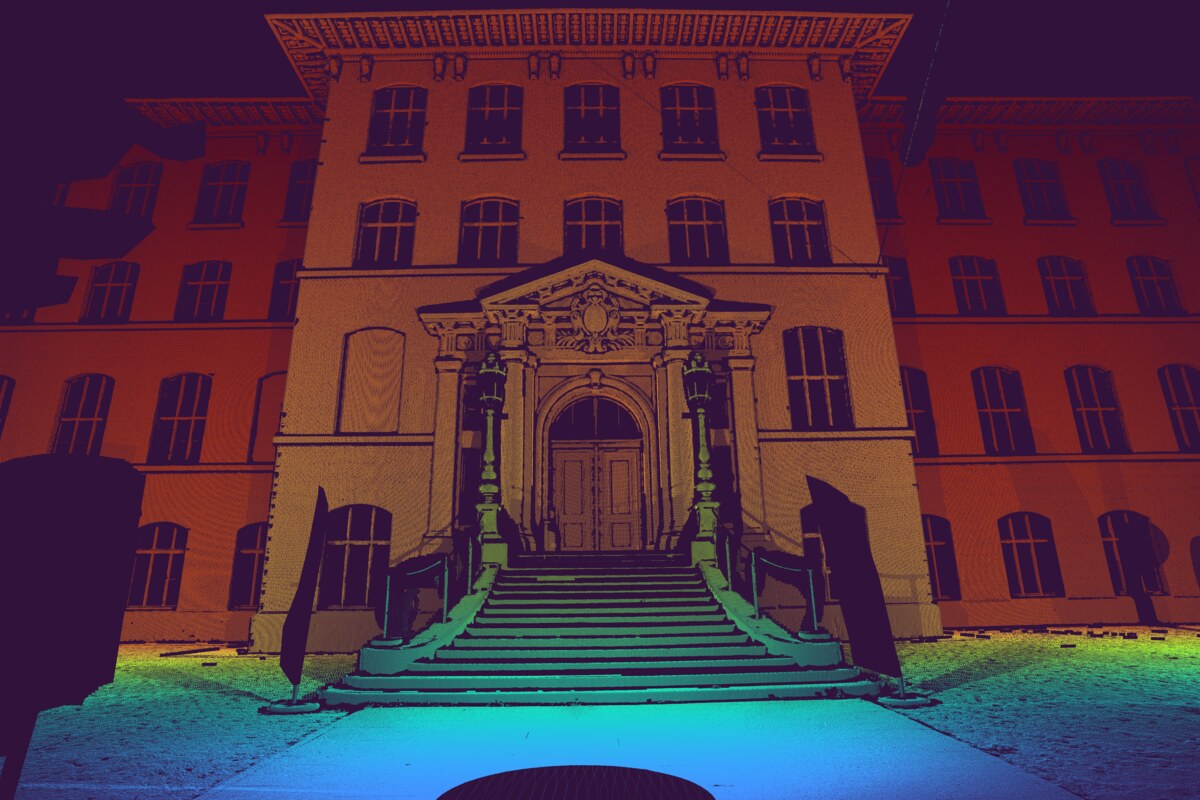} &
        \includegraphics[height=\turnheightnew]{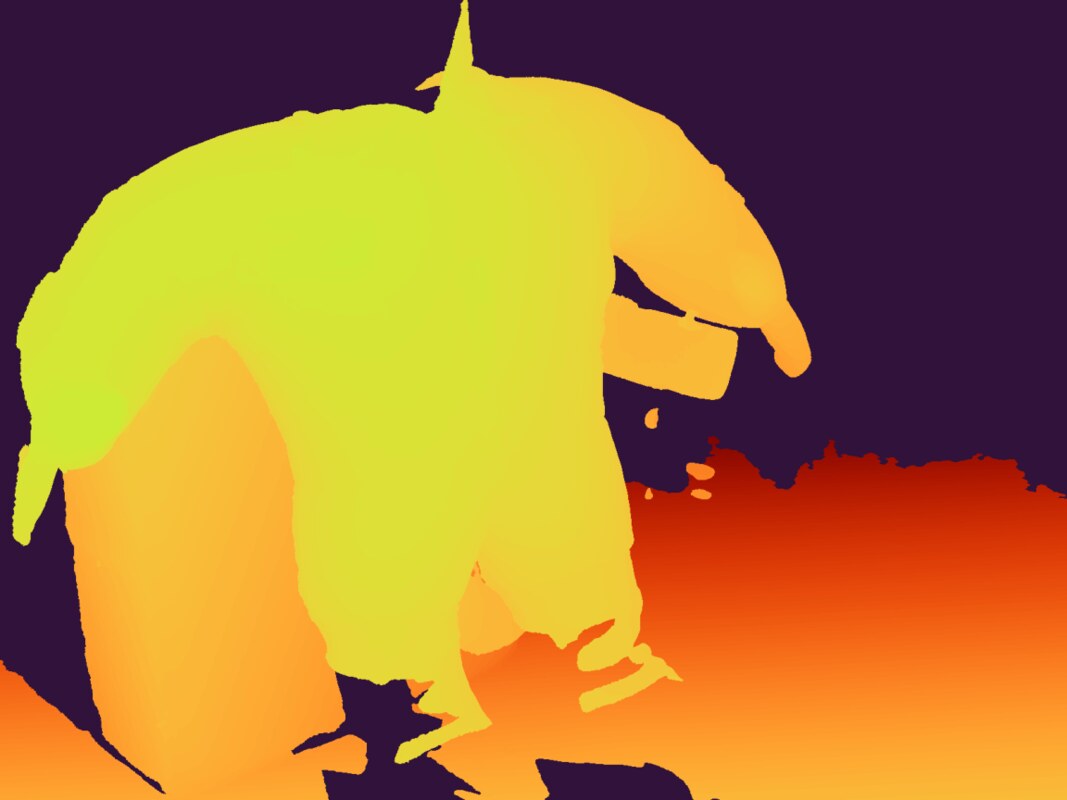} &
        \includegraphics[height=\turnheightnew]{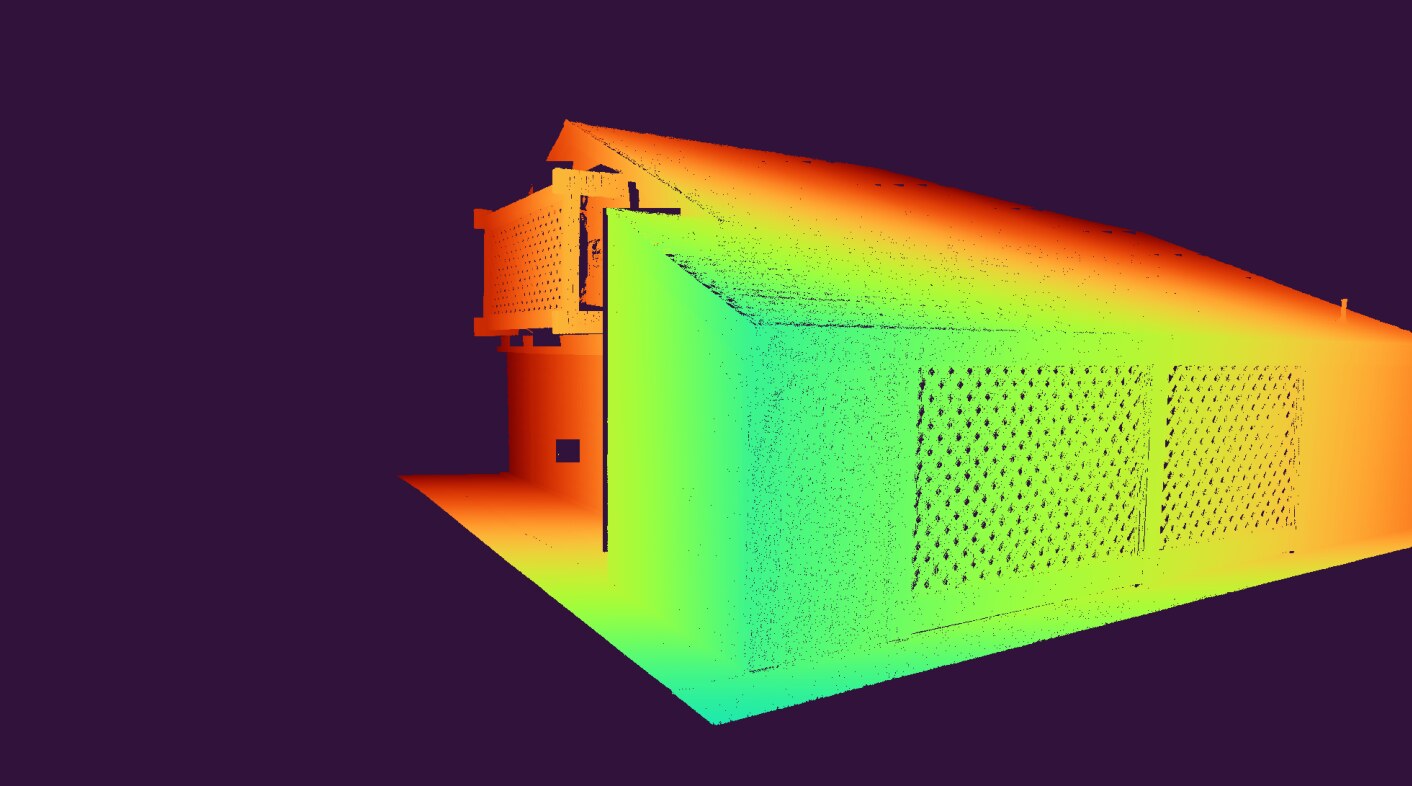}  \\
    \end{tabular}
    }

    \vspace{-10pt}
    \caption{
        \textbf{Qualitative comparison of depth prediction results across multiple datasets} (KITTI, ScanNet, ETH3D, DTU, and Tanks \& Temples). 
        Rows show different methods: Depth Pro~\cite{schroeppel2022robust}, rMVD baseline~\cite{schroeppel2022robust}, MAST3R (Triangulated)~\cite{mast3r_arxiv24}, and our MVSA model, along with RGB inputs ($I_r$) and ground-truth depths (GT). 
        Depth Pro provides sharp edges but often misestimates depth scale, while our MVSA model captures finer details than MAST3R and rMVD. 
        Depth maps are normalized to ground truth depth range for consistent visualization; see the supplementary material for unnormalized results, especially highlighting scale discrepancies in Depth Pro. 
    }
    \label{fig:qualitative_depths}
    \vspace{-12pt}
\end{figure*}

\begin{table}[t!]
    \centering
    \resizebox{1.0\columnwidth}{!}{
        \hspace{-10pt}
        \setlength{\tabcolsep}{0.8mm}
        
\newcommand{\gold}[1]{\textbf{#1}}
\newcommand{\silver}[1]{\underline{#1}}

\begin{tabular}{|ll
|c >{\columncolor{bgcolor}}c
|c >{\columncolor{bgcolor}}c
|c >{\columncolor{bgcolor}}c
|c >{\columncolor{bgcolor}}c
|c >{\columncolor{bgcolor}}c
|c >{\columncolor{bgcolor}}c|
}

\hline
    &
    & \multicolumn{2}{c|}{\textbf{\kittishort{}}}
    & \multicolumn{2}{c|}{\textbf{\scannetshort{}}}
    & \multicolumn{2}{c|}{\textbf{\ethdshort{}}}
    & \multicolumn{2}{c|}{\textbf{\dtushort{}}}
    & \multicolumn{2}{c|}{\textbf{\tanksandtemplesshort{}}}
    & \multicolumn{2}{c|}{\textbf{Average}}
    \\

    &
    & $\absrel\downarrow$ & $\threshI\uparrow$
    & $\absrel\downarrow$ & $\threshI\uparrow$
    & $\absrel\downarrow$ & $\threshI\uparrow$
    & $\absrel\downarrow$ & $\threshI\uparrow$
    & $\absrel\downarrow$ & $\threshI\uparrow$
    & $\absrel\downarrow$ & $\threshI\uparrow$ 
    \\
    \hline
    \hline

    \row{row:metadata_low_res} &

    Ours (no metadata, low res.)

        &  \silver{3.39}
        &  \gold{66.88}
        &  \silver{3.86}
        &  \silver{60.82}
        &  \gold{3.11}
        &  \gold{70.17}
        &  {2.43}
        &  \silver{92.05}
        &  \gold{2.23}
        &  \gold{88.38}
        &  \gold{3.00}
        &  \gold{75.66}
        \\

    \row{row:vit_small} &
    w/ ViT Small
        &  3.57
        &  64.34
        &  4.40
        &  56.57
        &  3.63
        &  64.61
        &  2.69
        &  91.52
        &  2.71
        &  84.51
        &  3.40
        &  72.31
        \\

    \row{row:sr_arch} &
    w/ \cite{sayed2022simplerecon}'s architecture 
        &  3.63
        &  65.94
        &  5.03
        &  51.76
        &  3.74
        &  63.09
        &  \textbf{1.77}
        &  90.97
        &  2.78
        &  \underline{87.90}
        &  3.39
        &  71.93
        \\

    \row{row:wo_noise} &
    w/o noise on GT range
     &   \gold{3.33}
        &   \silver{66.83}
        &  5.14
        &  52.77
        &  3.53
        &  66.32
        & 13.45  
        & 89.21
        & 2.34
        & 87.64
        &  5.32
        &   \silver{74.89}
        \\

    \row{row:wo_dav2_weights} &
    
    w/o DAV2 weights
        &  3.45
        &  65.42
        &  4.58
        &  57.14
        &  3.48
        &  65.59
        &  2.14
        &  \textbf{92.48}
        &  2.56
        &  86.01
        &  3.24
        &  73.33
        \\    

    \row{row:fixed_bins} &

    w/ fixed bins [0-100m] 
        &  3.41
        &  64.33
        &  \textbf{3.80}
        &  \gold{61.62}
        &  \silver{3.15}
        &  67.20
        &  4.11
        &  65.64
        &  2.36
        &  85.72
        &  3.37
        &  68.90
        \\

    \row{row:wo_mmcc_vit} &
    
     no MMCC ViT
        &  3.54
        &  65.32
        &  4.39
        &  56.94
        &  3.56
        &  65.27
        &  3.07
        &  90.49
        &  2.46
        &  87.54
        &  3.41
        &  73.11
        \\
    
    \row{row:wo_bin_refinement} &
    
    w/o bin refinement
        &  3.57
        &  63.39
        &  5.18
        &  51.05
        &  3.50
        &  \silver{67.93}
        &  6.80
        &  82.12
        &  \silver{2.27}
        &  87.04
        &  4.26
        &  70.31
        \\

    \row{row:naive_patchify} &
    
    Naive patchify 
        &  3.66
        &  62.61
        &  4.27
        &  58.86
        &  {3.18}
        &  {67.27}
        &  \underline{1.95}
        &  91.69
        &  2.46
        &  86.52
        &  \underline{3.11}
        &  73.39
        \\ \hline

\end{tabular}

    }
    \vspace{-8pt}
    \caption{\textbf{Ablation Study.} 
        Here we validate our design decisions on RMVDB~\cite{schroeppel2022robust} by ablating various components. See \cref{sec:ablations} for details. \textbf{First} and \underline{second} best scores are indicated.
    }
    \label{tab:robust-multi-view-depth-ablation}
    \vspace{-12pt}
\end{table}

\subsection{Meshing and 3D Reconstruction}
\vspace{-5pt}
To judge the 3D-consistency of our predictions we evaluate our model on ScanNet Mesh Evaluation benchmark using the protocol defined in~\cite{bozic2021transformerfusion} which also uses source frame selection from~\cite{duzceker2021deepvideomvs}. 
The benchmark uses a GT mesh collected with an active RGBD sensor captured in a video. The evaluation computes point-to-point vertex error from GT to predicted (as accuracy), from predicted to GT (as completion), and the average of the two (as chamfer). Additionally, 200k points are sampled uniformly over each mesh and point-to-point errors thresholded at 5cm distance are used to compute precision, recall and F-score.
Almost all competing methods are trained on ScanNet, however our method that was not trained on ScanNet performs comparatively, outperforming many of the methods in Table~\ref{table:scannet_mesh_evaluation}.
See \cref{fig:teaser} and the supplementary for qualitative meshing results.

\newcommand{\scannettrained}{\textbf{S}}

\begin{table}[t]%
    \newcommand{\gold}[1]{\textbf{#1}}
    \centering
    \setlength{\tabcolsep}{0.2em} %
    \resizebox{1.0\columnwidth}{!}{  %
    \newcommand{\rowindent}{\hspace{0pt}}
    \footnotesize
        \begin{tabular}{lccccccc}
            \toprule
            & \hspace{-15pt}  & Comp$\downarrow$  & Acc$\downarrow$ & Chamfer$\downarrow$ & Prec$\uparrow$ & Recall $\uparrow$ & F-Score $\uparrow$ \\
            \midrule
            \rowindent DeepVideoMVS~\cite{duzceker2021deepvideomvs} & \scannettrained & 10.68 & 6.90 & 8.79 & 0.541 & 0.592 & 0.563  \\
            \rowindent ATLAS~\cite{murez2020atlas} & \scannettrained & 7.16 & \cellcolor{thirdcolor}7.61 & 7.38 & 0.675 & 0.605 & 0.636 \\
            \rowindent NeuralRecon~\cite{sun2021neuralrecon} & \scannettrained & \cellcolor{secondcolor}5.09 & 9.13 & 7.11 & 0.630 & \cellcolor{thirdcolor}0.612 & 0.619   \\
            \rowindent 3DVNet~\cite{rich20213dvnet} & \scannettrained & 7.72 & \cellcolor{secondcolor}6.73 & 7.22 & 0.655 & 0.596 & 0.621 \\
            \rowindent TransformerFusion~\cite{bozic2021transformerfusion}  & \scannettrained & \cellcolor{thirdcolor}5.52 & 8.27 & \cellcolor{thirdcolor} 6.89 & \cellcolor{secondcolor}0.728 & 0.600 & \cellcolor{thirdcolor}0.655 \\
            \rowindent VoRTX~\cite{stier2021vortx} & \scannettrained & \gold{4.31} & 7.23 & \gold{5.77} & \gold{0.767} & 0.651 & \gold{0.703}   \\
            \rowindent SimpleRecon~\cite{sayed2022simplerecon} & \scannettrained & 5.53 & \gold{6.09} & \cellcolor{secondcolor}5.81 & \cellcolor{thirdcolor}0.686 & \gold{0.658} & \cellcolor{secondcolor}0.671 \\

        \midrule
            \rowindent COLMAP~\cite{schoenberger2016mvs} &  & 10.22 & 11.88 & 11.05 & 0.509 & 0.474 & 0.489  \\
            \rowindent MAST3R \cite{mast3r_arxiv24} (raw depth) & \scannettrained+ &
            12.35  & 12.69  & 12.52  & 0.265  & 0.283  & 0.272 \\

            \rowindent MAST3R \cite{mast3r_arxiv24} (+ triangulation) & \scannettrained+ & 5.38 & 6.78 & 6.08 & 0.572 & 0.655 & 0.608  \\

            \rowindent SimpleRecon~\cite{sayed2022simplerecon} (trained on our data) & & 8.07 & 6.67 & 7.37 & 0.501 & 0.597 & 0.544 \\
            \rowindent \textbf{MVSA} (Ours)  &  & \gold{4.93} & \gold{6.39} & \gold{5.66} & \gold{0.616} & \gold{0.696} & \gold{0.652} \\
        
            \bottomrule
        \end{tabular}
    }
    \hfill
    \vspace{-7pt}    
    \caption{
        \textbf{ScanNet Mesh Evaluation~\cite{bozic2021transformerfusion}}. 
        Scores adapted from~\cite{bozic2021transformerfusion,sayed2022simplerecon}.
        Rows marked with \scannettrained~were trained on ScanNet only, while those marked \scannettrained+ were trained on ScanNet and other datasets.
        Our MVSA model, which was not trained on ScanNet, outperforms many models which were, \eg \cite{rich20213dvnet,murez2020atlas,sun2021neuralrecon}.
    }
    \label{table:scannet_mesh_evaluation}
     \vspace{-10pt}
\end{table}

\paragraph{Limitations.}
While we use multi-view information to generate depths, we do not enforce or encourage temporal consistency.
Techniques for this~\cite{yang2024depthanyvideo,hu2024DepthCrafter,Luo2020VideoDepth} could work with \modelname.
Also, like traditional MVS, our method requires known camera intrinsics and poses; recent works suggest this requirement could be relaxed \cite{murai2025mast3rslam,wang2025spanner}.

\vspace{-3pt}
\section{Conclusions}
\vspace{-5pt}
We introduced MVSAnywhere, a new general-purpose MVS depth estimation approach. 
We addressed challenges associated with training on diverse MVS  datasets, such as how to best leverage ViT-based architectures, how to incorporate geometric metadata, and how to handle variable depth ranges. Through extensive experimentation, we compare to numerous existing and new baselines.
Our contributions result in state-of-the-art zero-shot performance on a range of challenging reconstruction and depth estimation test datasets, in some cases even outperforming models trained on the test domains.

\section*{Acknowledgements}
We are extremely grateful to Saki Shinoda, Jakub Powierza, and Stanimir Vichev for their invaluable compute infrastructure support.
S Izquierdo and J Civera are funded by the Spanish projects PID2021-127685NB-I00 and TED2021-131150B-I00, Aragón project T45\_23R, and the scholarship FPU20/02342. 

{
    \small
    \bibliographystyle{ieeenat_fullname}
    \bibliography{main}
}

\clearpage
\appendix 

\noindent{\LARGE \bf Appendix}

\setcounter{table}{0}
\renewcommand{\thetable}{A\arabic{table}}
\setcounter{figure}{0}
\renewcommand{\thefigure}{A\arabic{figure}}

\section{Additional baseline details}
\label{sec:supp_impl_details}

Most of the scores in Tab.~2 in the main paper (and Table~\ref{tab:robust-multi-view-depth-sup-mat}) are directly taken from prior works~\cite{schroeppel2022robust,dust3r_cvpr24}.
For completeness, we additionally evaluate other works which did not evaluate on RMVDB.
These were described briefly in the main paper, but  here we provide full details.

\paragraph{Single-view baselines.}
First we include general-purpose monocular depth models via the publicly available checkpoints, using the reference frame as input (\ie no multi-view signal). 
We include monocular models which predict metric depth \cite{yin2023metric,hu2024metric3d,piccinelli2024unidepth,bochkovskii2024depthpro}, both with and without median scaling to align the predictions to the GT.
We use the publically available code and pretrained models for these baselines.
For each of them, we use the mode which allows input of known intrinsics, and we provide the intrinsics given with each of the test datasets.
We additionally include Depth Anything V2 \cite{depth_anything_v2}, which is a large ViT-based model which predicts depths up to an unknown affine transform.
To account for the unknown affine transform, we align the predictions to the GT using least squares.
Without median scaling the performance of many of these monocular models is poor, even though they are trained to predict in metric space.
Even with median scaling, they rarely outperform our model, which benefits from multi-view information.

\paragraph{MAST3R~\cite{mast3r_arxiv24} (raw depth estimate).}
MAST3R  takes a pair of images as input and predicts a metric-scaled point cloud in the reference frame of the first image. 
The simplest method to extract depth is to input $I_r$, and one other frame, and take the $z$ component of the point cloud as the depth prediction.
When selecting the source frame to use, we use the `best' single source frame as provided by the benchmark.
We include this baseline as \emph{MAST3R (raw output)} in (b). 
We clarify here that MAST3R is trained on the Habitat dataset \cite{ramakrishnan2021hm3d}, which itself contains ScanNet~\cite{dai2017scannet}.
So \textbf{MAST3R trains on ScanNet}. They also train on MegaDepth~\cite{MegaDepthLi18}, which contains a subset of the Tanks and Temples dataset. Their scores are therefore in parentheses for these datasets.

\paragraph{MAST3R~\cite{mast3r_arxiv24} (plus our triangulation).}
The provided version of MAST3R is not able to make use of camera poses and extrinsics.
For an additional comparison with MAST3R, we extend their method so that it can use provided extrinsics and intrinsics, when available, to boost their results. 
For each of the available source images, we use MAST3R descriptors to match points with the reference frame. 
Then we use these matches and the known camera extrinsics and intrinsics to triangulate points, which results in a sparse depth map from the viewpoint of the reference image.
We then use these sparse depths to rescale the MAST3R raw depth predictions into a more `correct' range. 
Then, following DUST3R~\cite{dust3r_cvpr24}, we aggregate the point clouds $X^i$ from the different views, $i$, using a weighted sum and the predicted confidences, $C^i$ 
$$
D = \frac{\sum_{i=0}^{N}{X^i_z C^i}}{\sum_{i=0}^{N}{C^i}}
$$
Note, that this method requires one forward pass and thousands of triangulations per each source view, making it slow.
We denote this as \emph{MAST3R (plus our triangulation)} in part (d) of the results table.

\section{Additional  model and training details}

\subsection{Model architecture}
\label{sec:supp_architecture}

\paragraph{Feature extractor.} We use the first two blocks of a pretrained ResNet-18 network and apply an instance normalization to the features, following~\cite{sayed2022simplerecon}.

\paragraph{Reference image encoder.} We use a ViT-B initialized from the Depth Anything V2 relative depth weights.

\paragraph{Mono/Multi Cue Combiner.} For the cost volume patchifier we use three residual blocks with stride 1, 2, and 2 that progressively downsample the cost volume to 1/16 of the image resolution. After the first and the second of these blocks we concatenate the features with the output of the first two blocks of the reference image encoder, respectively. These outputs are first upsampled using transposed convolutions with kernel and stride size 4 and 2, and projected into the desired number of channels using a linear layer. The concatenated features are processed by two residual blocks.
Each of the resulting pixels at 1/16 resolution after the cost volume patchifier is a token which is then processed by a ViT-B initialized from DINOv2 weights. After blocks 2, 5, 9, and 11, we add these with their respective blocks from the reference image encoder. We linearly project the reference features before adding them.

\paragraph{Depth decoder.} Based on the decoder from \cite{Ranftl2021}, it reassembles the tokens into feature maps, and it fuses these to obtain fine-grained predictions. We adapt this module to produce outputs at full, 1/2, 1/4 and 1/8 resolutions.

\subsection{Metadata details}
\label{sec:supp_metadata}

In our full model we use `metadata' as components of the cost volume, following \cite{sayed2022simplerecon}.
For completeness, here we briefly re-state what these comprise.
Given pixel location $u_r, v_r$ in the reference image, and depth bin $k$, we include the following items of metadata in the cost volume:

\begin{description}

    \setlength\itemsep{0.15em}
    
    \item[Feature dot product:] 
        The dot product of the reference image features and each warped source image features, expressed for each source image $i$ as $\mathcal{F}^r \cdot \langle\mathcal{F}\rangle^i$, where  $\langle \rangle$ is the warping operation which warps features from the source image to the reference viewpoint at the depth corresponding to bin $k$.
        This value is often used as the sole matching affinity in cost volumes.
    
    \item[Ray directions $\mathbf{r}^0_{k, u_r, v_r}$ and $\mathbf{r}^i_{k, u_r, v_r} \in \mathbb{R}^3$:] 
        This is the normalized directions pointing from the camera origins to the 3D location of a point $(k, u_r, v_r)$ in the plane sweep cost volume. We create rays for the reference and all the source images.
    
    \item[Reference plane depth $z^r_{k, u_r, v_r}$:] 
        This is the depth of the point at position $(k, u_r, v_r)$ in the cost volume, measured perpendicularly from the reference camera. We normalize these values with the min and max depth.
    
    \item[Reprojected depths $z^i_{k, u_r, v_r}$:] 
        This is the perpendicular depth of the 3D point at position $(k, u_r, v_r)$ in the cost volume, relative to the source camera $n$. We normalize these values with the min and max depth.
    
    \item[Relative ray angles $\theta^{r, i}$:] 
        This is the angle between the ray directions $\mathbf{r}^r_{k, u_r, v_r}$ and $\mathbf{r}^i_{k, u_i, v_i}$.
    
    \item[Relative pose distance $p^{r,i}$:] 
        This is the relative pose distance between the reference camera and a source frame, as defined in \cite{duzceker2021deepvideomvs}:
        \begin{align} \label{eqn:pose_dist}
            p^{r,i} =
            \sqrt{\|\mathbf{t}^{r,i}\| + \frac{2}{3}\text{tr}
            (\mathbb{I}-\mathbf{R}^{r,i})},
        \end{align}
        where $\mathbf{t}^{r,i}$ and $\textbf{R}^{r,i}$ are the relative translation and rotation between views $i$ and $r$. The translation, $\mathbf{t}^{r,i}$, is normalized by the source frame with the biggest pose distance.
    
    \item[Depth validity masks $m^i_{k, u_r, v_r}$:] 
        This is a binary mask indicating whether the point $(k, u_r, v_r)$ in the cost volume projects in front of the source camera $i$ or not.
    
\end{description}

\subsection{Training details}

We train on two A100 GPUs for a total of 145k steps, which takes approximately 2.5 days.
We use a batch size of eight on each GPU for our $448 \times 336$ models, and a batch size of six for our $640 \times 480$ models.
We accumulate gradients over two consecutive batches, so we only backpropogate after two batches have been seen.

We use different learning rates for different components. 
The matching encoder and cost volume MLPs have an initial learning rate of $1\mathrm{e}-4$ until step 140k, and $1\mathrm{e}-5$ afterwards.
The decoder and  multi/mono cue combiner  have an initial learning rate of $5\mathrm{e}-5$ which is linearly decayed during training to $5\mathrm{e}-8$.
The reference image encoder has an initial learning rate of $5\mathrm{e}-6$ which is linearly decayed during training to $5\mathrm{e}-9$. 
We use weight decay of $1e-4$.

\subsection{Training losses}
\label{sec:supp_train_losses}

\paragraph{Log depth loss.}
We follow \cite{eigen2014depth,sayed2022simplerecon} and supervise predictions using log-depth across all pixels. 
At each decoder output scale $s$, we apply an absolute error on the log-depth:
\begin{align}
    \mathcal{L}_{\text{depth}} 
    &= 
    \frac{1}{HW} 
    \sum_{s=1}^{4}\sum_{i,j}
    \frac{1}{s^2} 
    \left| 
        \uparrow_{gt} \log \hat{D}_{r}^{i,j} - \log D^{i,j} 
    \right|,
\end{align}
where $\hat{D}_r$ is the network output depth map, $D$ is the ground truth depth map and $i,j$ superscripts indicate pixel indices.
Here each decoder output is aligned with the full-size ground truth depth map via nearest-neighbor upsampling~\cite{duzceker2021deepvideomvs} (denoted by $\uparrow_{gt}$).
We average this loss across pixels, scales, and batches. 

\paragraph{Gradient loss.}
Following \eg \cite{MegaDepthLi18, Ranftl2022, sayed2022simplerecon}, we apply a multi-scale gradient loss to the highest-resolution network output.
In contrast to those works, we apply the loss to $\frac{1}{D}$ rather than to $D$ directly, to help prevent this loss from dominating the overall loss.
Our gradient loss is therefore: 
\begin{align}
    \mathcal{L}_\text{grad}
    &= 
    \frac{1}{HW}
    \sum_{s=1}^{4}\sum_{i,j} 
    \left| 
        \nabla\downarrow_s \frac{1}{\hat{D}_r^{i,j} }
        - 
        \nabla\downarrow_s \frac{1}{D^{i,j} }
    \right|,
\end{align}
where $\nabla$ computes first-order spatial gradients, and $\downarrow_s$ represents downsampling to scale $s$.

\paragraph{Normals loss.}
We also use a simplified normal loss. 
\begin{align}
    \mathcal{L}_\text{normals} 
    &=
    \frac{1}{2HW} 
    \sum_{i,j} 
    \left( 
        1 - \hat{N}^{i,j} \cdot N^{i,j} 
    \right),
\end{align}
where $N$ is the normal map derived from depth and camera intrinsics.

\paragraph{Final loss.}
Our final loss is
\begin{align}
    \mathcal{L} = \mathcal{L}_{\text{depth}} + \mathcal{L}_\text{grad} + \mathcal{L}_\text{normals}. 
\end{align}

\section{Constructing 3D meshes}

In Fig.~1 and Sec.~4.4 in the main paper we convert our posed depth maps into 3D meshes for visualization and evaluation.
To do this, we broadly follow the procedure from prior work \eg \cite{sayed2022simplerecon}.
We fuse depths into a sparse TSDF volume using Open3D~\cite{zhou2018open3d}, use a voxel size of 4cm, and avoid fusing depth values over 3.5m.
We extract a final mesh from the TSDF using marching cubes \cite{lorensen1998marching}.

\section{Additional results}
\label{sec:supp_add_results}

Tab.~2 in the main paper omitted some earlier methods for space reasons.
In \cref{tab:robust-multi-view-depth-sup-mat} we present a more complete version of the table, with scores from the earlier methods re-instated.

In \cref{tab:robust-multi-view-depth-ablation-sup-mat} we present an extended version of the ablations from Tab.~4 from the main paper. 
We divide the ablations into subsections to validate the different design choices in \modelname. Except those experiments specifically ablating the metadata, all ablations are done without metadata for efficiency reasons. In \cref{tab:alternative-benchmark-sup-mat} we report results of these ablations on the improved tuples for ScanNet and undistorted images for ETH3D, as in Tab.~3 of the main paper.

\paragraph{Ablating the Mono/Multi Cue Combiner.} We show how both, using a CNN instead of a ViT (no MMCC ViT) and using a naive way to convert the cost volume into tokens (naive patchify) result in significantly degraded results.

\paragraph{Ablating metadata and normalization.} 
Using readily available metadata is overall beneficial for our model. 
The model without metadata can sometimes lead to better results, like in ScanNet, where the baseline between frames is many times too small and relying on the cost volume does not provide as much benefit. However the average results indicate that the using metadata is superior. Although using metadata without scale normalization can report better results on this metric benchmark, we show in \cref{tab:scannet_scaled} how this model struggles when using an arbitrary scale. 
We exemplify this by scaling the ScanNet poses by an arbitrary factor where we rescale ScanNet poses and depths (which are in metric scale) by a factor of 100. 
This simulates the type of `non-metric scene scale' we might see if, for example, we reconstructed a scene using  SfM package such as COLMAP which does not provide metric scaled reconstructions.

\paragraph{Ablating the cost volume.}
Not using noise on the cost volume range during training significantly impacts performance in cases where our range estimate using camera intrinsics and extrinsics is poor, like in DTU or ScanNet.
Unsurprisingly these are the datasets where not doing a refinement has the biggest impact. 
A fixed 0-100m range can be beneficial in some cases where the test data overlaps this range. However, the model fails in the case of very different ranges like on DTU, and would not work with non-metric poses. 

\paragraph{Ablating the overall architecture.} We compare our model without starting from DAv2 pretrained weights, using a smaller ViT and using a previous state-of-the-art architecture from~\cite{sayed2022simplerecon}. 
Each of these variants obtain worse results than using our architecture.

\paragraph{Effect of our training datasets.} We show in Table~\ref{tab:data} head-to-head network comparisons trained on the same data. In the first two rows, we show our network \vs MVSFormer++~\cite{cao2024mvsformerplus}, both of which were trained from scratch on DTU+BlendedMVS. In the bottom two rows, we fine-tuned\cite{cao2024mvsformerplus} for 40K steps with all MVSAnywhere datasets. We use their `regression loss option as we found that this gave them better scores when training with our data. However, this model is subpar to ours, particularly in casually captured datasets such as ScanNet.

\paragraph{Evaluating our depth range estimation.} In \ref{tab:range}, we show how our heuristic to estimate the depth range also works well in conjunction with MVSFormer++~\cite{cao2024mvsformerplus}'s architecture even though it was not trained to handle noisy ranges (but still it does not beat our full system).

\paragraph{Results on DTU.} Results using the point cloud evaluation on DTU are shown in Table~\ref{tab:dtu_recons}. We follow \cite{cao2024mvsformerplus} to fuse our depth maps using their confidence predictions and their default fusion hyperparameters.

\begin{table}
    \centering
    \resizebox{1.0\columnwidth}{!}{
    {\fontsize{8pt}{10pt}\selectfont

        \setlength{\tabcolsep}{0.8mm}
        \begin{tabular}{|l
        |c >{\columncolor{bgcolor}}c
        |c >{\columncolor{bgcolor}}c
        |}
        
        \hline
            \textbf{Approach}
            & \multicolumn{2}{c|}{\textbf{\scannetshort{}}}
            & \multicolumn{2}{c|}{\textbf{\scannetshort{} $\times 100$}}
            \\

            & $\absrel\downarrow$ & $\threshI\uparrow$
            & $\absrel\downarrow$ & $\threshI\uparrow$
            \\
            \hline
            \hline
            Ours (low res) \cite{schroeppel2022robust}
                & 4.22
                & 61.80
                & 4.22
                & 61.83
                \\
            Ours w/o norm. metadata w/o view count agnostic
                &  3.97
                &  61.23
                &  4.34
                &  57.59
                \\
                \hline
        
        \end{tabular}
    }}
    \vspace{-5pt}
    \caption{\textbf{Results on an arbitrary scale}. We scale the ScanNet poses by a factor of $100$ to evaluate robustness to arbitrary scales.
    \label{tab:scannet_scaled}
    }
    \vspace{-3pt}
\end{table}

\begin{table}
    \centering
    \resizebox{0.875\columnwidth}{!}{
    {\fontsize{8pt}{10pt}\selectfont

\setlength{\tabcolsep}{1pt}
\begin{tabular}{|l|c|c|c|c|c|}
    \hline
    & \multicolumn{3}{|c|}{\textbf{DTU (mm)}}  \\
        & Acc.$\downarrow$ & Compl.$\downarrow$   & Overall$\downarrow$ \\
     \hline
        \multicolumn{4}{|l|}{\textbf{Direct depth regression + cloud fusion}} \\
     Ours & 0.845 & 0.625 & 0.735 \\
     Robust MVD & 1.330 & 1.029 & 1.180 \\
     MAST3R + triangulation & 3.969 & 3.548 & 3.758 \\
     DUST3R & 2.677 & 0.805 & 1.741 \\
     MVSFormer++ [6] & (0.309) & (0.252) & (0.281) \\
     \hline
     \multicolumn{4}{|l|}{\textbf{Sub-pixel matching + triangulation}} \\
    MAST3R [39]  & 0.403 & 0.344 & 0.374 \\
     \hline
\end{tabular}
}}
\caption{\textbf{DTU Reconstruction.} We evaluate our method and other direct depth regression methods on DTU reconstruction.}
\label{tab:dtu_recons}
\end{table}

\begin{table*}
    
    \centering
    \resizebox{\textwidth}{!}{
        \setlength{\tabcolsep}{0.8mm}

\begin{tabular}{|l|c|c|c
|c >{\columncolor{bgcolor}}c
|c >{\columncolor{bgcolor}}c
|c >{\columncolor{bgcolor}}c
|c >{\columncolor{bgcolor}}c
|c >{\columncolor{bgcolor}}c
|c >{\columncolor{bgcolor}}c
|}

\hline
    \textbf{Approach}
    & \textbf{\scriptsize{GT}}
    & \textbf{\scriptsize{GT}} 
    & \textbf{Align}
    & \multicolumn{2}{c|}{\textbf{\kittishort{}}}
    & \multicolumn{2}{c|}{\textbf{\scannetshort{}}}
    & \multicolumn{2}{c|}{\textbf{\ethdshort{}}}
    & \multicolumn{2}{c|}{\textbf{\dtushort{}}}
    & \multicolumn{2}{c|}{\textbf{\tanksandtemplesshort{}}}
    & \multicolumn{2}{c|}{\textbf{Average}}
    \\

    & \textbf{\scriptsize{Poses}}
    & \textbf{\scriptsize{Range}}
    &
    & $\absrel\downarrow$ & $\threshI\uparrow$
    & $\absrel\downarrow$ & $\threshI\uparrow$
    & $\absrel\downarrow$ & $\threshI\uparrow$
    & $\absrel\downarrow$ & $\threshI\uparrow$
    & $\absrel\downarrow$ & $\threshI\uparrow$
    & $\absrel\downarrow$ & $\threshI\uparrow$
    \\
    \hline
    \hline

        MVSFormer++ \scriptsize{\dtu{}+BlendedMVG}~\cite{cao2024mvsformerplus}
	& \my
	& \my
	& \mn
        &  {4.4}
        &  {65.7}
        &  {7.9}
        &  {39.4}
        &  {7.8}
        &  {50.4}
        &  (0.9)
        &  (95.3)
        &  {3.2}
        &  {88.1}
        &  {4.8}
        &  {67.8}
        \\

       \textbf{\modelname} \scriptsize{\dtu{}+BlendedMVG}~\cite{cao2024mvsformerplus}
	& \my
	& \my
	& \mn
        &  {4.1}
        &  {57.8}
        &  {4.8}
        &  {53.7}
        &  {3.6}
        &  {61.5}
        &  (1.4)
        &  (90.7)
        &  {2.7}
        &  {80.1}
        &  {3.3}
        &  {68.8}
        \\

        \hline

         MVSFormer++ \scriptsize{\modelname data} %
	& \my
	& \my
	& \mn
        &  {3.7}
        &  {67.8}
        &  {5.4}
        &  {46.5}
        &  {4.9}
        &  59.1
        &  {1.3}
        &  {90.1}
        &  {2.0}
        &  {89.7}
        &  {3.5}
        &  {70.6}
        \\

         \textbf{\modelname} \scriptsize{\modelname data} %
	& \my
	& \mn
	& \mn
        &  {3.2}
        &  {68.8}
        &  {3.7}
        &  {62.9}
        &  {3.2}
        &  68.0
        &  {1.3}
        &  {95.0}
        &  {2.1}
        &  {90.5}
        &  {2.7}
        &  {77.0}
        \\

\hline
\end{tabular}

    }
    \vspace{-6pt}
    \caption{
        \textbf{Effect of training data.} We run a head-to-head comparison with MVSFormer++\cite{cao2024mvsformerplus}, first training our architecture on MVSFormer++ data, and then training MVSFormer++ on our data. In both cases, our architecture is better suited for the task of multi-view depth estimation.
    }
    \label{tab:data}
\end{table*}

\begin{table*}
    
    \centering
    \resizebox{\textwidth}{!}{
        \setlength{\tabcolsep}{0.8mm}

\begin{tabular}{|l|c|c|c
|c >{\columncolor{bgcolor}}c
|c >{\columncolor{bgcolor}}c
|c >{\columncolor{bgcolor}}c
|c >{\columncolor{bgcolor}}c
|c >{\columncolor{bgcolor}}c
|c >{\columncolor{bgcolor}}c
|}

\hline
    \textbf{Approach}
    & \textbf{\scriptsize{GT}}
    & \textbf{\scriptsize{GT}} 
    & \textbf{Align}
    & \multicolumn{2}{c|}{\textbf{\kittishort{}}}
    & \multicolumn{2}{c|}{\textbf{\scannetshort{}}}
    & \multicolumn{2}{c|}{\textbf{\ethdshort{}}}
    & \multicolumn{2}{c|}{\textbf{\dtushort{}}}
    & \multicolumn{2}{c|}{\textbf{\tanksandtemplesshort{}}}
    & \multicolumn{2}{c|}{\textbf{Average}}
    \\

    & \textbf{\scriptsize{Poses}}
    & \textbf{\scriptsize{Range}}
    &
    & $\absrel\downarrow$ & $\threshI\uparrow$
    & $\absrel\downarrow$ & $\threshI\uparrow$
    & $\absrel\downarrow$ & $\threshI\uparrow$
    & $\absrel\downarrow$ & $\threshI\uparrow$
    & $\absrel\downarrow$ & $\threshI\uparrow$
    & $\absrel\downarrow$ & $\threshI\uparrow$
    \\
    \hline
    \hline

        MVSFormer++~\cite{cao2024mvsformerplus}
	& \my
	& \my
	& \mn
        &  {4.4}
        &  {65.7}
        &  {7.9}
        &  {39.4}
        &  {7.8}
        &  {50.4}
        &  (0.9)
        &  (95.3)
        &  {3.2}
        &  {88.1}
        &  {4.8}
        &  {67.8}
        \\
        MVSFormer++~\cite{cao2024mvsformerplus} (0.25 - 100m)
	& \my
	& \mn
	& \mn
        &  {47.9}
        &  {13.0}
        &  {21.3}
        &  {33.3}
        &  {39.4}
        &  {18.7}
        &  (25.6)
        &  (51.5)
        &  {40.6}
        &  {30.0}
        &  {35.0}
        &  {29.3}
        \\

        MVSFormer++~\cite{cao2024mvsformerplus} (our heuristic)
	& \my
	& \mn
	& \mn
        &  {8.3}
        &  {49.3}
        &  {48.7}
        &  {11.3}
        &  {21.7}
        &  {31.8}
        &  (45.4)
        &  (59.7)
        &  {6.5}
        &  {75.6}
        &  {26.12}
        &  {45.5}
        \\

        MVSFormer++~\cite{cao2024mvsformerplus} (our heuristic + refinement)
	& \my
	& \mn
	& \mn
        &  {5.1}
        &  {64.6}
        &  {23.5}
        &  {27.3}
        &  {10.4}
        &  {48.6}
        &  (25.5)
        &  (62.6)
        &  {3.7}
        &  {87.2}
        &  {13.6}
        &  {58.1}
        \\

        \hline

         \textbf{\modelname} %
	& \my
	& \mn
	& \mn
        &  {3.2}
        &  {68.8}
        &  {3.7}
        &  {62.9}
        &  {3.2}
        &  68.0
        &  {1.3}
        &  {95.0}
        &  {2.1}
        &  {90.5}
        &  {2.7}
        &  {77.0}
        \\

\hline
\end{tabular}

    }
    \vspace{-6pt}
    \caption{
        \textbf{Depth range estimation.} We show how our heuristic to estimate a depth range can also work in other methods like MVSFormer++~\cite{cao2024mvsformerplus}. This effectively removes the GT Range input requirement.
    }
    \label{tab:range}
\end{table*}

\begin{table*}[t!]
    
    \newcommand{\baselinename}{Robust MVD Baseline}
    \newcommand{\absrelname}{Absolute Relative Error}
    \newcommand{\otherview}{\other{} view}
    \newcommand{\otherviews}{\otherview{}s}
    \newcommand{\threshIname}{Inlier Ratio}%
    \newcommand{\other}{source}
    \centering
    \resizebox{\textwidth}{!}{
        \setlength{\tabcolsep}{0.8mm}

\begin{tabular}{|l|c|c|c
|c >{\columncolor{bgcolor}}c
|c >{\columncolor{bgcolor}}c
|c >{\columncolor{bgcolor}}c
|c >{\columncolor{bgcolor}}c
|c >{\columncolor{bgcolor}}c
|c >{\columncolor{bgcolor}}c c
|}

\hline
    \textbf{Approach}
    & \textbf{\scriptsize{GT}}
    & \textbf{\scriptsize{GT}} 
    & \textbf{Align}
    & \multicolumn{2}{c|}{\textbf{\kittishort{}}}
    & \multicolumn{2}{c|}{\textbf{\scannetshort{}}}
    & \multicolumn{2}{c|}{\textbf{\ethdshort{}}}
    & \multicolumn{2}{c|}{\textbf{\dtushort{}}}
    & \multicolumn{2}{c|}{\textbf{\tanksandtemplesshort{}}}
    & \multicolumn{3}{c|}{\textbf{Average}}
    \\

    & \textbf{\scriptsize{Poses}}
    & \textbf{\scriptsize{Range}}
    &
    & $\absrel\downarrow$ & $\threshI\uparrow$
    & $\absrel\downarrow$ & $\threshI\uparrow$
    & $\absrel\downarrow$ & $\threshI\uparrow$
    & $\absrel\downarrow$ & $\threshI\uparrow$
    & $\absrel\downarrow$ & $\threshI\uparrow$
    & $\absrel\downarrow$ & $\threshI\uparrow$ & time [s] $\downarrow$
    \\
    \hline
    \hline

    \textbf{Classical SfM approaches}
	& 
	& 
	& 
	& 
	& 
	& 
	& 
	& 
	& 
	& 
	& 
	& 
	& 
	& 
	& 
	&
    \\

    \textsc{Colmap}~\cite{schoenberger2016mvs,schoenberger2016SfM}
	& \my
	& \mn
	& \mn
	& \bestresult{12.0}
	& \bestresult{58.2}
	& \bestresult{14.6}
	& \bestresult{34.2}
	& \bestresult{16.4}
	& \bestresult{55.1}
	& \bestresult{0.7}
	& \bestresult{96.5}
	& \bestresult{2.7}
	& \bestresult{95.0}
	& \bestresult{9.3}
	& \bestresult{67.8}
	& $\approx 180$
	\\ %

	\textsc{Colmap} {\scriptsize Dense}~\cite{schoenberger2016mvs,schoenberger2016SfM}
	& \my
	& \mn
	& \mn
	& 26.9
	& 52.7
	& 38.0
	& 22.5
	& 89.8
	& 23.2
	& 20.8
	& 69.3
	& 25.7
	& 76.4
	& 40.2
	& 48.8
	& $\approx 180$
	\\ %
	
\hline
\hline

    \multicolumn{17}{|l|}{\textbf{a) Depth from frames (w/o poses)}}
    \\

	DeMoN~\cite{Ummenhofer2016}
	& \mn
	& \mn
	& $\Vert \vect t \Vert$
	& 15.5
	& 15.2
	& \bestresult{12.0}
	& \bestresult{21.0}
	& 17.4
	& 15.4
	& 21.8
	& 16.6
	& 13.0
	& 23.2
	& 16.0
	& 18.3
	& \bestresult{0.08}
	\\ %

	DeepV2D \scriptsize{\kitti{}}~\normalsize{\cite{Teed2020Deepv2d}}
	& \mn
	& \mn
	& med
	& (\bestresult{3.1})
	& (\bestresult{74.9})
	& {23.7}
	& {11.1}
	& {27.1}
	& {10.1}
	& {24.8}
	& {8.1}
	& {34.1}
	& {9.1}
	& {22.6}
	& {22.7}
	& {2.07}
	\\ %

	DeepV2D \scriptsize{\scannet{}}\normalsize{~\cite{Teed2020Deepv2d}}
	& \mn
	& \mn
	& med
	& {10.0}
	& {36.2}
	& ({4.4})
	& ({54.8})
	& {11.8}
	& {29.3}
	& {7.7}
	& {33.0}
	& \bestresult{8.9}
	& \bestresult{46.4}
	& {8.6}
	& {39.9}
	& 3.57
	\\ %

 	MAST3R~\cite{mast3r_arxiv24} (raw output)
	& \mn
	& \mn
	& med
	& \bestresult{3.3}
	& \bestresult{67.7}
	& (\bestresult{4.3})
	& (\bestresult{64.0})
	& \bestresult{2.7}
	& \bestresult{79.0}
	& \bestresult{3.5}
	& \bestresult{66.7}
	& (\bestresult{2.4})
	& (\bestresult{81.6})
	& \bestresult{3.3}
	& \bestresult{71.8}
	& 0.07
	\\ %

        MAST3R~\cite{mast3r_arxiv24}  (raw output)  
	& \mn
	& \mn
	& \mn
    &  61.4 %
    &  0.4
    &  (12.8) %
    &  (19.4)
    &  43.8 %
    &  3.1
    &  145.8 %
    &  0.5
    &  (66.9) %
    &  (0.0)
    &  66.1
    &  4.7
    &  0.07
    \\

\hline
\hline

    \multicolumn{17}{|l|}{\textbf{b) Depth from frames and poses (with per-image range provided)}}
    \\

	MVSNet\normalsize{~\cite{yao2018mvsnet}}	& \my
	& \my
	& \mn
	& 22.7
	& 36.1
	& 24.6
	& 20.4
	& 35.4
	& 31.4
	& (1.8)
	& (86.0)
	& 8.3
	& 73.0
	& 18.6
	& 49.4
	& \bestresult{0.07}
	\\ %

	MVSNet \scriptsize{Inv. Depth}\normalsize{~\cite{yao2018mvsnet}}
	& \my
	& \my
	& \mn
	& 18.6
	& 30.7
	& 22.7
	& 20.9
	& 21.6
	& 35.6
	& (1.8)
	& (86.7)
	& 6.5
	& 74.6
	& 14.2
	& 49.7
	& 0.32
	\\ %

	CVP-MVSNet \normalsize{~\cite{Yang2020}}
	& \my
	& \my
	& \mn
	& {156.7}
	& {2.2}
	& {137.1}
	& {15.9}
	& {156.4}
	& {13.6}
	& ({4.0})
	& ({68.4})
	& {24.7}
	& {52.9}
	& {95.8}
	& {30.6}
	& {0.49}
	\\ %

	Vis-MVSNet~\cite{Zhang2020}
	& \my
	& \my
	& \mn
	& {9.5}
	& {55.4}
	& 8.9
	& 33.5
	& {10.8}
	& {43.3}
	& (1.8)
	& (87.4)
	& {4.1}
	& {87.2}
	& {7.0}
	& {61.4}
	& 0.70
	\\ %

	PatchmatchNet~\cite{Wang2020}
	& \my
	& \my
	& \mn
	& 10.8
	& 45.8
	& {8.5}
	& {35.3}
	& 19.1
	& 34.8
	& (2.1)
	& (82.8)
	& 4.8
	& 82.9
	& 9.1
	& 56.3
	& 0.28
	\\ %

	Fast-MVSNet~\cite{Yu2020}
	& \my
	& \my
	& \mn
	& 14.4
	& 37.1
	& 17.0
	& 24.6
	& 25.2
	& 32.0
	& (2.5)
	& (81.8)
	& 8.3
	& 68.6
	& 13.5
	& 48.8
	& 0.30
	\\ %

	MVS2D \scriptsize{\scannet{}}\normalsize{~\cite{Yang2022}}
	& \my
	& \my
	& \mn
	& 21.2
	& 8.7
	& (27.2)
	& (5.3)
	& 27.4
	& 4.8
	& 17.2
	& 9.8
	& 29.2
	& 4.4
	& 24.4
	& 6.6
	& \bestresult{0.04}
	\\ %

	MVS2D \scriptsize{\dtu{}}\normalsize{~\cite{Yang2022}}
	& \my
	& \my
	& \mn
	& 226.6
	& 0.7
	& 32.3
	& 11.1
	& 99.0
	& 11.6
	& (3.6)
	& (64.2)
	& 25.8
	& 28.0
	& 77.5
	& 23.1
	& 0.05
	\\ %

	MVSFormer++ \scriptsize{\dtu{}}~\cite{cao2024mvsformerplus}
	& \my
	& \my
	& \mn
	& 26.3
	& 42.8
	& 16.7
	& 28.0
	& 30.3
	& 40.1
	& (\bestresult{0.8})
	& (\bestresult{95.7})
	& 7.2
	& 82.3
	& 16.3
	& 57.8
	& 0.78
	\\ %

        MVSFormer++ \scriptsize{\dtu{}+BlendedMVG}~\cite{cao2024mvsformerplus}
	& \my
	& \my
	& \mn
        &  \bestresult{4.4}
        &  \bestresult{65.7}
        &  \bestresult{7.9}
        &  \bestresult{39.4}
        &  \bestresult{7.8}
        &  \bestresult{50.4}
        &  (0.9)
        &  (95.3)
        &  \bestresult{3.2}
        &  \bestresult{88.1}
        &  \bestresult{4.8}
        &  \bestresult{67.8}
        &  0.78
        \\

        \hline
        \hline  
            \multicolumn{17}{|l|}{\textbf{c) Single-view depth}}
    \\

        Depth Pro~\cite{bochkovskii2024depthpro} $\dagger$
        & \mn
        & \mn
        & med
        &  6.1
        &  39.6
        &  (4.3)
        &  (58.4)
        &  6.1
        &  53.5
        &  5.6
        &  49.6
        &  5.6
        &  57.5
        &  5.6
        &  51.7
        &  5.16
        \\

        Depth Pro~\cite{bochkovskii2024depthpro} $\dagger$
        & \mn
        & \mn
        & \mn
        &  13.6
        &  14.3
        &  9.2
        &  19.7
        &  28.5
        &  8.7
        &  161.8
        &  3.5
        &  38.3
        &  4.4
        &  50.3
        &  10.1
        &  5.16
        \\

        Metric3D~\cite{hu2024metric3d} $\dagger$
        & \mn
        & \mn
        & med
        &  5.1
        &  44.1
        &  \bestresult{2.4}
        &  \bestresult{78.3}
        &  4.4
        &  54.5
        &  10.1
        &  39.5
        &  6.2
        &  48.0
        &  5.6
        &  52.9
        &  0.46
        \\

        Metric3D~\cite{hu2024metric3d} $\dagger$
        & \mn
        & \mn
        & \mn
        &  8.7
        &  13.2
        &  6.2
        &  19.3
        &  12.7
        &  13.0
        &  890.5
        &  1.4
        &  16.7
        &  13.7
        &  187.0
        &  12.1
        &  0.46
        \\
    
        UniDepthV2~\cite{piccinelli2024unidepth} $\dagger$ 
        & \mn
        & \mn
        & med        
        &  \bestresult{4.0}
        &  \bestresult{55.3}
        &  (2.1)
        &  (82.6)
        &  \bestresult{3.7}
        &  \bestresult{66.2}
        &  3.2
        &  72.3
        &  \bestresult{3.6}
        &  \bestresult{68.4}
        &  \bestresult{3.3}
        &  \bestresult{68.9}
        &  0.29
        \\
    
        UniDepthV2~\cite{piccinelli2024unidepth} $\dagger$ 
        & \mn
        & \mn
        & \mn
        &  13.7
        &  4.8
        &  (3.2)
        &  (61.3)
        &  15.4
        &  11.9
        &  964.8
        &  1.3
        &  16.7
        &  12.7
        &  202.7
        &  18.4
        &  0.29
        \\

        UniDepthV1~\cite{piccinelli2024unidepth} $\dagger$ 
        & \mn
        & \mn
        & med 
            &  4.4
    &  51.6
    &  (\bestresult{1.9})
    &  (\bestresult{84.3})
    &  5.4
    &  48.4
    &  9.3
    &  31.8
    &  9.6
    &  38.7
    &  6.1
    &  51.0
    &  0.21
    \\  
        UniDepthV1~\cite{piccinelli2024unidepth} $\dagger$ 
        & \mn
        & \mn
        & \mn
            &  5.2
    &  39.5
    &  (2.7)
    &  (69.4)
    &  48.2
    &  1.8
    &  583.3
    &  1.0
    &  30.7
    &  4.2
    &  134.0
    &  23.2
    &  0.20
        \\

	DepthAnything V2 (ViT-B)~\cite{depth_anything_v2}
	& \mn
	& \mn
	& lstsq $\dagger$
	& 6.6
	& 38.6
	& 4.0
	& 58.6
	& 4.7
	& 56.5
	& \bestresult{2.6}
	& \bestresult{74.7}
	& 4.5
	& 57.5
	& 4.8
	& 54.1
	& \bestresult{0.05}
	\\ %

    \hline
    \hline
    
    \multicolumn{17}{|l|}{\textbf{d) Depth from frames and poses (w/o per-image range)}}
    \\
    
	DeMoN~\cite{Ummenhofer2016}
	& \my
	& \mn
	& \mn
	& 16.7
	& 13.4
	& 75.0
	& 0.0
	& 19.0
	& 16.2
	& 23.7
	& 11.5
	& 17.6
	& 18.3
	& 30.4
	& 11.9
	& 0.08
	\\ %

	DeepTAM~\cite{Zhou2018}
	& \my
	& \mn
	& \mn
	& 68.7
	& 0.4
	& \trainedsimto{6.7}
	& \trainedsimto{39.7}
	& 20.4
	& 19.8
	& 58.0
	& 9.1
	& 40.0
	& 12.9
	& 38.8
	& 16.4
	& 0.85
	\\ %
	
	DeepV2D \scriptsize{\kitti{}}\normalsize{~\cite{Teed2020Deepv2d}}
	& \my
	& \mn
	& \mn
	& \trainedsimto{{20.4}}
	& \trainedsimto{{16.3}}
	& {25.8}
	& {8.1}
	& {30.1}
	& {9.4}
	& {24.6}
	& {8.2}
	& {38.5}
	& {9.6}
	& {27.9}
	& {10.3}
	& 1.43
	\\ %

	DeepV2D \scriptsize{\scannet{}}\normalsize{~\cite{Teed2020Deepv2d}}
	& \my
	& \mn
	& \mn
	& 61.9
	& 5.2
	& \trainedsimto{\bestresult{3.8}}
	& \trainedsimto{60.2}
	& 18.7
	& 28.7
	& 9.2
	& 27.4
	& 33.5
	& 38.0
	& 25.4
	& 31.9
	& 2.15
	\\ %

	MVSNet\normalsize{~\cite{yao2018mvsnet}}
	& \my
	& \mn
	& \mn
	& 14.0
	& 35.8
	& 1568.0
	& 5.7
	& 507.7
	& 8.3
	& \trainedsimto{4429.1}
	& \trainedsimto{0.1}
	& 118.2
	& 50.7
	& 1327.4
	& 20.1
	& 0.15
	\\ %

	MVSNet \scriptsize{Inv. Depth}\normalsize{~\cite{yao2018mvsnet}}
	& \my
	& \mn
	& \mn
	& 29.6
	& 8.1
	& 65.2
	& 28.5
	& 60.3
	& 5.8
	& \trainedsimto{28.7}
	& \trainedsimto{48.9}
	& 51.4
	& 14.6
	& 47.0
	& 21.2
	& 0.28
	\\ %
	
	CVP-MVSNet\normalsize{~\cite{Yang2020}}
	& \my
	& \mn
	& \mn
	& 158.2
	& 1.2
	& 2289.0
	& 0.1
	& 1735.3
	& 1.2
	& \trainedsimto{8314.0}
	& \trainedsimto{0.0}
	& 415.9
	& 9.5
	& 2582.5
	& 2.4
	& 0.50
	\\ %

	Vis-MVSNet~\cite{Zhang2020}
	& \my
	& \mn
	& \mn
	& 10.3
	& 54.4
	& 84.9
	& 15.6
	& 51.5
	& 17.4
	& \trainedsimto{374.2}
	& \trainedsimto{1.7}
	& 21.1
	& 65.6
	& 108.4
	& 31.0
	& 0.82
	\\ %

	PatchmatchNet~\cite{Wang2020}
	& \my
	& \mn
	& \mn
	& 29.0
	& 16.3
	& 70.1
	& 16.7
	& 99.4
	& 3.5
	& (82.6)
	& (5.6)
	& 39.4
	& 19.3
	& 64.1
	& 12.3
	& 0.18
	\\ %

	Fast-MVSNet~\cite{Yu2020}
	& \my
	& \mn
	& \mn
	& 12.1
	& 37.4
	& 287.1
	& 9.4
	& 131.2
	& 9.6
	& (540.4)
	& (1.9)
	& 33.9
	& 47.2
	& 200.9
	& 21.1
	& 0.35
	\\ %

	MVS2D \scriptsize{\scannet{}}\normalsize{~\cite{Yang2022}}  %
	& \my
	& \mn
	& \mn
	& 73.4
	& 0.0
	& (4.5)
	& (54.1)
	& 30.7
	& 14.4
	& 5.0
	& 57.9
	& 56.4
	& 11.1
	& 34.0
	& 27.5
	& \bestresult{0.05}
	\\ %

	MVS2D \scriptsize{\dtu{}}\normalsize{~\cite{Yang2022}}  %
	& \my
	& \mn
	& \mn
	& 93.3
	& 0.0
	& 51.5
	& 1.6
	& 78.0
	& 0.0
	& (\bestresult{1.6})
	& (92.3)
	& 87.5
	& 0.0
	& 62.4
	& 18.8
	& 0.06
	\\ %

	Robust MVD Baseline \cite{schroeppel2022robust}
	& \my
	& \mn
	& \mn
	& {7.1}
	& 41.9
	& {7.4}
	& {38.4}
	& {9.0}
	& {42.6}
	& {2.7}
	& {82.0}
	& {5.0}
	& {75.1}
	& {6.3}
	& {56.0}
	& 0.06
	\\ %

  	MAST3R (plus our triangulation)
	& \my
	& \mn
	& \mn
	& 3.4
	& 66.6
	& (4.5)
	& (\bestresult{63.0})
	& \bestresult{3.1}
	& \bestresult{72.9}
	& 3.4
	& 67.3
	& (2.4)
	& (83.3)
	& 3.4
	& 70.1
	& 0.72
	\\ %

         \textbf{\modelname} %
	& \my
	& \mn
	& \mn
        &  \bestresult{3.2}
        &  \bestresult{68.8}
        &  \bestresult{3.7}
        &  \bestresult{62.9}
        &  {3.2}
        &  68.0
        &  \bestresult{1.3}
        &  \bestresult{95.0}
        &  \bestresult{2.1}
        &  \bestresult{90.5}
        &  \bestresult{2.7}
        &  \bestresult{77.0}
        &  0.12
        \\

\hline
\end{tabular}

    }
    \vspace{-6pt}
    \caption{
        \textbf{Full results on the RMVDB}.
        See Sec.~4 in the main paper for details of the metrics, baselines and groupings.
        Monocular methods with $\dagger$ are given ground truth intrinsics.
        The best result for each section appears in \bestresult{bold}, and (parentheses) indicates a given model is trained on data from the same dataset.
    }
    \label{tab:robust-multi-view-depth-sup-mat}
    \vspace{-12pt}
\end{table*}

\begin{table*}
    \centering
    \resizebox{1.0\textwidth}{!}{
        \setlength{\tabcolsep}{0.8mm}
        \newcommand{\indentrow}{\hspace{6pt}}
\newcommand{\bigindentrow}{\hspace{40pt}''\hspace{40pt}}
\newcommand{\fullres}{640 $\times$480}
\newcommand{\lowres}{448 $\times$ 336}
\newcommand{\yes}{\checkmark}
\newcommand{\winner}[1]{#1}

\begin{tabular}{|l|l|c|c|c|c|c
|c >{\columncolor{bgcolor}}c
|c >{\columncolor{bgcolor}}c
|c >{\columncolor{bgcolor}}c
|c >{\columncolor{bgcolor}}c
|c >{\columncolor{bgcolor}}c
|c >{\columncolor{bgcolor}}c|
}

\hline
    & \rotatebox{90}{Architecture}
    & \rotatebox{90}{Resolution}
    & \rotatebox{90}{Pretrn.~DAV2 weights}
    & \rotatebox{90}{Metadata}
    & \rotatebox{90}{Noise on GT range}
    & \rotatebox{90}{Bin refinement$\dagger$}
    & \multicolumn{2}{c|}{\winner{\kittishort{}}}
    & \multicolumn{2}{c|}{\winner{\scannetshort{}}}
    & \multicolumn{2}{c|}{\winner{\ethdshort{}}}
    & \multicolumn{2}{c|}{\winner{\dtushort{}}}
    & \multicolumn{2}{c|}{\winner{\tanksandtemplesshort{}}}
    & \multicolumn{2}{c|}{\winner{Average}}
    \\

    & & & & & &
    & $\absrel\downarrow$ & $\threshI\uparrow$
    & $\absrel\downarrow$ & $\threshI\uparrow$
    & $\absrel\downarrow$ & $\threshI\uparrow$
    & $\absrel\downarrow$ & $\threshI\uparrow$
    & $\absrel\downarrow$ & $\threshI\uparrow$
    & $\absrel\downarrow$ & $\threshI\uparrow$ 
    \\
    \hline
    \hline

    Ours full, high-res & Ours & \fullres & \yes  & \yes & \yes & \yes         &  \winner{3.2}
        &  68.8
        &  3.7
        &  62.9
        &  3.2
        &  68.0
        &  1.3
        &  95.0
        &  2.1
        &  90.5
        &  2.7
        &  77.0 \\

    Ours w/o metadata, high-res
        & Ours 
        &  \fullres
        & \yes
        &  
        &  \yes
        & \yes
        &  3.3
        &  68.3
        &  3.7
        &  63.8
        &  3.1
        &  69.3
        &  1.9
        &  93.5
        &  2.1
        &  89.9
        &  2.8
        &  77.0
        \\
    
    \hline 
    
    \hline 
    \hline
    \winner{a) Ablate Mono/Multi Cue Combiner} \\
    
    \indentrow \colorbox{yellow!30}{Ours w/o metadata} 
        & Ours 
        & \lowres 
        &  \yes
        &
        &  \yes
        &  \yes
        &  \winner{3.39}
        &  \winner{66.88}
        &  \winner{3.86}
        &  \winner{60.82}
        &  \winner{3.11}
        &  \winner{70.17}
        &  2.43
        &  \winner{92.05}
        &  \winner{2.23}
        &  \winner{88.38}
        &  \winner{3.00}
        &  \winner{75.66}
        \\

    \bigindentrow  no MMCC ViT
        & w/o MMCC ViT 
        &  \lowres 
        & \yes
        & 
        & \yes
        &\yes
        &  3.54
        &  65.32
        &  4.39
        &  56.94
        &  3.56
        &  65.27
        &  3.07
        &  90.49
        &  2.46
        &  87.54
        &  3.41
        &  73.11
        \\

    \bigindentrow w/ naive patchify 
        & Naive patchify 
        &  \lowres 
        & \yes
        &
        & \yes
        &\yes 
        &  3.66
        &  62.61
        &  4.27
        &  58.86
        &  3.18
        &  67.27
        &  \winner{1.95}
        &  91.69
        &  2.46
        &  86.52
        &  3.11
        &  73.39
        \\ \hline

    \winner{b) Ablate metadata and normalization} \\

    \indentrow Ours (low res.)
        & Ours 
        &  \lowres
        & \yes
        &  \yes
        &  \yes
        & \yes
        &  3.31
        &  67.63
        &  \winner{4.05}
        &  \winner{61.27}
        &  3.29
        &  66.66
        &  2.51
        &  93.15
        &  2.26
        &  88.66
        &  3.08
        &  75.47
        \\

    \indentrow w/o normalized metadata w/o view count agnostic
        & Ours 
        & \lowres
        & \yes
        & \yes
        & \yes
        & \yes
        & \winner{3.39}
        & \winner{66.23}
        & 3.97
        & 61.23
        & \winner{3.10 }
        & \winner{68.84}
        & \winner{2.25}
        & \winner{91.98}
        & \winner{2.24}
        & \winner{88.20}
        & \winner{2.99}
        & \winner{75.30}
        \\

    \indentrow \colorbox{yellow!30}{Ours w/o metadata} 
        & Ours 
        & \lowres 
        &  \yes
        &
        &  \yes
        &  \yes
        &  3.39
        &  66.88
        &  3.86
        &  60.82
        &  \winner{3.11}
        &  \winner{70.17}
        &  \winner{2.43}
        &  92.05
        &  2.23
        &  88.38
        &  3.00
        &  75.66
        \\

    \hline

    \winner{c) Ablate cost volume} \\

    \indentrow \colorbox{yellow!30}{Ours w/o metadata} 
        & Ours 
        & \lowres 
        &  \yes
        &
        &  \yes
        &  \yes
        &  3.39
        &  \winner{66.88}
        &  \winner{3.86}
        &  60.82
        &  \winner{3.11}
        &  \winner{70.17}
        &  \winner{2.43}
        &  \winner{92.05}
        &  \winner{2.23}
        &  \winner{88.38}
        &  \winner{3.00}
        &  \winner{75.66}
        \\

    \bigindentrow w/o noise on GT range
        & Ours 
        &  \lowres
        & \yes
        & 
        & 
        & \yes
        &   \winner{3.33}
        &   {66.83}
        &  5.14
        &  52.77
        &  3.53
        &  66.32
        & 13.45  
        & 89.21
        & 2.34
        & 87.64
        &  5.32
        &   {74.89}
        \\

    \bigindentrow w/o bin refinement
        & Ours 
        &  \lowres 
        & \yes
        & 
        &  \yes
        &
        &  3.57
        &  63.39
        &  5.18
        &  51.05
        &  3.50
        &  67.93
        &  6.80
        &  82.12
        &  {2.27}
        &  87.04
        &  4.26
        &  70.31
        \\

    \bigindentrow w/ fixed bins [0-100m] 
        & Ours 
        &  \lowres
        & \yes
        & 
        & \yes
        & \winner{F}
        &  3.41
        &  64.33
        &  {3.80}
        &  \winner{61.62}
        &  {3.15}
        &  67.20
        &  4.11
        &  65.64
        &  2.36
        &  85.72
        &  3.37
        &  68.90
        \\

    \hline

    \winner{d) Ablate overall architecture} \\

    \indentrow \colorbox{yellow!30}{Ours w/o metadata} 
        & Ours 
        & \lowres 
        &  \yes
        &
        &  \yes
        &  \yes
        &  \winner{3.39}
        &  \winner{66.88}
        &  \winner{3.86}
        &  \winner{60.82}
        &  \winner{3.11}
        &  \winner{70.17}
        &  {2.43}
        &  {92.05}
        &  \winner{2.23}
        &  \winner{88.38}
        &  \winner{3.00}
        &  \winner{75.66}
        \\

    \bigindentrow w/o DAV2 weights
        & Ours 
        &  \lowres
        & 
        & 
        & \yes
        &\yes
        &  3.45
        &  65.42
        &  4.58
        &  57.14
        &  3.48
        &  65.59
        &  2.14
        &  \winner{92.48}
        &  2.56
        &  86.01
        &  3.24
        &  73.33
        \\

    \bigindentrow w/ ViT Small
        & ViT Small 
        & \lowres
        & \yes
        & 
        &  \yes
        & \yes
        &  3.57
        &  64.34
        &  4.40
        &  56.57
        &  3.63
        &  64.61
        &  2.69
        &  91.52
        &  2.71
        &  84.51
        &  3.40
        &  72.31
        \\

    \bigindentrow w/ \cite{sayed2022simplerecon}'s architecture 
        & From \cite{sayed2022simplerecon} 
        & 512 $\times$ 384
        & \yes
        & 
        & \yes
        & \yes
        &  3.63
        &  65.94
        &  5.03
        &  51.76
        &  3.74
        &  63.09
        &  \winner{1.77}
        &  90.97
        &  2.78
        &   {87.90}
        &  3.39
        &  71.93
        \\
\hline
\end{tabular}

    }
    \vspace{-8pt}
    \caption{\textbf{Ablation on RMVDB.} 
        We validate our design decisions on RMVDB~\cite{schroeppel2022robust} by ablating various components. 
        Except otherwise indicated, our ablations are run on a low-resolution version of our network, trained without metadata (for tractability reasons). 
        The row with \textbf{F} uses a fixed set of bins (0-100m) for all images.
    }
    \label{tab:robust-multi-view-depth-ablation-sup-mat}
\end{table*}

\begin{table*}
    \newcommand{\indentrow}{\hspace{6pt}}
    \newcommand{\bigindentrow}{\hspace{40pt}''\hspace{40pt}}
    
    \centering
    {\fontsize{8pt}{10pt}\selectfont

        \setlength{\tabcolsep}{0.8mm}

\newcommand{\fullres}{640 $\times$448}
\newcommand{\lowres}{448 $\times$ 336}
\newcommand{\yes}{\checkmark}

\resizebox{0.9\textwidth}{!}{
\begin{tabular}{|l|l|c|c|c|c|c
|c >{\columncolor{bgcolor}}c
|c >{\columncolor{bgcolor}}c
|c >{\columncolor{bgcolor}}c
|c >{\columncolor{bgcolor}}c
|c >{\columncolor{bgcolor}}c
|c >{\columncolor{bgcolor}}c|
}

\hline
    & \rotatebox{90}{Architecture}
    & \rotatebox{90}{Resolution}
    & \rotatebox{90}{Pretrained DAV2 weights}
    & \rotatebox{90}{Metadata}
    & \rotatebox{90}{Noise on GT range}
    & \rotatebox{90}{Bin refinement$\dagger$}
    & \multicolumn{2}{c|}{\textbf{\scannetshort{}}}
    & \multicolumn{2}{c|}{\textbf{\ethdshort{}}}
    \\

    & & & & & &

            & $\absrel\downarrow$ & $\threshI\uparrow$
            & $\absrel\downarrow$ & $\threshI\uparrow$
            \\
            \hline
            \hline

            Ours full, high-res
            & Ours 
            & \fullres 
            & \yes  
            & \yes 
            & \yes 
            & \yes
            & 3.22
            & 69.45
            & 1.27
            & 93.24
            \\

        Ours w/o metadata, high-res
            & Ours 
            & \fullres
            & \yes
            & 
            & \yes
            & \yes
            & 3.28
            & 69.31
            & 1.25
            & 93.45
            \\ 

        \hline
        
        \textbf{a) Ablate cost volume patchifier}
        \\

        \indentrow \colorbox{yellow!30}{Ours w/o metadata} 
            & Ours 
            & \lowres
            & \yes
            & 
            & \yes
            & \yes
            &  3.22
            &  69.16
            &  1.52
            &  91.02
            \\

         \bigindentrow no MMCC ViT
            & w/o MMCC ViT 
            &  \lowres 
            & \yes
            & 
            & \yes
            &\yes
            & 3.42
            & 67.06
            & 1.75
            & 89.62
            \\

        \bigindentrow w/ naive patchify 
            & Naive patchify 
            &  \lowres 
            & \yes
            &
            & \yes
            &\yes 
            & 3.40
            & 68.05
            & 1.59
            & 90.06
            \\ 
            
        \hline

        \textbf{b) Ablate metadata and normalization} \\

        \indentrow Ours (low res)
            & Ours 
            & \lowres
            & \yes
            & \yes
            & \yes
            & \yes
            &   3.30
            &   69.70
            &  1.46
            &  91.65
            \\

        \bigindentrow  w/o normalized metadata w/o view count agnostic
            & Ours 
            & \lowres
            & \yes
            & \yes
            & \yes
            & \yes
            & 3.28
            & 68.87
            & 1.47
            & 91.30
            \\

        \hline 
        \textbf{c) Ablate cost volume}
        \\
        
        \indentrow \colorbox{yellow!30}{Ours w/o metadata} 
            & Ours 
            & \lowres
            & \yes
            & 
            & \yes
            & \yes
            &  3.22
            &  69.16
            &  1.52
            &  91.02
            \\ 
            
        \bigindentrow w/o noise on GT range
            & Ours 
            & \lowres
            & \yes
            & 
            & 
            & \yes
            & 3.93
            & 64.47
            & 1.86 
            & 90.36
            \\
    
        \bigindentrow w/o bin refinement
            & Ours 
            &  \lowres 
            & \yes
            & 
            &  \yes
            &
            & 4.13
            & 59.53
            & 1.92
            & 86.63
            \\

        \bigindentrow w/ fixed bins [0-100m] 
            & Ours 
            &  \lowres
            & \yes
            & 
            & \yes
            & \textbf{F}
            & 3.27
            & 69.03
            & 1.96
            & 86.78
            \\

        \hline

        \textbf{d) Ablate overall architecture} \\
        
        \indentrow \colorbox{yellow!30}{Ours w/o metadata} 
            & Ours 
            & \lowres
            & \yes
            & 
            & \yes
            & \yes
            &  3.22
            &  69.16
            &  1.52
            &  91.02
            \\ 
            
        \bigindentrow w/o DAV2 weights
            & Ours 
            &  \lowres
            & 
            & 
            & \yes
            &\yes
            &  3.33
            &  67.94
            &  1.71
            &  89.13
            \\

        \bigindentrow w/ ViT Small
            & ViT Small 
            & \lowres
            & \yes
            & 
            &  \yes
            & \yes
            &  3.43
            &  66.98
            & 1.89
            & 87.90
            \\
    
        \bigindentrow w/ \cite{sayed2022simplerecon}'s architecture 
            & From \cite{sayed2022simplerecon} 
            & 512 $\times$ 384
            & \yes
            & 
            & \yes
            & \yes
            & 4.02
            & 62.37
            & 1.57
            & 91.57
            \\
            \hline
    
        \end{tabular}
    }
    }
    \vspace{-5pt}
    \caption{\textbf{Ablation  \benchmarkname Variant}. We use better test-time tuples for ScanNet, and for ETH3D we use the undistorted test images. 
    \label{tab:alternative-benchmark-sup-mat}
    }
    \vspace{-3pt}
\end{table*}

\section{Gaussian Splat Regularization using MVSAnywhere}

\begin{figure*}
    \resizebox{1.0\textwidth}{!}{
    \includegraphics[]{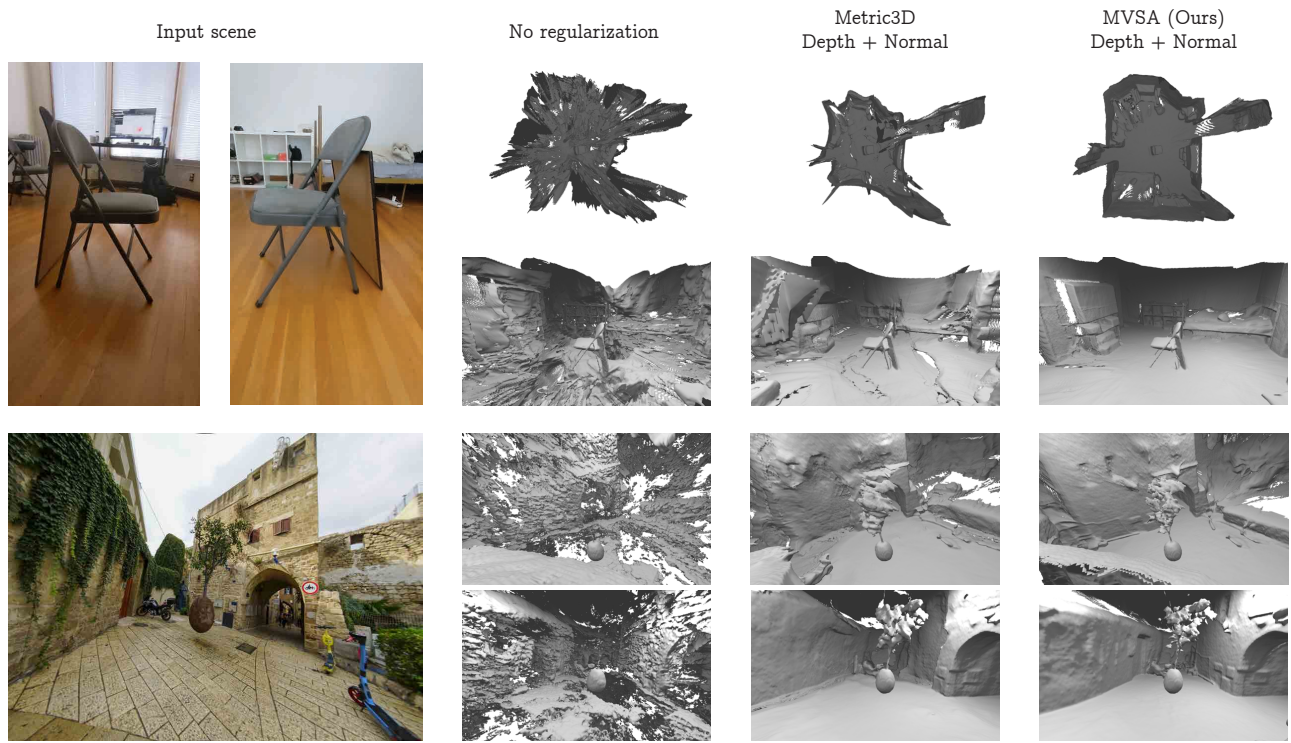}
    }
    \caption{
        \textbf{Qualitative comparison of Gaussian Splat meshes using Metric3D and MVSA as regularizers.} 
    }
    \label{fig:regsplatfacto}
\end{figure*}

We show a use case of our method as a regularizer for Gaussian Splats. We follow a similar approach as VCR-GauS~\cite{chen2024vcr} and DN-Splatter~\cite{turkulainen2024dn} regularizing depth and normals during training as additional losses. In Figure~\ref{fig:regsplatfacto}, we show qualitative results of the meshes obtained after using raw Splatfacto without regularization, and with normal and depth supervision from Metric3D~\cite{hu2024metric3d} and from MVSA. Note that we used an scale-invariant loss for Metric3D given that the used scenes lay in arbitrary scales. More details on this regularization are available on the code.

\section{Additional qualitative results}
\begin{figure*}
    \resizebox{1.0\textwidth}{!}{
    \newcommand{\turnheightnew}{60pt}
    \begin{tabular}{@{\hskip -2mm}c@{\hskip 1mm}c@{\hskip 1mm}c@{\hskip 1mm}c@{\hskip 1mm}c@{\hskip 1mm}c@{\hskip 1mm}c@{}}
        & KITTI & ScanNet & ETH3D & DTU & {T\&T} \\
    
        {\rotatebox{90}{\hspace{5mm}{\small RGB ($I_r$)}}} &
        \includegraphics[height=\turnheightnew, trim=4cm 0cm 5cm 0cm, clip]{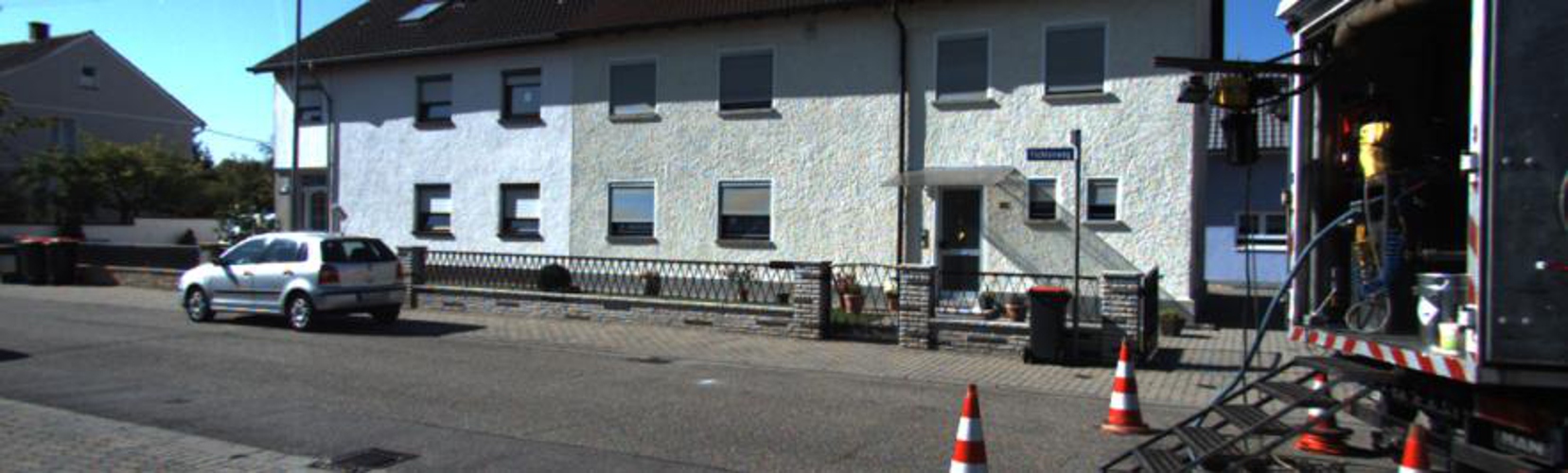} &
        \includegraphics[height=\turnheightnew]{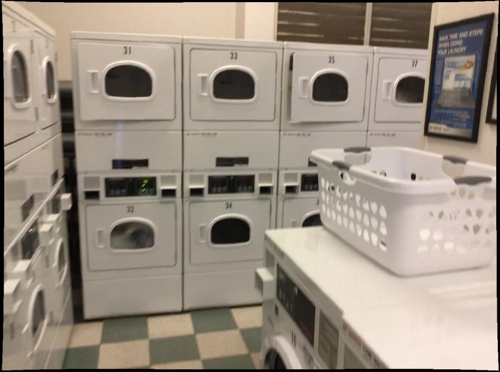} &
        \includegraphics[height=\turnheightnew]{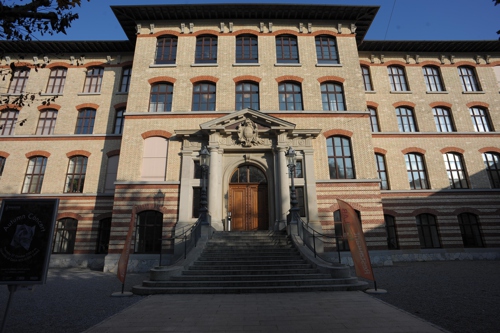} &
        \includegraphics[height=\turnheightnew]{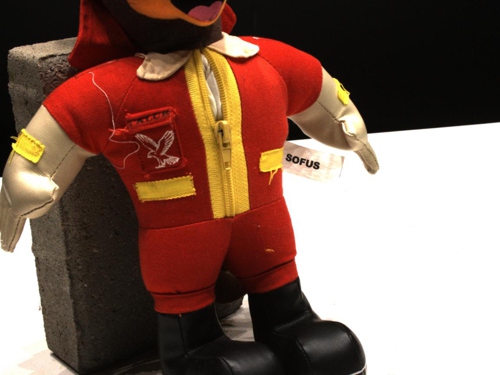} &
        \includegraphics[height=\turnheightnew]{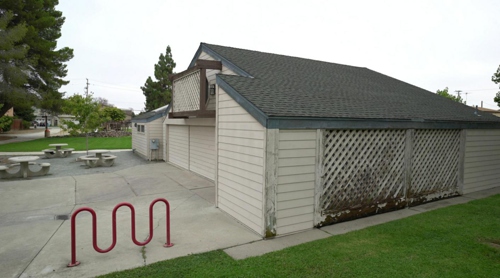} \\
        
        {\rotatebox{90}{\hspace{1mm}{\small Depth Pro~\protect\cite{bochkovskii2024depthpro}}}} &
        \includegraphics[height=\turnheightnew, trim=4cm 0cm 5cm 0cm, clip]{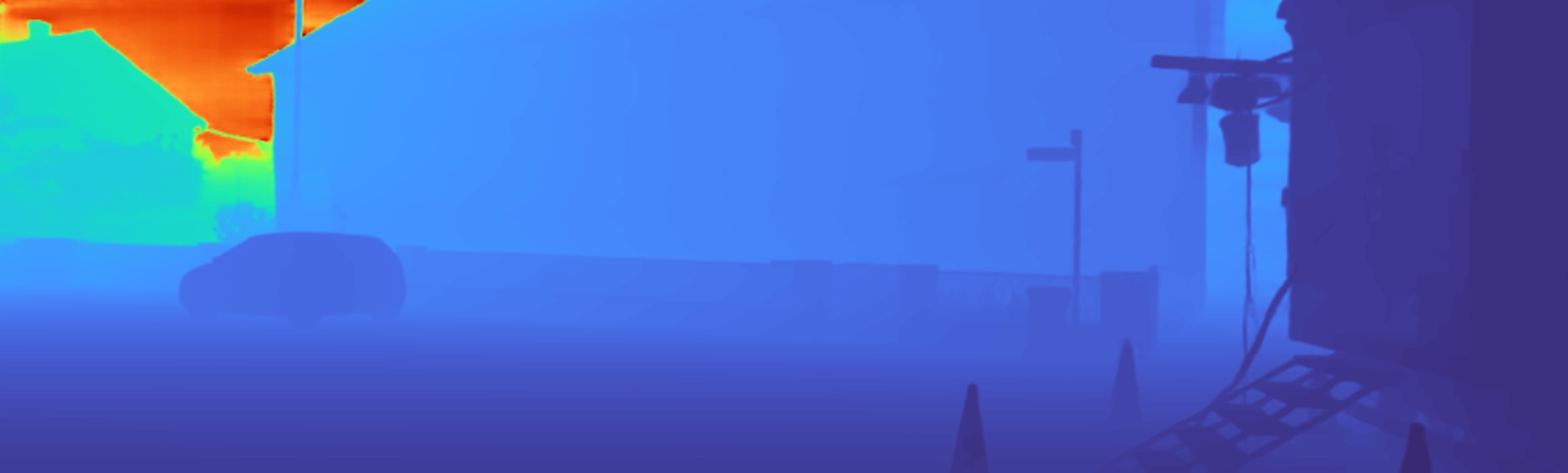} &
        \includegraphics[height=\turnheightnew]{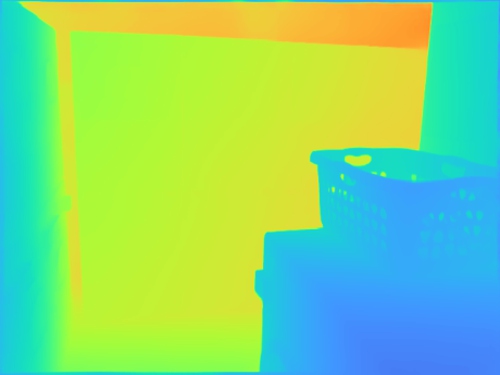} &
        \includegraphics[height=\turnheightnew]{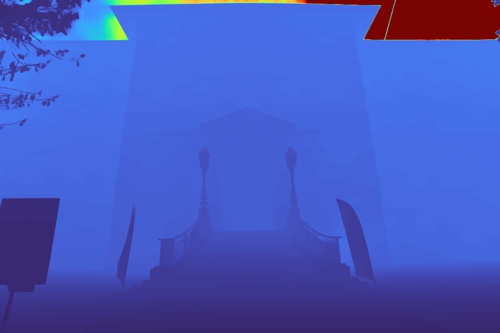} &
        \includegraphics[height=\turnheightnew]{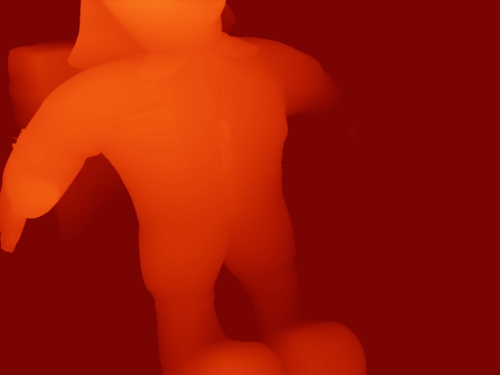} &
        \includegraphics[height=\turnheightnew]{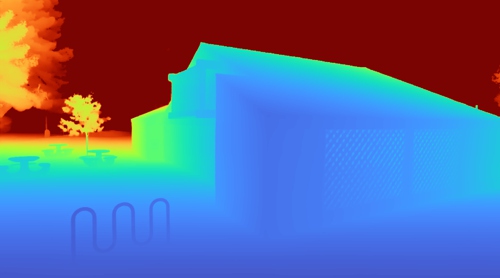} \\
        
        {\rotatebox{90}{\hspace{4mm}{\small rMVD~\protect\cite{schroeppel2022robust}}}} &
        \includegraphics[height=\turnheightnew, trim=4cm 0cm 5cm 0cm, clip]{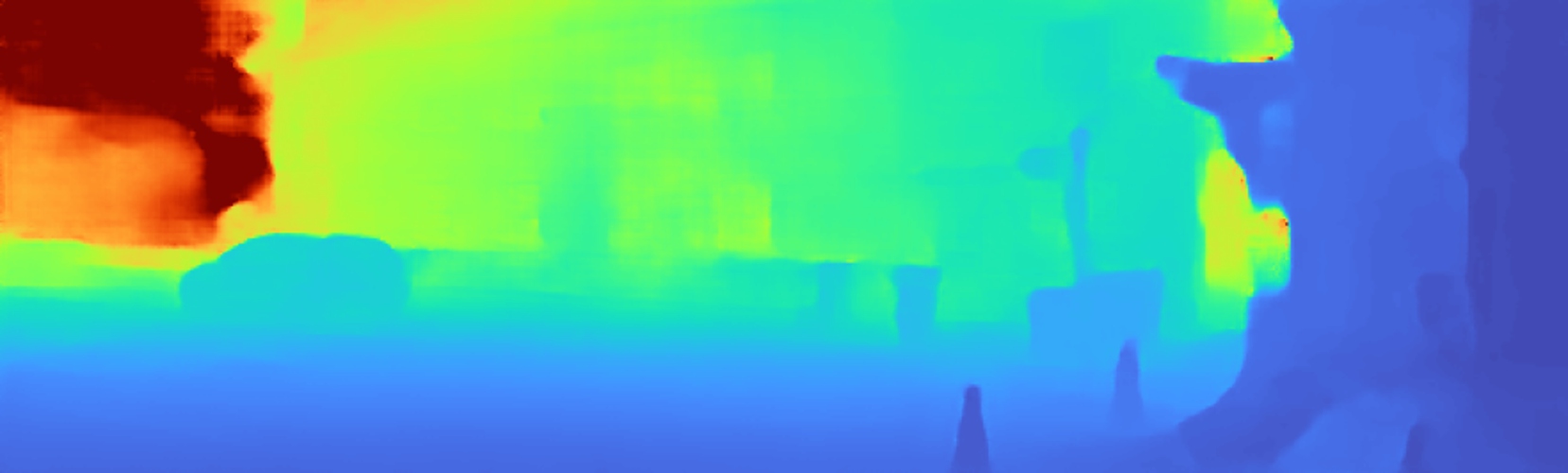} &
        \includegraphics[height=\turnheightnew]{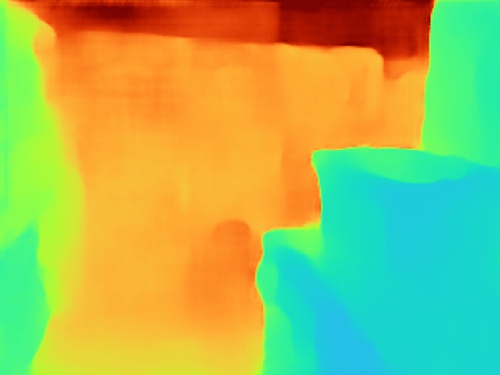} &
        \includegraphics[height=\turnheightnew]{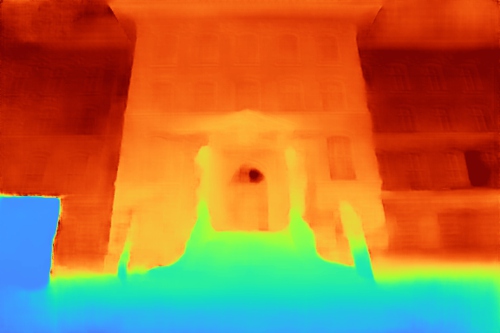} &
        \includegraphics[height=\turnheightnew]{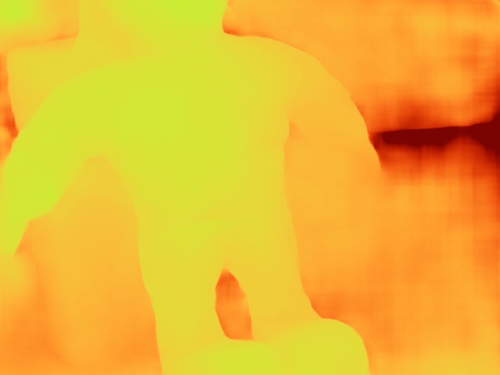} &
        \includegraphics[height=\turnheightnew]{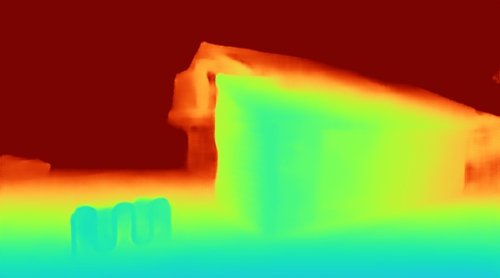} \\

        {\rotatebox{90}{\hspace{1mm}{\small MAST3R~\protect\cite{mast3r_arxiv24}}}} &
        \includegraphics[height=\turnheightnew, trim=4cm 0cm 5cm 0cm, clip]{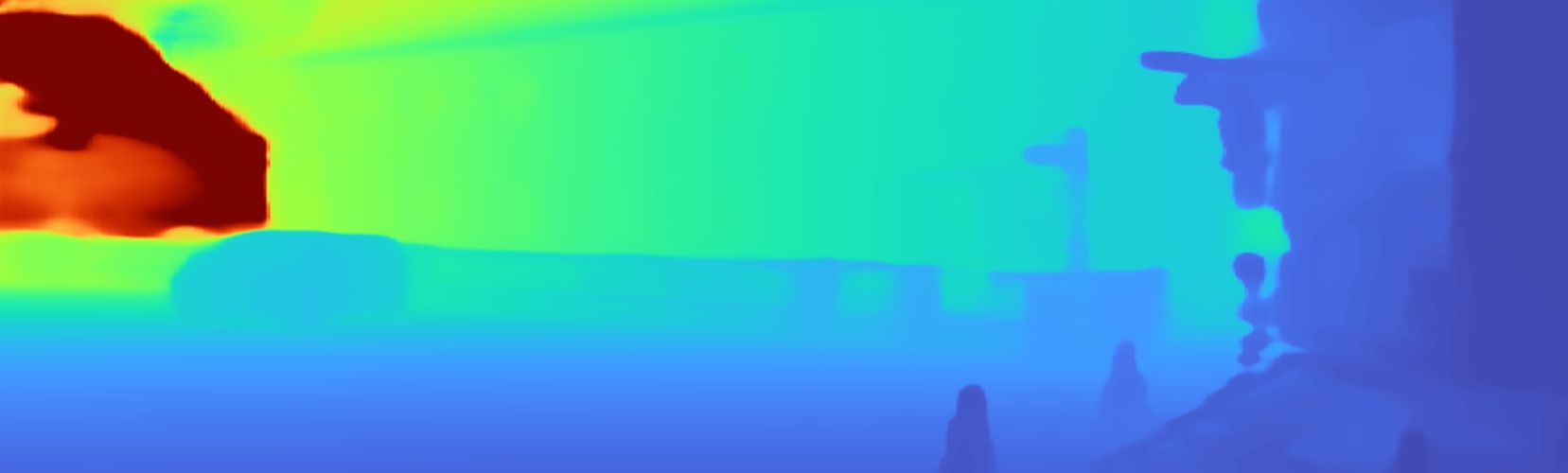} &
        \includegraphics[height=\turnheightnew]{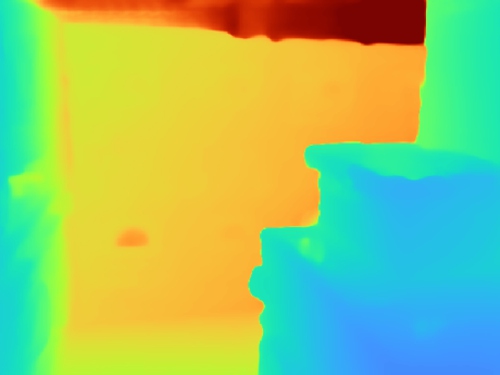} &
        \includegraphics[height=\turnheightnew]{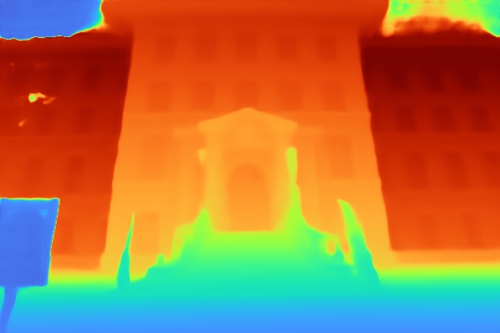} &
        \includegraphics[height=\turnheightnew]{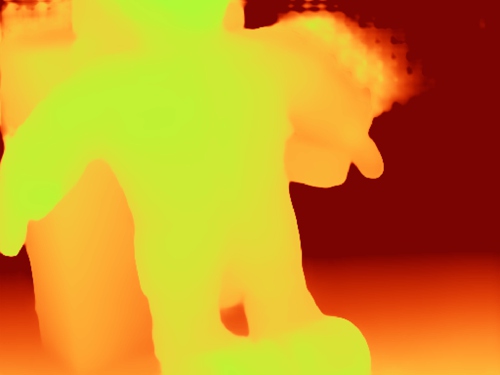} &
        \includegraphics[height=\turnheightnew]{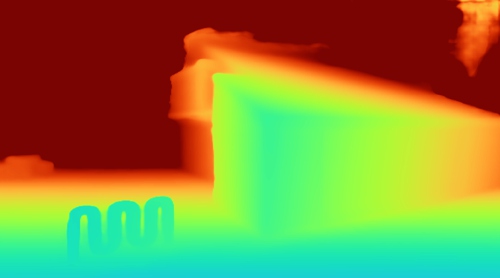} \\
        
        {\rotatebox{90}{\hspace{0.5mm}{\small MVSA (Ours)}}} &
        \includegraphics[height=\turnheightnew, trim=4cm 0cm 5cm 0cm, clip]{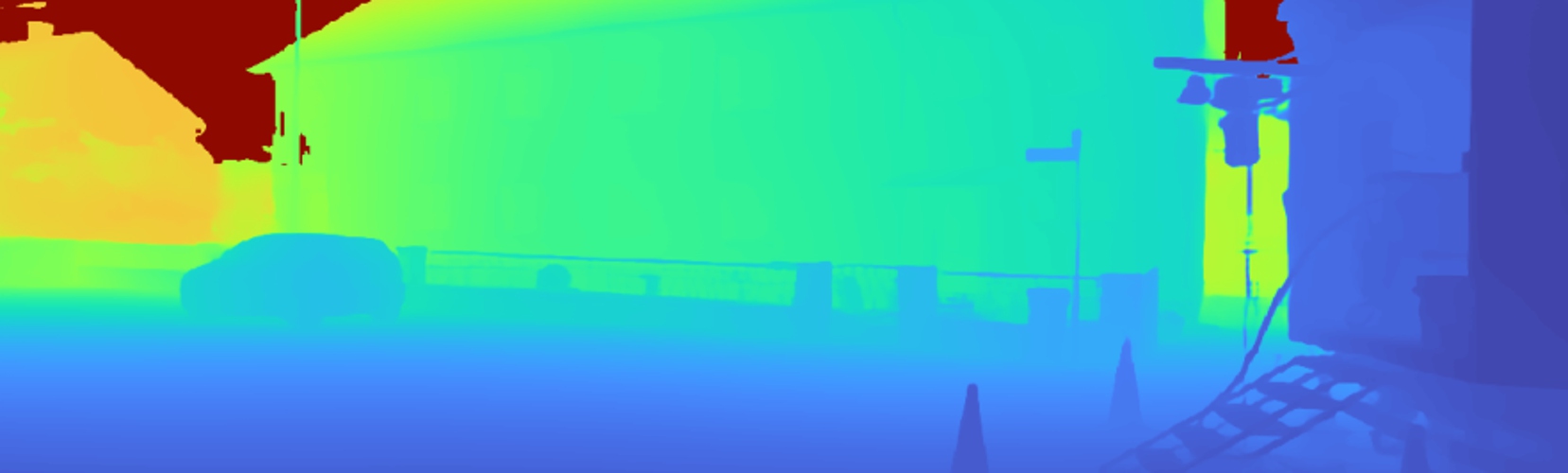} &
        \includegraphics[height=\turnheightnew]{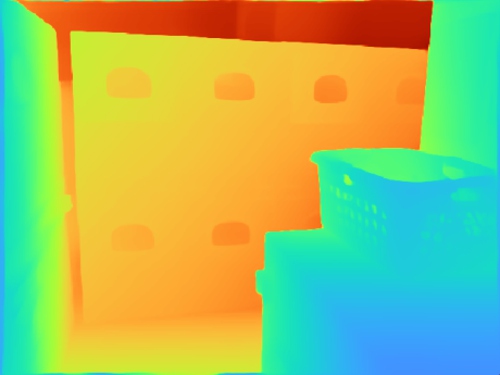} &
        \includegraphics[height=\turnheightnew]{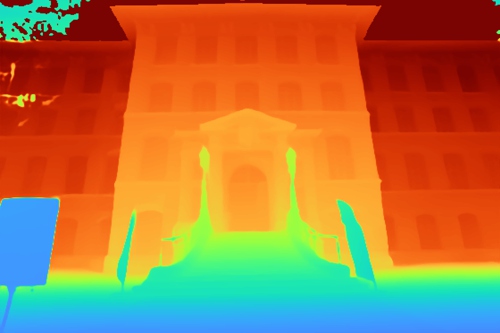} &
        \includegraphics[height=\turnheightnew]{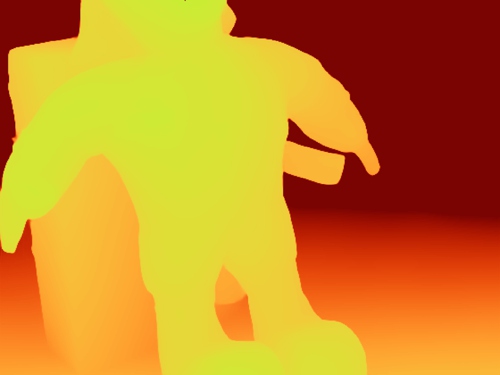} &
        \includegraphics[height=\turnheightnew]{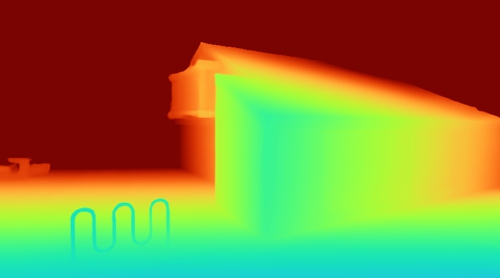} \\

        {\rotatebox{90}{\hspace{7mm}{\small GT}}} &
        \includegraphics[height=\turnheightnew, trim=4cm 0cm 5cm 0cm, clip]{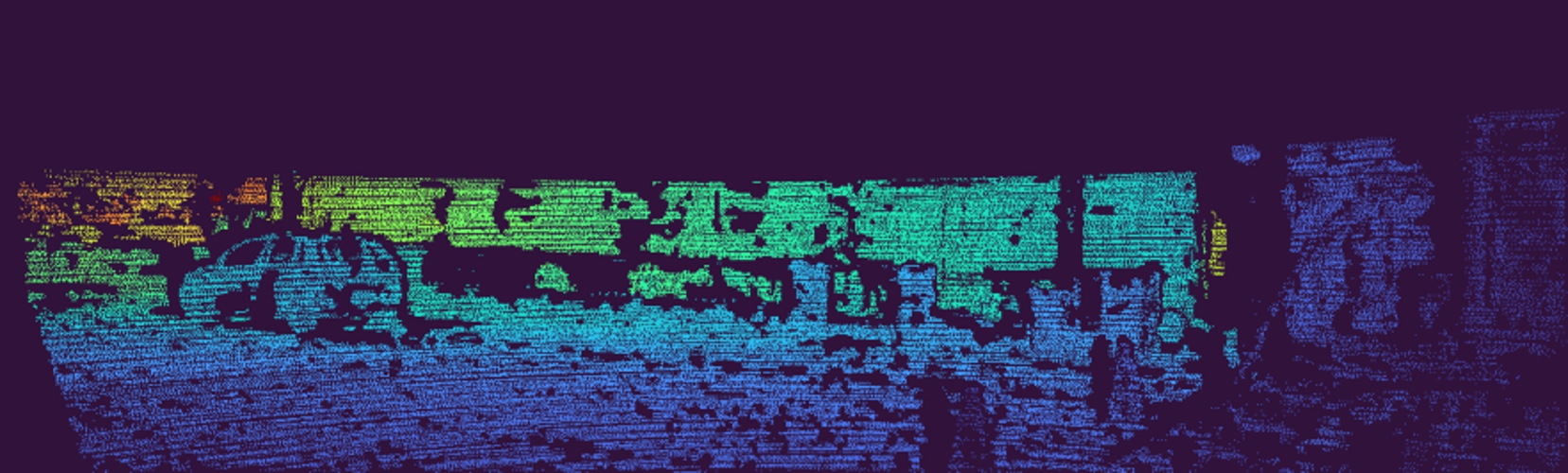} &
        \includegraphics[height=\turnheightnew]{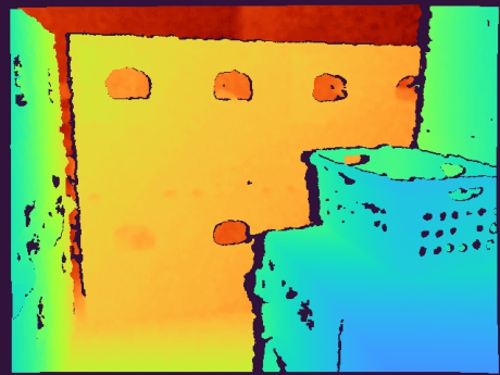} &
        \includegraphics[height=\turnheightnew]{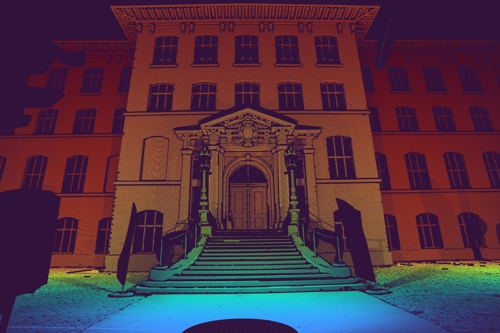} &
        \includegraphics[height=\turnheightnew]{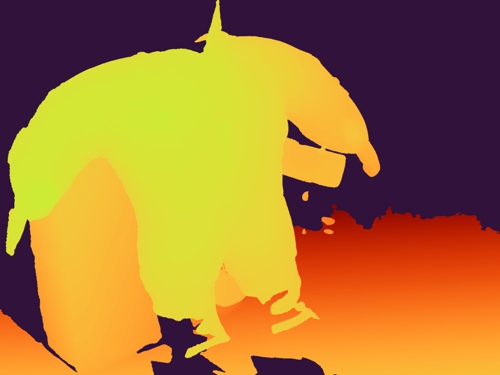} &
        \includegraphics[height=\turnheightnew]{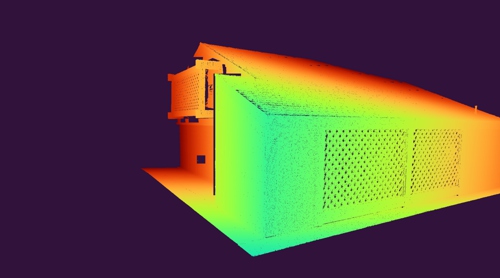}  \\
    \end{tabular}
    }
    \vspace{-10pt}
    \caption{
        \textbf{Qualitative comparison of depth prediction results across multiple datasets (Normalized). This is a duplicate from main paper, for completeness.} (KITTI, ScanNet, ETH3D, DTU, and Tanks \& Temples). 
        Rows show different methods: Depth Pro~\cite{schroeppel2022robust}, rMVD baseline~\cite{schroeppel2022robust}, MAST3R (Triangulated)~\cite{mast3r_arxiv24}, and our MVSA model, along with RGB inputs ($I_r$) and ground-truth depths (GT). 
        Depth Pro provides sharp edges but often misestimates depth scale, while our MVSA model captures finer details than MAST3R and rMVD. 
        Depth maps are normalized to ground truth depth range for consistent visualization.
    }
    \label{fig:qualitative_depths_main}
\end{figure*}

\begin{figure*}
    \resizebox{1.0\textwidth}{!}{
    \newcommand{\turnheightnew}{60pt}
    \begin{tabular}{@{\hskip -2mm}c@{\hskip 1mm}c@{\hskip 1mm}c@{\hskip 1mm}c@{\hskip 1mm}c@{\hskip 1mm}c@{\hskip 1mm}c@{}}
        & KITTI & ScanNet & ETH3D & DTU & {T\&T} \\
    
        {\rotatebox{90}{\hspace{5mm}{\small RGB ($I_r$)}}} &
        \includegraphics[height=\turnheightnew, trim=4cm 0cm 5cm 0cm, clip]{figs_jpeg/qualitative/supplementary/main/kitti/color.jpg} &
        \includegraphics[height=\turnheightnew]{figs_jpeg/qualitative/supplementary/main/scannet/color.jpg} &
        \includegraphics[height=\turnheightnew]{figs_jpeg/qualitative/supplementary/main/eth3d/color.jpg} &
        \includegraphics[height=\turnheightnew]{figs_jpeg/qualitative/supplementary/main/dtu/color.jpg} &
        \includegraphics[height=\turnheightnew]{figs_jpeg/qualitative/supplementary/main/tanks_and_temples/color.jpg} \\
        
        {\rotatebox{90}{\hspace{1mm}{\small Depth Pro~\protect\cite{bochkovskii2024depthpro}}}} &
        \includegraphics[height=\turnheightnew, trim=4cm 0cm 5cm 0cm, clip]{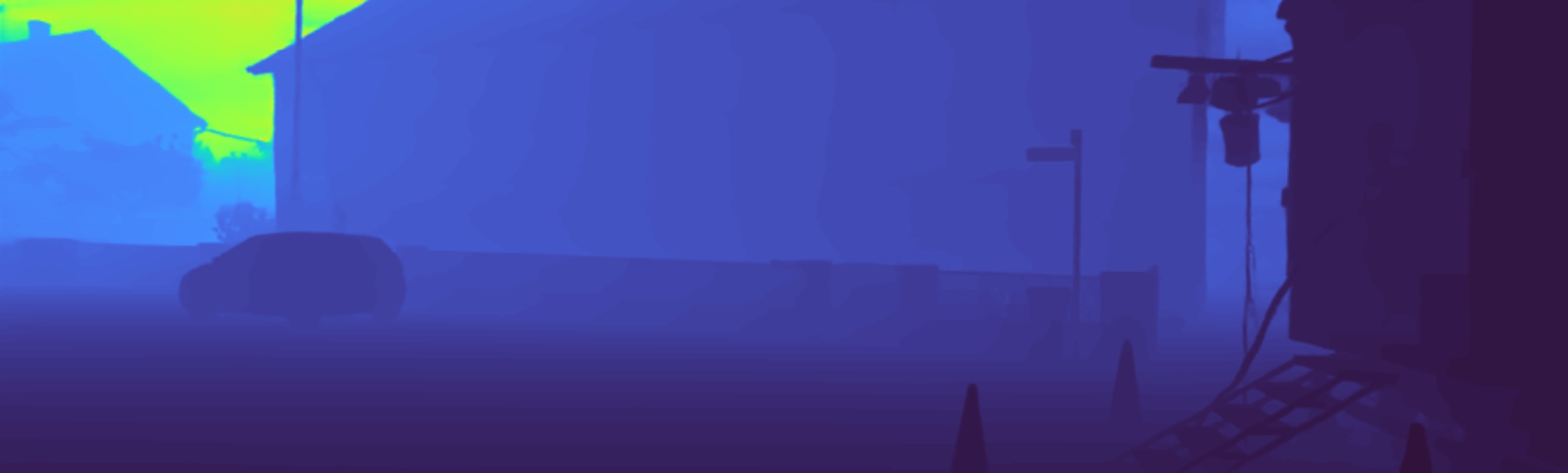} &
        \includegraphics[height=\turnheightnew]{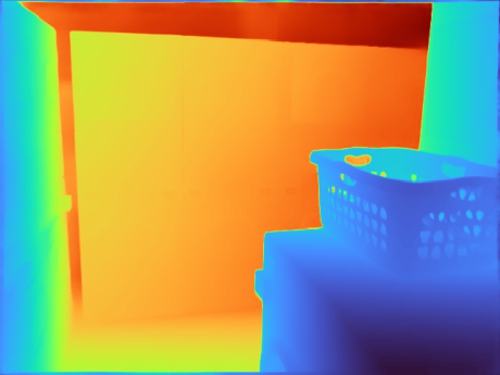} &
        \includegraphics[height=\turnheightnew]{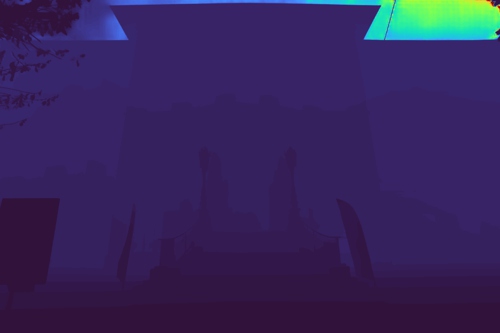} &
        \includegraphics[height=\turnheightnew]{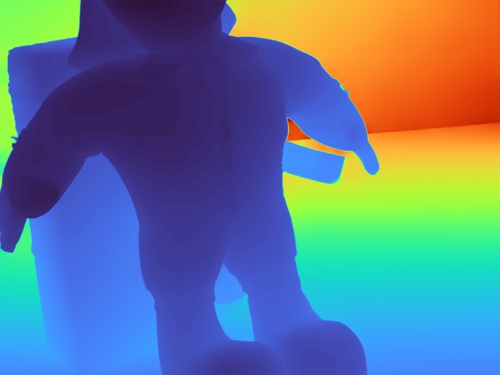} &
        \includegraphics[height=\turnheightnew]{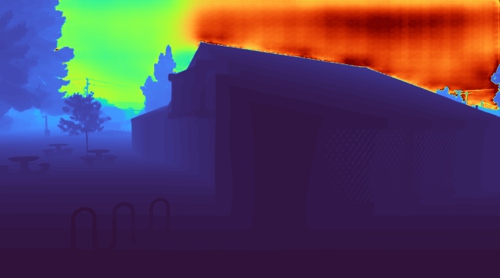} \\
        
        {\rotatebox{90}{\hspace{4mm}{\small rMVD~\protect\cite{schroeppel2022robust}}}} &
        \includegraphics[height=\turnheightnew, trim=4cm 0cm 5cm 0cm, clip]{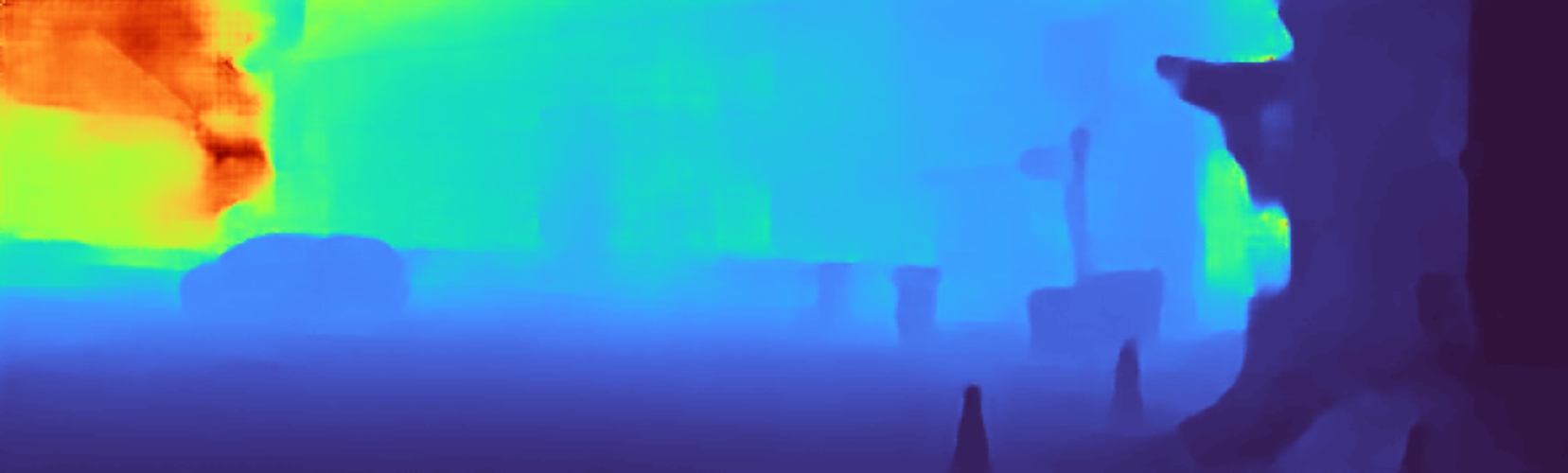} &
        \includegraphics[height=\turnheightnew]{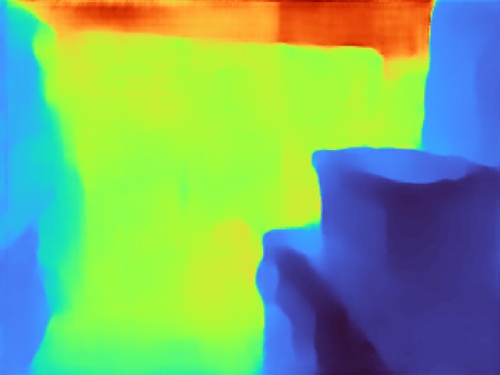} &
        \includegraphics[height=\turnheightnew]{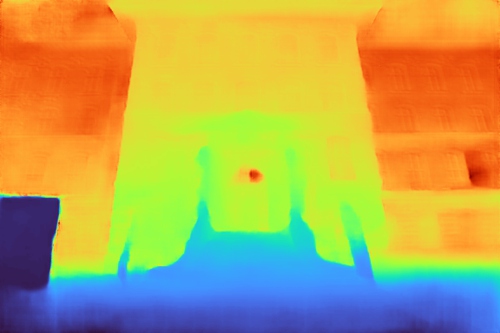} &
        \includegraphics[height=\turnheightnew]{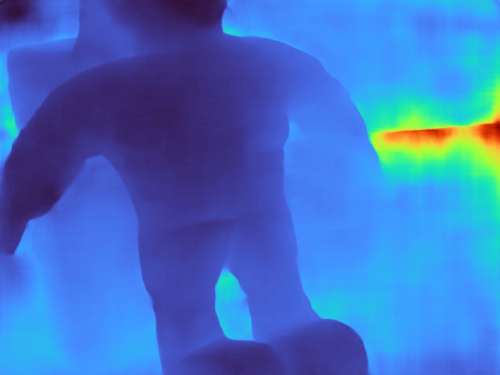} &
        \includegraphics[height=\turnheightnew]{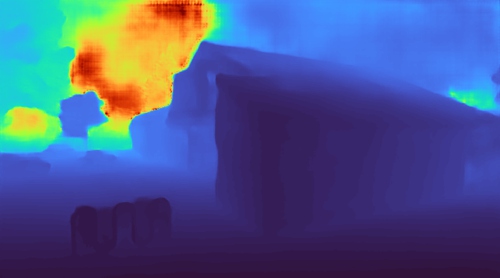} \\

        {\rotatebox{90}{\hspace{1mm}{\small MAST3R~\protect\cite{mast3r_arxiv24}}}} &
        \includegraphics[height=\turnheightnew, trim=4cm 0cm 5cm 0cm, clip]{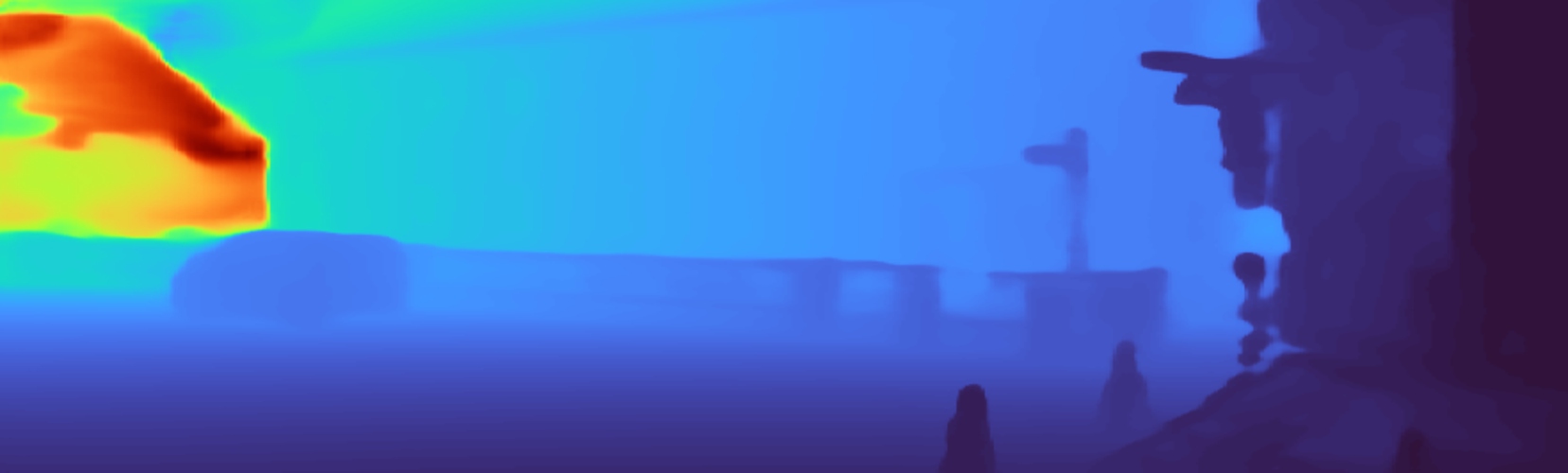} &
        \includegraphics[height=\turnheightnew]{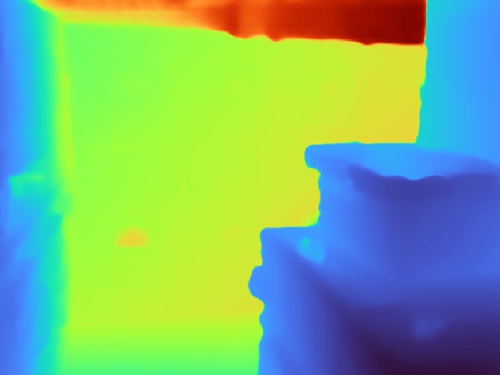} &
        \includegraphics[height=\turnheightnew]{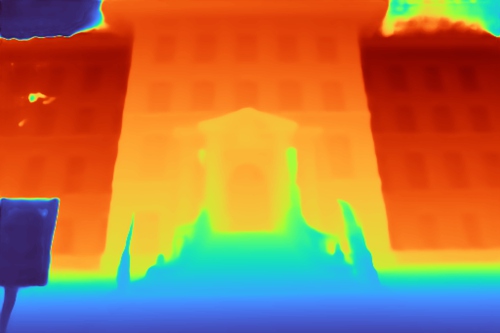} &
        \includegraphics[height=\turnheightnew]{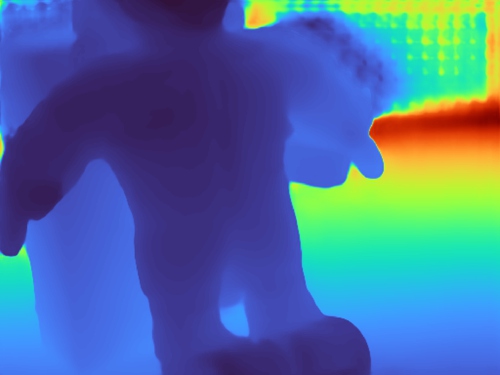} &
        \includegraphics[height=\turnheightnew]{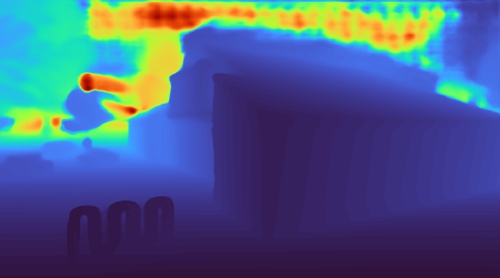} \\
        
        {\rotatebox{90}{\hspace{0.5mm}{\small MVSA (Ours)}}} &
        \includegraphics[height=\turnheightnew, trim=4cm 0cm 5cm 0cm, clip]{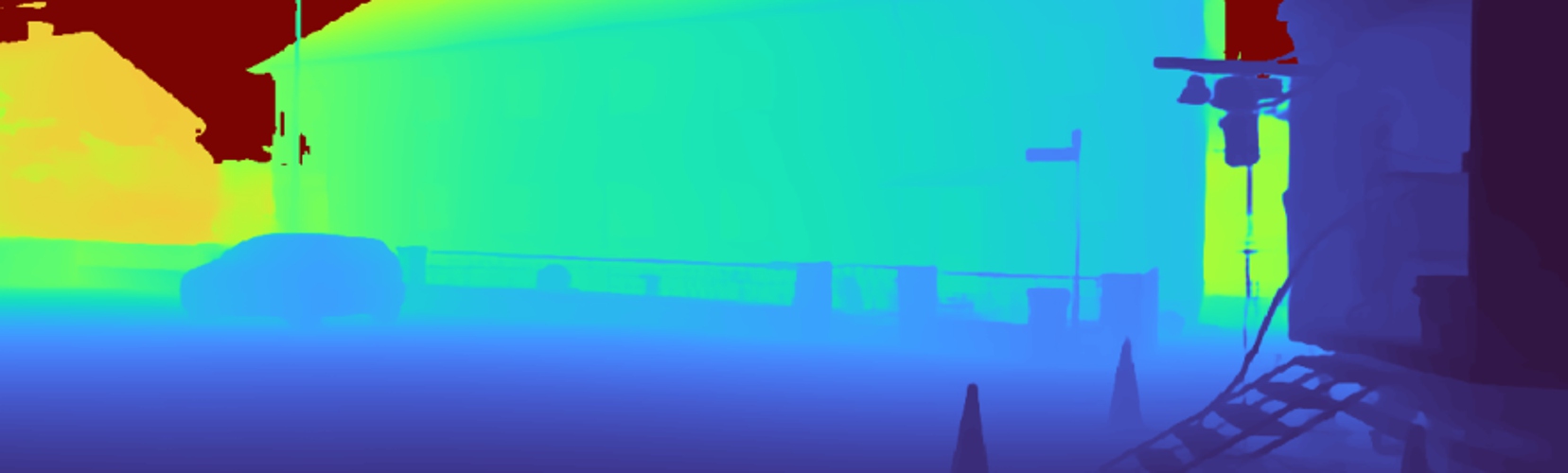} &
        \includegraphics[height=\turnheightnew]{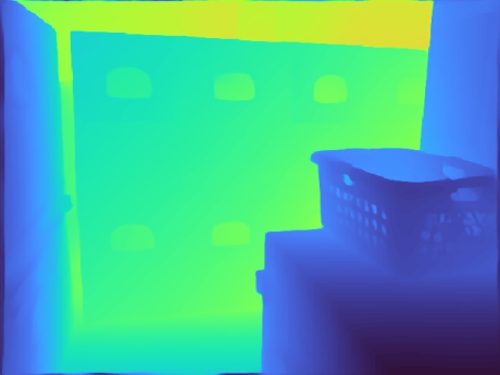} &
        \includegraphics[height=\turnheightnew]{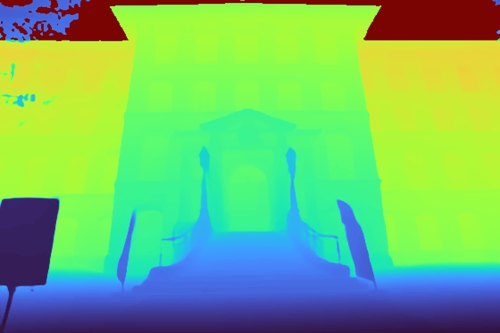} &
        \includegraphics[height=\turnheightnew]{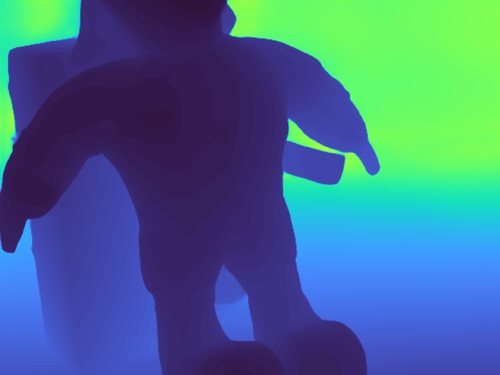} &
        \includegraphics[height=\turnheightnew]{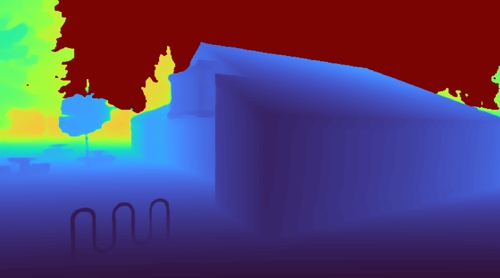} \\

        {\rotatebox{90}{\hspace{7mm}{\small GT}}} &
        \includegraphics[height=\turnheightnew, trim=4cm 0cm 5cm 0cm, clip]{figs_jpeg/qualitative/supplementary/main/kitti/depth_gt.jpg} &
        \includegraphics[height=\turnheightnew]{figs_jpeg/qualitative/supplementary/main/scannet/depth_gt.jpg} &
        \includegraphics[height=\turnheightnew]{figs_jpeg/qualitative/supplementary/main/eth3d/depth_gt.jpg} &
        \includegraphics[height=\turnheightnew]{figs_jpeg/qualitative/supplementary/main/dtu/depth_gt.jpg} &
        \includegraphics[height=\turnheightnew]{figs_jpeg/qualitative/supplementary/main/tanks_and_temples/depth_gt.jpg}  \\
    \end{tabular}
    }
    \vspace{-10pt}
    \caption{
        \textbf{Qualitative comparison of depth prediction results across multiple datasets (Unnormalized)} (KITTI, ScanNet, ETH3D, DTU, and Tanks \& Temples). 
        Rows show different methods: Depth Pro~\cite{schroeppel2022robust}, rMVD baseline~\cite{schroeppel2022robust}, MAST3R (Triangulated)~\cite{mast3r_arxiv24}, and our MVSA model, along with RGB inputs ($I_r$) and ground-truth depths (GT). 
        Depth maps are normalized per image.
    }
    \label{fig:qualitative_depth_unnorms_sup_main}
\end{figure*}

\begin{figure*}
    \resizebox{1.0\textwidth}{!}{
    \newcommand{\turnheightnew}{60pt}
    \begin{tabular}{@{\hskip -2mm}c@{\hskip 1mm}c@{\hskip 1mm}c@{\hskip 1mm}c@{\hskip 1mm}c@{\hskip 1mm}c@{\hskip 1mm}c@{}}
        & KITTI & ScanNet & ETH3D & DTU & {T\&T} \\
    
        {\rotatebox{90}{\hspace{5mm}{\small RGB ($I_r$)}}} &
        \includegraphics[height=\turnheightnew, trim=4cm 0cm 5cm 0cm, clip]{figs_jpeg/qualitative/supplementary/main/kitti/color.jpg} &
        \includegraphics[height=\turnheightnew]{figs_jpeg/qualitative/supplementary/main/scannet/color.jpg} &
        \includegraphics[height=\turnheightnew]{figs_jpeg/qualitative/supplementary/main/eth3d/color.jpg} &
        \includegraphics[height=\turnheightnew]{figs_jpeg/qualitative/supplementary/main/dtu/color.jpg} &
        \includegraphics[height=\turnheightnew]{figs_jpeg/qualitative/supplementary/main/tanks_and_temples/color.jpg} \\
        
        {\rotatebox{90}{\hspace{1mm}{\small Depth Pro~\protect\cite{bochkovskii2024depthpro}}}} &
        \includegraphics[height=\turnheightnew, trim=4cm 0cm 5cm 0cm, clip]{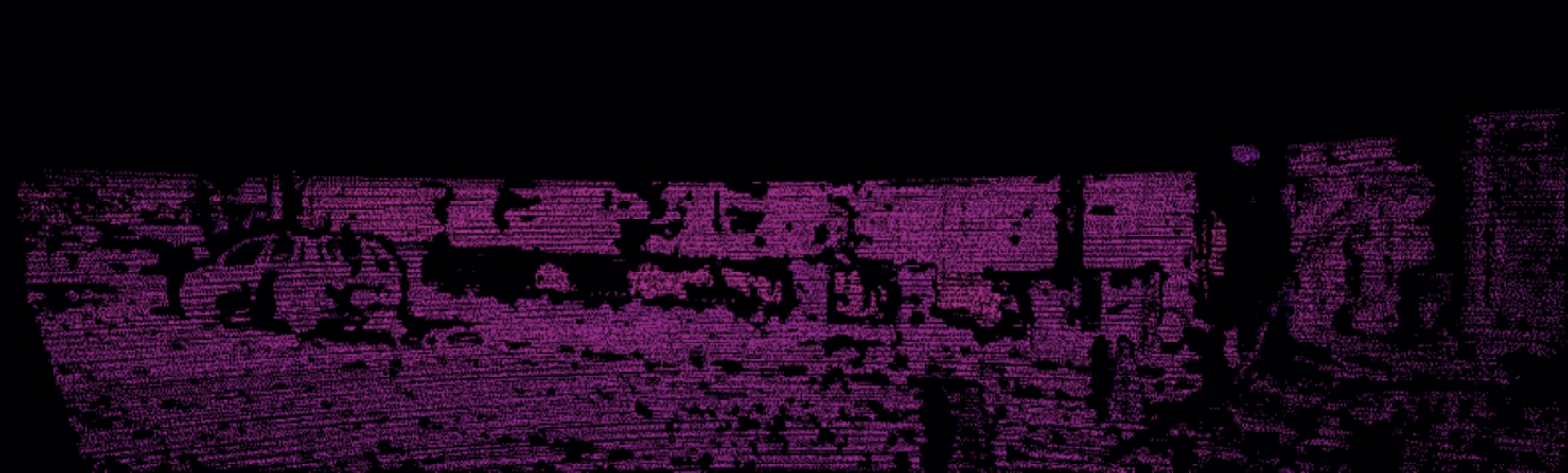} &
        \includegraphics[height=\turnheightnew]{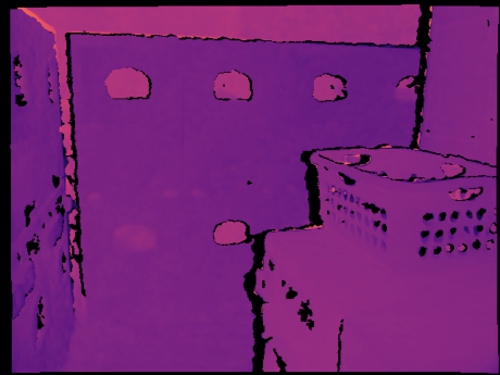} &
        \includegraphics[height=\turnheightnew]{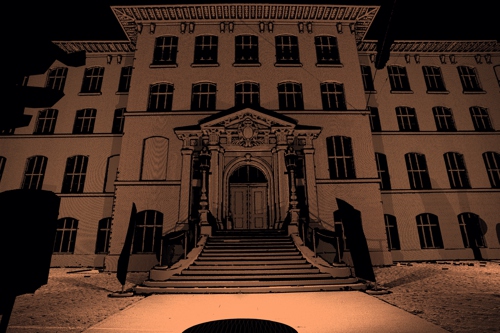} &
        \includegraphics[height=\turnheightnew]{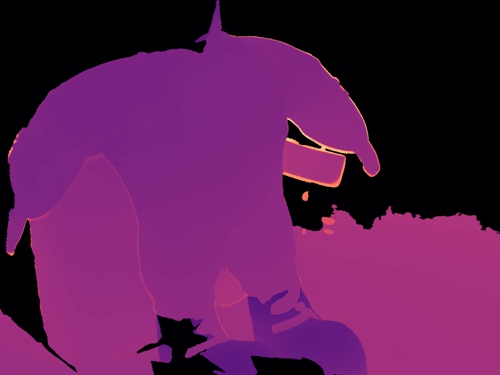} &
        \includegraphics[height=\turnheightnew]{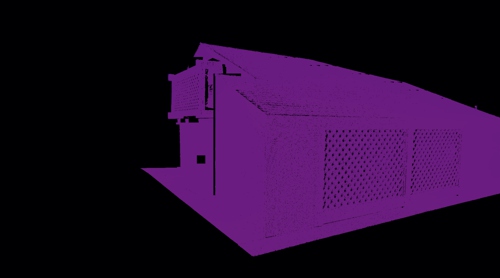} \\
        
        {\rotatebox{90}{\hspace{4mm}{\small rMVD~\protect\cite{schroeppel2022robust}}}} &
        \includegraphics[height=\turnheightnew, trim=4cm 0cm 5cm 0cm, clip]{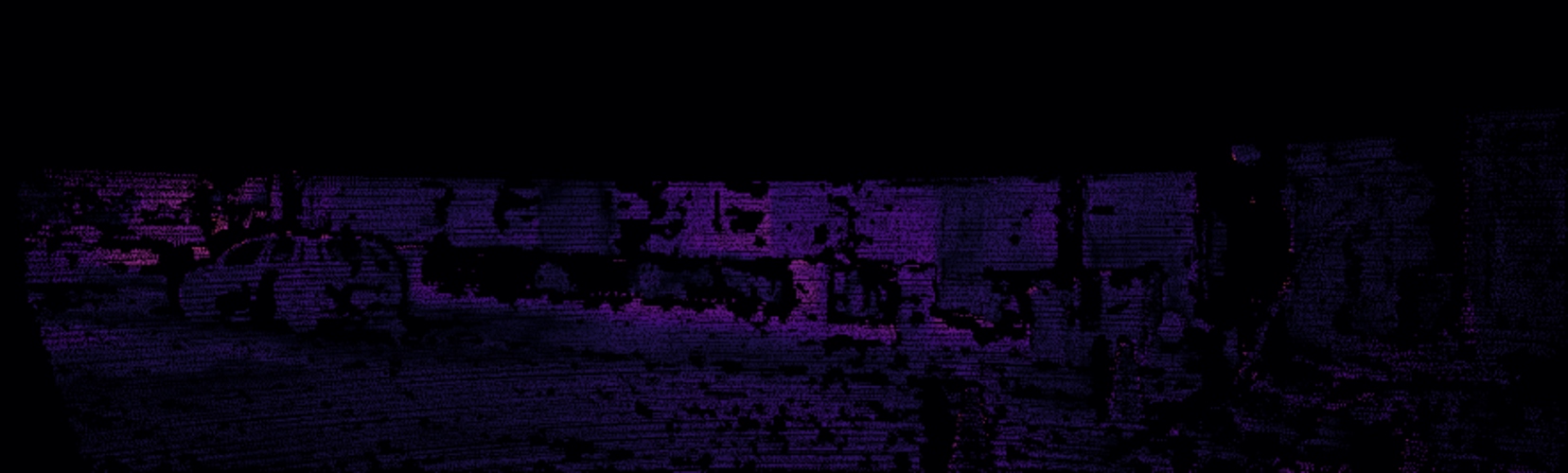} &
        \includegraphics[height=\turnheightnew]{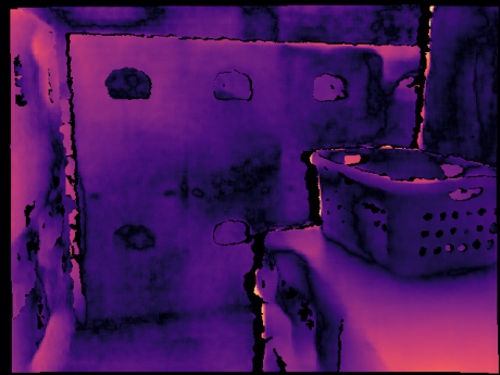} &
        \includegraphics[height=\turnheightnew]{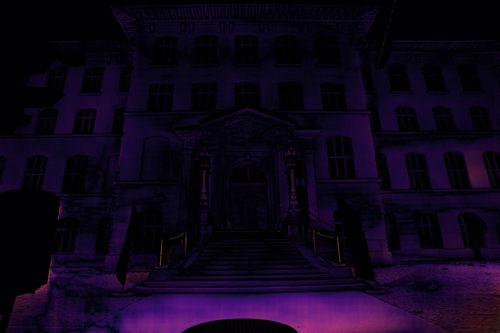} &
        \includegraphics[height=\turnheightnew]{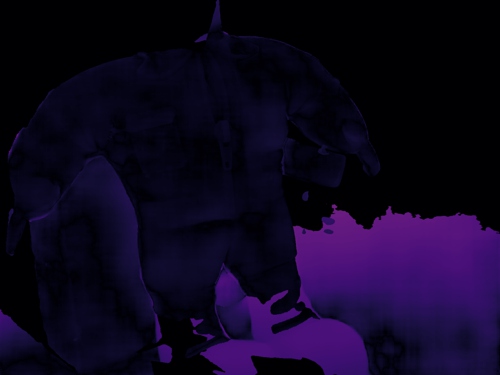} &
        \includegraphics[height=\turnheightnew]{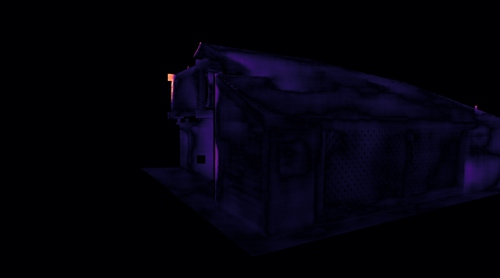} \\

        {\rotatebox{90}{\hspace{1mm}{\small MAST3R~\protect\cite{mast3r_arxiv24}}}} &
        \includegraphics[height=\turnheightnew, trim=4cm 0cm 5cm 0cm, clip]{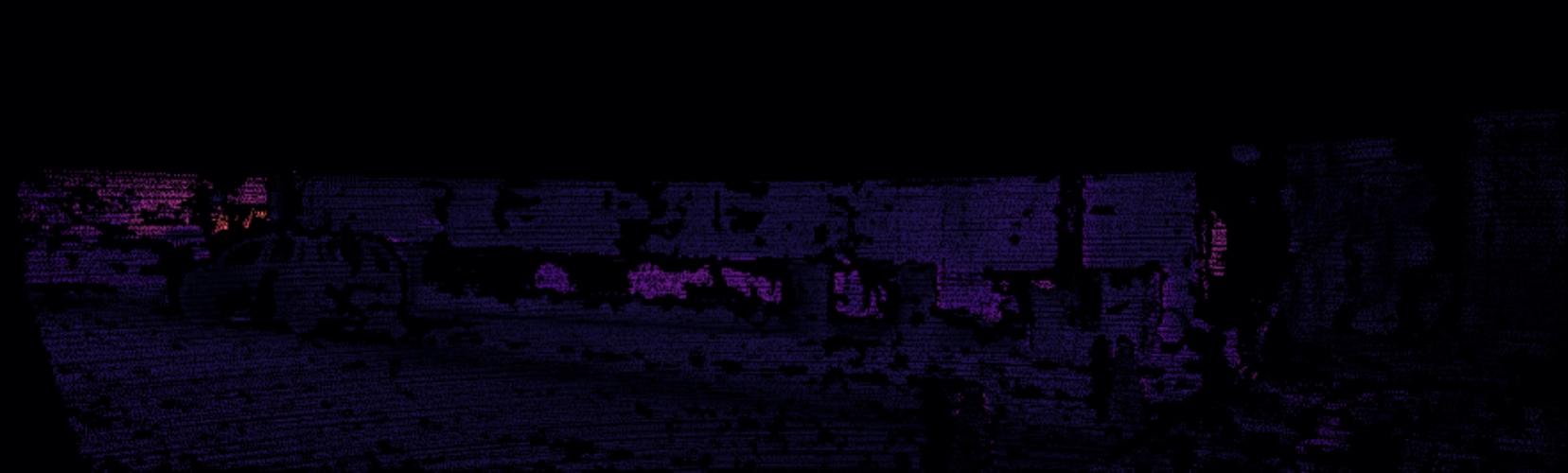} &
        \includegraphics[height=\turnheightnew]{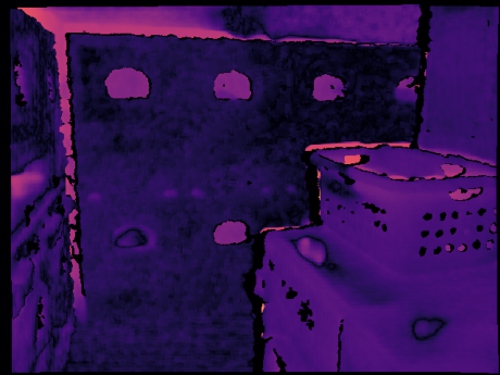} &
        \includegraphics[height=\turnheightnew]{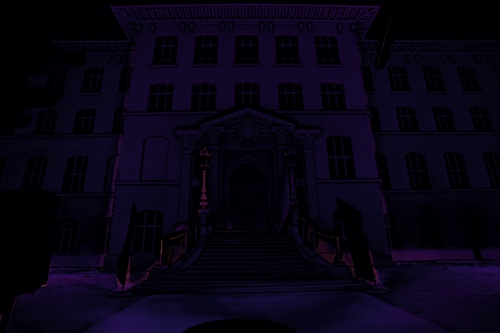} &
        \includegraphics[height=\turnheightnew]{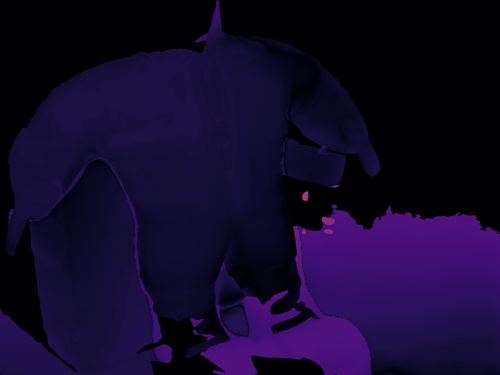} &
        \includegraphics[height=\turnheightnew]{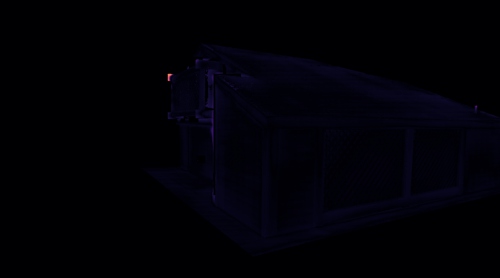} \\
        
        {\rotatebox{90}{\hspace{0.5mm}{\small MVSA (Ours)}}} &
        \includegraphics[height=\turnheightnew, trim=4cm 0cm 5cm 0cm, clip]{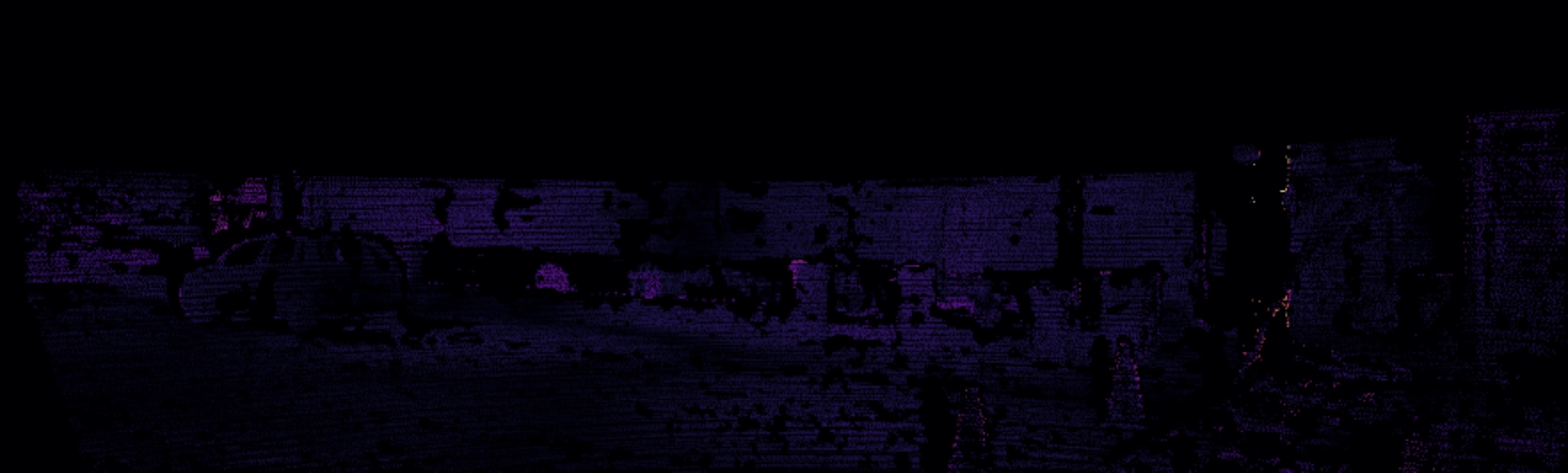} &
        \includegraphics[height=\turnheightnew]{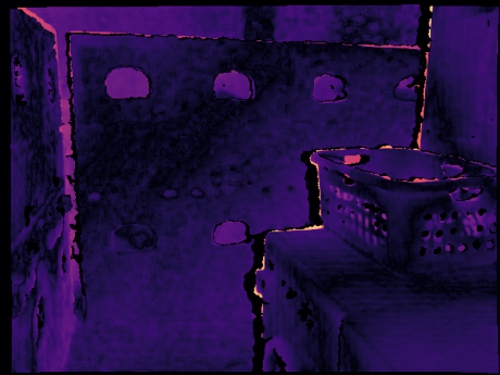} &
        \includegraphics[height=\turnheightnew]{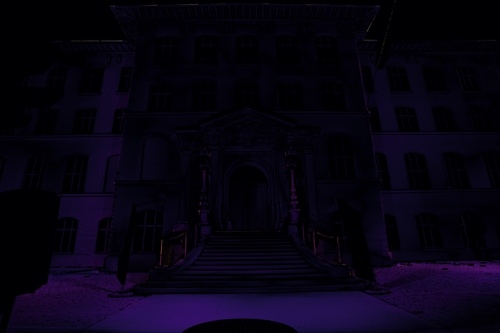} &
        \includegraphics[height=\turnheightnew]{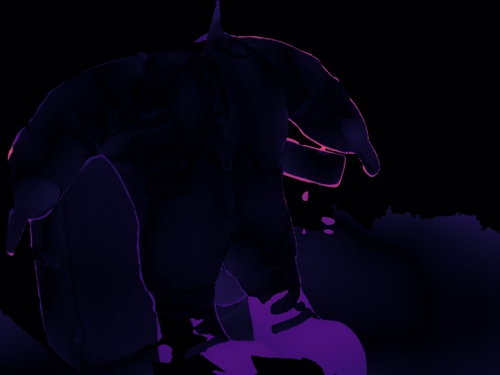} &
        \includegraphics[height=\turnheightnew]{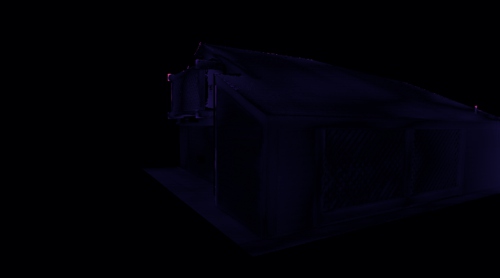} \\

        {\rotatebox{90}{\hspace{5mm}{\small Depth GT}}} &
        \includegraphics[height=\turnheightnew, trim=4cm 0cm 5cm 0cm, clip]{figs_jpeg/qualitative/supplementary/main/kitti/depth_gt.jpg} &
        \includegraphics[height=\turnheightnew]{figs_jpeg/qualitative/supplementary/main/scannet/depth_gt.jpg} &
        \includegraphics[height=\turnheightnew]{figs_jpeg/qualitative/supplementary/main/eth3d/depth_gt.jpg} &
        \includegraphics[height=\turnheightnew]{figs_jpeg/qualitative/supplementary/main/dtu/depth_gt.jpg} &
        \includegraphics[height=\turnheightnew]{figs_jpeg/qualitative/supplementary/main/tanks_and_temples/depth_gt.jpg}  \\
    \end{tabular}
    }
    \vspace{-10pt}
    \caption{
        \textbf{Qualitative comparison of depth prediction errors across multiple datasets} (KITTI, ScanNet, ETH3D, DTU, and Tanks \& Temples). 
        Rows show different methods: Depth Pro~\cite{schroeppel2022robust}, rMVD baseline~\cite{schroeppel2022robust}, MAST3R (Triangulated)~\cite{mast3r_arxiv24}, and our MVSA model, along with RGB inputs ($I_r$) and ground-truth depths (GT). 
        Depth Pro provides sharp edges but often misestimates depth scale, while our MVSA model captures finer details than MAST3R and rMVD. 
        Error maps are normalized to maximum error among methods for each scene.
    }
    \label{fig:qualitative_errors_sup_main}
\end{figure*}

\begin{figure*}
    \resizebox{1.0\textwidth}{!}{
    \newcommand{\turnheightnew}{60pt}
    \begin{tabular}{@{\hskip -2mm}c@{\hskip 1mm}c@{\hskip 1mm}c@{\hskip 1mm}c@{\hskip 1mm}c@{\hskip 1mm}c@{\hskip 1mm}c@{}}
        & KITTI & ScanNet & ETH3D & DTU & {T\&T} \\
    
        {\rotatebox{90}{\hspace{5mm}{\small RGB ($I_r$)}}} &
        \includegraphics[height=\turnheightnew, trim=4cm 0cm 5cm 0cm, clip]{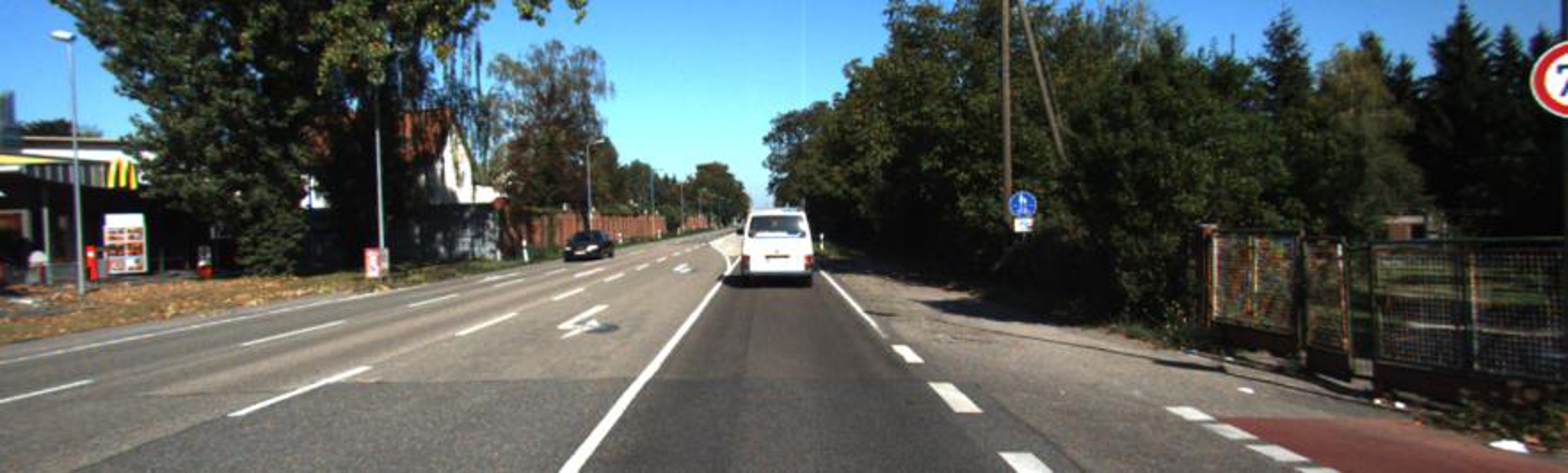} &
        \includegraphics[height=\turnheightnew]{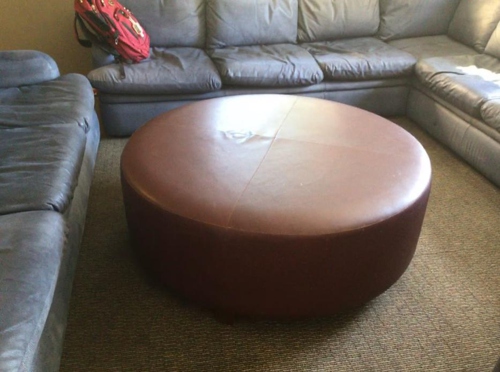} &
        \includegraphics[height=\turnheightnew]{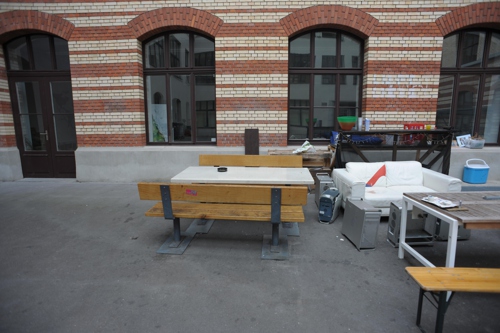} &
        \includegraphics[height=\turnheightnew]{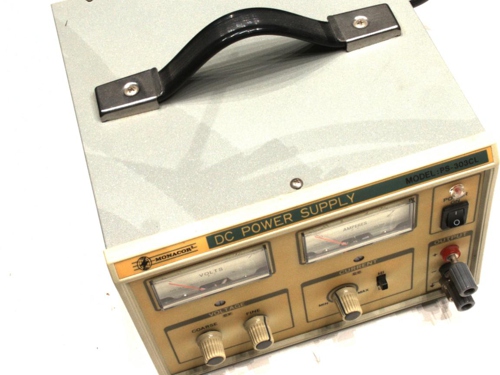} &
        \includegraphics[height=\turnheightnew]{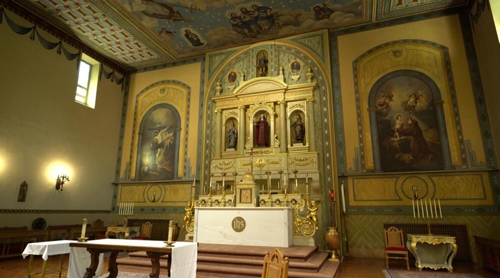} \\
        
        {\rotatebox{90}{\hspace{1mm}{\small Depth Pro~\protect\cite{bochkovskii2024depthpro}}}} &
        \includegraphics[height=\turnheightnew, trim=4cm 0cm 5cm 0cm, clip]{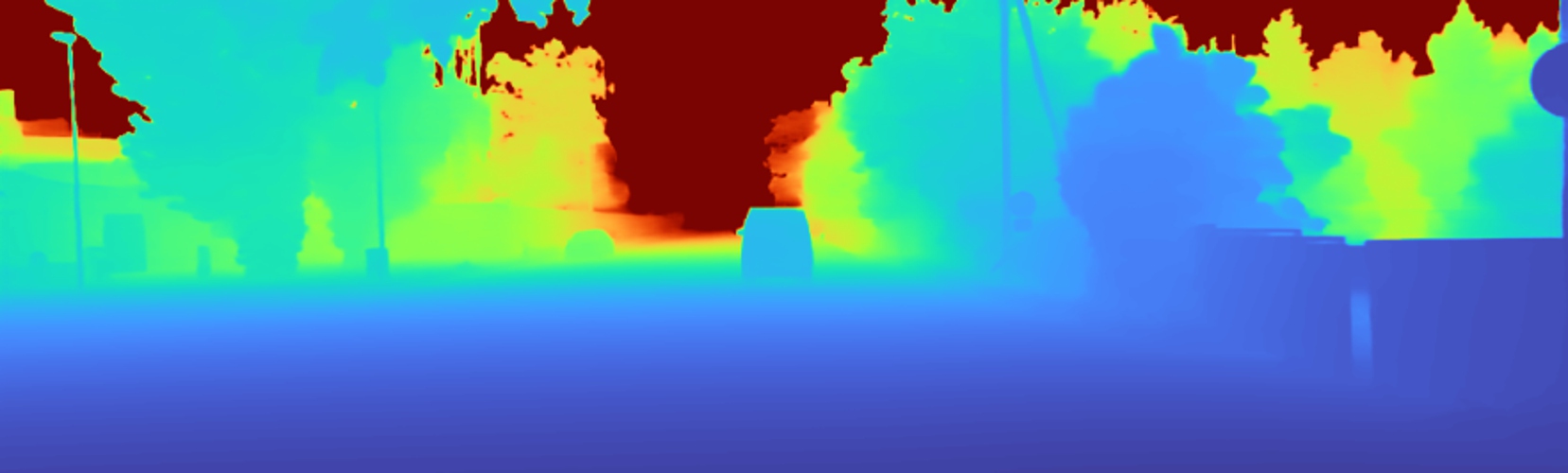} &
        \includegraphics[height=\turnheightnew]{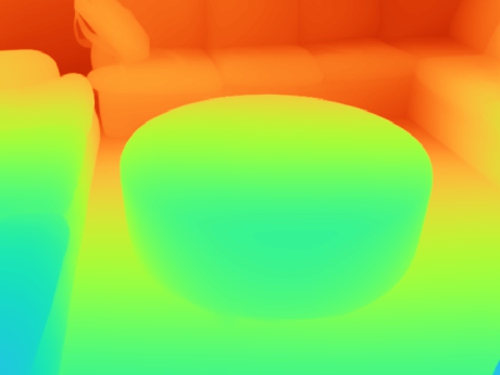} &
        \includegraphics[height=\turnheightnew]{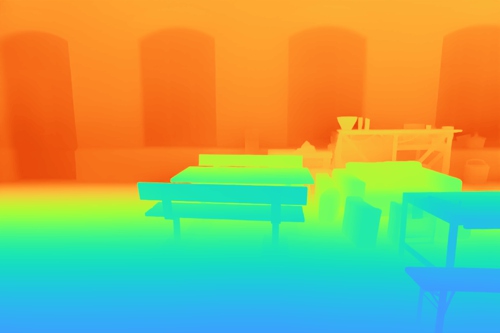} &
        \includegraphics[height=\turnheightnew]{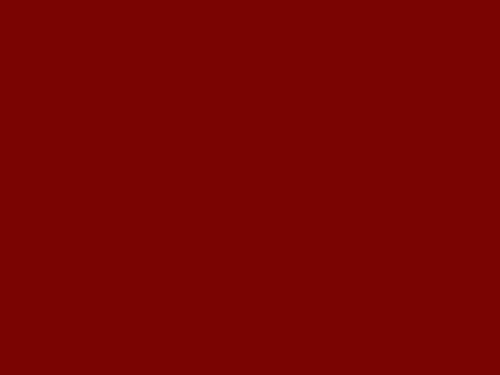} &
        \includegraphics[height=\turnheightnew]{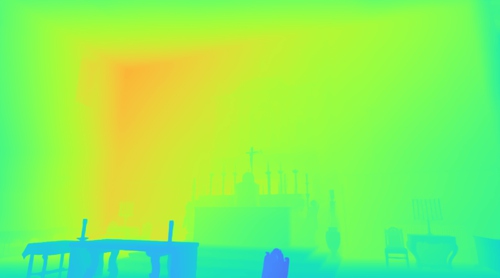} \\
        
        {\rotatebox{90}{\hspace{4mm}{\small rMVD~\protect\cite{schroeppel2022robust}}}} &
        \includegraphics[height=\turnheightnew, trim=4cm 0cm 5cm 0cm, clip]{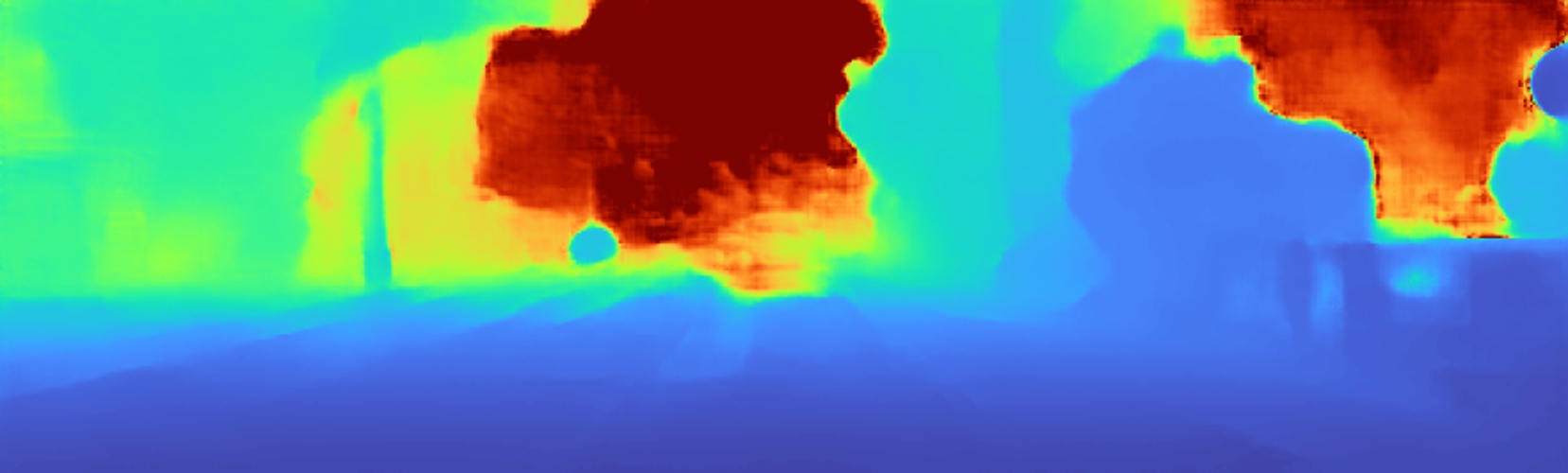} &
        \includegraphics[height=\turnheightnew]{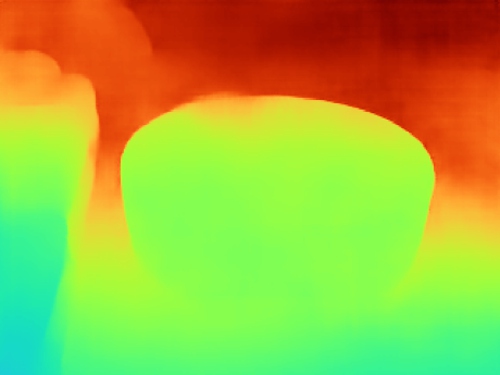} &
        \includegraphics[height=\turnheightnew]{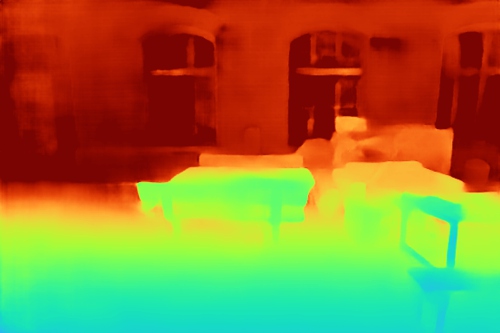} &
        \includegraphics[height=\turnheightnew]{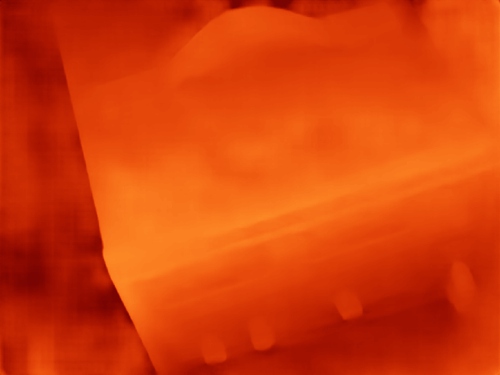} &
        \includegraphics[height=\turnheightnew]{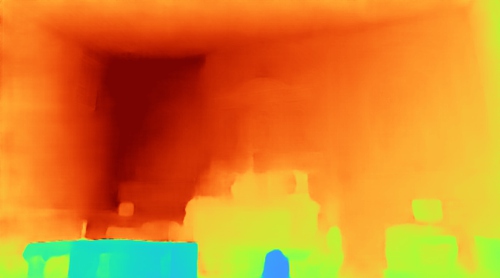} \\

        {\rotatebox{90}{\hspace{1mm}{\small MAST3R~\protect\cite{mast3r_arxiv24}}}} &
        \includegraphics[height=\turnheightnew, trim=4cm 0cm 5cm 0cm, clip]{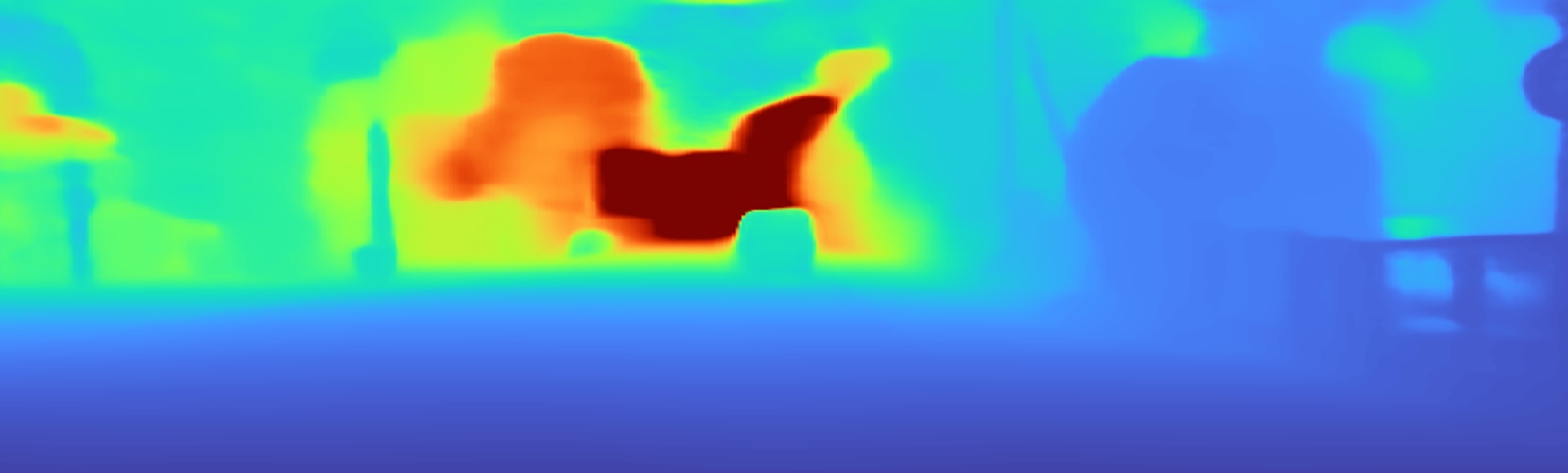} &
        \includegraphics[height=\turnheightnew]{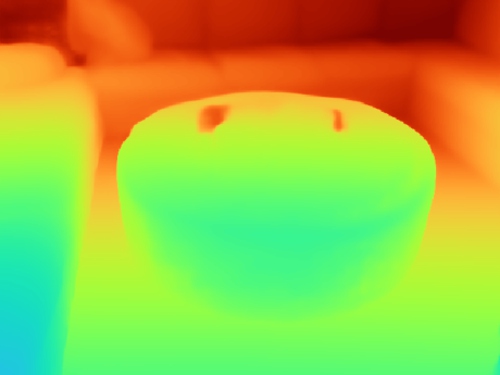} &
        \includegraphics[height=\turnheightnew]{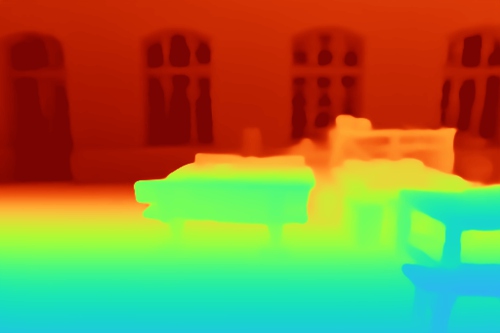} &
        \includegraphics[height=\turnheightnew]{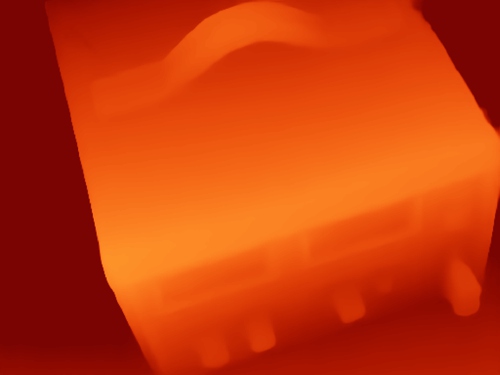} &
        \includegraphics[height=\turnheightnew]{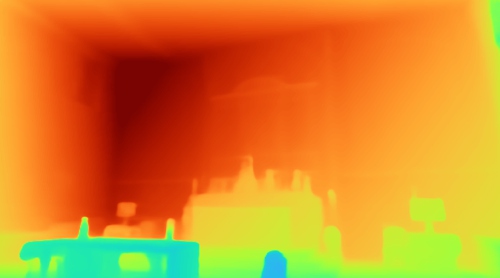} \\
        
        {\rotatebox{90}{\hspace{0.5mm}{\small MVSA (Ours)}}} &
        \includegraphics[height=\turnheightnew, trim=4cm 0cm 5cm 0cm, clip]{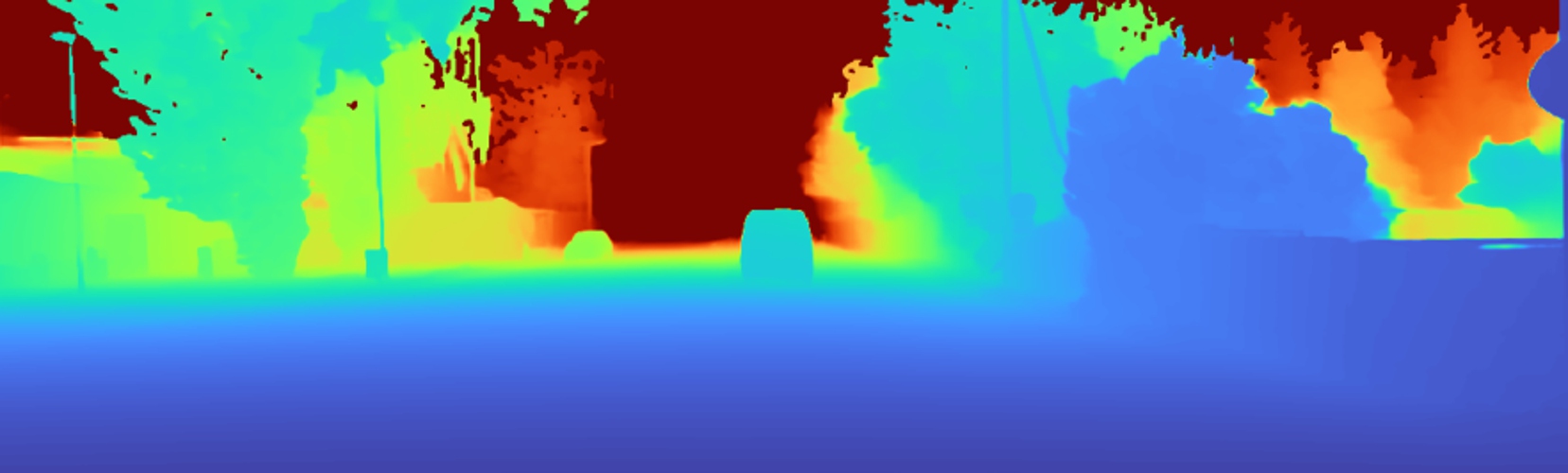} &
        \includegraphics[height=\turnheightnew]{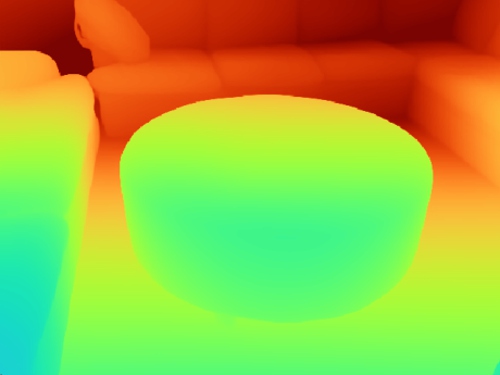} &
        \includegraphics[height=\turnheightnew]{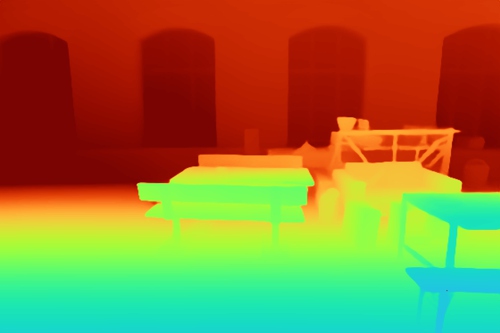} &
        \includegraphics[height=\turnheightnew]{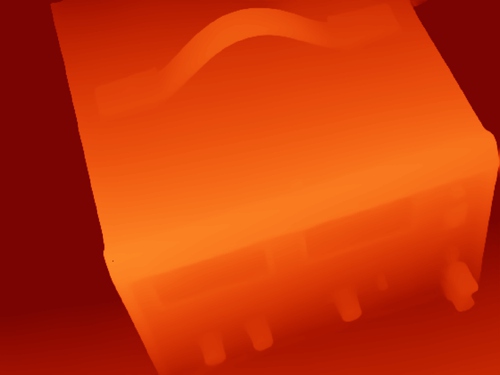} &
        \includegraphics[height=\turnheightnew]{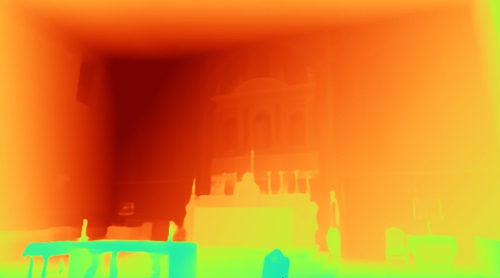} \\

        {\rotatebox{90}{\hspace{7mm}{\small GT}}} &
        \includegraphics[height=\turnheightnew, trim=4cm 0cm 5cm 0cm, clip]{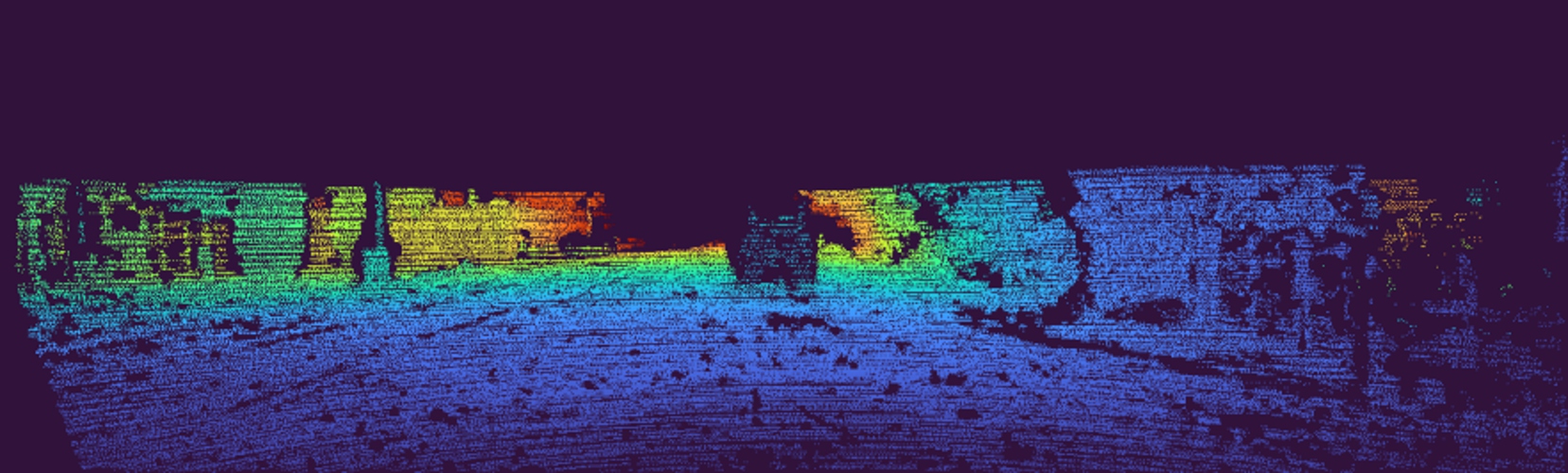} &
        \includegraphics[height=\turnheightnew]{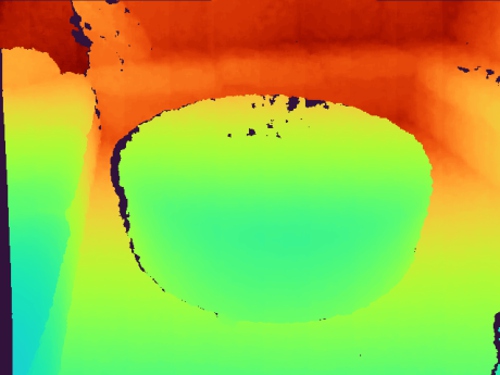} &
        \includegraphics[height=\turnheightnew]{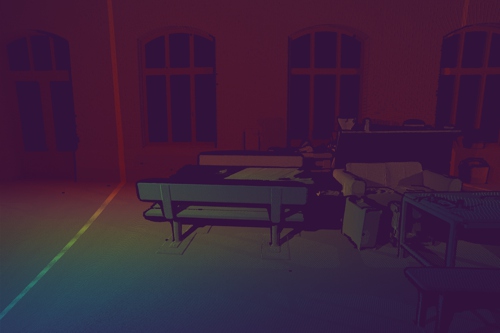} &
        \includegraphics[height=\turnheightnew]{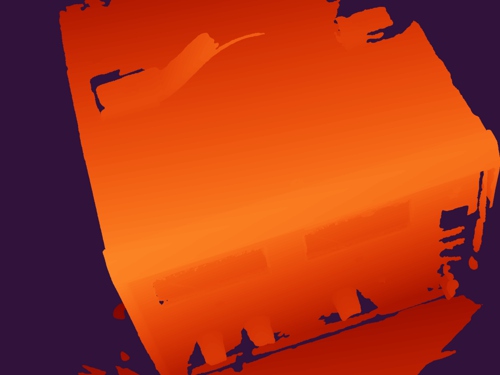} &
        \includegraphics[height=\turnheightnew]{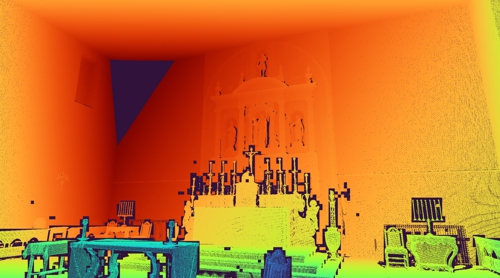}  \\
    \end{tabular}
    }
    \vspace{-10pt}
    \caption{
        \textbf{Qualitative comparison of depth prediction results across multiple datasets (Normalized)} (KITTI, ScanNet, ETH3D, DTU, and Tanks \& Temples). 
        Rows show different methods: Depth Pro~\cite{schroeppel2022robust}, rMVD baseline~\cite{schroeppel2022robust}, MAST3R (Triangulated)~\cite{mast3r_arxiv24}, and our MVSA model, along with RGB inputs ($I_r$) and ground-truth depths (GT). 
        Depth Pro provides sharp edges but often misestimates depth scale, while our MVSA model captures finer details than MAST3R and rMVD. 
        Depth maps are normalized to ground truth depth range for consistent visualization.
    }
    \label{fig:qualitative_depths_sup}
\end{figure*}

\begin{figure*}
    \resizebox{1.0\textwidth}{!}{
    \newcommand{\turnheightnew}{60pt}
    \begin{tabular}{@{\hskip -2mm}c@{\hskip 1mm}c@{\hskip 1mm}c@{\hskip 1mm}c@{\hskip 1mm}c@{\hskip 1mm}c@{\hskip 1mm}c@{}}
        & KITTI & ScanNet & ETH3D & DTU & {T\&T} \\
    
        {\rotatebox{90}{\hspace{5mm}{\small RGB ($I_r$)}}} &
        \includegraphics[height=\turnheightnew, trim=4cm 0cm 5cm 0cm, clip]{figs_jpeg/qualitative/supplementary/1/kitti/color.jpg} &
        \includegraphics[height=\turnheightnew]{figs_jpeg/qualitative/supplementary/1/scannet/color.jpg} &
        \includegraphics[height=\turnheightnew]{figs_jpeg/qualitative/supplementary/1/eth3d/color.jpg} &
        \includegraphics[height=\turnheightnew]{figs_jpeg/qualitative/supplementary/1/dtu/color.jpg} &
        \includegraphics[height=\turnheightnew]{figs_jpeg/qualitative/supplementary/1/tanks_and_temples/color.jpg} \\
        
        {\rotatebox{90}{\hspace{1mm}{\small Depth Pro~\protect\cite{bochkovskii2024depthpro}}}} &
        \includegraphics[height=\turnheightnew, trim=4cm 0cm 5cm 0cm, clip]{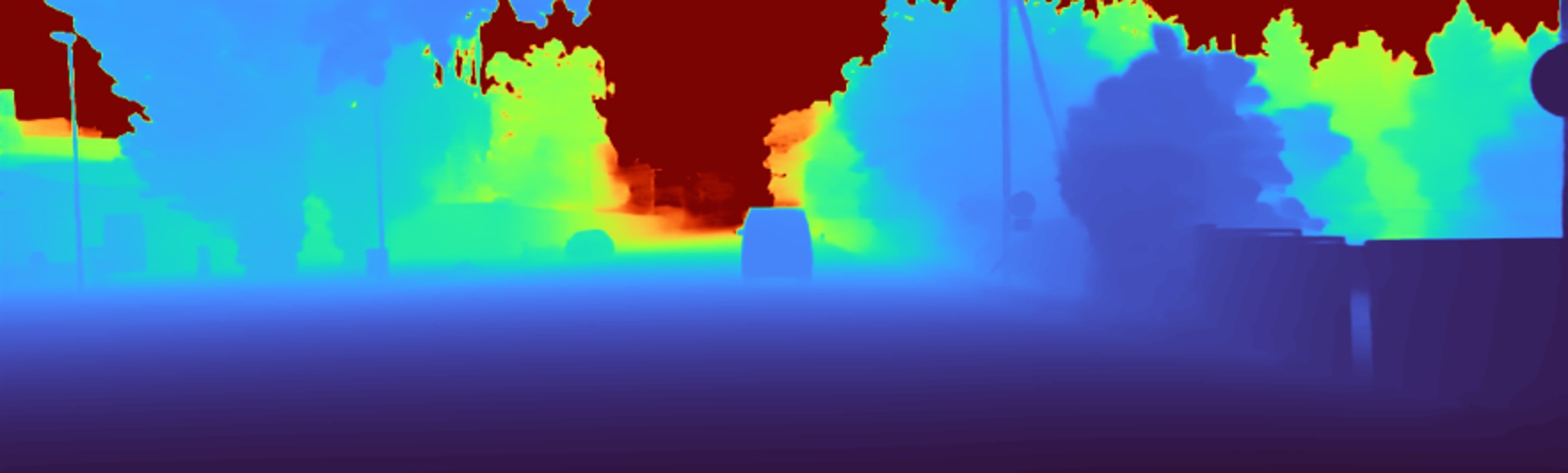} &
        \includegraphics[height=\turnheightnew]{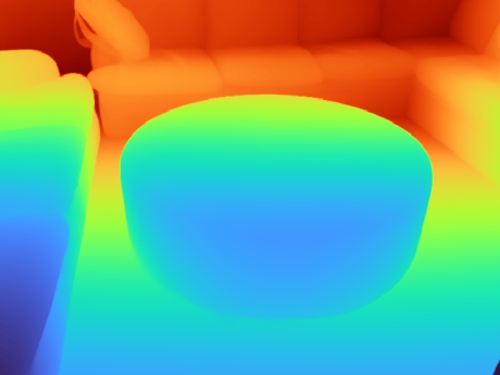} &
        \includegraphics[height=\turnheightnew]{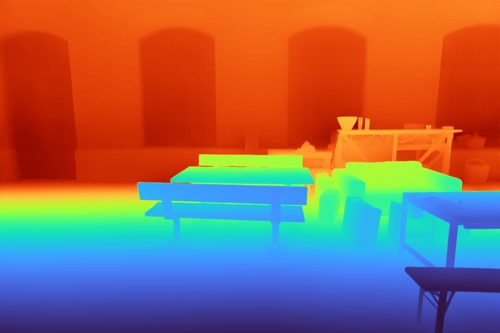} &
        \includegraphics[height=\turnheightnew]{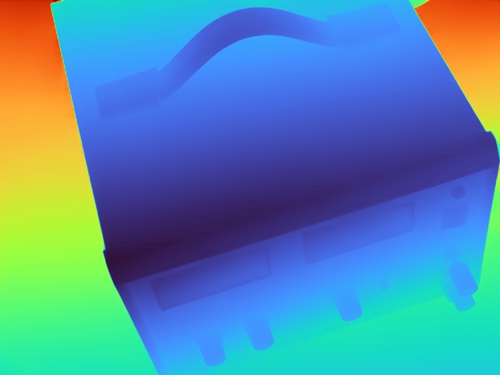} &
        \includegraphics[height=\turnheightnew]{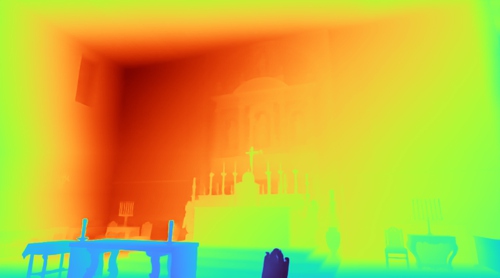} \\
        
        {\rotatebox{90}{\hspace{4mm}{\small rMVD~\protect\cite{schroeppel2022robust}}}} &
        \includegraphics[height=\turnheightnew, trim=4cm 0cm 5cm 0cm, clip]{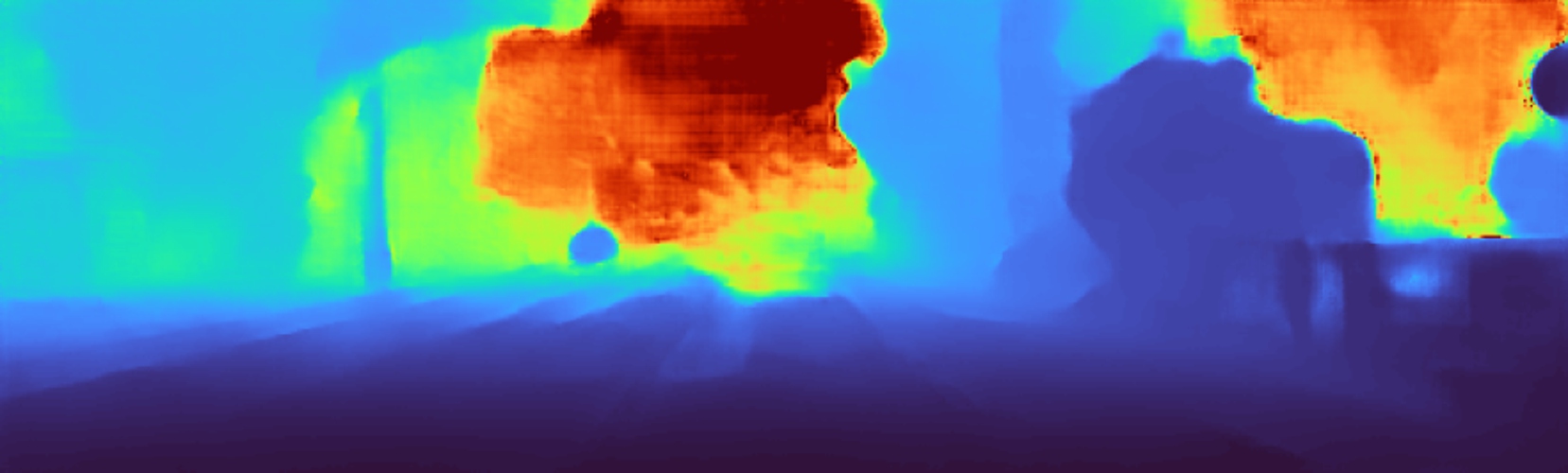} &
        \includegraphics[height=\turnheightnew]{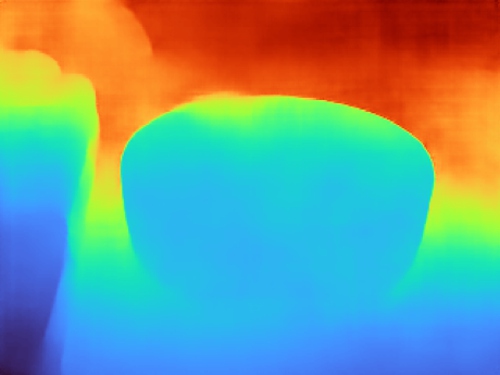} &
        \includegraphics[height=\turnheightnew]{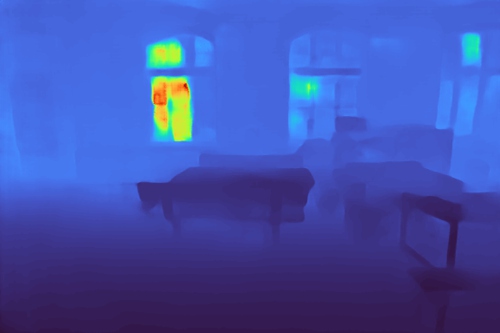} &
        \includegraphics[height=\turnheightnew]{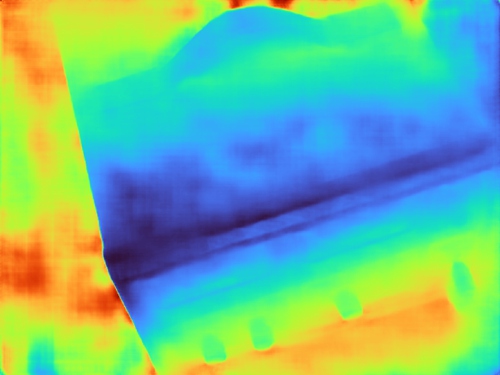} &
        \includegraphics[height=\turnheightnew]{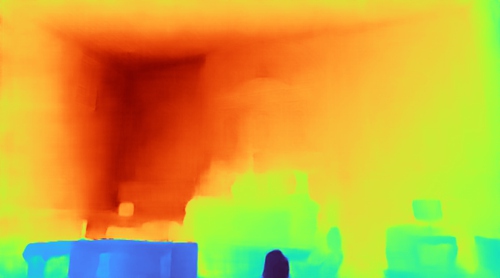} \\

        {\rotatebox{90}{\hspace{1mm}{\small MAST3R~\protect\cite{mast3r_arxiv24}}}} &
        \includegraphics[height=\turnheightnew, trim=4cm 0cm 5cm 0cm, clip]{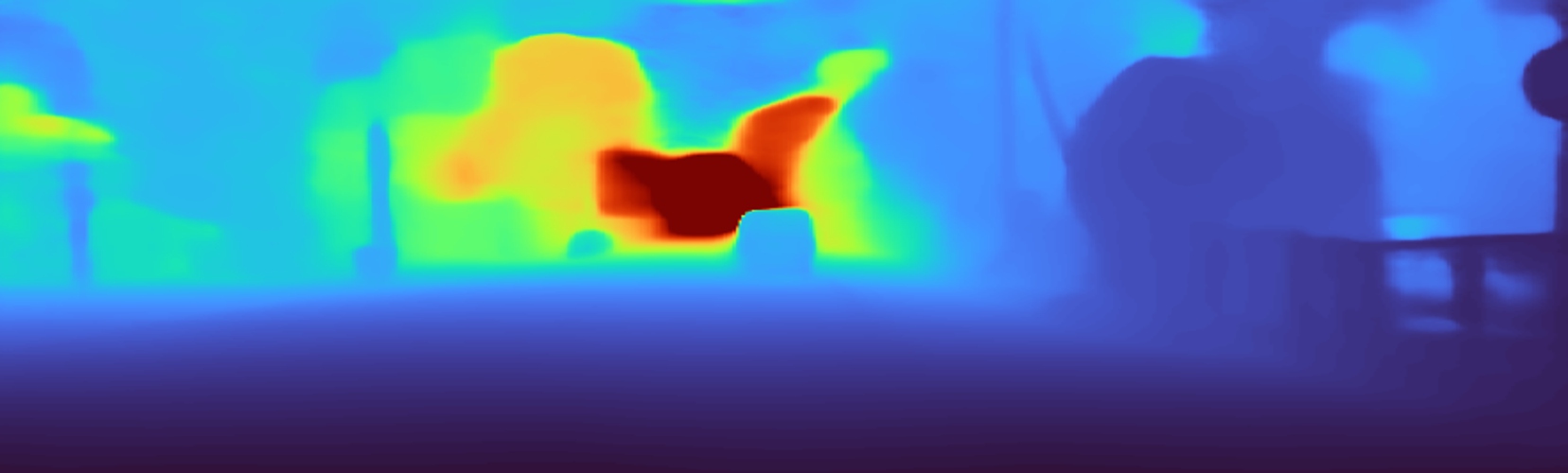} &
        \includegraphics[height=\turnheightnew]{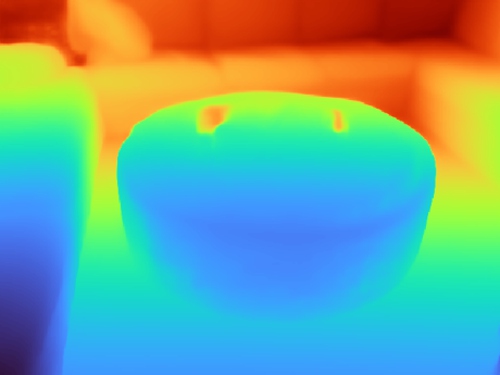} &
        \includegraphics[height=\turnheightnew]{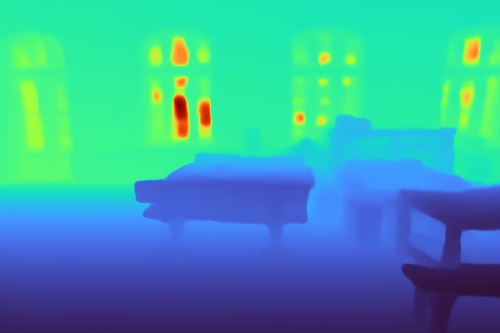} &
        \includegraphics[height=\turnheightnew]{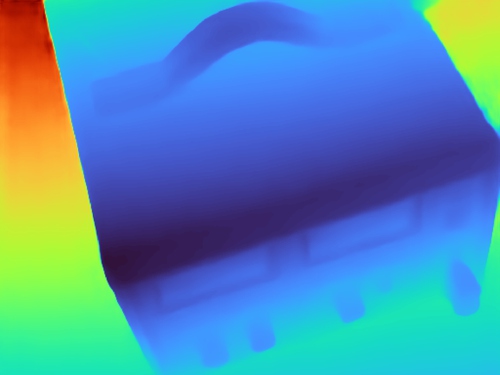} &
        \includegraphics[height=\turnheightnew]{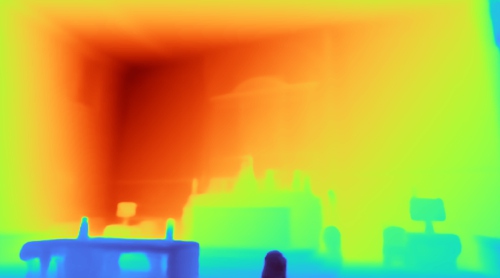} \\
        
        {\rotatebox{90}{\hspace{0.5mm}{\small MVSA (Ours)}}} &
        \includegraphics[height=\turnheightnew, trim=4cm 0cm 5cm 0cm, clip]{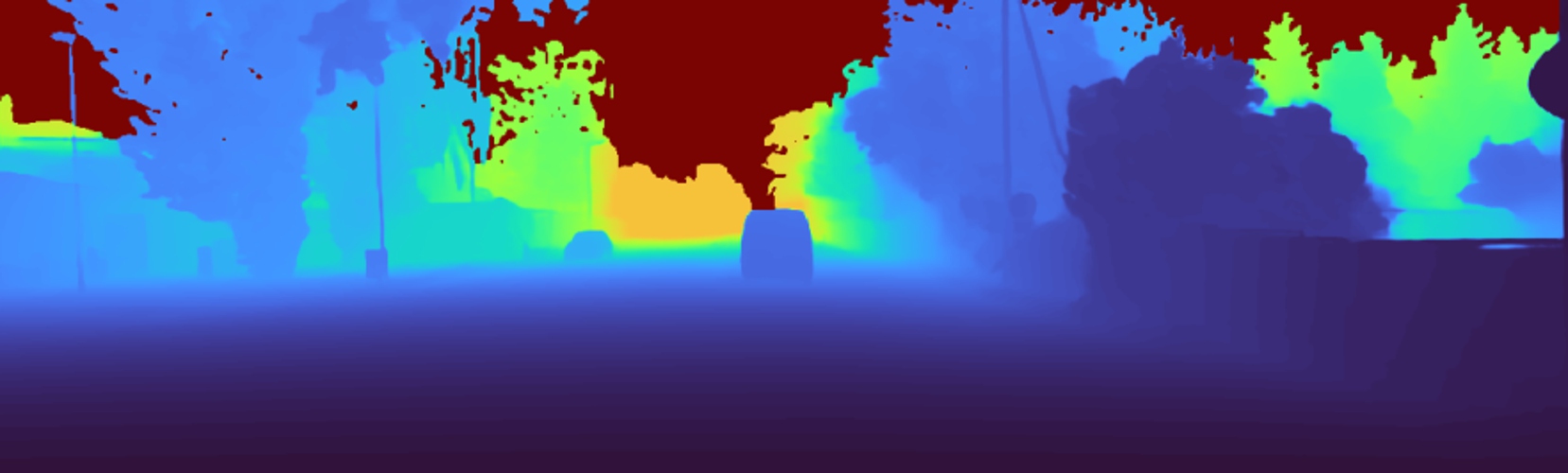} &
        \includegraphics[height=\turnheightnew]{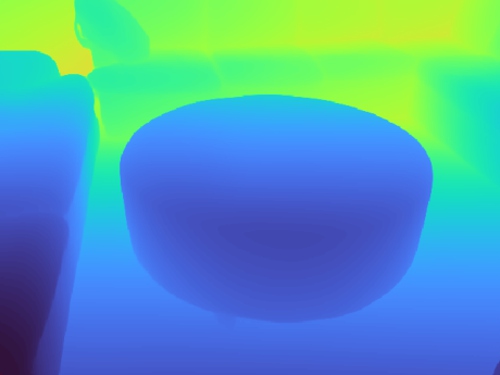} &
        \includegraphics[height=\turnheightnew]{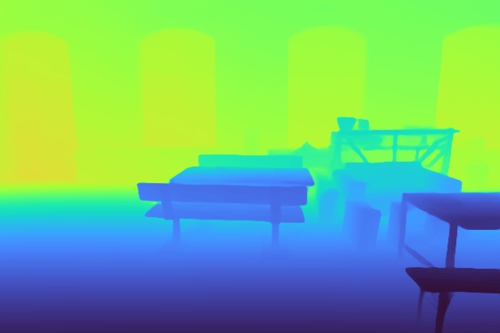} &
        \includegraphics[height=\turnheightnew]{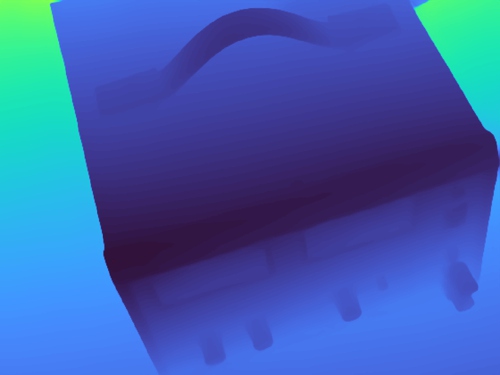} &
        \includegraphics[height=\turnheightnew]{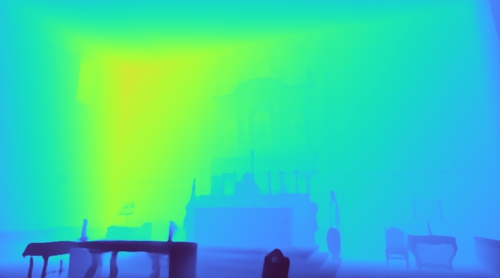} \\

        {\rotatebox{90}{\hspace{7mm}{\small GT}}} &
        \includegraphics[height=\turnheightnew, trim=4cm 0cm 5cm 0cm, clip]{figs_jpeg/qualitative/supplementary/1/kitti/depth_gt.jpg} &
        \includegraphics[height=\turnheightnew]{figs_jpeg/qualitative/supplementary/1/scannet/depth_gt.jpg} &
        \includegraphics[height=\turnheightnew]{figs_jpeg/qualitative/supplementary/1/eth3d/depth_gt.jpg} &
        \includegraphics[height=\turnheightnew]{figs_jpeg/qualitative/supplementary/1/dtu/depth_gt.jpg} &
        \includegraphics[height=\turnheightnew]{figs_jpeg/qualitative/supplementary/1/tanks_and_temples/depth_gt.jpg}  \\
    \end{tabular}
    }
    \vspace{-10pt}
    \caption{
        \textbf{Qualitative comparison of depth prediction results across multiple datasets (Unnormalized)} (KITTI, ScanNet, ETH3D, DTU, and Tanks \& Temples). 
        Rows show different methods: Depth Pro~\cite{schroeppel2022robust}, rMVD baseline~\cite{schroeppel2022robust}, MAST3R (Triangulated)~\cite{mast3r_arxiv24}, and our MVSA model, along with RGB inputs ($I_r$) and ground-truth depths (GT). 
        Depth maps are normalized per image.
    }
    \label{fig:qualitative_depth_unnorms_sup}
\end{figure*}

\begin{figure*}
    \resizebox{1.0\textwidth}{!}{
    \newcommand{\turnheightnew}{60pt}
    \begin{tabular}{@{\hskip -2mm}c@{\hskip 1mm}c@{\hskip 1mm}c@{\hskip 1mm}c@{\hskip 1mm}c@{\hskip 1mm}c@{\hskip 1mm}c@{}}
        & KITTI & ScanNet & ETH3D & DTU & {T\&T} \\
    
        {\rotatebox{90}{\hspace{5mm}{\small RGB ($I_r$)}}} &
        \includegraphics[height=\turnheightnew, trim=4cm 0cm 5cm 0cm, clip]{figs_jpeg/qualitative/supplementary/1/kitti/color.jpg} &
        \includegraphics[height=\turnheightnew]{figs_jpeg/qualitative/supplementary/1/scannet/color.jpg} &
        \includegraphics[height=\turnheightnew]{figs_jpeg/qualitative/supplementary/1/eth3d/color.jpg} &
        \includegraphics[height=\turnheightnew]{figs_jpeg/qualitative/supplementary/1/dtu/color.jpg} &
        \includegraphics[height=\turnheightnew]{figs_jpeg/qualitative/supplementary/1/tanks_and_temples/color.jpg} \\
        
        {\rotatebox{90}{\hspace{1mm}{\small Depth Pro~\protect\cite{bochkovskii2024depthpro}}}} &
        \includegraphics[height=\turnheightnew, trim=4cm 0cm 5cm 0cm, clip]{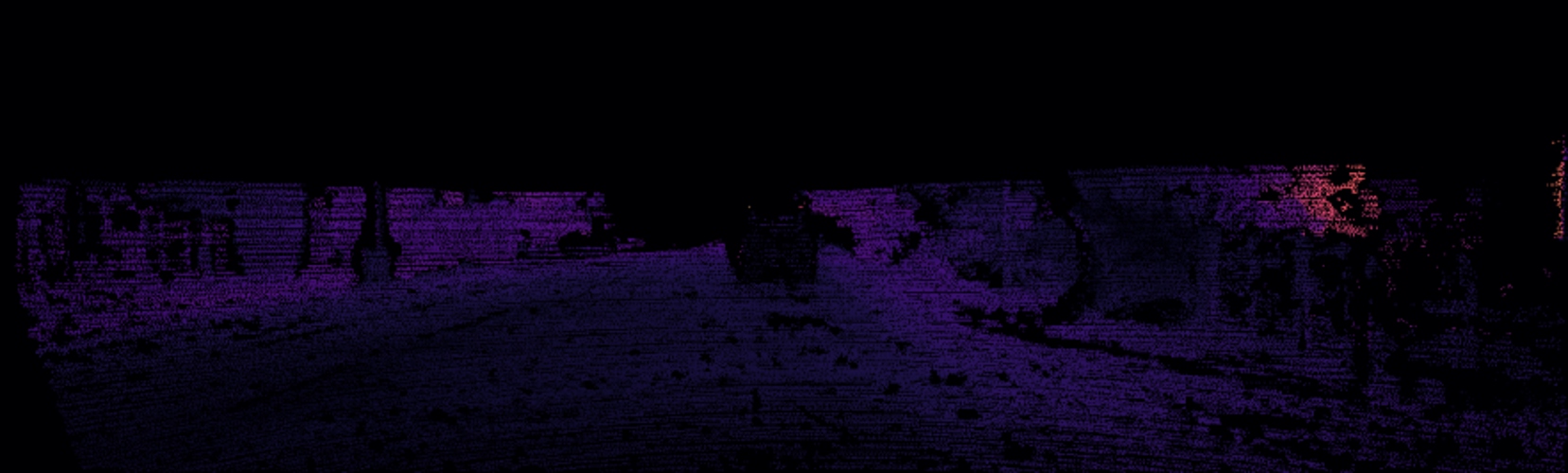} &
        \includegraphics[height=\turnheightnew]{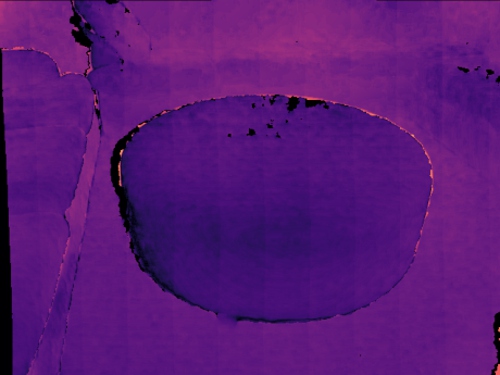} &
        \includegraphics[height=\turnheightnew]{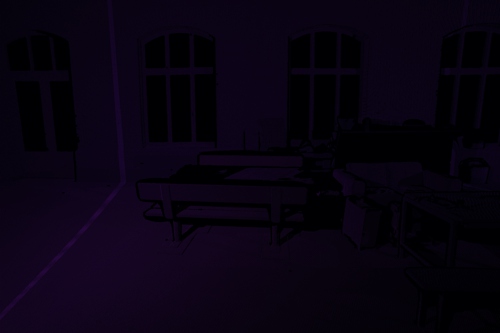} &
        \includegraphics[height=\turnheightnew]{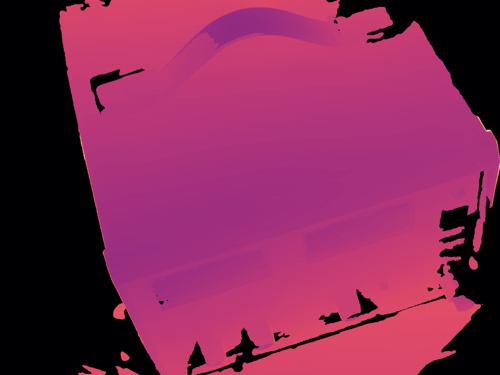} &
        \includegraphics[height=\turnheightnew]{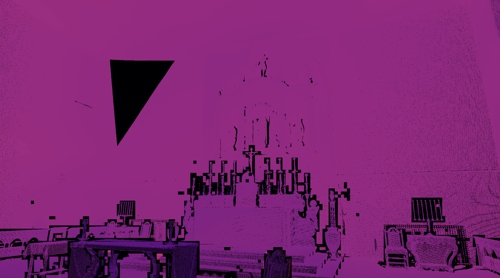} \\
        
        {\rotatebox{90}{\hspace{4mm}{\small rMVD~\protect\cite{schroeppel2022robust}}}} &
        \includegraphics[height=\turnheightnew, trim=4cm 0cm 5cm 0cm, clip]{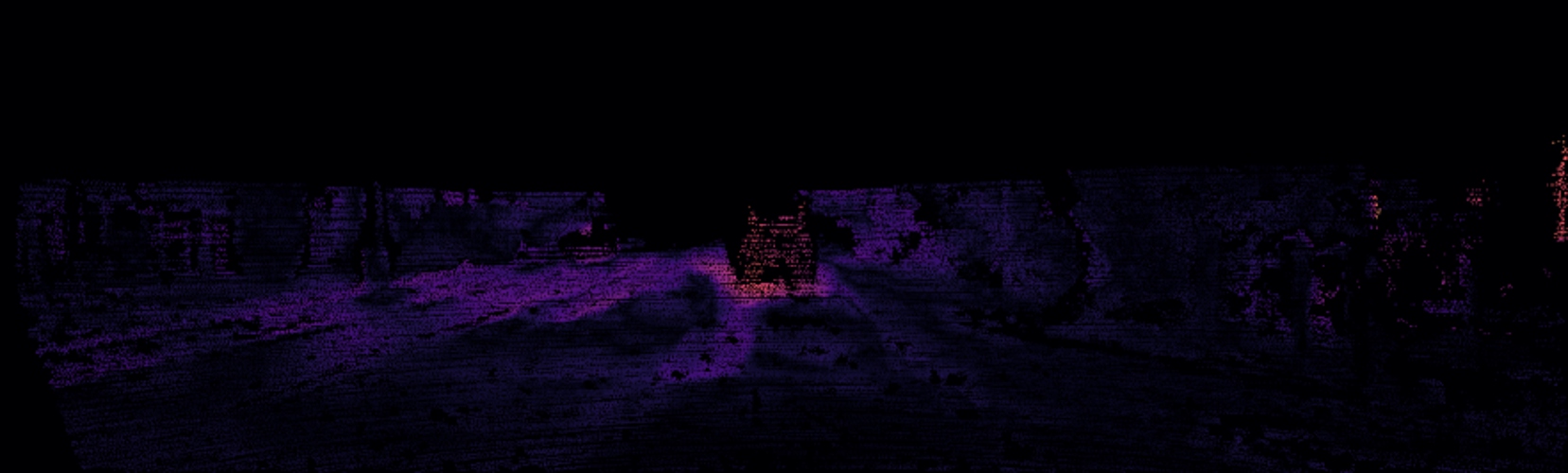} &
        \includegraphics[height=\turnheightnew]{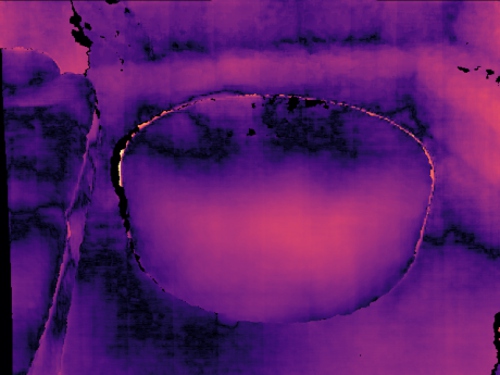} &
        \includegraphics[height=\turnheightnew]{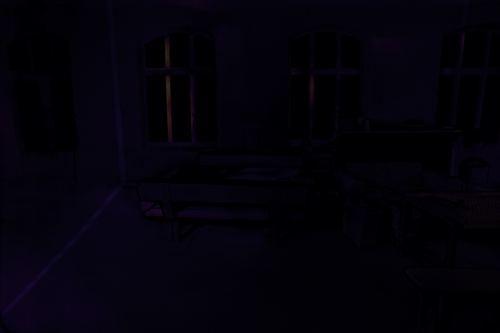} &
        \includegraphics[height=\turnheightnew]{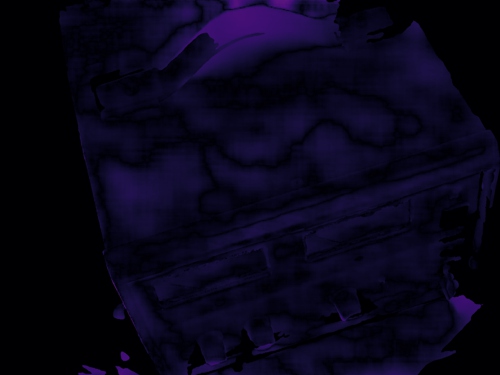} &
        \includegraphics[height=\turnheightnew]{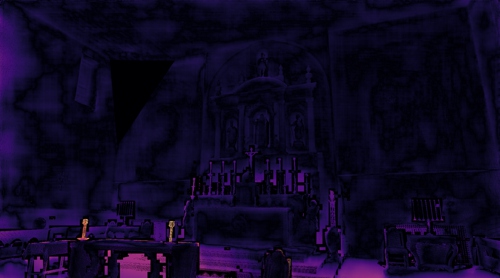} \\

        {\rotatebox{90}{\hspace{1mm}{\small MAST3R~\protect\cite{mast3r_arxiv24}}}} &
        \includegraphics[height=\turnheightnew, trim=4cm 0cm 5cm 0cm, clip]{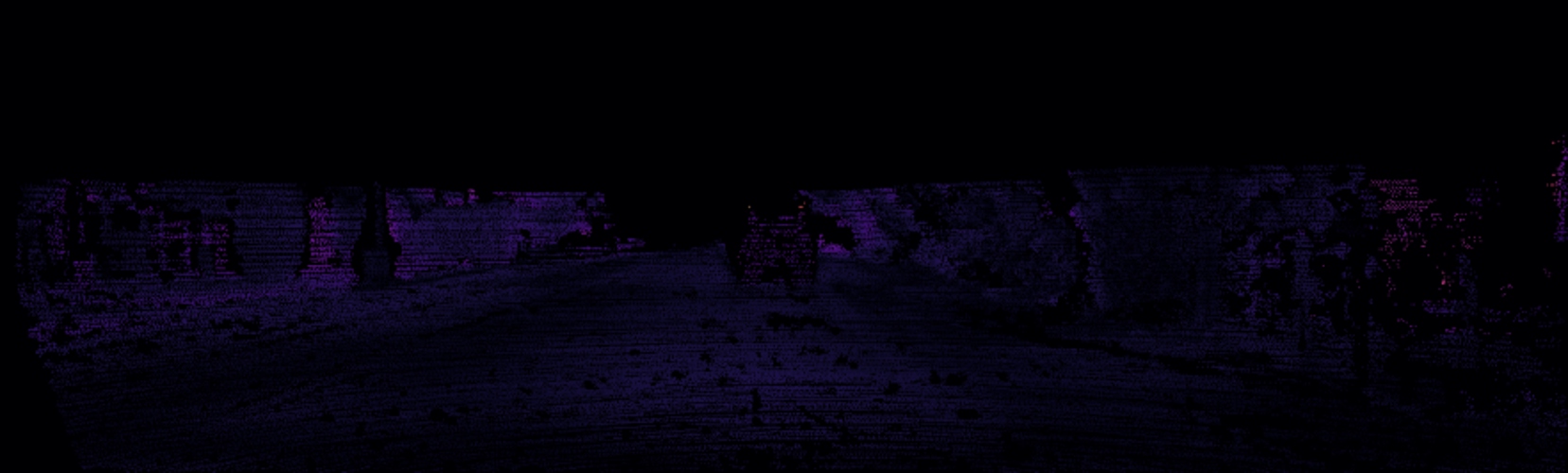} &
        \includegraphics[height=\turnheightnew]{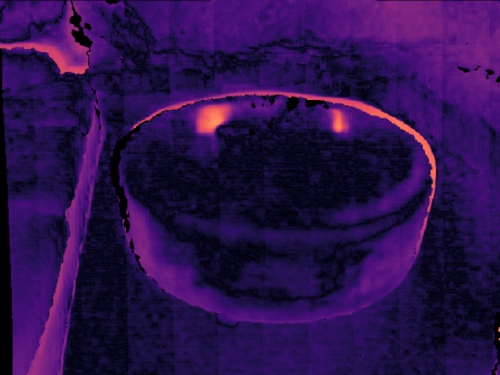} &
        \includegraphics[height=\turnheightnew]{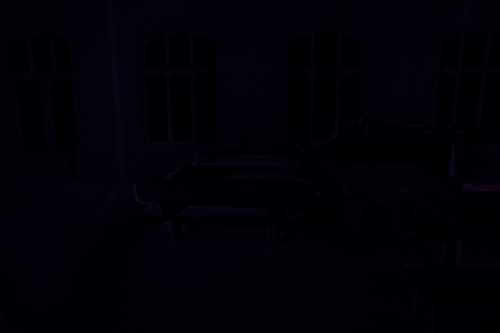} &
        \includegraphics[height=\turnheightnew]{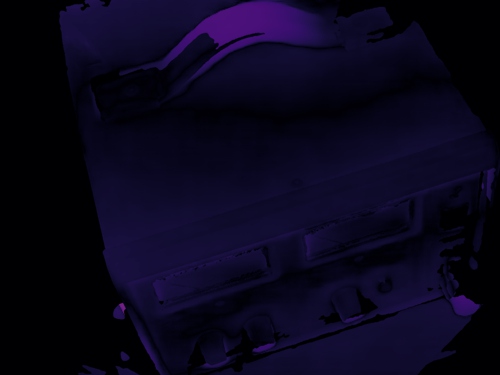} &
        \includegraphics[height=\turnheightnew]{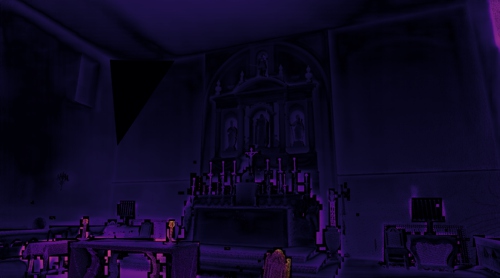} \\
        
        {\rotatebox{90}{\hspace{0.5mm}{\small MVSA (Ours)}}} &
        \includegraphics[height=\turnheightnew, trim=4cm 0cm 5cm 0cm, clip]{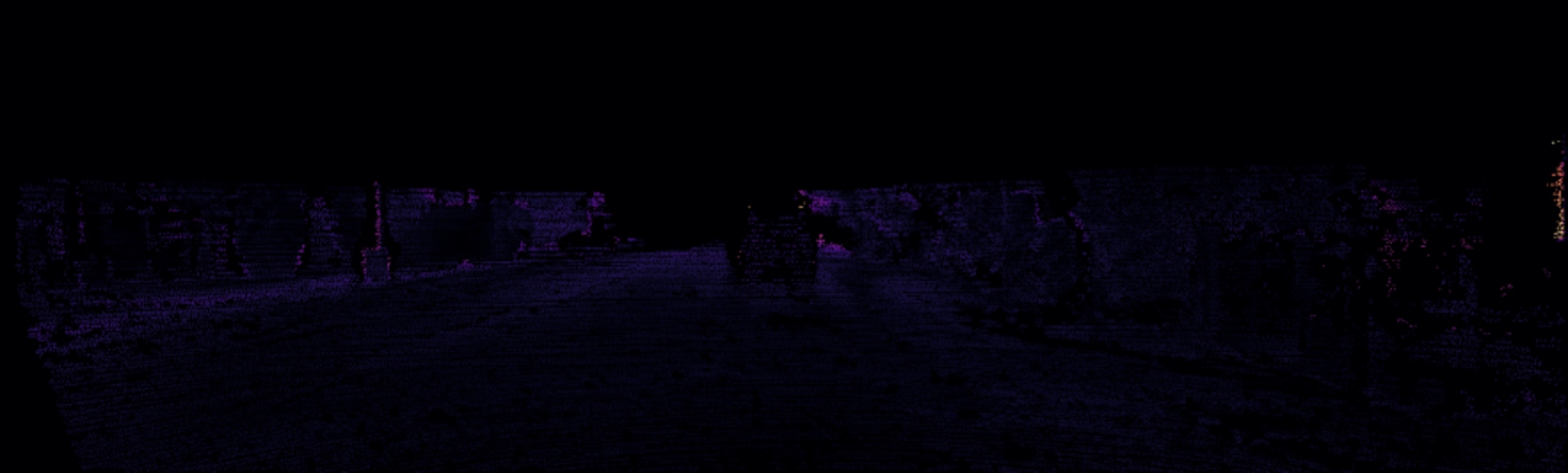} &
        \includegraphics[height=\turnheightnew]{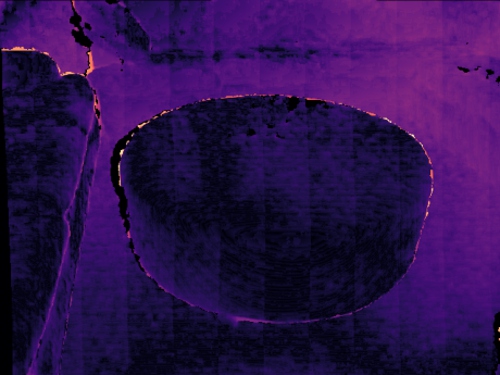} &
        \includegraphics[height=\turnheightnew]{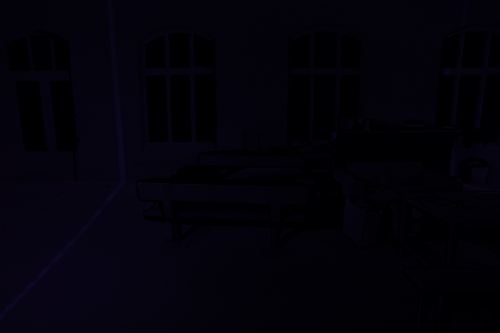} &
        \includegraphics[height=\turnheightnew]{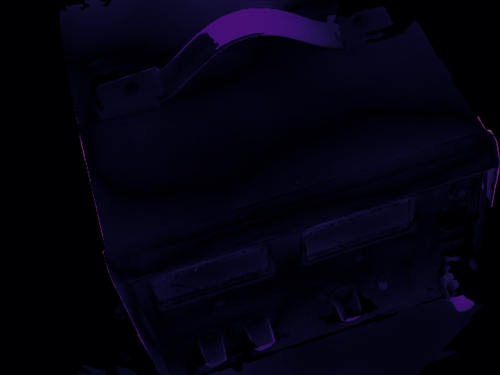} &
        \includegraphics[height=\turnheightnew]{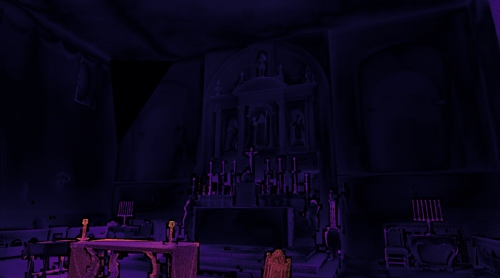} \\

        {\rotatebox{90}{\hspace{7mm}{\small Depth GT}}} &
        \includegraphics[height=\turnheightnew, trim=4cm 0cm 5cm 0cm, clip]{figs_jpeg/qualitative/supplementary/1/kitti/depth_gt.jpg} &
        \includegraphics[height=\turnheightnew]{figs_jpeg/qualitative/supplementary/1/scannet/depth_gt.jpg} &
        \includegraphics[height=\turnheightnew]{figs_jpeg/qualitative/supplementary/1/eth3d/depth_gt.jpg} &
        \includegraphics[height=\turnheightnew]{figs_jpeg/qualitative/supplementary/1/dtu/depth_gt.jpg} &
        \includegraphics[height=\turnheightnew]{figs_jpeg/qualitative/supplementary/1/tanks_and_temples/depth_gt.jpg}  \\
    \end{tabular}
    }
    \vspace{-10pt}
    \caption{
        \textbf{Qualitative comparison of depth prediction errors across multiple datasets} (KITTI, ScanNet, ETH3D, DTU, and Tanks \& Temples). 
        Rows show different methods: Depth Pro~\cite{schroeppel2022robust}, rMVD baseline~\cite{schroeppel2022robust}, MAST3R (Triangulated)~\cite{mast3r_arxiv24}, and our MVSA model, along with RGB inputs ($I_r$) and ground-truth depths (GT). 
        Depth Pro provides sharp edges but often misestimates depth scale, while our MVSA model captures finer details than MAST3R and rMVD. 
        Error maps are normalized to maximum error among methods for each scene.
    }
    \label{fig:qualitative_errors_sup1}
\end{figure*}

\begin{figure*}
    \resizebox{1.0\textwidth}{!}{
    \newcommand{\turnheightnew}{60pt}
    \begin{tabular}{@{\hskip -2mm}c@{\hskip 1mm}c@{\hskip 1mm}c@{\hskip 1mm}c@{\hskip 1mm}c@{\hskip 1mm}c@{\hskip 1mm}c@{}}
        & KITTI & ScanNet & ETH3D & DTU & {T\&T} \\
    
        {\rotatebox{90}{\hspace{5mm}{\small RGB ($I_r$)}}} &
        \includegraphics[height=\turnheightnew, trim=4cm 0cm 5cm 0cm, clip]{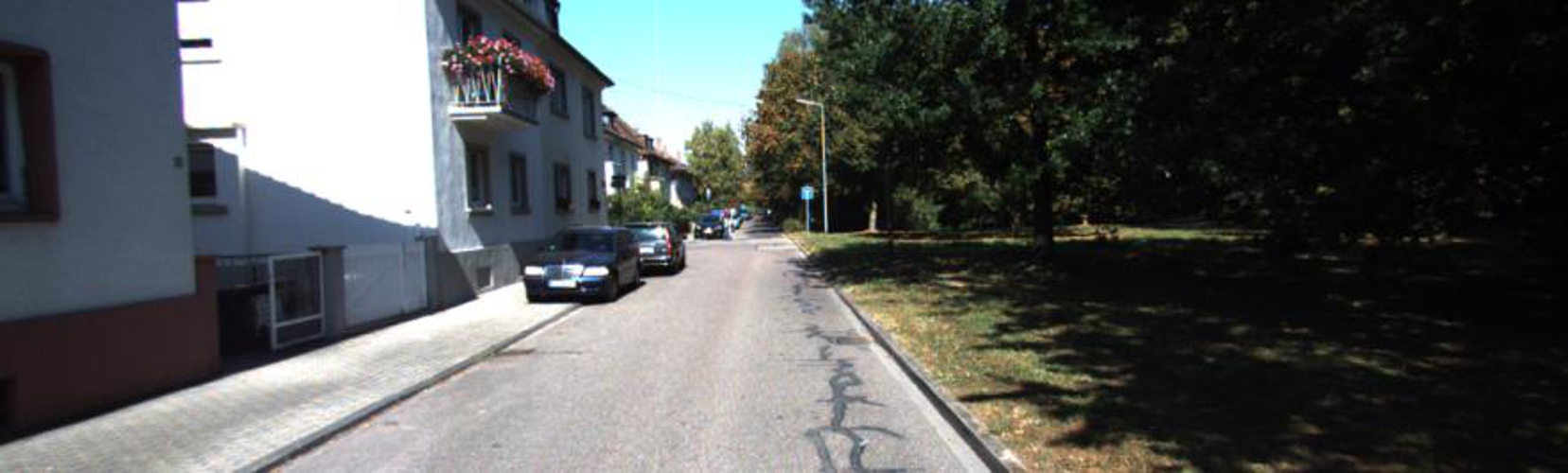} &
        \includegraphics[height=\turnheightnew]{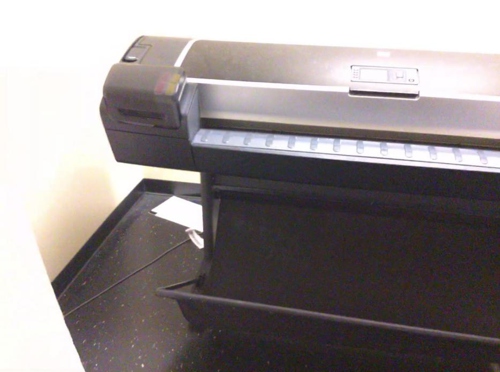} &
        \includegraphics[height=\turnheightnew]{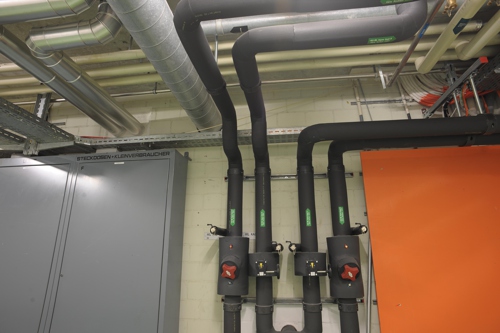} &
        \includegraphics[height=\turnheightnew]{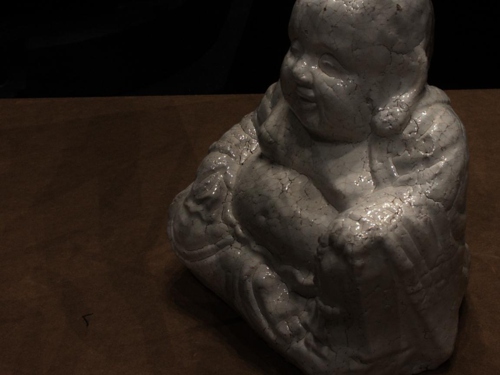} &
        \includegraphics[height=\turnheightnew]{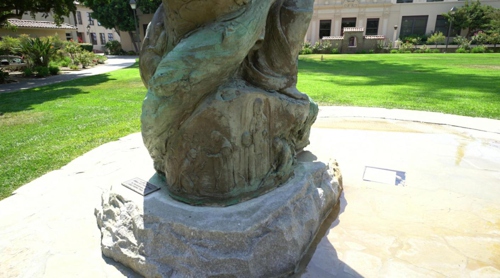} \\
        
        {\rotatebox{90}{\hspace{1mm}{\small Depth Pro~\protect\cite{bochkovskii2024depthpro}}}} &
        \includegraphics[height=\turnheightnew, trim=4cm 0cm 5cm 0cm, clip]{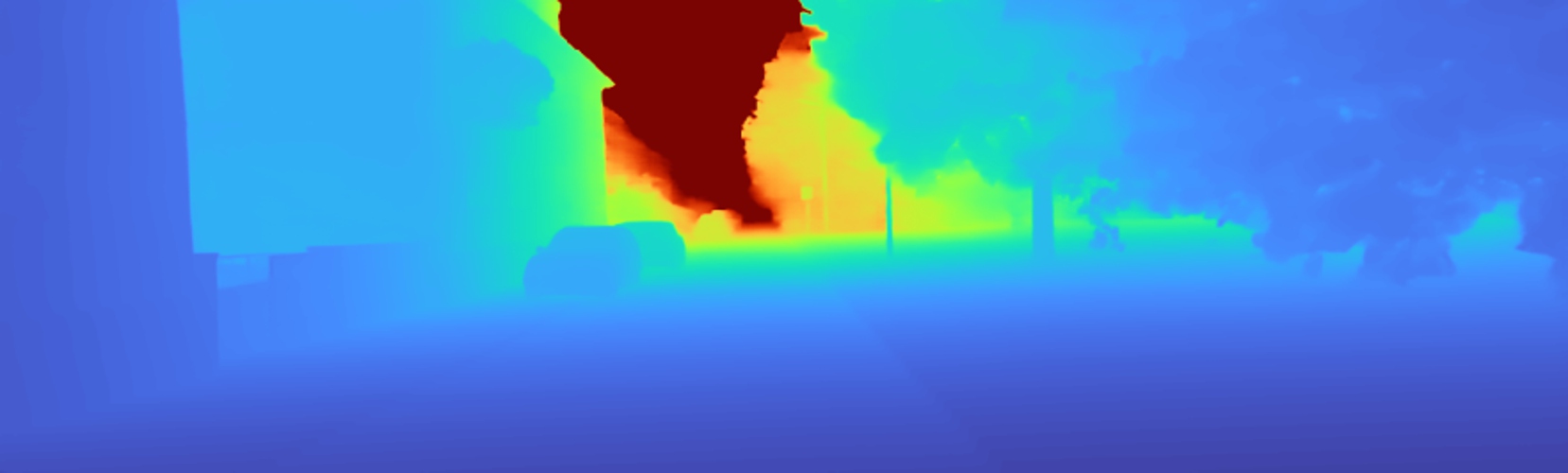} &
        \includegraphics[height=\turnheightnew]{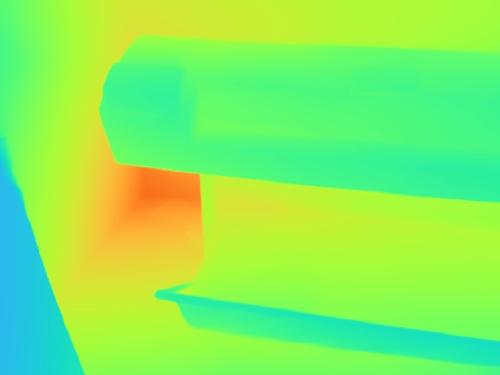} &
        \includegraphics[height=\turnheightnew]{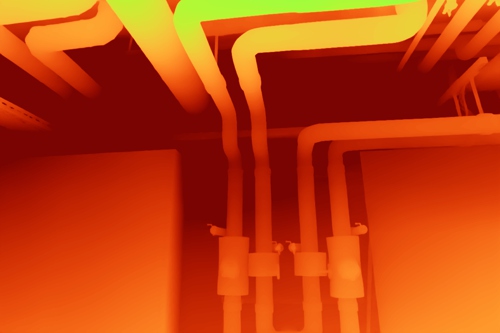} &
        \includegraphics[height=\turnheightnew]{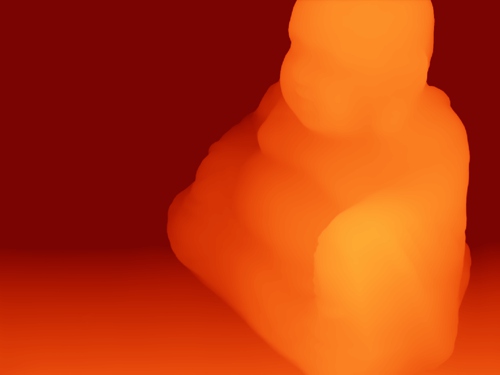} &
        \includegraphics[height=\turnheightnew]{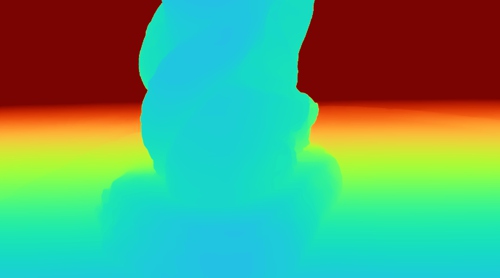} \\
        
        {\rotatebox{90}{\hspace{4mm}{\small rMVD~\protect\cite{schroeppel2022robust}}}} &
        \includegraphics[height=\turnheightnew, trim=4cm 0cm 5cm 0cm, clip]{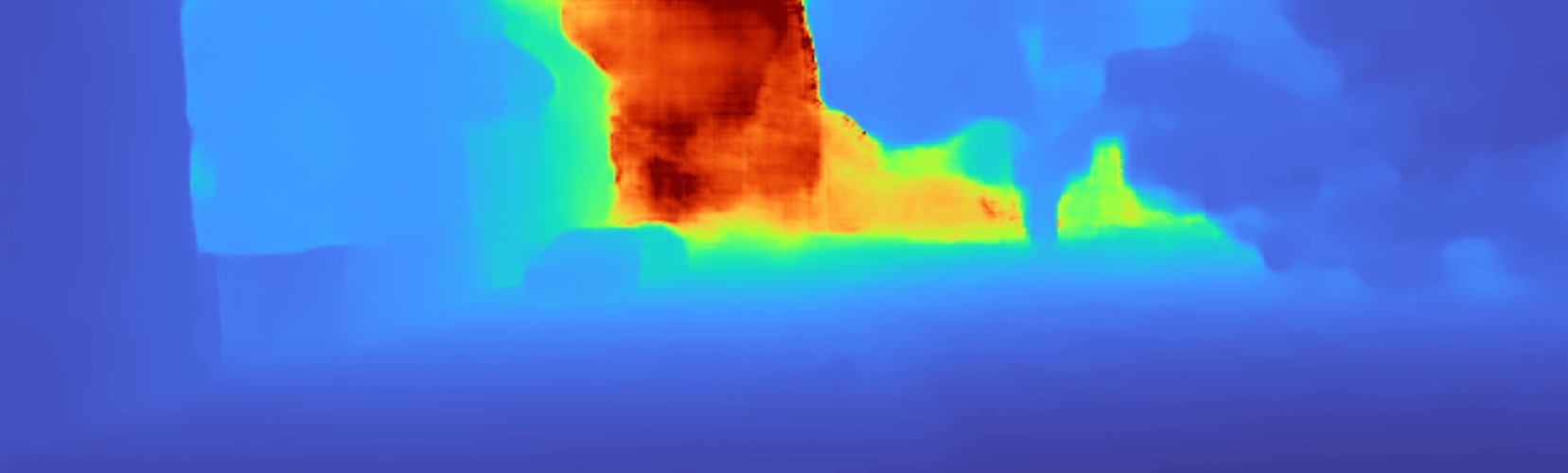} &
        \includegraphics[height=\turnheightnew]{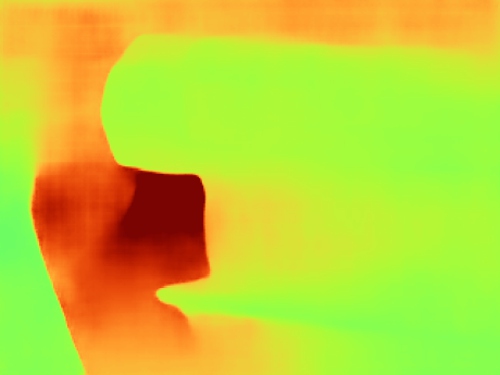} &
        \includegraphics[height=\turnheightnew]{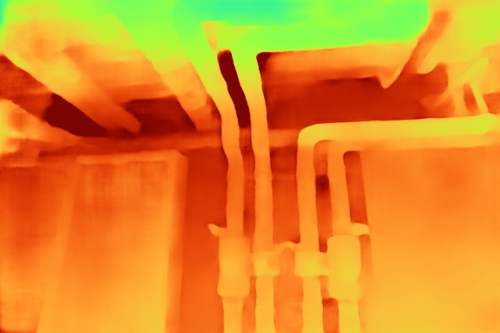} &
        \includegraphics[height=\turnheightnew]{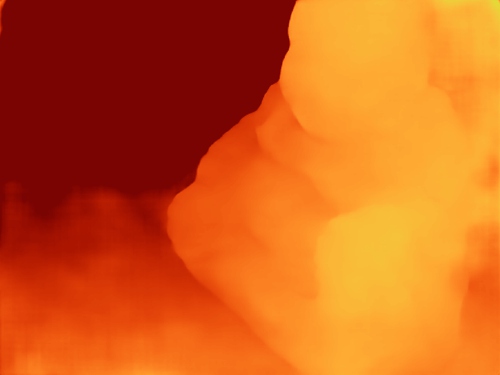} &
        \includegraphics[height=\turnheightnew]{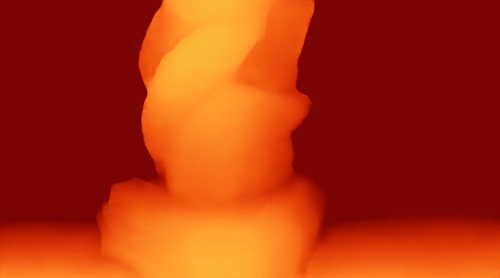} \\

        {\rotatebox{90}{\hspace{1mm}{\small MAST3R~\protect\cite{mast3r_arxiv24}}}} &
        \includegraphics[height=\turnheightnew, trim=4cm 0cm 5cm 0cm, clip]{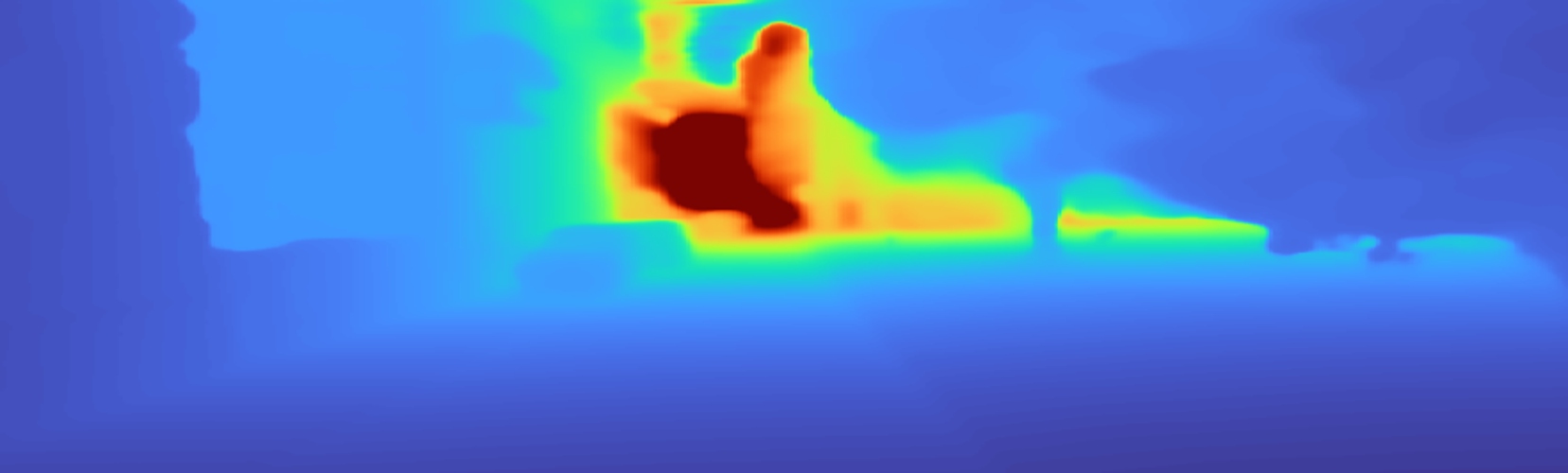} &
        \includegraphics[height=\turnheightnew]{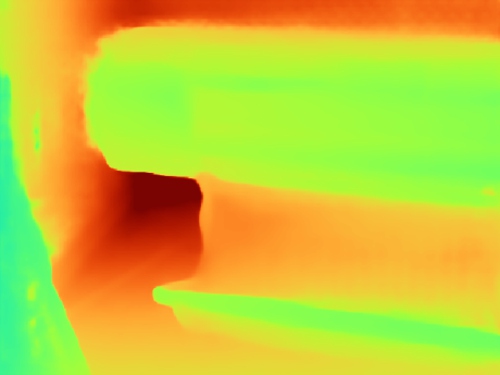} &
        \includegraphics[height=\turnheightnew]{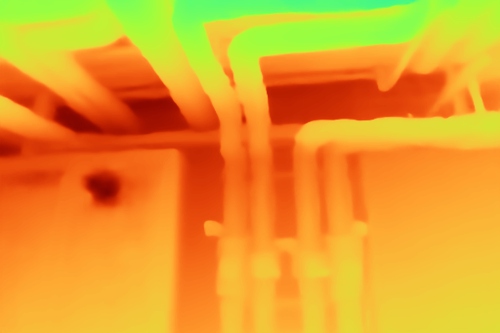} &
        \includegraphics[height=\turnheightnew]{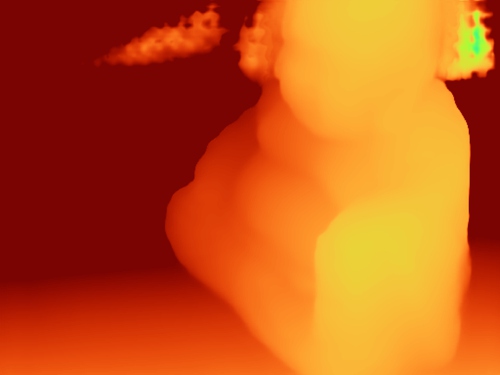} &
        \includegraphics[height=\turnheightnew]{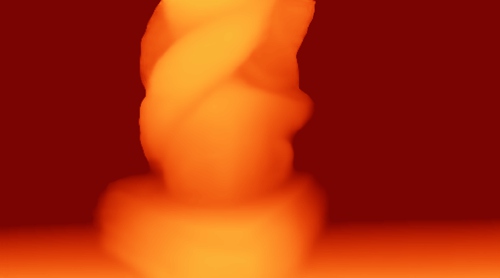} \\
        
        {\rotatebox{90}{\hspace{0.5mm}{\small MVSA (Ours)}}} &
        \includegraphics[height=\turnheightnew, trim=4cm 0cm 5cm 0cm, clip]{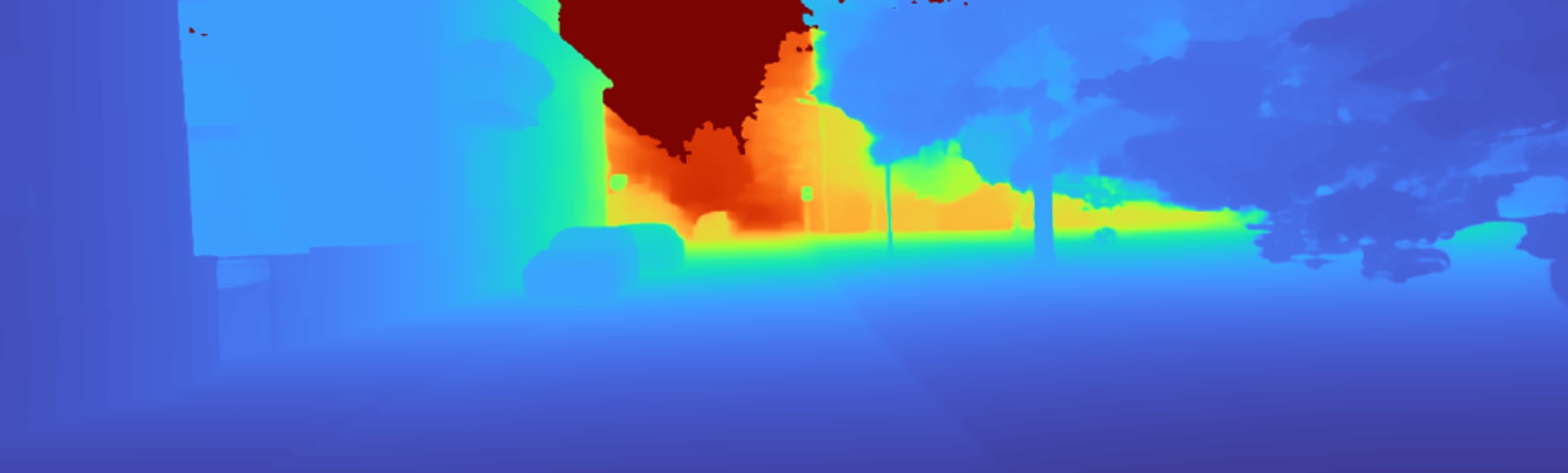} &
        \includegraphics[height=\turnheightnew]{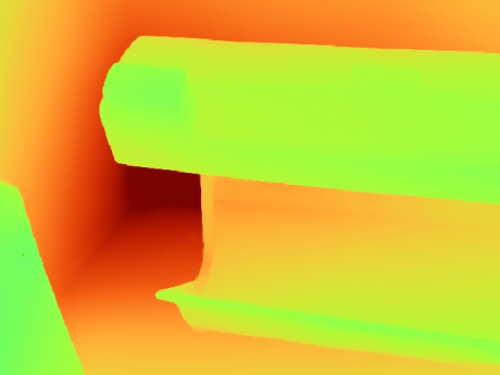} &
        \includegraphics[height=\turnheightnew]{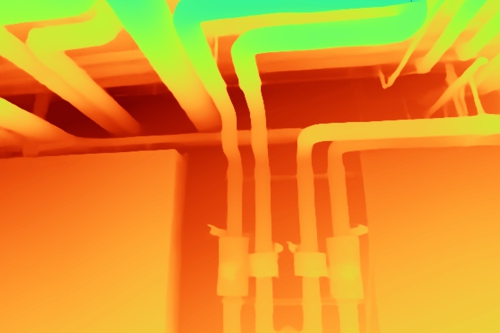} &
        \includegraphics[height=\turnheightnew]{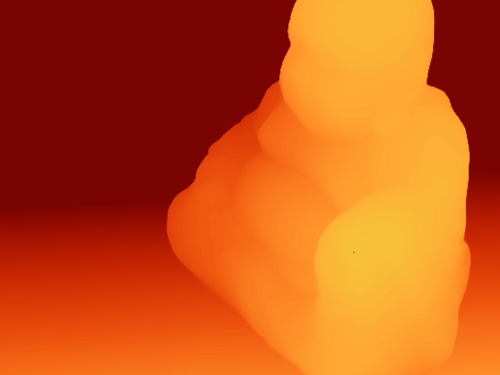} &
        \includegraphics[height=\turnheightnew]{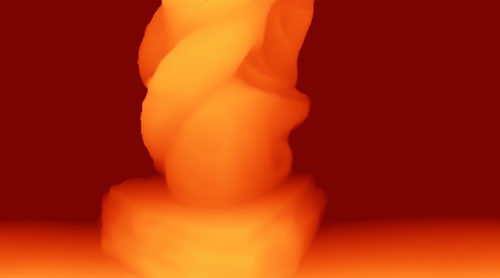} \\

        {\rotatebox{90}{\hspace{7mm}{\small GT}}} &
        \includegraphics[height=\turnheightnew, trim=4cm 0cm 5cm 0cm, clip]{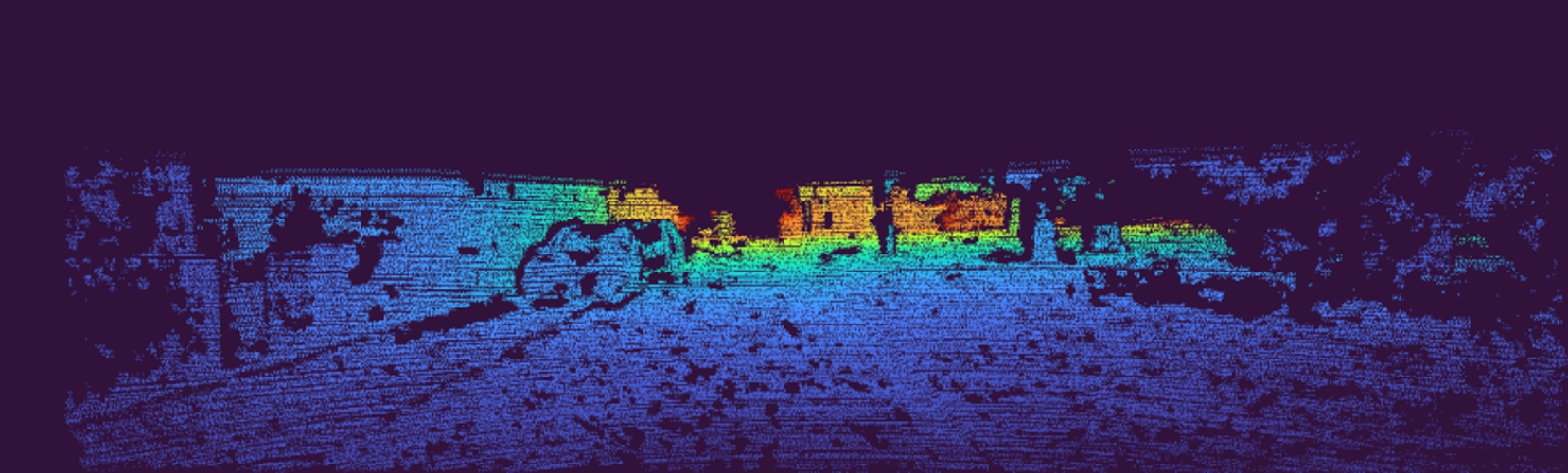} &
        \includegraphics[height=\turnheightnew]{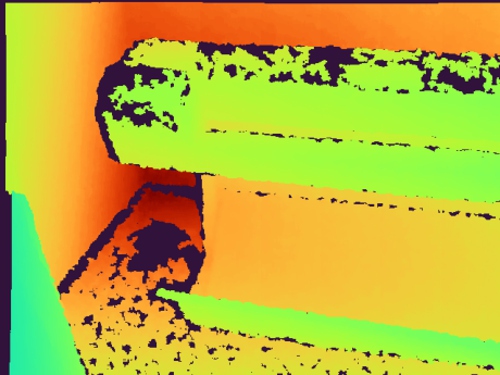} &
        \includegraphics[height=\turnheightnew]{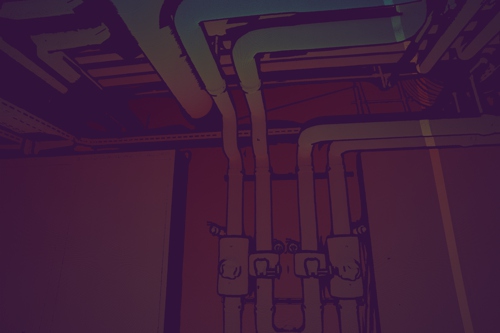} &
        \includegraphics[height=\turnheightnew]{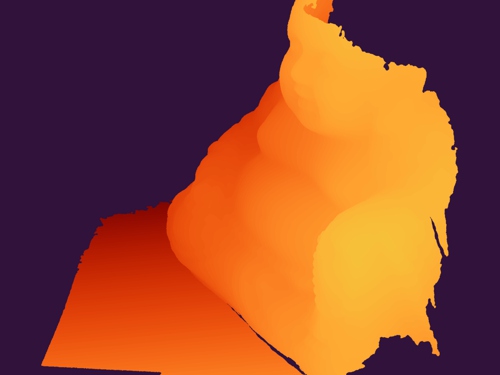} &
        \includegraphics[height=\turnheightnew]{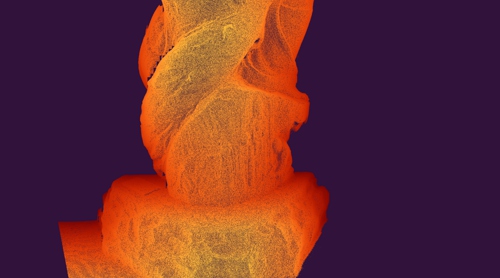}  \\
    \end{tabular}
    }
    \vspace{-10pt}
    \caption{
        \textbf{Qualitative comparison of depth prediction results across multiple datasets (Normalized)} (KITTI, ScanNet, ETH3D, DTU, and Tanks \& Temples). 
        Rows show different methods: Depth Pro~\cite{schroeppel2022robust}, rMVD baseline~\cite{schroeppel2022robust}, MAST3R (Triangulated)~\cite{mast3r_arxiv24}, and our MVSA model, along with RGB inputs ($I_r$) and ground-truth depths (GT). 
        Depth Pro provides sharp edges but often misestimates depth scale, while our MVSA model captures finer details than MAST3R and rMVD. 
        Depth maps are normalized to ground truth depth range for consistent visualization.
    }
    \label{fig:qualitative_depths_sup2}
\end{figure*}

\begin{figure*}
    \resizebox{1.0\textwidth}{!}{
    \newcommand{\turnheightnew}{60pt}
    \begin{tabular}{@{\hskip -2mm}c@{\hskip 1mm}c@{\hskip 1mm}c@{\hskip 1mm}c@{\hskip 1mm}c@{\hskip 1mm}c@{\hskip 1mm}c@{}}
        & KITTI & ScanNet & ETH3D & DTU & {T\&T} \\
    
        {\rotatebox{90}{\hspace{5mm}{\small RGB ($I_r$)}}} &
        \includegraphics[height=\turnheightnew, trim=4cm 0cm 5cm 0cm, clip]{figs_jpeg/qualitative/supplementary/2/kitti/color.jpg} &
        \includegraphics[height=\turnheightnew]{figs_jpeg/qualitative/supplementary/2/scannet/color.jpg} &
        \includegraphics[height=\turnheightnew]{figs_jpeg/qualitative/supplementary/2/eth3d/color.jpg} &
        \includegraphics[height=\turnheightnew]{figs_jpeg/qualitative/supplementary/2/dtu/color.jpg} &
        \includegraphics[height=\turnheightnew]{figs_jpeg/qualitative/supplementary/2/tanks_and_temples/color.jpg} \\
        
        {\rotatebox{90}{\hspace{1mm}{\small Depth Pro~\protect\cite{bochkovskii2024depthpro}}}} &
        \includegraphics[height=\turnheightnew, trim=4cm 0cm 5cm 0cm, clip]{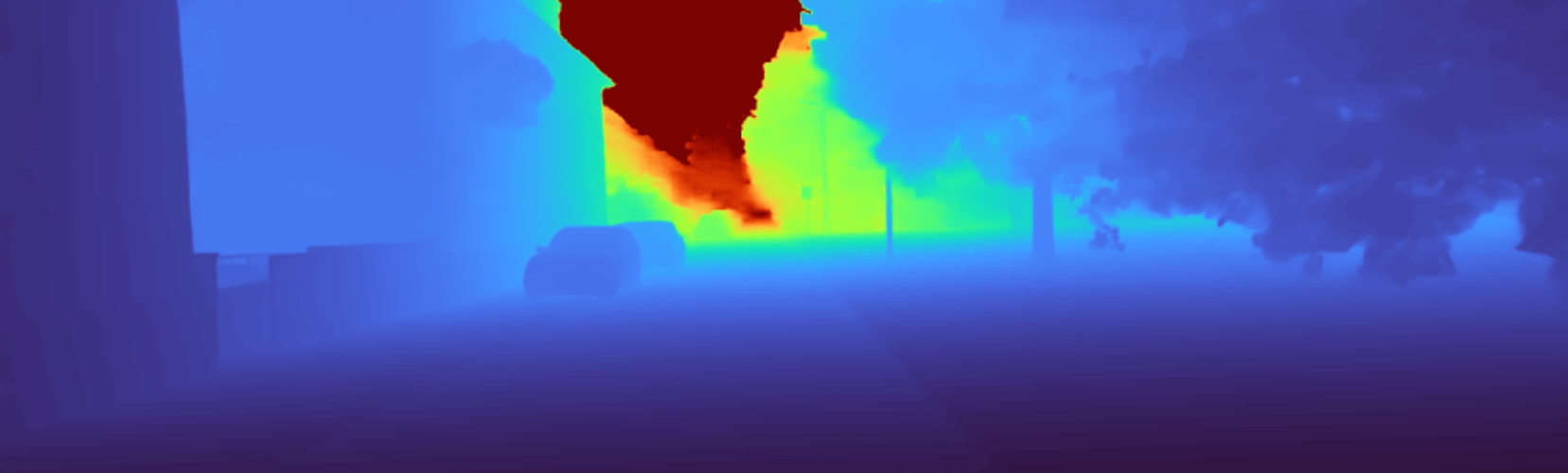} &
        \includegraphics[height=\turnheightnew]{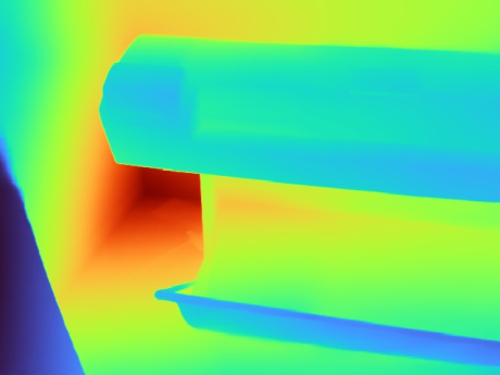} &
        \includegraphics[height=\turnheightnew]{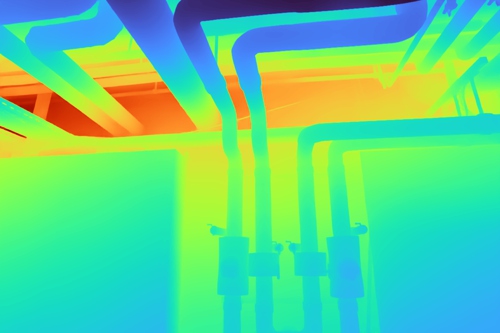} &
        \includegraphics[height=\turnheightnew]{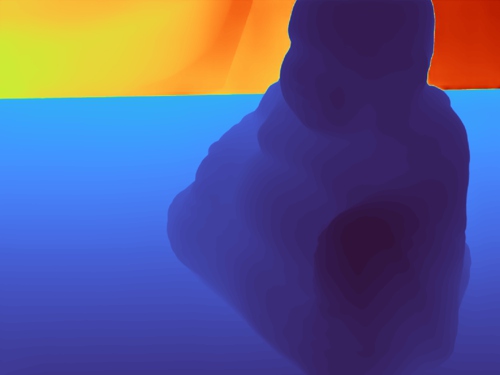} &
        \includegraphics[height=\turnheightnew]{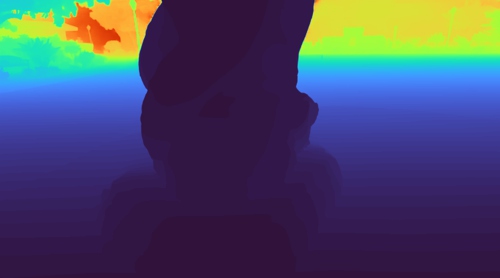} \\
        
        {\rotatebox{90}{\hspace{4mm}{\small rMVD~\protect\cite{schroeppel2022robust}}}} &
        \includegraphics[height=\turnheightnew, trim=4cm 0cm 5cm 0cm, clip]{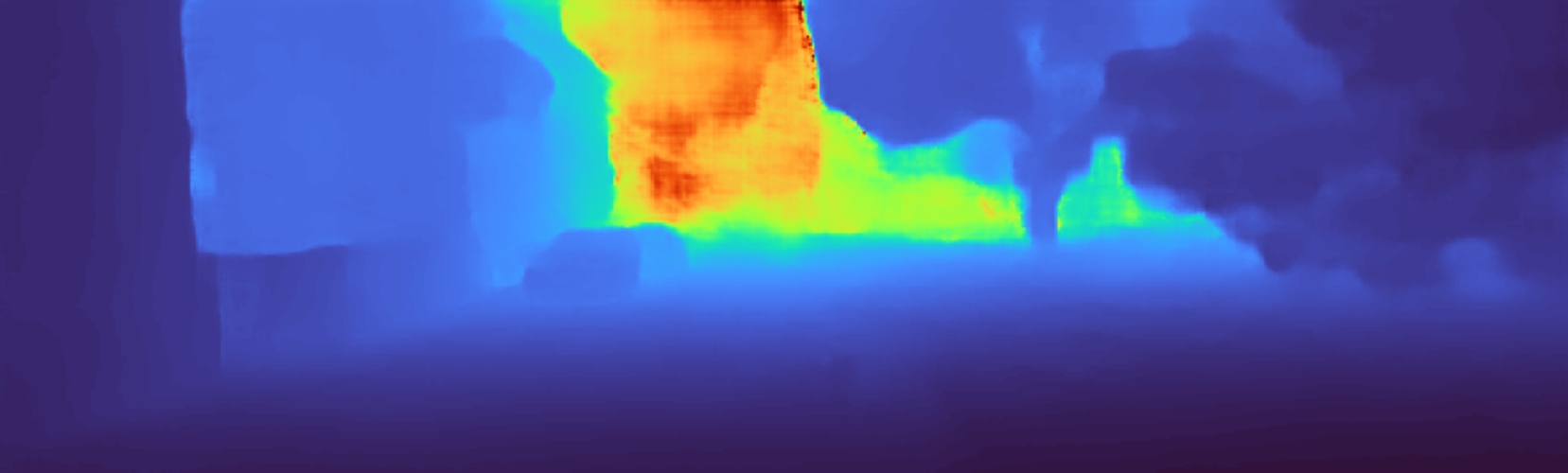} &
        \includegraphics[height=\turnheightnew]{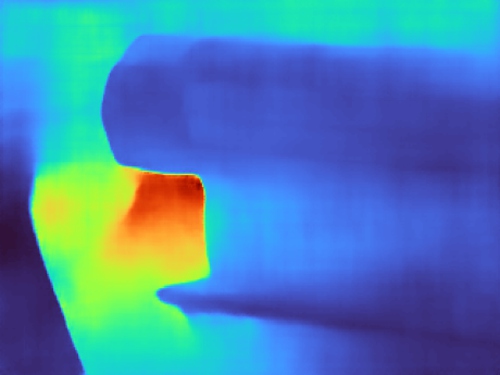} &
        \includegraphics[height=\turnheightnew]{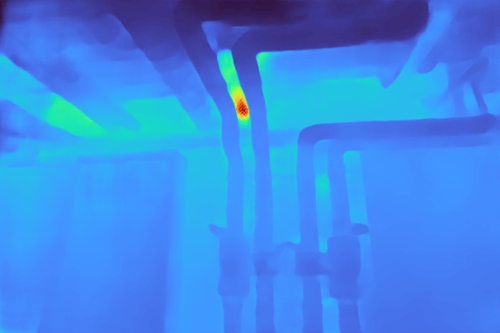} &
        \includegraphics[height=\turnheightnew]{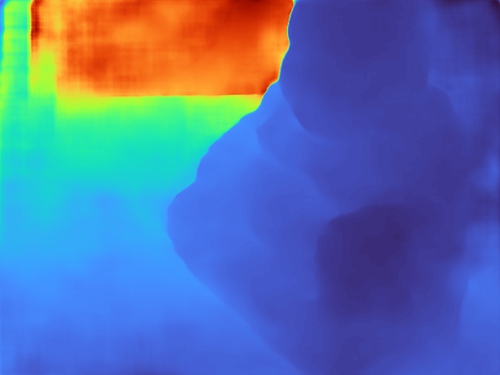} &
        \includegraphics[height=\turnheightnew]{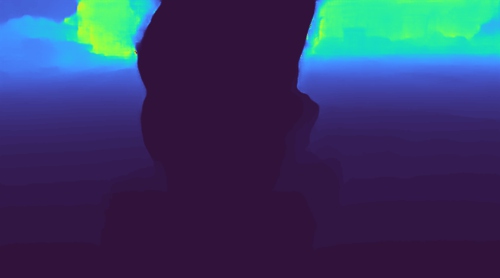} \\

        {\rotatebox{90}{\hspace{1mm}{\small MAST3R~\protect\cite{mast3r_arxiv24}}}} &
        \includegraphics[height=\turnheightnew, trim=4cm 0cm 5cm 0cm, clip]{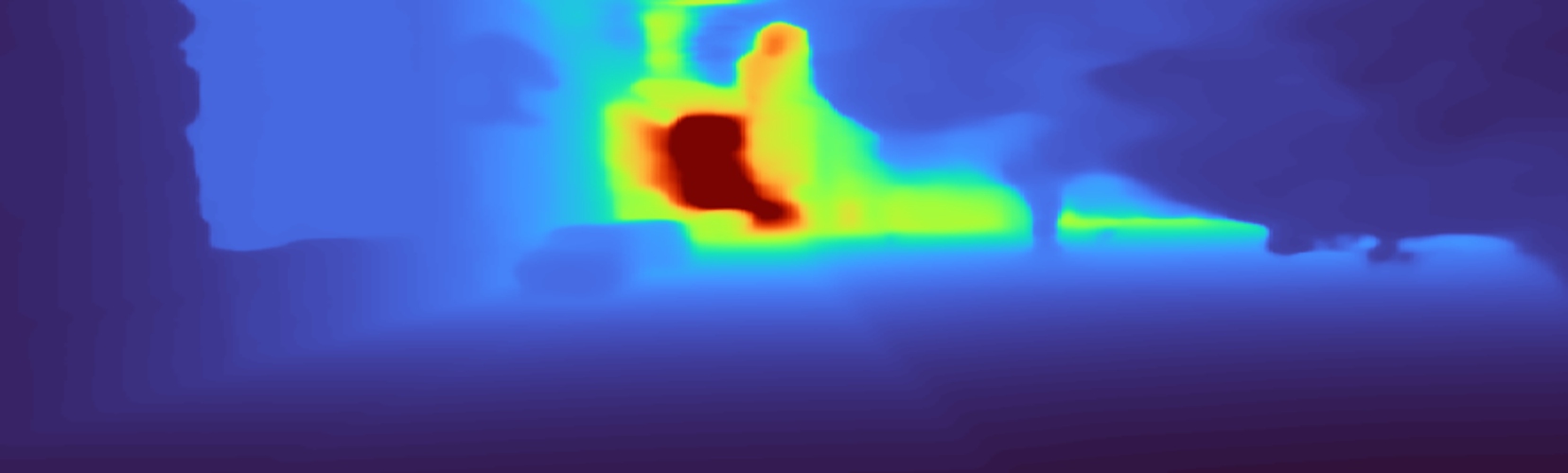} &
        \includegraphics[height=\turnheightnew]{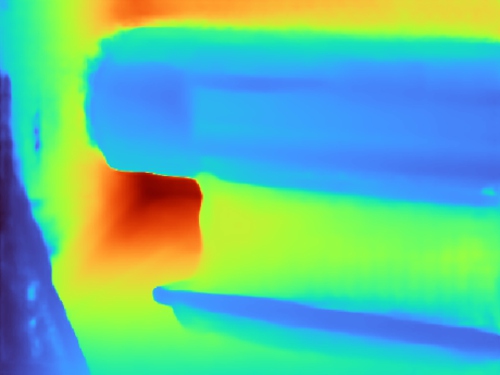} &
        \includegraphics[height=\turnheightnew]{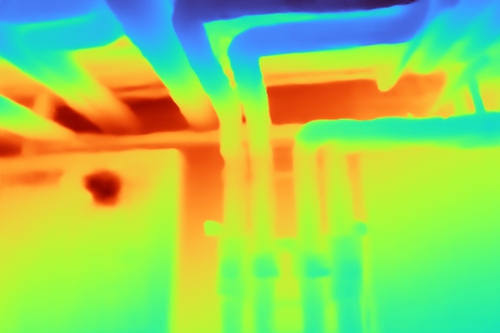} &
        \includegraphics[height=\turnheightnew]{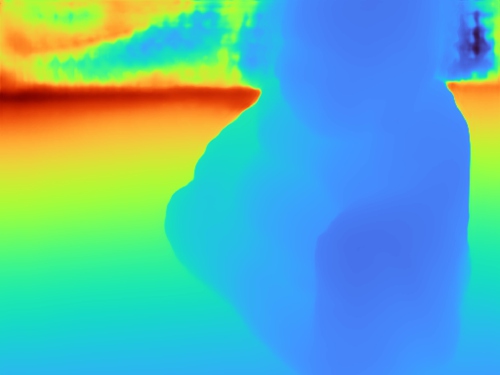} &
        \includegraphics[height=\turnheightnew]{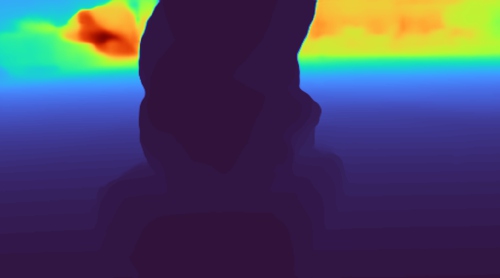} \\
        
        {\rotatebox{90}{\hspace{0.5mm}{\small MVSA (Ours)}}} &
        \includegraphics[height=\turnheightnew, trim=4cm 0cm 5cm 0cm, clip]{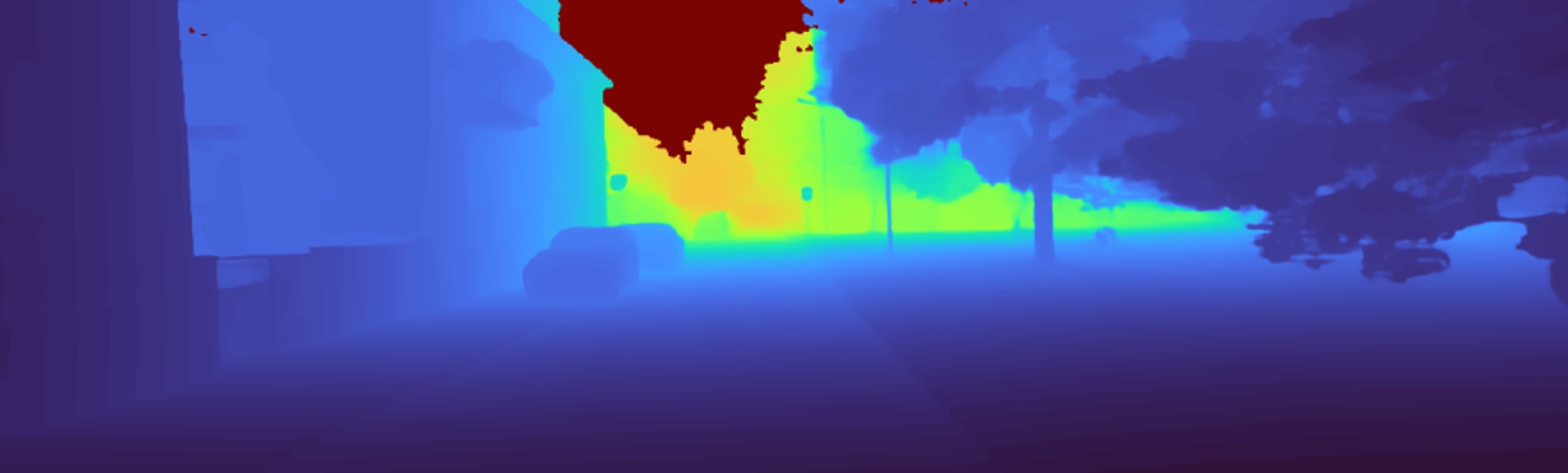} &
        \includegraphics[height=\turnheightnew]{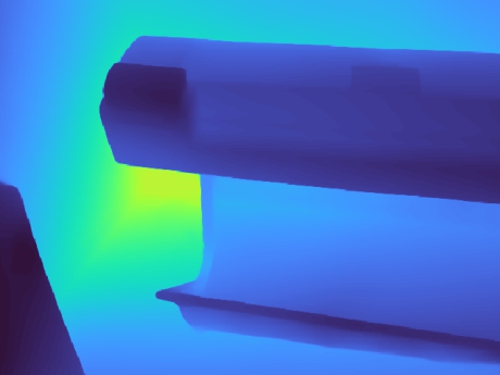} &
        \includegraphics[height=\turnheightnew]{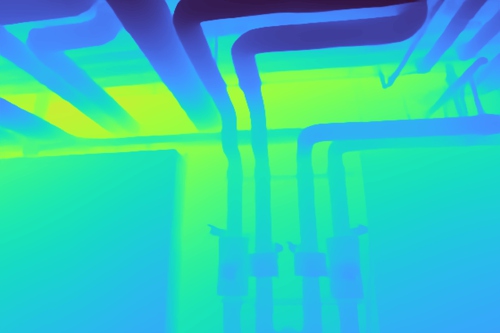} &
        \includegraphics[height=\turnheightnew]{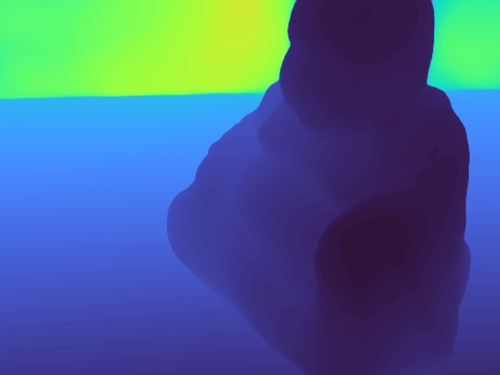} &
        \includegraphics[height=\turnheightnew]{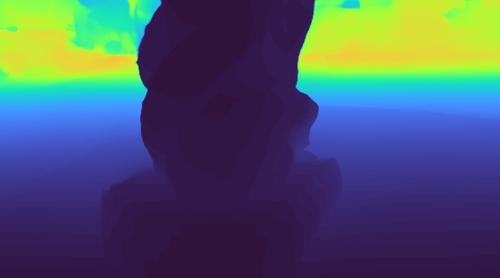} \\

        {\rotatebox{90}{\hspace{7mm}{\small GT}}} &
        \includegraphics[height=\turnheightnew, trim=4cm 0cm 5cm 0cm, clip]{figs_jpeg/qualitative/supplementary/2/kitti/depth_gt.jpg} &
        \includegraphics[height=\turnheightnew]{figs_jpeg/qualitative/supplementary/2/scannet/depth_gt.jpg} &
        \includegraphics[height=\turnheightnew]{figs_jpeg/qualitative/supplementary/2/eth3d/depth_gt.jpg} &
        \includegraphics[height=\turnheightnew]{figs_jpeg/qualitative/supplementary/2/dtu/depth_gt.jpg} &
        \includegraphics[height=\turnheightnew]{figs_jpeg/qualitative/supplementary/2/tanks_and_temples/depth_gt.jpg}  \\
    \end{tabular}
    }
    \vspace{-10pt}
    \caption{
        \textbf{Qualitative comparison of depth prediction results across multiple datasets (Unnormalized)} (KITTI, ScanNet, ETH3D, DTU, and Tanks \& Temples). 
        Rows show different methods: Depth Pro~\cite{schroeppel2022robust}, rMVD baseline~\cite{schroeppel2022robust}, MAST3R (Triangulated)~\cite{mast3r_arxiv24}, and our MVSA model, along with RGB inputs ($I_r$) and ground-truth depths (GT). 
        Depth maps are normalized per image.
    }
    \label{fig:qualitative_depth_unnorms_sup_2}
\end{figure*}

\begin{figure*}
    \resizebox{1.0\textwidth}{!}{
    \newcommand{\turnheightnew}{60pt}
    \begin{tabular}{@{\hskip -2mm}c@{\hskip 1mm}c@{\hskip 1mm}c@{\hskip 1mm}c@{\hskip 1mm}c@{\hskip 1mm}c@{\hskip 1mm}c@{}}
        & KITTI & ScanNet & ETH3D & DTU & {T\&T} \\
    
        {\rotatebox{90}{\hspace{5mm}{\small RGB ($I_r$)}}} &
        \includegraphics[height=\turnheightnew, trim=4cm 0cm 5cm 0cm, clip]{figs_jpeg/qualitative/supplementary/2/kitti/color.jpg} &
        \includegraphics[height=\turnheightnew]{figs_jpeg/qualitative/supplementary/2/scannet/color.jpg} &
        \includegraphics[height=\turnheightnew]{figs_jpeg/qualitative/supplementary/2/eth3d/color.jpg} &
        \includegraphics[height=\turnheightnew]{figs_jpeg/qualitative/supplementary/2/dtu/color.jpg} &
        \includegraphics[height=\turnheightnew]{figs_jpeg/qualitative/supplementary/2/tanks_and_temples/color.jpg} \\
        
        {\rotatebox{90}{\hspace{1mm}{\small Depth Pro~\protect\cite{bochkovskii2024depthpro}}}} &
        \includegraphics[height=\turnheightnew, trim=4cm 0cm 5cm 0cm, clip]{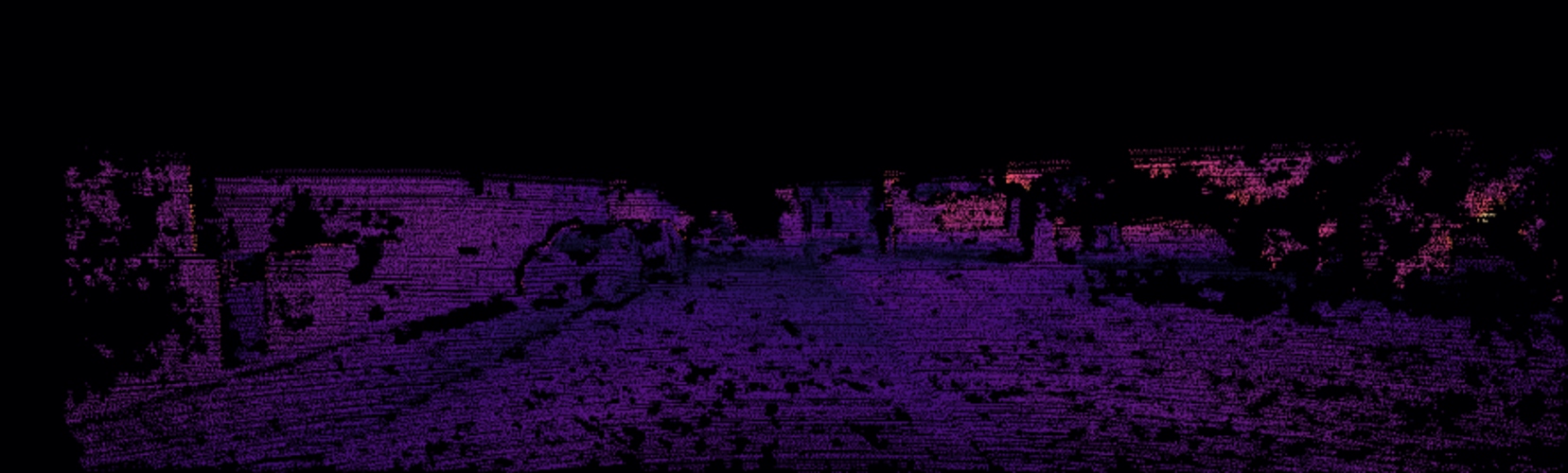} &
        \includegraphics[height=\turnheightnew]{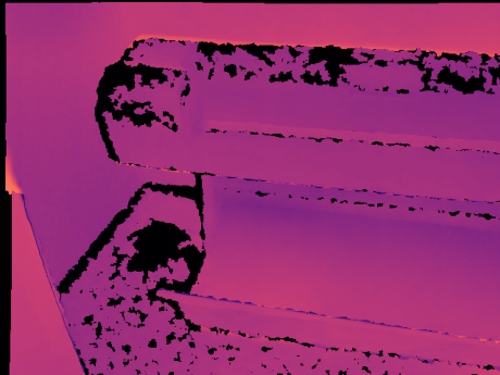} &
        \includegraphics[height=\turnheightnew]{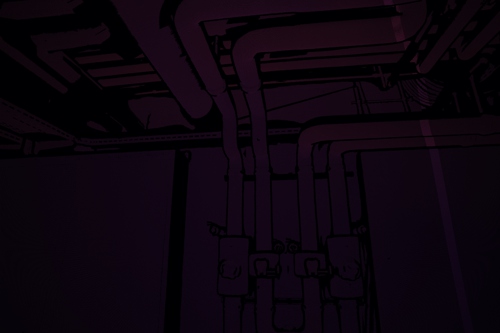} &
        \includegraphics[height=\turnheightnew]{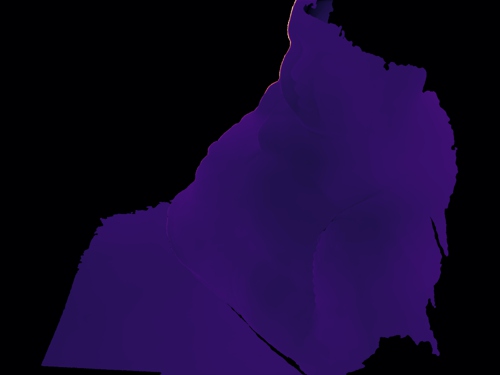} &
        \includegraphics[height=\turnheightnew]{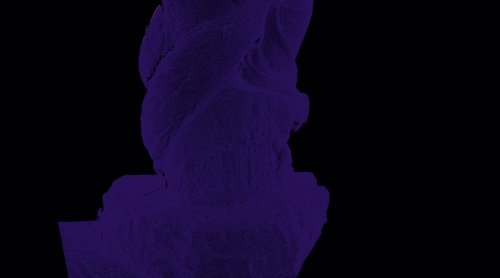} \\
        
        {\rotatebox{90}{\hspace{4mm}{\small rMVD~\protect\cite{schroeppel2022robust}}}} &
        \includegraphics[height=\turnheightnew, trim=4cm 0cm 5cm 0cm, clip]{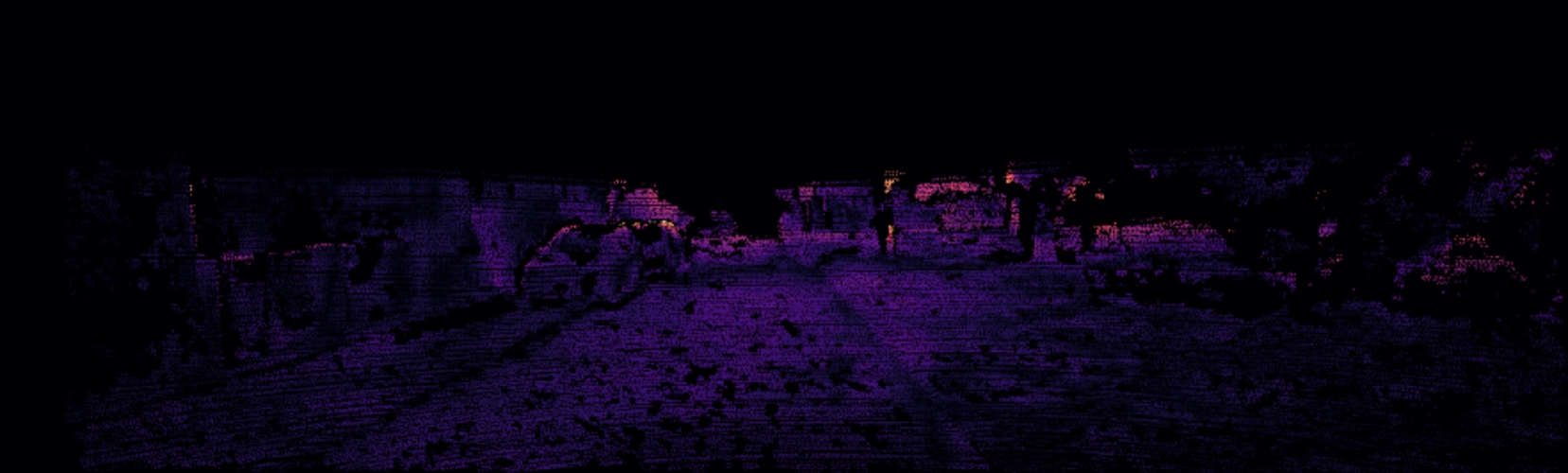} &
        \includegraphics[height=\turnheightnew]{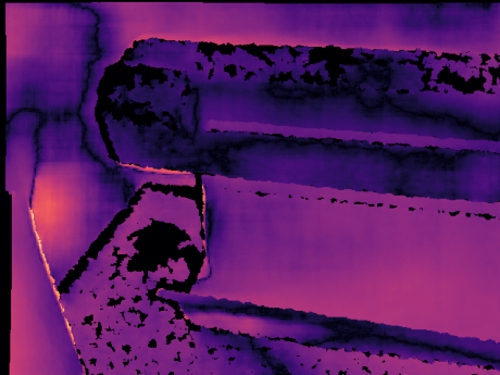} &
        \includegraphics[height=\turnheightnew]{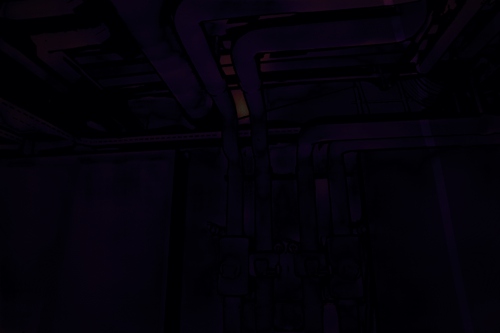} &
        \includegraphics[height=\turnheightnew]{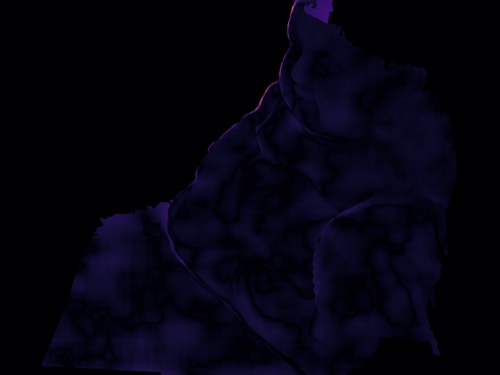} &
        \includegraphics[height=\turnheightnew]{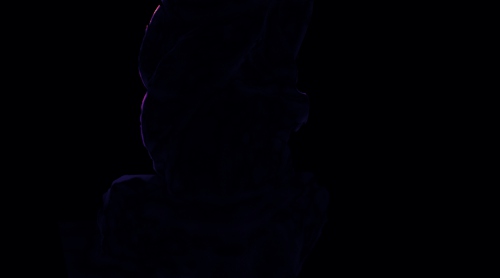} \\

        {\rotatebox{90}{\hspace{1mm}{\small MAST3R~\protect\cite{mast3r_arxiv24}}}} &
        \includegraphics[height=\turnheightnew, trim=4cm 0cm 5cm 0cm, clip]{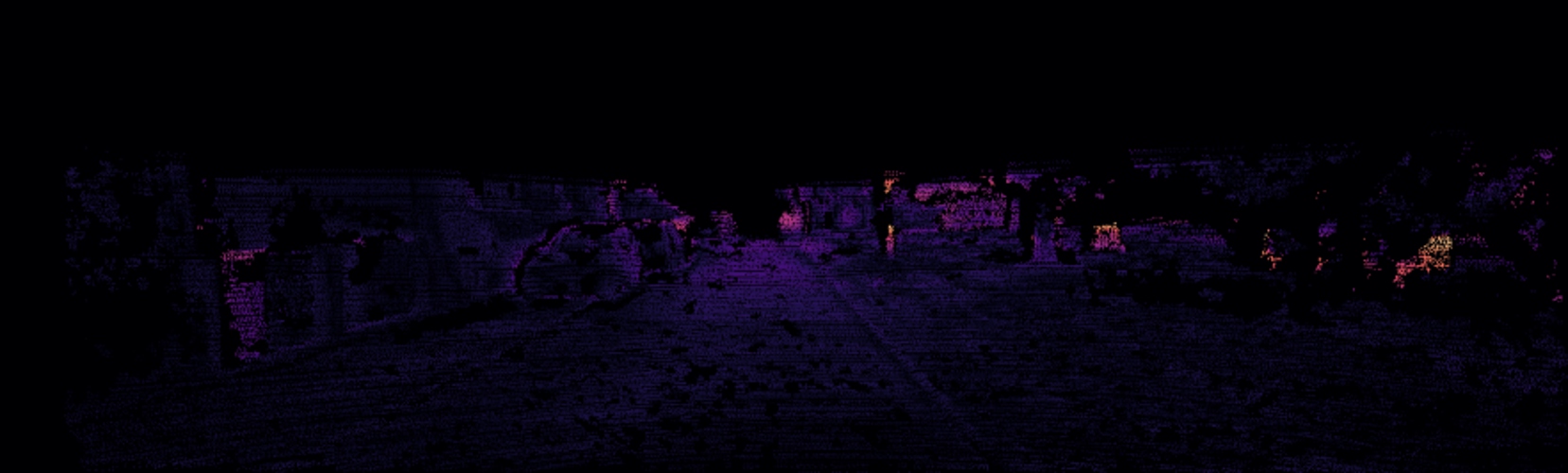} &
        \includegraphics[height=\turnheightnew]{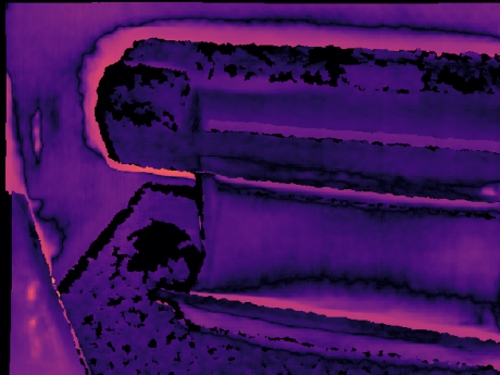} &
        \includegraphics[height=\turnheightnew]{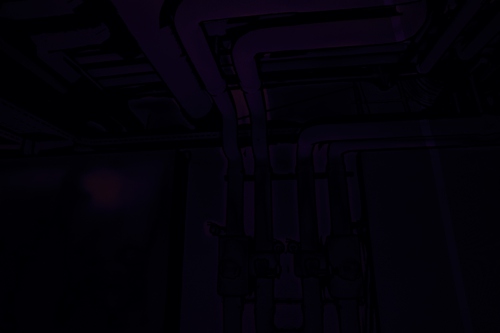} &
        \includegraphics[height=\turnheightnew]{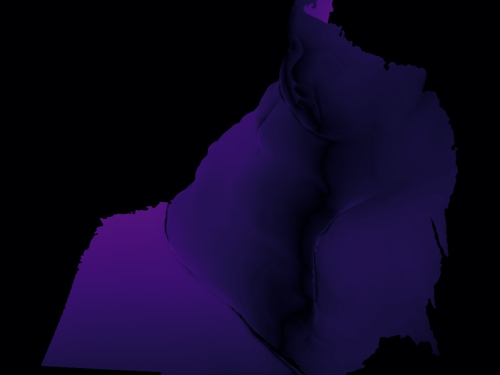} &
        \includegraphics[height=\turnheightnew]{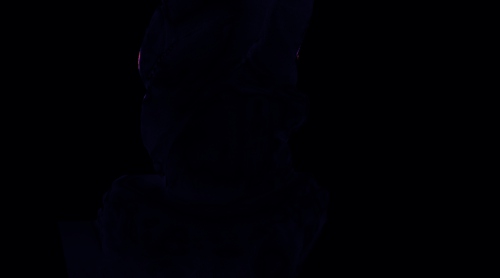} \\
        
        {\rotatebox{90}{\hspace{0.5mm}{\small MVSA (Ours)}}} &
        \includegraphics[height=\turnheightnew, trim=4cm 0cm 5cm 0cm, clip]{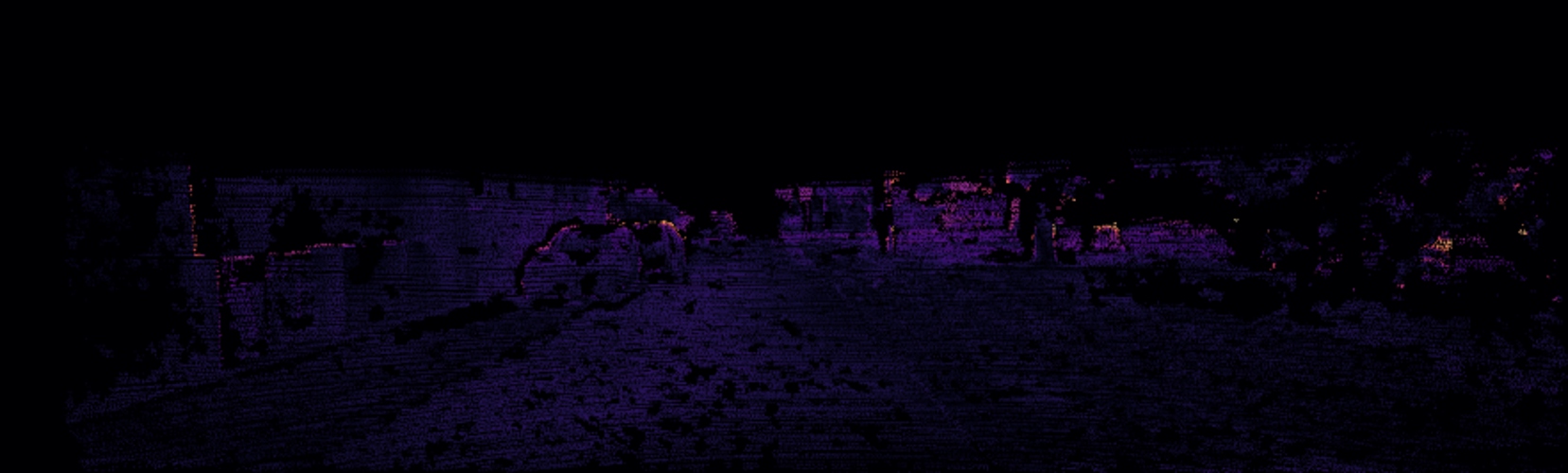} &
        \includegraphics[height=\turnheightnew]{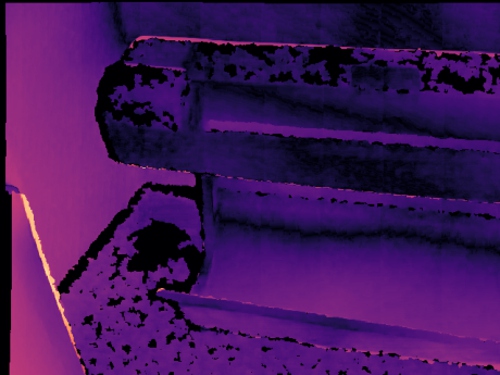} &
        \includegraphics[height=\turnheightnew]{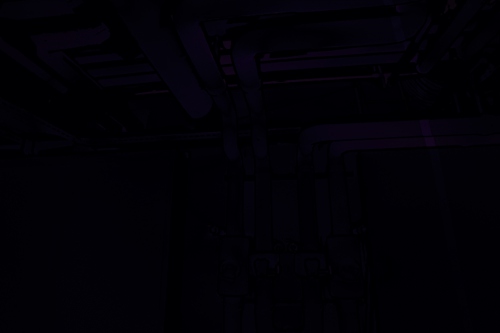} &
        \includegraphics[height=\turnheightnew]{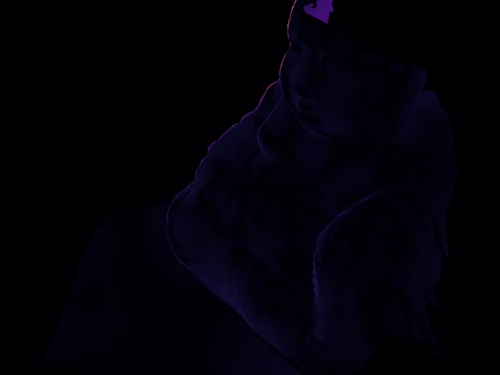} &
        \includegraphics[height=\turnheightnew]{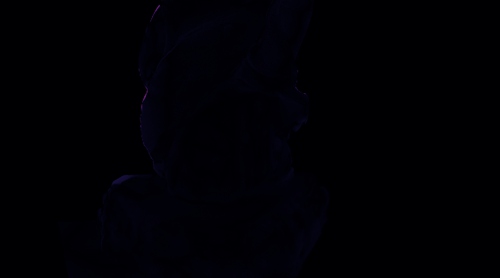} \\

        {\rotatebox{90}{\hspace{5mm}{\small Depth GT}}} &
        \includegraphics[height=\turnheightnew, trim=4cm 0cm 5cm 0cm, clip]{figs_jpeg/qualitative/supplementary/2/kitti/depth_gt.jpg} &
        \includegraphics[height=\turnheightnew]{figs_jpeg/qualitative/supplementary/2/scannet/depth_gt.jpg} &
        \includegraphics[height=\turnheightnew]{figs_jpeg/qualitative/supplementary/2/eth3d/depth_gt.jpg} &
        \includegraphics[height=\turnheightnew]{figs_jpeg/qualitative/supplementary/2/dtu/depth_gt.jpg} &
        \includegraphics[height=\turnheightnew]{figs_jpeg/qualitative/supplementary/2/tanks_and_temples/depth_gt.jpg}  \\
    \end{tabular}
    }
    \vspace{-10pt}
    \caption{
        \textbf{Qualitative comparison of depth prediction errors across multiple datasets} (KITTI, ScanNet, ETH3D, DTU, and Tanks \& Temples). 
        Rows show different methods: Depth Pro~\cite{schroeppel2022robust}, rMVD baseline~\cite{schroeppel2022robust}, MAST3R (Triangulated)~\cite{mast3r_arxiv24}, and our MVSA model, along with RGB inputs ($I_r$) and ground-truth depths (GT). 
        Depth Pro provides sharp edges but often misestimates depth scale, while our MVSA model captures finer details than MAST3R and rMVD. 
        Error maps are normalized to maximum error among methods for each scene.
    }
    \label{fig:qualitative_errors_sup2}
\end{figure*}

We present additional qualitative results in \cref{fig:qualitative_depth_unnorms_sup_main,fig:qualitative_depths_main,fig:qualitative_errors_sup_main,fig:qualitative_depth_unnorms_sup,,fig:qualitative_errors_sup1,fig:qualitative_depth_unnorms_sup_2,fig:qualitative_depths_sup2,fig:qualitative_errors_sup2}. \cref{fig:qualitative_depths_main,fig:qualitative_depths_sup,fig:qualitative_depths_sup2} depicts depth maps where the range has been normalized using the ground truth range. 
It demonstrates how our \modelname approach is able to predict very similar scale to the ground truth, whereas Depth Pro struggles to do so. 
In \cref{fig:qualitative_depth_unnorms_sup_main,fig:qualitative_depth_unnorms_sup,fig:qualitative_depth_unnorms_sup_2} the depth maps are normalized per-image. Our model and Depth Pro produce the sharpest results. In \cref{fig:qualitative_errors_sup_main,fig:qualitative_errors_sup1,fig:qualitative_errors_sup2} the error maps reveal how our approach achieves the best alignment with the ground truth. Although MAST3R's scale is good, it has high errors around object boundaries. 

\end{document}